# Neural Networks for Chess
## The magic of deep and reinforcement learning revealed

Dominik Klein

June 11, 2022

# Contents

# Preface

This book is a brief introduction into modern computer chess. Chess engines using artificial intelligence based on *deep-learning* made quite an impact in the chess community. AlphaZero, the chess monster developed by Google's research subsidiary *DeepMind* suddenly seemed to be able to play beautiful, almost human-like chess. Some games showed piece sacrifices with no immediate tactical gain but positional long-term advantages that common chess engines were unable to find. Some even proclaimed jokingly the return of the *era of Romantic chess*.[1]

From the perspective of a chess player, the inner workings of a chess engine are often a mystery. They are usually understood as black boxes that only genius programmers and researchers can understand. In the famous Deep Blue versus Garry Kasparov (re)match in 1997, where the world chess champion Kasparov lost to an IBM-developed super computer dubbed Deep Blue, he made allegations that the Deep Blue team cheated. He hinted that a specific move in a position in Game 2 of the match was not the result of computer calculation but that the move was the result of human intervention.

He stated that it seemed very unlikely that a chess computer could have made that particular move. His argument was that the computer abstained from winning a pawn in order to get a better piece placement instead. At the time he simply could not comprehend that a chess engine was capable of such a move.

---

[1] referring to the style of chess that was common in the 18th century, namely wild but beautiful battles and sacrifices.

With some distance Kasparov recounts the match from his perspective [KG18] and corrects his misconceptions somewhat. Contrasted to that, Deep Blue developer Feng-hsiung Hsu's technically detailed recount of the events paints a very different story [Hsu04]. Both are very interesting reads and I recommend giving them both a try.

Nevertheless one thing must be stated clearly: There is no reason to believe that DeepBlue was not able to find the move in question in that position. This is even acknowledged by Kasparov in his book where he states that he was wrong and owes the Deep Blue team an apology. He notes that *current* computers find the best move in seconds. It takes great courage to admit one's faults. But even engines released in 1997 were able to find the correct move as we will see in Chapter 3.

Kasparov dominated chess over decades. How come that such a chess genius completely misjudged the situation? We can only speculate but it seems he assessed the chess engine purely from a chess player's perspective neglecting its inner technical workings. Moreover he apparently relied on the chess program Fritz to train and assess chess computers [KG18] even though Fritz at that time was known to be somewhat of a "pawn snatcher" with high tactical vision but subpar positional evaluation compared to other engines.

With more technical understanding and by taking the engineering perspective into account we can pose the question if Kasparov would have been able to prolong the inevitable loss of humans versus chess computers for some more time. We can only speculate, but the psychological side of things, namely being surprised and stunned by engine moves should not be underestimated: After the fifth game, Kasparov said he had not been in the mood for playing, and when asked to elaborate on his outlook, he said: "I'm a human being. [...] When I see something that is well beyond my understanding, I'm afraid."[2]

Allegedly Kasparov was perplexed after 44... Rd1 in Game 1 of the 1997 rematch and he and his team deeply analyzed why DeepBlue preferred it to

---

[2]`https://www.latimes.com/archives/la-xpm-1997-05-14-ls-58417-story.html` and `https://www.nytimes.com/1997/05/12/nyregion/swift-and-slashing-computer-topples-ka sparov.html`, accessed 14.07.2021

44...Rf5; coming to the conclusion that DeepBlue must have seen mate in 20 or more [Hsu04].

According to Hsu [Hsu04] it was just a bug in the program. With a different perspective on the technical challenges involved in designing and programming a massively parallel system with custom hardware, Kasparov's team might have given this possibility a much higher probability.

This book is not about the Kasparov versus Deep Blue match. But this story clearly illustrates that it is interesting and maybe even worthwhile to look behind the curtain and figure out the inner workings of chess engines, even if one is not a computer scientist but just a chess player enjoying the game. It's one thing to look in awe and admire the *game changer* AlphaZero, but it's another one to figure out how AlphaZero and similar engines work and get a better understanding what they are capable of.

Chess programming is a quite technical domain. It is hard to get started and the entry barrier is seemingly high. Famous Science Fiction author Arthur C. Clarke once wrote that "any sufficiently advanced technology is indistinguishable from magic" [Cla99]. Here we want to reveal that magic: This book is for all those who do not want to put up with shallow and superficial explanations such as "This engine works better because it now has artificial intelligence". This book is for those who want to look deeper, love to take things apart and dig in the mud to figure out how it all works.

We will make a journey into computer chess. In this book you will learn what a neural network is and how it is trained, how classical alpha-beta searchers work and how all this was combined to create the strongest chess engine available today, Stockfish NNUE.

Confused by all this lingo? That's ok since we will all sort it out step by step. Yet this book is not an graduate textbook. We will cover all the central ideas with lots of examples but try to abstract away unnecessary details and make sure you will (hopefully) not get lost in mathematical details.

Indeed, there is math involved to understand neural networks, but it is really not that much. Also, there is surprisingly little prior technical knowledge required.

In fact, there are only two major mathematical concepts that you have to know. First, there is basic calculus, namely how to calculate a derivative. Second you need to understand matrix multiplications. You most likely remember these things from high school, and if not, I suggest that you dig up your old high-school math textbook and do a quick brush up. But fortunately that's mostly it, really.

This book contains programming examples in Python. This is a really newbie friendly and simple programming language. But make no mistake — by no means it is required to use these programming examples or become a programmer to understand the concepts described in this book. To illustrate this you can think of it like being an architect: It helps if you are familiar with masonry and maybe even worked in construction, but it is by no means a must to understand how houses are designed. If you like programming though, these example support learning the presented concepts and serve as a starting point for your own experiments.

All source code presented in this book is also available online at

```
http://github.com/asdfjkl/neural_network_chess
```

If you are a chess player and work carefully through this book, you will have a better understanding how chess engines work, what their limitations are, and how to interpret their results. If new engines and approaches appear, you will have a better understanding what their advancements imply for the game. You will also be able to evaluate the marketing lingo by commercial vendors.

And if you encounter yet another newspaper article about the "revolutions of artificial intelligence and deep learning", you will have a much better ability to judge on whether the claims made there are accurate or a dubious science fiction fantasy at best.

Have a lot of fun!

# 1

# Introduction

I think we should all say things
but pretend that other, very smart
people said them and we are
merely quoting them

Aristotle

Chess is made up of tactics and strategy. For tactics you calculate moves and replies and look if any of the move sequences leads to forced advantage or disadvantage. Essentially you *search* among all possible variations. For strategy you judge a position without too much calculation.

My father taught me the rules of chess, probably at the age of eight or nine, but I later lost interest in the game. I re-discovered chess during the 2014 world chess championship between Carlsen and Anand, started playing and searched for high quality teaching material. Then I stumbled upon Jeremy Silman's *Reassess Your Chess* [Sil10]. There it was — all aspects to *evaluate* a position listed and explained: pawn structures, strong and weak squares, open lines, good and bad squares for your pieces, king safety — everything was there and formulated in a way such that an amateur can understand it. Silman explains how to characterizes a position by *imbalances*, i.e. advantages and disadvantages that characterize the position. He advocates to use this kind of evaluation technique to find good moves in a position.

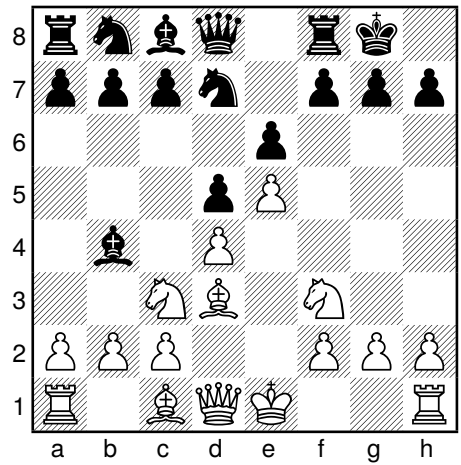

Figure 1.1: White to move.

I later stumbled upon Willy Hendriks' book *Move First, Think Later* [Hen12]. Hendriks is very critical of Silman's method of evaluation by imbalances. His argument is that we rarely find good moves by just looking at a position, and that we should instead immediately *search* and apply moves that come to our mind — hence the title — in order to come to an understanding of the position at hand and find the best next move.

From a computer scientist's perspective, this is however a pseudo-debate: Of course we need both, search[1] *and* evaluation.

Let's have look at the famous Greek gift sacrifice shown in Figure 1.1. How do we evaluate this position? There is equal material of course. White has maybe a small space advantage. Thus without searching we would evaluate the position with a small advantage for White. But after *searching* through several candidate moves, we quickly find the sequence 1.Bxh7+! Kxh7 2.Ng5+. Now we have to search deeper for all possible replies after each move and find that in the resulting positions, either White checkmates or wins material. In other words, the resulting positions are very easy to evaluate with high confidence. Based on these evaluations, we re-evaluate the original position as "White has a significant advantage".

---

[1]Chess players often refer to *searching* as *tactics*, or calculation. But since we focus on computer chess, let's stick to *search* for now.

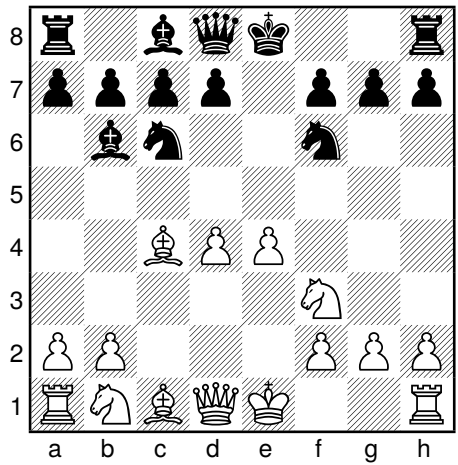

Figure 1.2: M. Euwe - K. Jutte, Amsterdam 1927. White to move.

However often when we calculate, we cannot reach a checkmating position or one where the evaluation is easy due to significant material advantages. Of course, every chess game ends in a final position *eventually*, but it is in general impossible to search so deeply. Thus we have to stop after a certain number of moves, and then evaluate the resulting position by structural features, such as Silman's imbalances.

For example, take a look at Figure 1.2[2]. We can go through several move sequences, in particular d5! and all subsequent replies by Black; consider especially Ne7 followed by e5. Only by using Silman's imbalance criteria and without any further search, we can evaluate the resulting position(s): There is a significant positional advantage in all subsequent variations for White, but not an immediate material gain or checkmating sequence. We can then use the evaluations of the resulting positions to (re)evaluate the original position of Figure 1.2. In other words, White has a significant advantage in the current position, due to his space advantage in the center.

---

[2]This is an interesting position, since even modern chess engines require some time to find d5! Stockfish 12 for example looks at Nc3 first until it finally switches to d5 and notices the significant advantage.

I take it that Willy Hendriks' major point is that humans rarely do it in the structural way that is often taught, i.e.

- find a number of candidate moves. Then for each of them sequentially...
- calculate all subsequent positions and variation trees that result from those candidate moves as deep as possible
- evaluate all the resulting positions
- finally, based on all these evaluations, (re)evaluate the current position and select the best move

Because really, that's how computers do it! Humans probably jump back and forth, forget candidate moves, focus on one specific line, then decide by instinct instead of by rational thought and become angry when realizing that a piece has just been lost.

At least that's how I do it.

In the end however, it all centers around *searching* among (candidate) moves, and *evaluating* positions. And since we cannot always search so deep that we end up in a final state of a game, we need to stop after a number of moves, and evaluate a *non-final position*.

For a long time, humans were always better in *evaluating* positions and worse at *searching*. This was especially true for chess positions that are difficult to judge, i.e. where no specific criteria exist, so that we can calculate or prove an advantage. Typical examples are very closed positions like the one in Figure 1.3. White has a material advantage here. Rybka wants to avoid a draw by repetition in a position with a material advantage, and will finally make a bad move, misjudging the corresponding positional disadvantage. Black won the game eventually. It's the specific pawn structure that will result in White's eventual loss if he tries to force a win, and the rooks cannot prevent this.

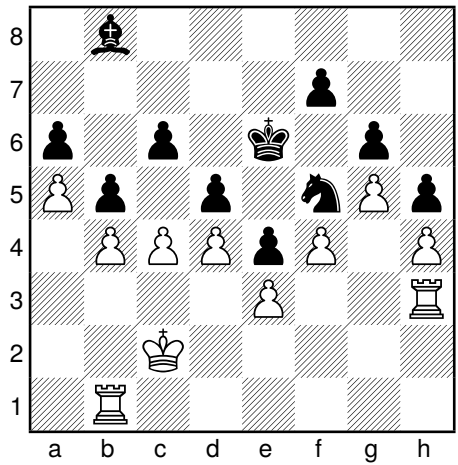

Figure 1.3: Rybka (Computer) vs Hikaru Nakamura, after move 174. ICC, 2008. Black to move.

The advantage in searching that computers have is very obvious: Computers are just better at number crunching. Any pocket calculator is faster at multiplying numbers than a human.

The weakness in evaluating positions is something that have plagued computers for a long time. Humans learn how to judge a position by intuition. It is often difficult to express in words *why* a position is better or worse. Stefan Meyer-Kahlen, author of the chess program Shredder, once told this story:[3] In some positions, Shredder made a particularly bad positional move. He asked a strong player what the problem was in order to generate a potential fix, i.e. he had to figure out what the exact problem was. But he only got the reply "Nah, you just don't play like that".

---

[3]Eric van Reem: Let's talk about chess #12: Stefan Meyer-Kahlen, `https://letscast.fm/site s/let-s-talk-about-chess-753d92ea/episode/12-stefan-meyer-kahlen-shredderchess`

The revolution that AlphaZero brought, the real game changer is generating that ability to intuitively evaluate a chess position like a human, and to not rely on a fixed rule set like e.g. pawn counting. Technically, chess engines use *neural networks* to create an abstract model of a human's brain, in order to process and judge a chess position.

But how do good players *become* so good at chess? If we listen to Grandmaster's advice, frequently mentioned is the following strategy:

- Look at your own games without an engine. Look careful at the mistakes you and your opponent made and learn from these mistakes to play better next time.

- Look carefully at games of Grandmasters. Improve by looking at their approaches and plans in various types of positions.

Interestingly, following this strategy is precisely what computer chess programs do nowadays to gain a (super-)human understanding of chess positions. Either by taking large databases of games of chess Grandmasters, or by looking and learning from their own games.

For the first step we obviously have to play a lot of games. In the famous novel Schachnovelle by Stefan Zweig, a chess player becomes very good at chess (and almost insane) by splitting his personality and playing endless games of chess between these two personalities. Computers have no insanity issues (so far), and hence, *learning by self-play* has become the de-facto standard in training chess programs or rather the underlying neural networks. This is also mostly the preferred approach compared to looking at large databases of games of human players.

You could also say that computers finally figured out how humans learn and play and this is why they get so strong. What more of a compliment could there be to us mere mortals?

# How To Read This Book

Chapter 2 is all about neural networks. Here we discuss the mathematical concepts and lay down the foundation for all subsequent discussions about the recent improvements based on artificial intelligence.

If this is too bothersome to read due to the math involved, you can also simply chose to understand a neural network as a black box that, given a chess position, answers with a positional evaluation; skip this chapter, and come back to it later. If you chose to go with it, you will learn how to approximate unknown, complex functions, detect *Baikinman* in images, and construct fancy complex neural networks for which you unfortunately very likely lack the computational power to train them.

In Chapter 3 we look at search methods that chess engines employ. We learn how "classic" chess engines employ alpha-beta search, and how Monte-Carlo tree search provides an alternative for chess and also other, more difficult games such as Shogi or Go. As an example, we will also implement alpha-beta search and Monte-Carlo tree search and test it with a simple chess position.

Chapter 4 is where it all comes together. We take a deep look at how AlphaZero and Leela Chess Zero work technically, and how they evolved from AlphaGo. We also take a look at the latest revolution in computer chess, namely the combination of classic alpha-beta searchers in chess and Shogi with simpler neural networks. These are current state-of-the art and are able to beat Leela Chess Zero. And they can even run on a standard home computer!

In Chapter 5 we will implement our own neural network based game engine, similar to AlphaZero or Leela Chess Zero. Since the game of chess is too complex to train effectively on standard PC, we will limit ourselves to a simpler variant of chess called Hexapawn. Still, since all methods are there and since the AlphaZero approach is quite agnostic to the particular rules of a game, this implementation can easily be extended to more complex problems including chess — provided, the huge required computational resources are available.

Finally in Chapter 6 we conclude the recent developments and wildly speculate on what's there to come.

How to read this book then? If you are

- mostly interested in AlphaZero, then read Chapter 2 until multilayer perceptrons in order to get a brief understanding about neural networks. Then head to Chapter 3 and read the section about Monte Carlo tree search and proceed to Chapter 4. Skip the section about AlphaGo and proceed directly to the sections on both AlphaGo Zero and AlphaZero.

- mostly interested in efficiently updatable neural networks, then also read Chapter 2 until multilayer perceptrons in order to get a brief understanding about neural networks, read the section about alpha-beta search in Chapter 3 and the section about NNUE in Chapter 4.

- are mostly interested in implementing your own version of AlphaZero and have a programmer's perspective, then just head over to Chapter 5 and have look at HexapawnZero. Then look-up everything on-the go while you make your way through the code.

Ideally however you carefully read this book from start to finish!

**2**

# A Crash Course into Neural Networks

> There must be some kind of way outta here
> Said the joker to the thief
> There's too much convolution
> I can't get no relief
>
> Yann LeCun

In the movie Pi (1998), math genius and Go enthusiast Max Cohen is convinced that everything in the world can be understood through numbers. The more he delves into his theories, the more he is getting at the verge of becoming insane.

It is now our task to study the mathematical and programming concepts of neural networks that power modern strong Go and chess engines. Since we do not want to become insane, obviously great care must be taken and we should not take this task lightly.

Max Cohen by the way prevented himself from getting insane by self-performing an impromptu trepanning. Or maybe that scene was just supposed to be a hal-lucination, I am not sure. Anyway just to be on the safe side, grab yourself a cordless screwdriver, put it next to this book, and now let's start!

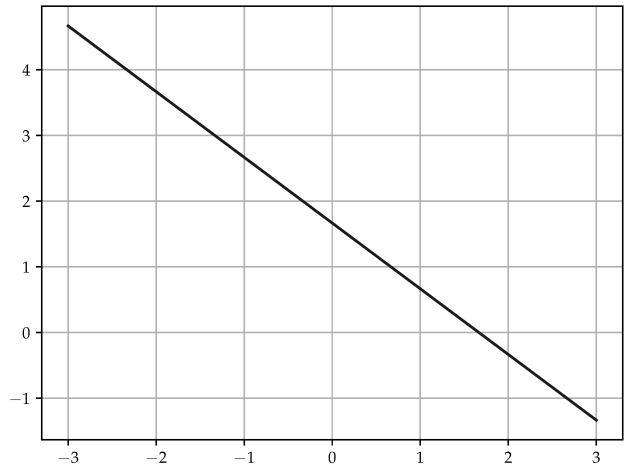

Figure 2.1: The line $y = -x + \frac{5}{3}$

## 2.1 The Perceptron

Neural networks are all about learning functions - in particular difficult, non-linear functions. Roughly said, a function is a description on how to map every value from one set to a value from another set. You might remember from high school linear equations of the form

$$y = ax + b,$$

where $x$ is a variable and $a$ and $b$ are fixed constants. You might also remember the following problem: Given two (or more) pairs $(x, y)$, determine the constant values $a$ and $b$ such that the generated line hits all pairs $(x, y)$. Consider for example the two points $(0, \frac{5}{3})$ and $(1, \frac{2}{3})$. In order to find suitable $a$ and $b$ we just put the two points as $x$ and $y$ into $y = ax + b$ and obtain two equations:

$$\frac{5}{3} = a0 + b \quad \text{and} \quad \frac{2}{3} = a1 + b.$$

By the first equation we get $b = \frac{5}{3}$, and the second equation leads to $a = \frac{2}{3} - b = \frac{2}{3} - \frac{5}{3} = -1$. This line is plotted in Figure 2.1.

Given some pairs of values it is of course easy to determine the underlying function if you know the form of the function in advance - like $y = ax + b$. But it is difficult to learn the function if you do not know the form of the function at all, and if the function is complex. Consider the following examples:

- Given a person's income, her age, her job education, her place of living and probably other variables, what is the most likely amount of credit debt that she can repay?

- Given two facial images, do they show the same person or not?

- Given an arbitrary chess position and assuming perfect play, will White win the game eventually, will Black win eventually, or is it a draw?

Note that all of these questions can be expressed as mathematical functions over sets of numbers. In the first case all input variables are numbers, and the credit debt is a number as well. In the second example you could consider images as being composed of pixels where each pixel is a triple of red, green and blue values in the range between 0.0 and 1.0, and the function maps to one if we have the same person, and to zero otherwise. In the third example you could encode a chess position as a number — we will see later how this can be done — and the function outputs 1 if White wins, −1 if Black wins, and 0 if we have a draw.

The goal is now to find forms of functions that are general enough and can be adjusted to simulate other, very complex functions whose true definition we do not know — like for the above mentioned examples.

For this, researchers took inspiration from the brain. A brain is made up of a web of neurons that are interconnected via so-called synapses. Depending on the signals that the synapses send, a neuron then itself fires signals to other synapses.

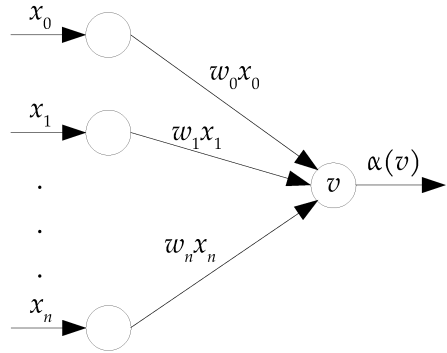

Figure 2.2: The Perceptron.

A mathematical approximation of such a neuron — dubbed a perceptron — is depicted in Figure 2.2. It has $n + 1$ *inputs*, however the first input $x_0$ is always set to 1. With each input associated is a *weight*. The neuron then calculates

$$v = 1 * w_0 + x_1 * w_1 + x_2 * w_2 + \cdots + x_n * w_n$$

Afterwards, an *activation function* $a(v)$ is applied on $v$. A perceptron is the simplest form of a *neural network*.

Let's consider a small example and let's try to simulate a very simple function, the logical AND. To make that more approachable, imagine you're a pretty bad chess player[1] and suppose there are two conditions. Depending on these conditions, the function tells you whether you will win the game or not. Condition one is that you're a queen up. Condition two is that your enemy is a pretty bad player. We will write 1 if a condition is true, and 0 otherwise. The function outputs 1 if you will win, and 0 otherwise. Table 2.1 lists all possible combinations. In other words, you'll only win if you are a queen up AND your enemy is pretty weak. In all other cases you'll screw up the game eventually and loose. Now consider the perceptron in Figure 2.3. It has three inputs, but the first input $x_0$ is always set to 1, as mentioned above. So variables $x_1$ and $x_2$ correspond to our inputs *queen up* and *enemy pretty weak*. Let's check whether the output of the perceptron actually calculates our function, i.e. when and if we will win the game:

$$x_1 = 0 \text{ and } x_2 = 0 : \quad v = -10 + 0 * 6 + 0 * 6 = -10$$
$$x_1 = 1 \text{ and } x_2 = 0 : \quad v = -10 + 1 * 6 + 0 * 6 = -4$$
$$x_1 = 0 \text{ and } x_2 = 1 : \quad v = -10 + 0 * 6 + 1 * 6 = -4$$
$$x_1 = 1 \text{ and } x_2 = 1 : \quad v = -10 + 1 * 6 + 1 * 6 = 2$$

---

[1]This example is directly inspired by the author's over-the-board experience and not meant to insult the reader.

Table 2.1: Logical AND

| Queen Up | Weak Enemy | Will Win |
|:---:|:---:|:---:|
| 0 | 0 | 0 |
| 0 | 1 | 0 |
| 1 | 0 | 0 |
| 1 | 1 | 1 |

Turns out this perceptron does not simulate our desired function. But it's close: Whenever we desire the output of the AND function to be 1, the output of our simulation is a positive value. On the other hand whenever we want the output of the AND function to be 0, the output of our simulation is strictly smaller than zero. So let's define our *activation function* to be a simple thresholding:

$$a(v) = \begin{cases} 0 & \text{if } v < 0 \\ 1 & \text{if } v \geq 0 \end{cases}$$

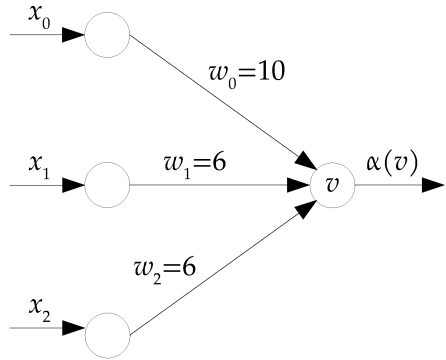

Figure 2.3: Perceptron simulating the logical AND.

Now we have

$$x_1 = 0 \text{ and } x_2 = 0 : \quad a(-10 + 0 * 6 + 0 * 6) = a(-10) = 0$$
$$x_1 = 1 \text{ and } x_2 = 0 : \quad a(-10 + 1 * 6 + 0 * 6) = a(-4) = 0$$
$$x_1 = 0 \text{ and } x_2 = 1 : \quad a(-10 + 0 * 6 + 1 * 6) = a(-4) = 0$$
$$x_1 = 1 \text{ and } x_2 = 1 : \quad a(-10 + 1 * 6 + 1 * 6) = a(2) = 1$$

Great – our perceptron successfully simulates the AND function! The reason that this simulation works and the perceptron computes AND obviously is rooted in the carefully chosen weights $w_0 = 10, w_1 = 6$ and $w_2 = 6$. But where do these come from? Turns out that we can automatically *learn* suitable weights by looking at a lot of examples, or short *samples*. The algorithm for this is called *back-propagation* based on *gradient descent*; and that's where all the magic of neural networks happens.

## 2.2 Back-Propagation and Gradient Descent

First, let's switch to another activation function. Hard thresholding is easy and fast to implement, but in a lot of scenarios a more granular approach is required.

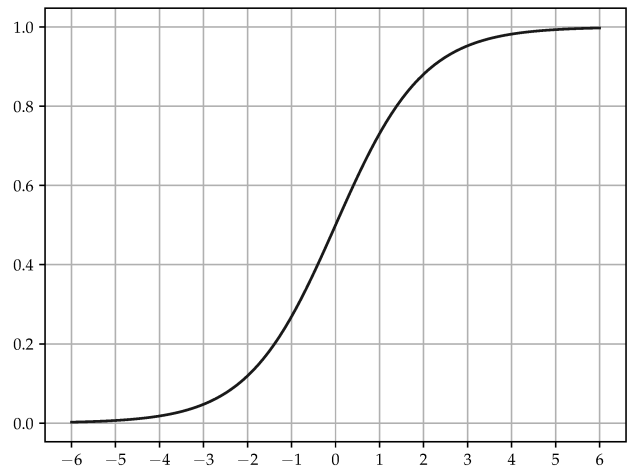

Figure 2.4: The sigmoid function $\text{sigmoid}(x) = \frac{1}{1+e^{-x}}$

Consider the aforementioned example of comparing two facial images. Do they depict the same person? If the network for example outputs a value of 0.001, this could indicate that it's rather unlikely that these two images show the same person. Or there is at least a comparatively high level of doubt. However, using hard thresholding as the activation function as we did for the logical AND would output a 1. This might be misleading.

Instead we'll use the sigmoid function, which is defined as

$$\text{sigmoid}(x) = \frac{1}{1 + e^{-x}}$$

and depicted in Figure 2.4. The important property of the sigmoid functions is that no matter what the input value is, the output value is always in the range 0...1. For the task of facial image comparison, imagine again a network output of 0.001. We have $\text{sigmoid}(0.001) \approx 0.5$. Since we have a range from 0 to 1 we could think of this as some kind of percentage or probability, i.e. it's 50 percent chance that this is a match. Then that's probably not good enough to call it a safe match and thus the result that we would expect. Back to the question of how to find suitable weights for a network. Let's go back to the AND-network of the previous section, and let's just initialize it randomly, say with weights $w_0 = 1$, $w_1 = 2$ and $w_2 = 3$, and evaluate the network performance for the two examples $(x_1 = 0, x_2 = 1)$ and $(x_1 = 1, x_2 = 1)$. These two examples should result in 0 and 1, respectively. We have:

$$x_1 = 1 \text{ and } x_2 = 0: \quad \text{sigmoid}(1 + 2 * 1 + 3 * 0) = \text{sigmoid}(3) \approx 0.95$$
$$x_1 = 1 \text{ and } x_2 = 1: \quad \text{sigmoid}(1 + 2 * 1 + 3 * 1) = \text{sigmoid}(6) \approx 0.99$$

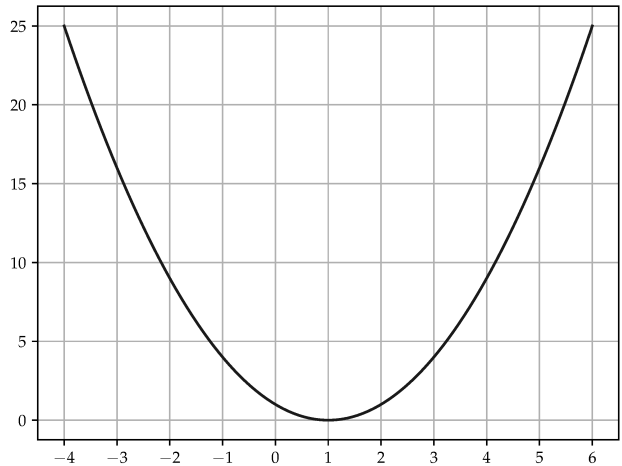

Figure 2.5: Square Error Function for expected outcome 1

That intuitively seems pretty bad: For $(x_1 = 1, x_2 = 1)$ we have 0.99 which is almost 1 and thus the result we want. For $(x_1 = 0, x_2 = 1)$ however we have 0.95, which is also almost 1, but the correct result is 0.

How can we objectively measure the current performance of the network w.r.t.

the chosen weights? An obvious way is to take the difference between the output of the network and the expected result. Let's write $\text{out}_i$ for the output of the network of example $i$, and let's write $r_i$ for the expected outcome of example $i$. Then one way to measure the error of the network is to take the square of the difference, i.e. $(\text{out}_i - r_i)^2$.

For the second example ($x_1 = 1, x_2 = 1$) we have an expected outcome $r_2 = 1$, i.e. given the output $\text{out}_2$ of the network, the error is $(\text{out}_2 - 1)^2$. This function is depicted in Figure 2.5.

As one can see, the square of differences puts emphasis on larger errors: A small differences of 0.1 between $\text{out}_i$ and $r_i$ results in an error value of 0.01, whereas a large difference of 2 results in an error value of 4.

Our network successfully simulates the intended function if the error of the network is *minimal*. Thus, given an example, we would like to *minimize* the error of the network. Ideally we'd have an error of 0. Taking a look at Figure 2.5 or at the expression $(\text{out}_2 - 1)^2$, it is not difficult to see that the error function has a (global) minimum for $\text{out}_2 = 1$.

Now, how do we find the minimum of a function? You might remember the following three steps from high-school:

1. Compute the first derivative and check where it zeros

2. Compute the second derivative and check whether you actually found a local minimum and not a maximum.

3. We now have one or more local minima. Now compare the function value for all those local minima identified in the first two steps in order to get the global minimum.

That's a valid approach, but difficult to implement in practice. Given a value $x_0$, a much easier approach – especially if we deal with very complex functions – is this: to find a (local) minimum of a function $f(x)$ we

- Compute the first derivative $\frac{d}{dx}f(x)$

- Evaluate the first derivative at $x_0$. If $\frac{d}{dx}f(x_0) > 0$, then $f(x)$ decreases if

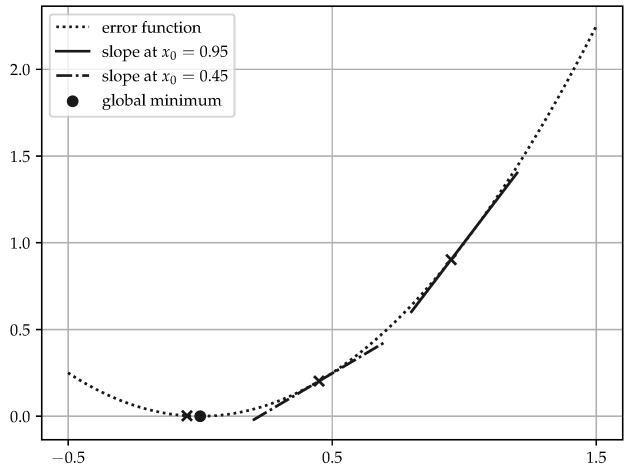

Figure 2.6: Minimizing the error w.r.t. expected outcome 1.

we decrease $x_0$. If $\frac{d}{dx}f(x_0) < 0$, then $f(x)$ decreases if we increase $x_0$. If $\frac{d}{dx}f(x_0) = 0$ or if $\frac{d}{dx}f(x_0) \approx 0$, we found a local minimum.

- Just *hope* that this local minimum is a global one.

This is illustrated in Figure 2.6. Let's consider the example $x_1 = 1$ and $x_2 = 0$. This has an expected outcome of 0, but the network outputs 0.95 which is our starting value for $x_0$. For $x_0 = 0.95$ we have an error of $(0.95 - 0)^2 = 0.9025$ — the rightmost point marked in Figure 2.6. Taking the derivative, we have $\frac{\partial x^2}{\partial x} = 2x$. But for that particular point the derivative provides the slope in form of a line. The idea is now to take a look at the slope and move into the direction that minimizes the error, i.e. here to the left. Numerically, we have $\frac{\partial x^2}{\partial x}(0.95) = 2 * 0.95 = 1.9$. Since the first derivative is positive at this point, we need to decrease the input argument to reach the minimum. Let's subtract some small value from $x_0$, say 0.5. Our new value is $x_0 = 0.95 - 0.5 = 0.45$; again marked in Figure 2.6. The error for this new value $x_0$ is $(0.45 - 0)^2 \approx 0.2$, so we already decreased the error significantly.

We can repeat this another time, again considering the slope at this point: $\frac{\partial x^2}{\partial x}(0.45) = 0.9$, i.e. the derivative is again positive, and we need to decrease $x_0$ to move in the direction of the minimum. Now we have $0.45 - 0.5 = -0.05$. This is the leftmost point marked in Figure 2.6. The error is then $(-0.05-0)^2 = 0.0025$. We slightly overshot the global minimum of $x_0 = 0$ a bit, but we are very near the minimum as the error is *almost* zero.

Now we have a simple method to compute the local minimum of the error function. But how does this relate to the network? Sure, we can minimize the error function, but the *input* to the error function is the *output* of the network, i.e. out$_i$ which we cannot directly change - we can only adjust the weights of the network! Also, we just took a look at the error function for the first example, which was $(\text{out}_1 - 0)^2$. For the second example the error function is slightly different with $(\text{out}_2 - 1)^2$.

To measure the error of the network w.r.t. $N$ examples, we can simply take the *average* of all individual errors, i.e. $\sum_{i=1}^{N} \frac{1}{\text{err}_i}$ to get the *total* error.

Next, to *minimize* the total error, let's take a look at the definition of the input of the error function, namely the output of the network. This is per definition:

$$\text{out}_i = \text{sigmoid}(1 * w_0 + x_1 * w_1 + x_2 * w_2)$$

In order to estimate how the error changes if we make slight changes to $w_1$, assume everything else in that function is constant, including $w_0$ and $w_2$. But that's precisely the *partial derivative* w.r.t. $w_1$. The partial derivative answers the question: Assuming everything is constant, should we slightly increase or decrease $w_1$ in order to lower the global overall error? And for $w_2$, the same question is answered by using the partial derivative w.r.t. $w_2$.

We now have a plan to *train* a neural network:

1. initialize the network with random weights

2. take a batch of examples, and compute the error of the network w.r.t. these examples

3. compute the *global error* by averaging over all individual errors

4. for all weights $w_i$, compute the *partial derivative* w.r.t. $w_i$

5. depending on the partial derivative, increase or decrease slightly each weight $w_i$

Let's put this into practice for our example. We will restrict ourselves to the non-bias weights $w_1$ and $w_2$. Using the definition of the net together with the activation and error functions, what we compute for our two examples is:

$$E_{\text{total}}(w_1, w_2) = \frac{1}{2} \sum_{i=1}^{2} \left( \frac{1}{1 + e^{-(1*w_0 + x_1^i * w_1 + x_2^i * w_2)}} - r_i \right)^2$$

and we want to compute $\frac{\partial E_{\text{total}}}{\partial w_1}$ and $\frac{\partial E_{\text{total}}}{\partial w_2}$. Note here that $E_{\text{total}}$ can be expressed as a *composition* of several functions. We leave out the function arguments on the left side for better readability:

$$E_{\text{total}} = \frac{1}{2} \left( (\text{out}_1 - r_1)^2 + (\text{out}_2 - r_2)^2 \right)$$

$$\text{out}_i = \frac{1}{1 + e^{-\text{net}_i}}$$

$$\text{net}_i = 1 * w_0 + x_1 * w_1 + x_2 * w_2$$

Now recall the *chain rule* for computing derivatives. We have

$$\frac{\partial E_{\text{total}}}{\partial w_1} = \frac{\partial E_{\text{total}}}{\partial \text{out}_1} * \frac{\partial \text{out}_1}{\partial \text{net}_1} * \frac{\partial \text{net}_1}{\partial w_1}, \quad \text{and} \quad \frac{\partial E_{\text{total}}}{\partial w_2} = \frac{\partial E_{\text{total}}}{\partial \text{out}_2} * \frac{\partial \text{out}_2}{\partial \text{net}_2} * \frac{\partial \text{net}_2}{\partial w_2}$$

Let's compute $\frac{\partial E_{\text{total}}}{\partial w_1}$ as an example:

$$\frac{\partial E_{\text{total}}}{\partial \text{out}_1} = \frac{\partial}{\partial \text{out}_1} \left( \frac{1}{2} \left( (\text{out}_1 - r_1)^2 + (\text{out}_2 - r_2)^2 \right) \right)$$

$$= \left( 2 * \frac{1}{2} (\text{out}_1 - r_1)^{2-1} \right) * 1$$

$$= \text{out}_1 - r_1 = 0.95 - 0 = 0.95$$

Note here that since we take the partial derivative w.r.t. $\text{out}_1$ we can consider $\text{out}_2$ as a constant whose derivative is zero. Also note that the 1 in the second line at the very right stems from the inner derivative of $\text{out}_1$.

Let's tackle the next derivative from the chain. The derivative of the sigmoid function is a tricky one, but note that $\frac{\partial \text{sigmoid}(x)}{\partial x} = \text{sigmoid}(x)(1 - \text{sigmoid}(x))$. Indeed, we have

$$\frac{\partial \text{out}_1}{\partial \text{net}_1} = \frac{\partial}{\partial \text{net}_1}\left(\frac{1}{1 + e^{-net_1}}\right) = \frac{e^{\text{net}_1}}{(1 + e^{net_1})^2}$$
$$= \text{out}_1 * (1 - \text{out}_1) = 0.95 * (1 - 0.95) \approx 0.048$$

Last, we have

$$\frac{\partial \text{net}_1}{\partial w_1} = \frac{\partial}{\partial w_1}(w_0 * 1 + w_1 * 1 + w_2 * 0) = w_1 = 2$$

Putting all this together we have $\frac{\partial E_{\text{total}}}{\partial w_1} = 0.95 * 0.048 * 2 = 0.0912$. The partial derivative is positive, so we should decrease $w_1$ a little bit.

By how much? That's a difficult question. If your remember Figure 2.6 you notice that if we take too large steps, we could overstep the minimum. If we take very small steps this should not happen, but it could take quite some iterations until we come close to the minimum. A general rule of thumb is to update a weight as $w_i' = w_i - \eta \frac{\partial E_{\text{total}}}{\partial w_1}$ where $\eta$ is just some heuristically chosen positive constant factor, called the *learning rate*. Note how the partial derivative takes care of the correct sign (e.g. here the derivative is positive, thus $\eta \frac{\partial E_{\text{total}}}{\partial w_1}$ is also positive, and we correctly subtract from $w_1$. Let's take $\eta = 10$, and we have $w_1 = 2 - 0.912 = 1.088$.

Then for $\frac{\partial E_{\text{total}}}{\partial w_2}$ we can do the same computation. You can verify yourself that we obtain $-0.01 * 0.0099 * 3 = -0.000297$. We have $w_2 = 3 + 0.0002978 = 3.000297$.

Using the network with the updated weights we get

$$x_1 = 1 \text{ and } x_2 = 0: \quad \text{sigmoid}(1 + 1.088 * 1 + 3.000297 * 0) \approx 0.75$$
$$x_1 = 1 \text{ and } x_2 = 1: \quad \text{sigmoid}(1 + 1.088 * 1 + 3.000297 * 1) \approx 0.99$$

As we can see, the network evaluates the example $x_1 = 1, x_2 = 1$ still almost at 1, but significantly reduced the output of the example of $x_1 = 1, x_2 = 0$ towards the desired output of 0. One can expect that after some more iterations the network will yield weights which correctly simulate the desired AND function.

Before putting this into practice, there is a small amount of terminology and observations that we need to make sense of. The algorithm is called *back propagation* by *gradient descent*. If we think of one calculation of the network from left to right as a forward pass, then after all, we are propagating the error *back* — i.e. from right to left — through the network to update the weights. Roughly, a gradient is just the vector of all partial derivatives, and this is precisely what we employ here for the update rule to *descent* towards a minimum of the error function.

Since we only updated two weights, we can visualize the error as a function of the two weights in a three-dimensional plot. That is precisely the function $E_{\text{total}}$ that describes the whole network, and depicted in Figure 2.7.

Gradient descent works by computing the current error w.r.t. the initial random weights $w_1$ and $w_2$. The error is a point located somewhere in the plot of Figure 2.7 "on top of the hill". It then computes the gradient in order to make a small step in a direction where the error slightly decreases. Making a lot of these small steps results in reaching the minimum.

In the example, we had actually three inputs, however the first one was set to a fixed value of one. The weight corresponding to that input creates a small *bias*. If we look at the network equation we notice that if all inputs are zero, the output of the network before applying the activation function is exactly the bias weight. That weight is still subject to change by gradient descent though. The role of a bias is just as the name implies — without caring for specific input values, it tunes the overall network to output values that minimize the error.

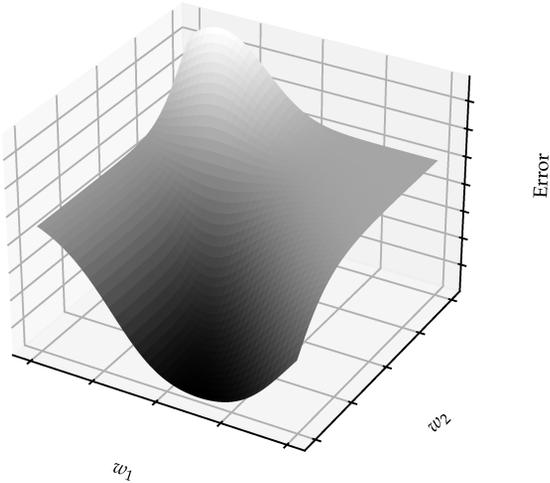

Figure 2.7: 3D plot of the error function.

In the example, we took two samples — namely $x_1 = 1, x_2 = 0$ and $x_1 = 1, x_2 = 1$ or shorter $[1, 0]$ and $[1, 1]$ —, computed the output of the network, computed the errors for each sample and averaged the error over these two samples. Then we applied gradient descent to update the weights of the network. Instead we could have also employed a different strategy: Just take one sample, compute the error and update the weights of the network immediately. Yet another possibility is to take *all* samples — here $[0, 0], [1, 0], [0, 1]$ and $[1, 1]$. The number of samples one takes to update the weights of the network once is referred to as the *batch size*.

In practice, one usually has lots, if not millions of samples. In such a case it is not efficient to apply all samples at once. It has also been observed that the network does not learn very well in such a case. On the other hand taking only one example is slow. Typical batch sizes are 32, or 64, 128, and 256, but in general it is difficult for a particular problem to choose the optimal batch size in advance.

In our example there are only four samples. Suppose we choose a batch size of 4. One step of gradient descent is not enough to significantly change the weights of the network. Instead, we have to repeat the four examples over and over for the network to reach a minimal error. We refer to one application of a batch as one *epoch*. In the example, we might need at least a few hundred number of epochs until the network learns the AND function. Of course this depends on both the initial randomly chosen weights as well as the learning rate.

We chose the *mean squared error* as the error function, but as mentioned there are other ways to quantify the error of the networks as well, e.g. by taking the absolute difference of the output and the expected result of a sample. In our example for the AND function, the output of the network was a single numerical value. If we do classification instead of regression, we might want to somehow measure the correctly versus incorrectly classified objects that we put in the network. We will see a suitable error function for that scenario in the next section. In general, machine learning frameworks refer to the value that is to be minimized during learning not as the error but as the *loss* of the network.

The activation function used here was the *sigmoid* function. But there is a plethora of other functions as well, each with its own mathematical and practical properties. Without diving into details, a popular current choice is a *rectifier*, or *rectified linear unit*, or short *ReLU*. It is defined simply as $f(x) = \max(0, x)$. Since ReLU's play quite an important role in modern neural networks, we will revisit them in a later section.

The learning rate $\eta = 10$ in our example was chosen ad-hoc. As can be observed in Figure 2.6, a too large learning rate can result in that we step over a minimum again and again and never reach it. On the other hand a low learning rate will result in only slight changes of the weights of the network, and thus progress is slow and learning takes forever. There are other ways to update the network. Quite efficient is an algorithm dubbed *Adam* [KB15]. Here the underlying idea is to employ individual learning rates for each weight depending on properties of the gradients. Adam usually converges much faster to the minimum than gradient descent, but can also fail to converge in cases where gradient descent eventually converges.

In general, choosing the best parameters to train a network effectively is an active area of research. Both from a mathematical standpoint in that researchers try to identify mathematical properties that make training the network more powerful and effective, but also from an engineering standpoint in that some parameters work better for certain classes of problems than others.

## 2.3 Classification vs Regression

In the previous section we discussed a problem of learning a specific value (the result of the AND function). Another problem that was stated before is: Given a person's income, her age, her job education, her place of living and probably other variables, what is the most likely amount of credit debt that she can repay? In such problems we want to learn to predict one specific *value*. Hence, a corresponding neural network has one specific output node. This kind of problem is referred to as *regression*.

What about other problems, such as: Given images of three known persons and one image with an unknown person, is the unknown person one of the three persons? Or is the person not one of these known persons?

A straight-forward way of encoding this as a *regression* problem would be: A numerical output value of 0 represents an unknown person, a value of 1 represents the first person, 2 the second, and 3 the third person. Train the network with the images of the known persons, and get a prediction for the image with the unknown person. Then, if the predicted value is close to say 1 with a value of 1.1, we conclude it is the first person. If the prediction is around 1.5 we could conclude it is either the first or the second person.

But what if the image shows a person looking quite close to both the first and third, but not the second person? A value of 2 would be the average value, but then we would falsely conclude the second person is shown.

Such problems are about *classification*: Given a choice among $n$ several categories, which one fits best? A much better network architecture is then to create an output layer with $n$ nodes. On each node the output value indicates how probable/good the category fits. For training, we extend the expected outcomes for samples to a vector. For example for the images of the (known) first person, an expected outcome is the vector $[1.0, 0.0, 0.0]$, i.e. the first output node should result in a 1.0 (= category first person fits a 100%), and the second and third output nodes should be 0.0 (= category second and third person do not fit). For training with images of the second and third person the expected outcome would be $[0.0, 1.0, 0.0]$ and $[0.0, 0.0, 1.0]$. A prediction of the network for an unknown image with values $[0.7, 0.1, 0.2]$ would then be: With a high confidence 70 percent this image shows the first person, it's very likely not the second person (10 percent), and the chance that it's the third person is about 20 percent.

If we want to interpret the output of the network as probabilities we need to make sure that the sum of all outputs equals 1, otherwise these numbers don't make a lot of sense. We can ensure this by applying the *softmax* function. Let $[o_1, \ldots, o_n]$ be the outputs of the network. The softmax function $\sigma$ is defined as

$$\sigma(\mathbf{o}) = \frac{e^{o_i}}{\sum_{j=1}^{n} o_j} \quad \text{for } i = 1, \ldots n$$

How do we measure the error, i.e. loss of our network? Softmax is often used together with negative *log likelihood*: We take the negated logarithm of the output probability w.r.t. the correct label.

Suppose we have a batch of two examples. One shows an image of the first person, and the output vector is $[0.7, 0.1, 0.2]$, and another shows the image of the second person with an output vector of $[0.6, 0.3, 0.1]$. We sum over the negated logarithms for the correct labels and divide by two, i.e.

$$\frac{-\log(0.7) - \log(0.3)}{2} \approx 0.78$$

This is our loss value. Now let's assume that we updated the network via gradient descent, and suppose the network now outputs some slightly more accurate classification for the second example, say $[0.3, 0.6, 0.1]$. We then have

$$\frac{-\log(0.7) - \log(0.6)}{2} \approx 0.43$$

Indeed, if the network becomes moor accurate, our loss decreases.

One thing to note here is that instead of decreasing the loss, we could also omit the negation part and use directly the logarithm. Now we do no longer seek to minimize the error, but to *maximize* the (log) probabilities. Gradient descent updates all weights of the network by subtracting a fraction of the gradient, i.e. partial derivatives. In this scenario we can simply add a fraction of the gradient and perform gradient *ascent*.

Yet another way to measure loss is to take the output of the network and apply *categorical cross entropy*. Assume again there are $n$ outputs $o_1, \ldots, o_n$. For each we have an expected outcome $e_1, \ldots, e_n$. Then we define as the loss

$$\text{loss} = -\sum_{i=1}^{n} e_i \cdot \log o_i$$

For the first example above we would obtain $1 \cdot \log(0.7) + 0 \cdot \log(0.1) + 0 \cdot \log(0.2) \approx -0.35$. Note that we might not always have binary values for the expected outcome. Depending on what you are training for, the expected outcome could be a vector of probabilities itself that sums up to one. Imagine for example probabilities for moves in a specific chess position. There could be more than one good move. An expected outcome might be $[0.5, 0.4, 0.1]$: Out of three possible moves, the first two are almost equally good moves, but the last one looses.

## 2.4 Putting it into Practice

Let's put the previous section into practice and create a small neural network to learn the AND function. This is depicted in Listing 2.1. Let's go quickly through the code:

**Listing 2.1: Neural Network for AND**

```
1  import numpy as np
2  import tensorflow as tf
3  from tensorflow import keras
4  import random
5
6  random.seed(42)
7  np.random.seed(42)
8  tf.random.set_seed(42)
9
10 x = []
11 y = []
12
13 x = [[0,0],[0,1],[1,0],[1,1]]
14 y = [ 0, 0, 0, 1]
15
16 model = tf.keras.models.Sequential()
17 model.add(tf.keras.Input(shape=(2,)))
18 model.add(tf.keras.layers.Dense(units=1, activation='sigmoid'))
19 print(model.summary())
20
21 model.compile(optimizer=keras.optimizers.SGD(learning_rate=0.1),
       loss=keras.losses.MeanSquaredError())
22
23 print(np.array(x).shape)
```

```
24
25 model.fit(np.array(x), np.array(y),batch_size=4, epochs=50000)
26
27 q = model.predict( np.array( [[0,1],[1,1]] )  )
28 print(q)
```

- In lines 1 to 4 we import several packages: `numpy` is a general numeric package for python, `tensorflow` is a high-performance, low-level neural network framework and `keras` is a high-level python interface for `tensorflow`.

- As mentioned, neural network weights are initialized randomly. In order for you to reproduce the results mentioned here, we also import `random` and seed all random number generators from python, numpy and tensorflow with a fixed value, so that we always generate the same random numbers.

- In lines 10 to 14 we create samples and expected outcomes. The array $x$ contains all four possible pairs of input values. The array $y$ contains the corresponding desired output. Note that only for input $[1, 1]$ we expect an output of 1.

- In lines 16 to 18 we define the neural network architecture. Considering Figure 2.3, we start from left to right, i.e. we first define the input layer. Our input layer has precisely two inputs, i.e. one for $x_1$ and one for $x_2$. Note that we do not count $x_0$, as it is always added automatically by *keras* itself.

- In line 18 we add the output layer. The output layer consists only of a single node (unit). We use the sigmoid function as the activation function.

- In line 21 we compile the model. For learning we use *stochastic gradient descent (SGD)*, which denotes the back-propagation algorithm by using gradient descent. The loss function is mean squared error.

- In line 25 we train our network. As a batch size we use 4 - after all, we only have four samples - and we repeat our batch 50000 times. On my computer this takes a few minutes to compute

- Finally, in line 27 we feed the network some samples and get the prediction of the network. Note that these could be potentially unknown examples. Here the network predicts approximately 0.033 for sample $[0, 1]$ and 0.96 for sample $[1, 1]$. That's quite near the expected outcomes of 0 and 1. Training our network succeeded!

As you can see we specify a lot of parameters when we define and compile our network. You can try different parameters and check the outcome yourself. For example, replace `keras.optimizers.SGD` by another optimizer such as `keras.optimizers.Adam`, change the learning rate, change the number of epochs, decrease the batch size, or change the loss function to the absolute error (`MeanAbsoluteError`). All these parameters have an effect how *fast* and how *good* the network trains.

You can also try to compute another simple binary function, the *exclusive or*, or XOR. This function is defined as: The output is 1, if the input arguments differ, and 0 otherwise. Namely

- 0 XOR 0 maps to 0

- 0 XOR 1 maps to 1

- 1 XOR 0 maps to 1

- 1 XOR 1 maps to 0

To train the network for XOR, we only have to change the array of expected outcomes by setting $y = [0, 1, 1, 0]$. I challenge you to find parameters on which the network succeeds in training. It seems to be impossible. But why?

## 2.5   Inherent Limits of a Single Perceptron

Let's have a deeper look. What does a perceptron actually calculate? According to our definition, we have

$$v = 1 * w_0 + x_1 * w_1 + x_2 * w_2 + \cdots + x_n * w_n$$

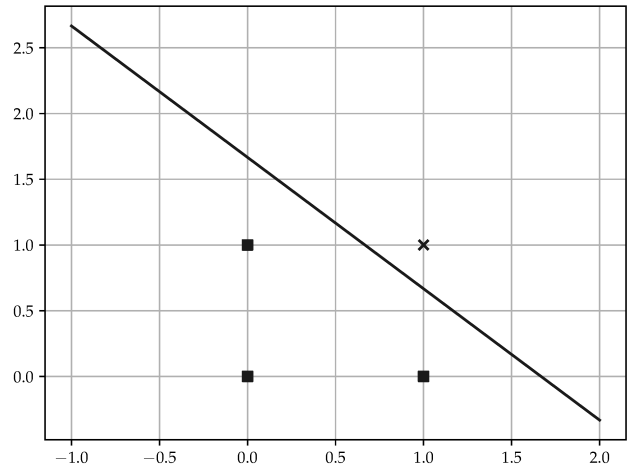

Figure 2.8: A simple line separates all points of the AND function.

and here with the weights to simulate the AND function

$$v = 1 * -10 + x_1 * 6 + x_2 * 6 = 6x_1 + 6x_2 - 10$$

If we take simple thresholding as the activation function $a(v)$, we then check if $6x_1 + 6x_2 - 10 \geq 0$ holds. Let's solve this for $x_2$, then we have $x_2 = -x_1 + \frac{10}{6}$. And that's a line – coincidentally the one from Figure 2.1!

Now let's also plot the AND function. We can plot the inputs $x_1$ and $x_2$ as points in a 2D coordinate system. And if the result of the AND function is 0, we mark the point as a black square, and if the AND function is 1, then we mark the point with an x. As can be seen in Figure 2.8, the line $x_2 = -x_1 + \frac{10}{6}$ separates these two kinds of points. And that's precisely how this perceptron is able to simulate the logical AND function — the perceptron computes a line, that separates the different kinds of points that make up the AND function. What is the limitation of a single perceptron then? Imagine the two categories would not be separable with a simple line. Consider for example Figure 2.9. No matter how you draw a line into that plot, it is impossible to separate the two kinds of points. In such case, a single perceptron will not do.

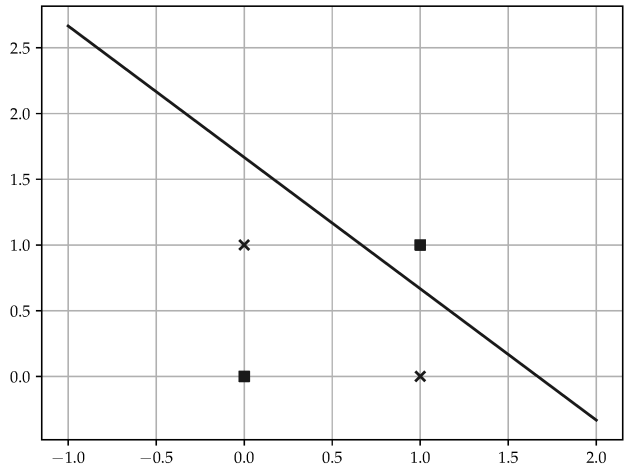

Figure 2.9: Impossible to separate all points of XOR by a line.

Luckily however we can chain several perceptrons to form more complex neural networks and thus simulate more difficult functions.

## 2.6   Multilayer Perceptrons

How do we add complexity to a network? We can simply add one or more several *layers* between the input and output layer. Each layer can have several *nodes*. This is depicted in Figure 2.10. Extending a forward computation through the network is simple. We just have to calculate lots and lots of values. Let's denote with an upper index the layer (counting from left to right), and with a lower index the node (from top to bottom). As before we write $w$ for the weights, and $x$ for the inputs to one layer. Here the input to the leftmost layer

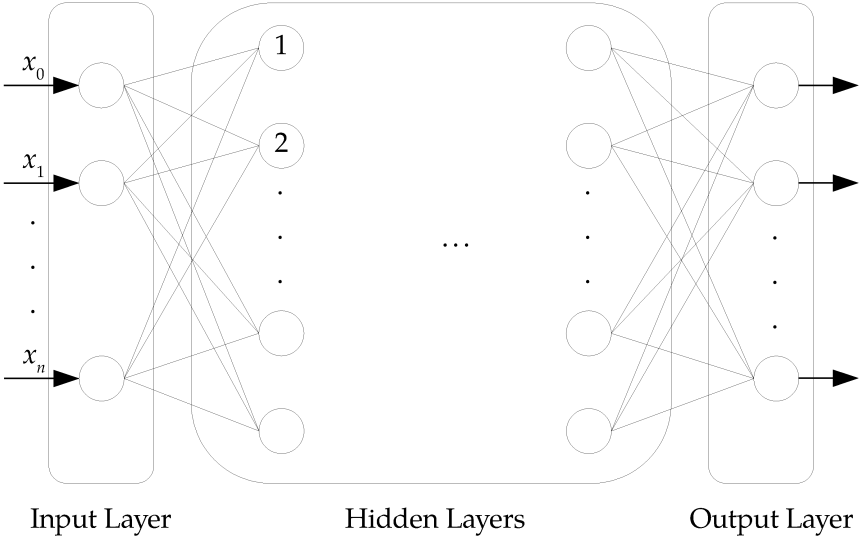

Figure 2.10: A multi-layer perceptron.

are just our training data, and otherwise it is the result of a computation of one node.

Consider for example the node marked 1 in Figure 2.10. Suppose we have $n$ nodes in the input layer. The value at 1 is then computed as

$$w_0^1 * x_0 + w_1^1 * x_1 + \cdots + w_n^1 * x_n$$

and similar for node 2 we have

$$w_0^2 * x_0 + w_1^2 * x_1 + \cdots + w_n^2 * x_n$$

Note that at each node we could also apply an activation function that modifies the sum that we just computed before passing it to the next layer. We'll skip writing this out, as we would soon be lost in indices!

We can also extend back-propagation w.r.t. multiple layers. The chain rule for the partial derivatives simply contains more chains, and again more computation has to be done, but the underlying principle is the same: We start from right to left, compute the partial derivative for each node of the rightmost layer, then compute the partial derivative for each node of the layer before, and continue until we reach the leftmost input layer.

Let's use our newly gained knowledge and revisit the XOR function. We just need to replace a few lines of our simple perceptron and insert another layer inbetween with lots of (8) nodes. We simply change the lines of Listing 2.2 to the ones shown in Listing 2.3.

**Listing 2.2: Neural Network for AND**

```
1 model = tf.keras.models.Sequential()
2 model.add(tf.keras.Input(shape=(2,)))
3 model.add(tf.keras.layers.Dense(units=1, activation='sigmoid'))
```

**Listing 2.3: Neural Network for XOR**

```
1 model = tf.keras.models.Sequential()
2 model.add(tf.keras.Input(shape=(2,)))
3 model.add(tf.keras.layers.Dense(units=8, activation='sigmoid'))
4 model.add(tf.keras.layers.Dense(units=1, activation='sigmoid'))
```

Success! After a few minutes of training, the network predicts 0.97 for input [0, 1] and 0.02 for input 1, 1.

The term *deep learning* was coined by using *deep* networks — that is, having a lot of layers inbetween input and output. They are also called *hidden* layers, since we can not directly observe the inputs to and outputs from these layers — our training data consists only of inputs to the first layer, and we compare the expected outcome with the prediction of the network that we get from the output of the last layer. It is precisely the addition of many hidden layers that makes deep learning so powerful. Of course this comes at a cost, namely the required computational effort.

Before we continue, lets have a look at another function. Let's try to create a network that learns the function $f(x) = x^2$. The source code is shown in Listing 2.4.

**Listing 2.4: Neural Network for $x^2$**

```
1  import numpy as np
2  import tensorflow as tf
3  from tensorflow import keras
4  import matplotlib.pyplot as plt
5  import random
6
7  random.seed(42)
8  np.random.seed(42)
9  tf.random.set_seed(42)
10
11 x = []
12 y = []
13 for i in range(0,10000):
14     xi = random.randint(0,5000)
15     xi = xi/100
16     if(xi==2.0 or xi==4.0):
17         continue
18     yi = xi**2
19     x.append([xi])
20     y.append(yi)
21
22 print(np.array(x)[0:20])
```

```
23
24 model = tf.keras.models.Sequential()
25 model.add(tf.keras.Input(shape=(1,)))
26 model.add(tf.keras.layers.Dense(units=8, activation="relu"))
27 model.add(tf.keras.layers.Dense(units=8, activation="relu"))
28 model.add(tf.keras.layers.Dense(units=1, activation="linear"))
29 print(model.summary())
30
31 model.compile(optimizer=keras.optimizers.Adam(),
32               loss=keras.losses.MeanSquaredError())
33
34
35 print(np.array(x).shape)
36
37 model.fit(np.array(x), np.array(y),
38           batch_size=256, epochs=5000)
39
40
41 q = model.predict( np.array( [[2],[4]] )  )
42 print(q)
43
44 original = [ x**2 for x in range(0,50)]
45
46 plt.figure()
47 plt.plot(original)
48 plt.show()
49
50 predicted = model.predict(np.array([ x for x in range(0,50) ]))
51
52 plt.figure()
53 plt.plot(predicted)
54 plt.show()
```

The network definitions should be clear by now, but there are a few differences in the source code compared to the other examples:

- In lines 13 to 20 we create some input values in the range 0...50 and add them to the array $x$. We put the corresponding squared value in the array $y$. We explicitly avoid to add values 2 and 4 to the list of training data, because we want to evaluate how well the network predicts unknown values. We of course expect predictions near 4 and 16 respectively.

- In line 41 we compute the prediction of the network for the input values 2 and 4. We get prediction values around 3.2 and 23 which is close enough.

- In lines 42 to 54 we plot the original function in the range 0...50 and we also plot the predictions of the network in that range. The plot essentially shows the function $y = x^2$. Apparently, our network works quite well.

All in all, we succeed in training a network to compute the function $f(x) = x^2$. Let's investigate some more. Try to change the lines of Listing 2.5 to the ones of Listing 2.6.

**Listing 2.5: Out of Domain Predictions**

```
1 original = [ x**2 for x in range(0,50)]
2 ...
3 predicted = model.predict(np.array([ x for x in range(0,50) ]))
```

**Listing 2.6: Out of Domain Predictions**

```
1 original = [ x**2 for x in range(50,100)]
2 ...
3 predicted = model.predict(np.array([ x for x in range(50,100) ]))
```

In other words, let's investigate how well our network predicts the completely different range of input values 50...100 for which there was no training data whatsoever — all our samples were in the range 0...50. One might expect that the networks performs well on all ranges of our function $f(x) = x^2$. But as can be seen in the generated plot, this is not the case. In general, networks perform quite bad if they are fed with data that is completely out of their domain.

More intuitively spoken: Suppose you create a network that identifies chess pieces in an image, and train it with pictures of chess pieces. You would not expect it to be good at identifying Shogi pieces. It does not mean that the network cannot recognize chess pieces from images that it has not seen during training — after all, our network was good at predicting input values 2 and 4

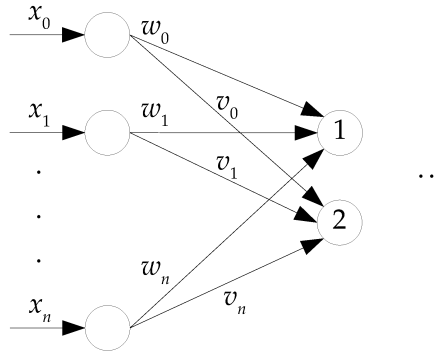

Figure 2.11: Network Vectorization.

— but it simply means that it is important to have a large enough number of samples that cover the *whole domain* of values that you want to predict later.

## 2.7   Vectorization and GPUs

The many calculations to train a neural network require significant amounts of computational power. Why are *graphical processing units* or *tensor processing units (TPUs)* — the chips developed by Google — so much better than general computer CPUs for training neural networks?

We will not go into details of GPU programming here — after all, frameworks such as `tensorflow` fortunately abstract away such nifty details — but only give an intuition.

Consider the network in Figure 2.11, and let's just focus on how to compute the inputs of the first hidden layer. We have at node 1 and 2

$$w_0 x_0 + w_1 * x_1 + \cdots + w_n * x_n$$
$$v_0 x_0 + v_1 * x_1 + \cdots + v_n * x_n$$

We can rewrite both the inputs $x_1$ and $x_2$ and all the weights of the input layer

as matrices

$$\mathbf{x} = \begin{bmatrix} x_0 \\ \vdots \\ x_n \end{bmatrix} \text{ and } \mathbf{w} = \begin{bmatrix} w_0 & w_1 & \cdots & w_n \\ v_0 & v_1 & \cdots & v_n \end{bmatrix},$$

If we then compute the matrix product **wx** we have

$$\mathbf{wx} = \begin{bmatrix} w_0 * x_0 + w_1 * x_1 + \cdots + w_n * x_n \\ v_0 * x_0 + v_1 * x_1 + \cdots + v_n * x_n \end{bmatrix}, \tag{2.1}$$

and this is precisely the input of the hidden layer! In other words: Computing forward passes of a neural network can be reformulated into computations over (potentially large) matrices. This is also known as *vectorizing* a computational problem.

Note that each entry of the matrix in Equation 2.1 is completely independent of each other. That's why it is easy to compute each entry of the result matrix in parallel. CPUs are intended as general processing units, i.e. no matter what problem is given, they almost always perform with decent speed. But for computer graphics, we mostly need fast matrix operations, and that's what GPUs are highly optimized for – at the cost of not being very universal computing machines. This is why having a fast GPU will help you train networks much faster, often in the dimension of a two-digit factor.

Needless to says that back-propagation can also be reformulated in terms of matrix operations. So if you really want to program and experiment with neural networks, a fast graphic card is a must. Even better is of course working at a company that is capable and willing to design and produce their own dedicated processors just for accelerating neural network training.

## 2.8 Convolutional Layers

Let's again reconsider the problem of image recognition. As mentioned, suppose you are given a facial image that is digitally stored on a passport and want to compare it to a live image taken during border control. Do these show the same person?

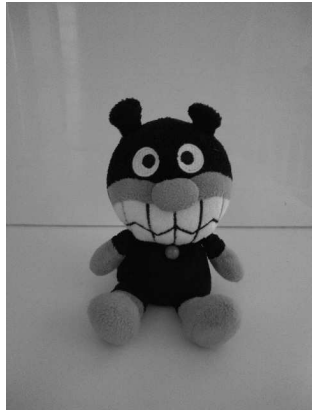

Figure 2.12: Baikinman. Leader of the Viruses.

How would one intuitively approach such a problem? A natural idea is to look for unique *features* of the person's face like the shape of the head, eye, nose or ears, or the person's eye and hair color. Maybe there are also birth marks, slight deviations in skin color, wrinkles... the list goes on and on. The next step would then to come up with a list of most important features, and try to write computer programs that extract these information from the image.

This is exactly what researches tried to do in the early days of image processing and face recognition. Let's start with a small example of edge detection. After all, when looking for the outline of the head, we need to separate it from the background of the image.

Consider Figure 2.12. The stuffed animal in the grayscale image is *Baikinman*, the evil villain from the Japanese kid's show *Anpanman*. Let's try to construct a filter that after processing the image with that filter gives us the shape of Baikinman. Remember that a grayscale image is nothing else that a two-dimensional matrix of values, usually in the range 0...255. We can use an edge detection filter that is given by the matrix

$$\begin{bmatrix} 1 & 0 & -1 \\ 1 & 0 & -1 \\ 1 & 0 & -1 \end{bmatrix}$$

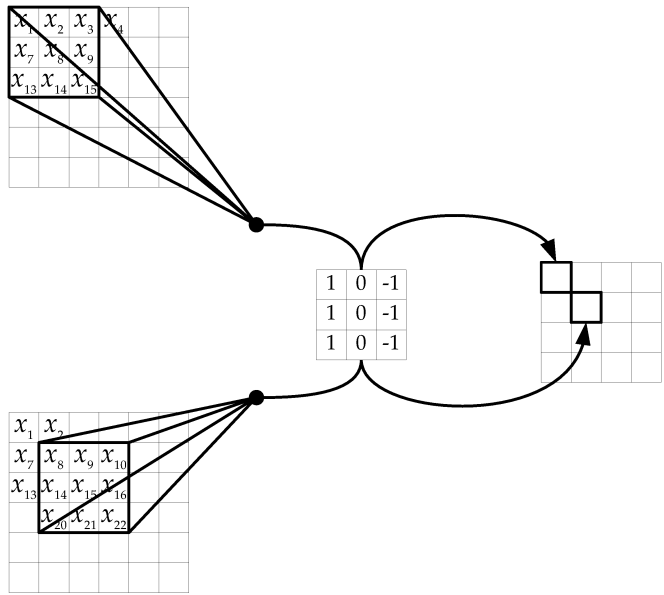

Figure 2.13: Applying a convolutional filter.

We apply this filter by iterating over the image, each time taking a $3x3$ sub-image. Then we multiply each element from the sub-image with the corresponding value from the filter matrix, and sum up the results. This sum is then written back as a value in the filtered image.

This is depicted in Figure 2.13. Our source image is $6x6$ pixels, and we start in the upper left corner, taking the sub-image given by the pixel values

$$\begin{bmatrix} x_1 & x_2 & x_3 \\ x_7 & x_8 & x_9 \\ x_{13} & x_{14} & x_{15} \end{bmatrix}$$

Then we multiply each element and sum this up as

$$1x_1 + 0x_2 - 1x_3 + 1x_7 + 0x_8 - 1x_9 + 1x_{13} + 0x_{14} - 1x_{15} \qquad (2.2)$$

and this makes up the upper left pixel value of our filtered image. We continue, as illustrated in Figure 2.13, ending up with a slightly smaller image of size $4x4$. If we want to, we can further filter the image for better illustration by setting each pixel value that is above a certain threshold to 0 and below to 255, thereby creating a strictly black/white image.

It's straight-forward to implement this in Python, and the source-code is shown in Listing 2.7.

**Listing 2.7: Edge Detection for Baikinman**

```
1  import imageio
2  import numpy as np
3  from skimage import img_as_ubyte
4
5  im = imageio.imread('baikinman.jpg')
6
7  yy = im.shape[0]
8  xx = im.shape[1]
9
10 filter1 = [ [1, 1, 1],
11            [0, 0, 0],
12            [-1, -1, -1]]
13
14 sobel = [ [1, 2, 1],
15            [0, 0, 0],
16            [-1, -2, -1]]
17
18 filtered_image = []
19 for y in range(0,yy-3):
20     row = []
21     for x in range(0,xx-3):
22         val = 0.0
23         for i in range(0,3):
24             for j in range(0,3):
25                 val += im[y+i, x+j] * filter1[i][j]
26         row.append(val)
27     filtered_image.append(row)
28
29 img1 = np.array(filtered_image)
30 imageio.imwrite("baikinman_filter_no_thresh.png", img1)
```

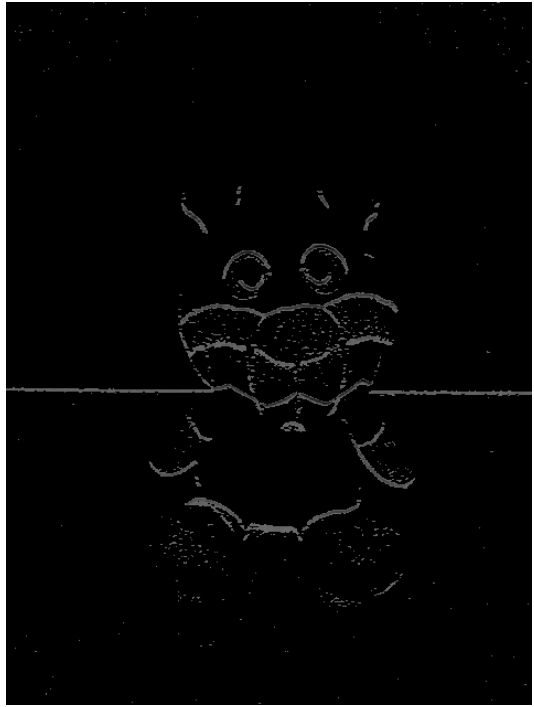

Figure 2.14: Baikinman, applied filter and thresholding.

First we read the image using the `imageio` library, then loop through the image, and write the result back to a file. Thresholding is here omitted for brevity.

Figure 2.14 shows the filtered and thresholded image of Baikinman. It seems that after all, Baikinman's unique features are his scary teeth!

Can we do better? Well there is another filter based on the so-called *Sobel* operator. The difference is a subtle change in the filter matrix, here we use

$$\begin{bmatrix} 1 & 0 & -1 \\ 2 & 0 & -2 \\ 1 & 0 & -1 \end{bmatrix}$$

How did researchers initially came up with these values? Actually, I have no clue. Probably by analysing the pixel values of example images and then analysing how a sharp threshold looks like and then finally coming up with numbers for the filter matrix.

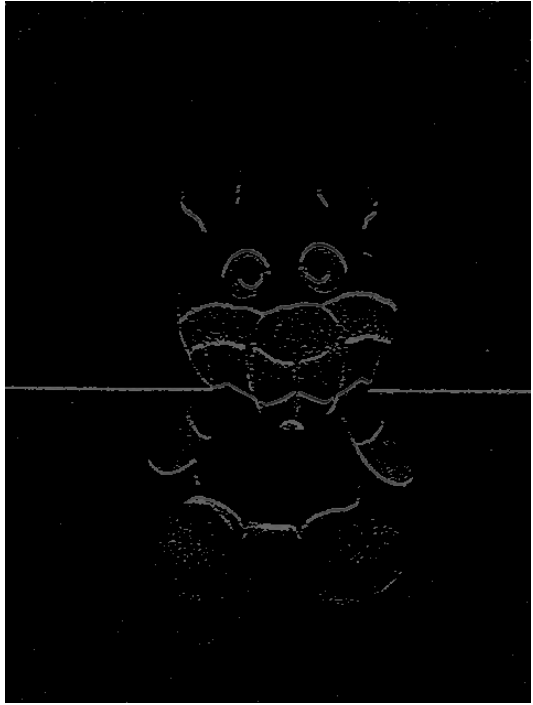

Figure 2.15: Baikinman, applied sobel operator and thresholding.

So, which one is better? Actually, it's hard to see (cf. Figure 2.15) and personally I see almost no difference here. Selecting a suitable filter also certainly depends on what we want to do with the filtered image (here: face recognition). Can we come up with better values?

You probably already figured where I am going at. Equation (2.2) looks suspiciously close to the computation behind one layer of a neural network. In other words: We can simply translate this filter operation — *convolution* — to a neural network, and let the network figure out suitable weights — the entries of the filter matrix — by itself using gradient descent and backpropagation. The translation of this filter operation to a *convolutional layer* is illustrated in Figure 2.16. We take each pixel as a potential input value.

Then we just wire the nine values of each sub-image and connect them to an output value. Each arrow here has one weight associated which corresponds to one entry of the above mentioned filter matrix. The associated weights then make up the actual filter. Or rather, the filter is designed specifically for the particular network, namely by training the network and thus updating the weights. Next we do this for all other combinations of pixel values. Note that we reuse the same weights thereby significantly reducing their total number, as illustrated for example for weight $w_8$ — here we apply *one* filter to *all* sub-images. We can of course always increase the number of filters.

There are four natural extensions that we will discuss just very briefly:

1. First, we can extend a filter to handle color images. We simply apply it to either all or some of the red, green and blue color channels and extend the summation by applying it to the channels as well.

2. Second, we can of course apply (train) several filters in parallel within one layer. We just have to add another filter matrix, and output nodes at the output layer, and wire the weights that represent the filter matrix accordingly.

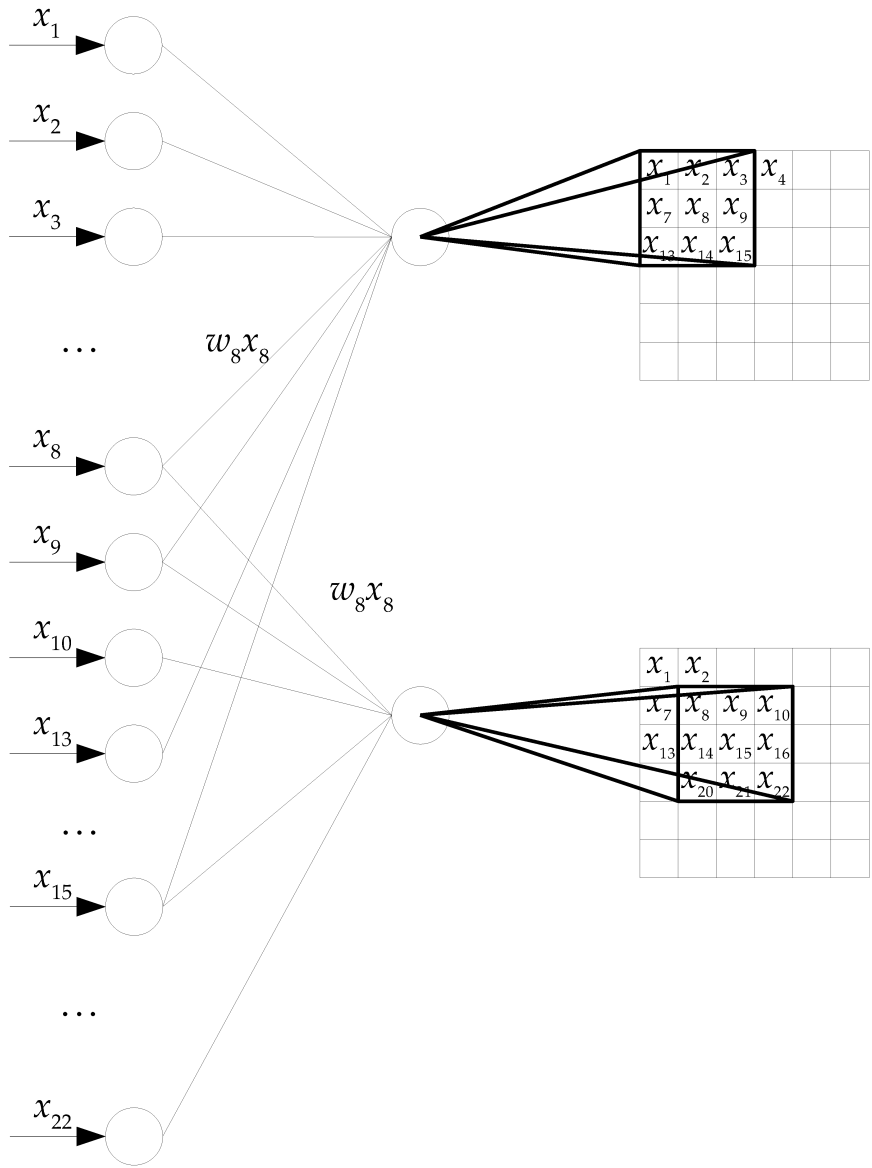

Figure 2.16: Convolution Filter encoded as Neural Network.

3. Similar to a standard perceptron, we can add a small *bias* value to the result of the summation, and apply a (non-linear) activation function like sigmoid afterwards.

4. When iterating over the source image, we stepped through the image one pixel at a time. Of course we can use larger steps and consider only i.e. every second, every third or every tenth pixel, and simply not consider the pixels in between. Of course this comes as the cost of precision, but can be much faster. The step-size is referred to as *stride*.

We will skip the specific details of these extensions though. What is rather interesting though is to visualize what the learned filter actually does. Let's consider a small example of the VGG16 model. This is a neural network for image recognition with pre-trained weights released by Visual Geometry Group (VGG), at the Department of Engineering Science, University of Oxford under the Creative Commons Attribution 4.0 International Public License (CC BY 4.0). It is one of the examples delivered with the Keras framework, and ideally suited to play around with and investigate a complex network without having to actually train it yourself. We can now visualize the output of the first convolutional layer. Let's first have a look on the numerical values of the filters with Listing 2.8.

**Listing 2.8: Visualize Convolution Filter**

```
from tensorflow.keras.applications.vgg16 import VGG16
model = VGG16()
print(model.summary())
filt, bias = model.layers[1].get_weights()
print(filt.shape)
print(filt[:,:,0,0])
```

In that listing, we first load the network including its weights, and then print a summary of the model. You can verify that there is indeed an input layer followed by the first convolutional layer. We can use the Keras function get_weights() to get the actual weights of that layer, which are equivalent to the filter values. The array of filter values has shape $(3, 3, 3, 64)$, i.e. we have 64 filters of dimension $(3, 3)$ — like the filter for the edge detector —, and we have three of those for each color channel, namely red, green and blue. The last line gives the numerical values of the first filter for the first color channel as[2]

$$\begin{bmatrix} 0.42 & 0.37 & -0.06 \\ 0.27 & 0.03 & -0.36 \\ -0.05 & -0.26 & -0.35 \end{bmatrix}$$

It's even more fun to see what the convolutional layer does to Baikinman. We can visualize this by creating a new neural network, which consists of only the input layer and the first convolutional layer of the original network. The output of the network is then the output of the convolutional layer, i.e. the filters applied to our input image of Baikinman.

This is shown in Listing 2.9.

**Listing 2.9: Visualize Several Convolution Filters**

```
1  from  tensorflow.keras.applications.vgg16 import VGG16
2  from  tensorflow.keras.applications.vgg16 import preprocess_input
3  from  tensorflow.keras.preprocessing.image import load_img
4  from  tensorflow.keras.preprocessing.image import img_to_array
5  from  tensorflow.keras.models import Model
6  from numpy import expand_dims
7  import  matplotlib.pyplot as plt
8
9  model = VGG16()
10 model = Model(inputs=model.inputs, outputs=model.layers[1].output)
11 model.summary()
12
13 image = load_img('baikinman.jpg', target_size=(224, 224))
14
15 image = img_to_array(image)
16 image = expand_dims(image, axis=0)
17 image = preprocess_input(image)
18
19 filtered_image = model.predict(image)
20
21 filter_index = 3
22
```

[2]This was at the time of writing. If the model is re-trained at any stage, your values might differ

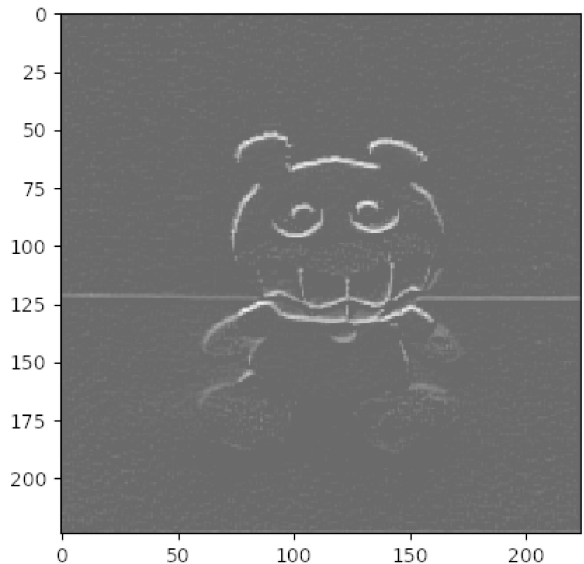

Figure 2.17: The third filter matrix applied to the image of Baikinman.

```
23 plt.figure()
24 plt.imshow(filtered_image[0, :, :, filter_index])
25 plt.show()
```

First we load the model, and then immediately overwrite the model by creating a new one by re-using the input layer and first convolutional layer of the old network. Next we load the image of Baikinman, and resize it slightly — the input images of VGG16 are all of dimension (224, 224). Next we have to do some pre-processing of our image to make sure it has a similar form as those images with which the network was trained. Last we predict the output of the network, i.e. apply the convolutional layer. Next we select the third filter and the first color channel, and plot the result, which is depicted in Figure 2.17. It looks as if this is an edge detector for Bakinman's face. Maybe this one helps to detect face images? Who knows!

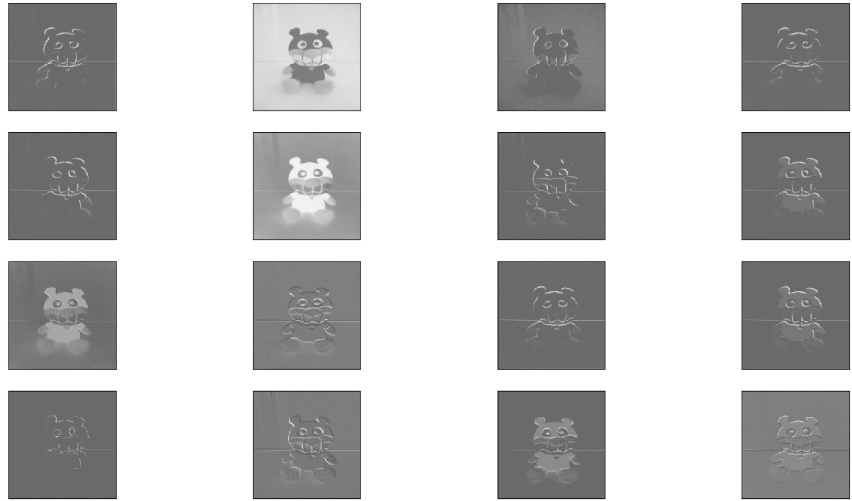

Figure 2.18: First 16 filters applied to the image of Baikinman.

It's even more fun to plot more images of these filters. The first 16 of all 64 are shown in Figure 2.18 Again, it's hard to figure out *what* these filters actually do. But it looks both interesting and beautiful in its own way! That is by the way one of the drawbacks of deep neural networks — we can evaluate and understand that they work, but in general it is difficult to understand *why* and *how* they work.

You're probably already wondering what all this has to do with chess. The other network structures we discussed so far were very technical and geared towards optimizing the network behaviour as a whole. Convolutional layers seem to be geared very much towards image recognition.

And this is exactly the field which they were developed and invented for. However it is not that hard to imagine that they are beneficial for chess, too. Imagine a network that, given a chess position, assess this position.

How would a Grandmaster assess a chess position? Of course she would look at tactics in a position, like winning a pawn or a piece. But of great importance is also the (visual) structure. Are there open files? Where are the knights placed? Do the bishops have open lines, or are they blocked by pawns? Entire books haven been written about pawn structures [Sha18]. While it of course does not make sense to feed directly literal *images* of chess positions into a network, the logical 8x8 representation of a chessboard has already an image-like structure. Thus we can hope that a convolutional layer is suitable to learn filters that extract abstract, structural information about a position. It is precisely this pattern recognition task that humans are very good at, but which are very hard to formulate as a small set of objectively computable rules. We will see later when discussing specific network structures of successful neural network approaches that convolutional layers are a crucial tool to tackle this pattern recognition task.

## 2.9 Squeeze-and-Excitation Networks

We only briefly mentioned how to extend convolutional layers to handle different channels. When we think of images, this is naturally color channels. In other machine learning contexts a channel might be something completely different. For example in chess, we might encode a chess position by creating one image of all white rooks, one image of all white knights, and continue like that for all white and all black pieces. One channel encodes here the position of one piece type of one player. But let's stick to color images for now, as that's probably more intuitive.

We noted how an extension of convolutional layers can be done for multiple (color channels), namely apply a filter to either all or some of the red, green and blue color channels and extend the summation accordingly.

But this also means that we weigh each channel equally. Now imagine we are trying to create a network that is able to predict whether a color image contains an image of Baikinman or not. Baikinman has very distinct colors: A dark blue, almost completely black body, bright white teeth, purple hands and feet, and some parts of his face are purple, too. It is not difficult to imagine that some color channels are more important than other color channels. Therefore it might make sense to weight each channel differently when computing the overall output (i.e. the sum) of the convolutional layer.

However we can even go one step further. There could be interdependencies between the channels. Consider the purple hands and feet of Baikinman. If we have a color image encoded by cyan, magenta and yellow color channels, the purple is encoded with a high value of cyan, a high value of magenta and a medium/low value of yellow. Then for a filter that detects the hands and feet of Baikinman, there is clearly an interdependency between the color channels.

Squeeze and Excitation Networks are motivated by this idea. Quite a recent improvement suggested by Hu et al. [HSA+20], they are built upon convolutional or residual layers by explicitly modelling interdependencies between channels. Squeeze and Excitation Networks improved image recognition tasks on certain benchmarks by 25 percent. At the time of writing they are not yet available as standardized building blocks in typical deep learning frameworks such as tensorflow, but they are also not too difficult to construct either.

From a broader perspective, Squeeze and Excitation networks are more of an evolution than a revolution in deep learning. A significant one nonetheless, but still more of an evolution. It therefore makes sense to skip the mathematical details of constructing Squeeze and Excitation networks and instead just understand it as a slightly advanced version of convolutional or residual layers.

## 2.10   Fully Connected Layers

A fully connected layer is just as what the name implies. Every neuron of one layer is connected to all neurons of the next layer. Such a structure is often found right before the output. For example in Figure 2.10, if the output layer is used for some classification task, and each node represents some possible category, we want to make sure that we use all the information that the network has computed so far to produce the final results. Therefore — as depicted — the rightmost hidden-layer connects every node with every node from the output layer.

## 2.11 Batch normalization and Rectified Linear Units

Let's start this section by quoting a quite recent (2018) research paper by Santurkar et al. [STIM18].

> Batch Normalization (BatchNorm) is a widely adopted technique that enables faster and more stable training of deep neural networks (DNNs). Despite its pervasiveness, the exact reasons for Batch-Norm's effectiveness are still poorly understood.

Indeed, whereas in the previous chapters there were small examples that illustrate the underlying principles and even allow to calculate things by hand, for batch normalization it is hard to give a convincing example. Instead we'll only show the general motivating idea, and skip the mathematical details.

This difficulty of illustrating actually holds true for most of the remaining network layers presented in this chapter. These layers, often introduced very recently — Batch normalization was proposed for example in 2015 by Ioffe and Szegedy [IS15] — represent the engineering advances and refinements that made the general concept work out in practice. Fortunately there are only few layers left to cover.

We'll step back a moment and discuss first *input normalization*. Let's go back to our example on predicting credit worthiness, i.e. taken a bank's perspective, how much money should we at most lend to someone based on some input parameters. Suppose now, the input parameters are age and monthly income. Suppose we have only three training examples, namely Jack, Jill and Hipp. We investigated them thoroughly and know exactly what maximum credit score in the range of 0...1 they should be given. Their data are depicted in Table 2.2. Are these data realistic? Probably not, but let's still stick with it. Suppose we have the same network structure as in our AND example. We initialize the network with random weights $w_0 = 1$, $w_1 = 2$ and $w_2 = 3$. If we input the training data of Jack, the network outputs

$$\text{sigmoid}(1*1 + 26*2 + 2500*3) = \text{sigmoid}(7553) \approx 1$$

Table 2.2: Credit Prediction, Training Data

| Name | Age | Income (Month) | Max. Credit |
|------|-----|----------------|-------------|
| Jack | 26 | 2,500 | 0.4 |
| Jill | 37 | 3,000 | 0.5 |
| Hipp | 65 | 26,000 | 0.9 |

We immediately notice that the output value is completely dominated by the salary. It's as if we almost disregard the age! Let's continue with just the training data of Jack and look how the network learns from this example. We have $\frac{\partial E_{\text{total}}}{\partial \text{out}_1} = 1 - 0.4 = 0.6$, and $\frac{\partial \text{out}_1}{\partial \text{net}_1} = 0.6 * (1 - 0.6) = 0.24$. We could continue to calculate the update of the weights, but let skip this part. Instead, how would the next learning step from this example look like after updating the weights slightly? Well, unless the weight $w_2$ changes dramatically, we will have again the sigmoid over some very large number, resulting in almost 1 for the output of the net. Nothing will change for quite some iterations. We summarize

- the network *output* will be dominated by the salary

- the network *update* will be dominated by the salary

- the network will learn very slowly

We can mitigate this effect by *normalizing* the inputs of the network to a standardized scale.

Consider for example the *age* in the example. We first compute the mean age as $\frac{26+37+65}{3} = \frac{128}{3} \approx 42.66$. The variance is given by

$$\frac{1}{3}[(26 - 42.66)^2 + (37 - 42.66)^2 + (65 - 42.66)^2] = 269.55 \approx 16.41^2$$

Next we normalize the ages by subtracting the mean and dividing through the standard deviation, which is defined as the square-root of the variance. We get for Jack $\frac{26-42.66}{16.41} \approx -1.015$; for Jill we have $\frac{37-42.66}{16.41} \approx -0.34$, and finally for Hipp we get $\frac{65-42.66}{16.41} \approx 1.36$.

When we have a very deep network, there will be several layer with activation functions before the network gives an output. The above described effect may not only occurs at the input level, but also at all the immediate values. That is, if the input to some immediate layer has a very uneven range, this might affect the performance of the network.

The underlying idea of *batch normalization* is to create a network layer that normalizes intermediate data. Such batch normalization layers can then be inserted between other layers of the network.

Usually, batch normalization is applied after computing the linear combination of the outputs but before applying the activation function of a layer. Suppose **z** is such an immediate layer, and that there are $m$ examples in the current batch that we want to learn. Then we compute the mean $\mu = \frac{1}{m} \sum_{i=1}^{m} z_i$ and standard deviation $\sigma^2 = \frac{1}{m} \sum_{i=1}^{m} (z_i - \mu)^2$. Given input $z_i$ we then normalize it by $z_i' = \frac{z_i - \mu}{\sigma}$. This is all similar as for input normalization.

However after applying this normalization step, the mean is exactly zero, and the variance is exactly one. That might not be true for this specific layer, e.g. it could be that at this particular layer the values that come in have a different distribution. What we want to make sure by batch normalization is that we also learn how these values are distributed. This is especially true since we learn in batches, and looking at *all* training data, the values might be very different than looking at the current batch. Therefore one usually adds parameters $\alpha$ and $\beta$, and computes $z_i'' = \alpha z_i' + \beta$, which is then the final output after batch normalization that is fed to the activation function.

Again, we will skip the mathematical details on how to integrate this with back-propagation. After all, there is no need to implement this manually. In Deep-Learning Frameworks, inserting a batch normalization layer is just one line of code during network creation.

There is yet another way to look at this example. The slope of the sigmoid function is such that for very large values, the slope (i.e. the derivative or gradient) is almost zero. If we have a very deep network, we have a very long chain of derivatives due to the chain rule. If each step in the chain creates a derivative that is a small value almost near zero, and we multiply all these small values, the resulting value will be so small, that it is very difficult to numerically distinguish it from zero. The gradient *vanishes*, and this is known as the *vanishing gradient problem*. How can we mitigate that?

1. If we look at the update rule $w_i = \nu * ..$, we might consider increasing the learning rate $\nu$ by a lot and hope that a large $\nu$ multiplied by a value that is almost zero gives a sensible output. However for a high learning rate we've already seen that could lead to overstepping a minima. Thus, this is not really a good option.

2. Apply *input normalization* and *batch normalization*. Note that it's now clear why we apply batch normalization before the activation function: To make sure we feed the activation function values where there is a nonzero slope.

3. Choose a better activation function!

Indeed, the third option is important. The sigmoid function is very suited for the output of the network, for example if we want to have something akin to a probabilistic output as in "these photos might show the same person with confidence $0.8$". For the intermediate layers however it is not so suitable, because it's range is very limited: For large values, we essentially have $0$ as the value of the slope.

A more suitable activation function for deep networks is adding a rectified linear activation function or *ReLU* for short. It is defined as

$$\alpha(x) = \begin{cases} x, & \text{if } x > 0 \\ 0, & \text{otherwise} \end{cases}$$

This helps a lot with the vanishing gradient problem, since for the derivative of the linear part of the function (for $x > 0$) we always have sensible values for the derivative, and when multiplying all these — larger than zero — values via the chain rule we help to circumvent the vanishing gradient problem. Note that a pure linear function however is not enough to learn complex functions, and the zeroization part fulfills this requirement of having non-linearity.

## 2.12 Residual Layers

There is one last difficult layer that we have to discuss. I think the motivating quote for this layer is that

> The difference between theory and practice is that in theory it works, but in practice it doesn't.

If we make a network deeper, it should learn better. After all, we have more layers, and therefore can construct a better approximation of the original function.

However what researchers observed in practice with deep networks is that initially the error goes down and down, then reaches a kind of minimum, and with lots and lots of examples often gets worse. This is counterintuitive, as according to theory, the more examples we use for training, the better the network should perform.

One reason is the above mentioned problem with vanishing gradients (the opposite effect can occur, too, then dubbed exploding gradients). But there is another problem. Suppose the network was already trained with lots of examples. There is a small error left, but it is very small. Next we encounter a training example which does not really contain new information. The weights in the network actually don't really need an update, they should more or less be the same – the output of the network for these examples is already pretty much correct. What we are learning at that point for the majority of examples is the identity function $f(x) = x$. But it turns out that the identity function is actually *difficult* to learn for deep networks, as information is not passed straight down from one layer to the following layers.

Let's say you are an ambitious chess amateur, but at one point *don't* want to make progress any more. Well the best way to do this is probably just to skip any lesson! And that's exactly the intuition between residual layers (resp. residual networks): Once in a while, just skip some layers and directly pass information down to deeper layers.

Consider the network illustrated in Figure 2.19. At the very left, we start with some inputs $\mathbf{x_0} = x_1, \ldots, x_n$. Then we multiply by some weights, let's abbreviate that in matrix notation with $\mathbf{W_1}$, i.e. we get $\mathbf{W_1 x_0}$. This is then the input to the activation function $\alpha$; here we suppose a rectifier linear unit. Let's call this output $\mathbf{x_1}$, i.e. $\mathbf{x_1} = \alpha(\mathbf{W_1 x_0})$), which is then again multiplied by some weights $\mathbf{W_2}$, and passed to another activation function to finally obtain $\alpha(\mathbf{W_2 x_1})$. However just before the second application of the activation function, a *residual network* uses a skip connection and just *adds* some of the original input $\mathbf{x_0}$. Instead of $\alpha(\mathbf{W_2 x_1})$ the output is changed to $\alpha(\mathbf{W_2 x_1} + \mathbf{x_0})$.

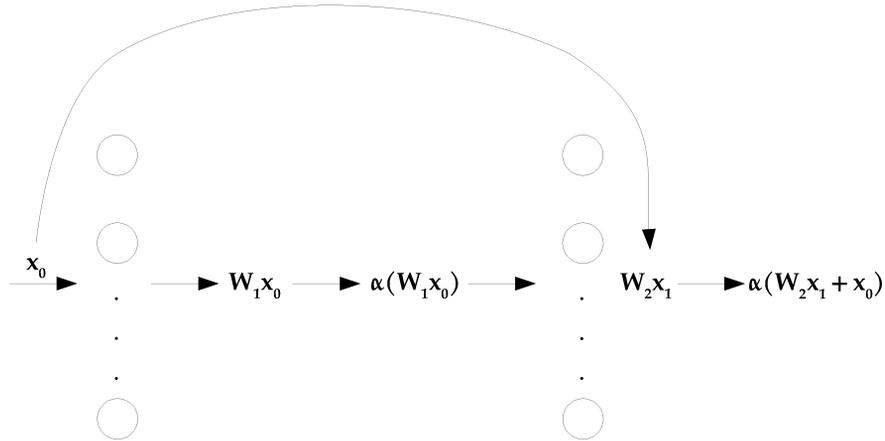

Figure 2.19: Residual Network.

What is the motivation for that? Suppose due to vanishing gradients the weight matrix $\mathbf{W_2}$ is almost zero. Then the output $\alpha(\mathbf{W_2 x_1} + \mathbf{x_0}) = \alpha(\mathbf{x_0})$. That's the identity function - the two layer don't affect the output at all, and $x_0$ dominates. Thus it is very easy for the network to learn the identity function, i.e. not to change anything for the worse in the deeper layers.

How does this fit with back-propagation? This is actually quite straight-forward. First let's simplify notation, and write $F(\mathbf{x_0})$ instead of $\mathbf{W_2 x_1}$. Remember from the previous section that we want to compute the partial derivative $\frac{\partial E_{\text{total}}}{\partial \mathbf{x_0}}$ [3]. We

---

[3]technically, we want the partial derivative of each $x_0, \ldots x_n$ of the vector $\mathbf{x}$, but this doesn't affect any of the following steps

have

$$\frac{\partial E_{\text{total}}}{\partial \mathbf{x_0}} = \frac{\partial E_{\text{total}}}{\partial \alpha} \frac{\partial \alpha}{\partial \mathbf{x_0}} = \frac{\partial E_{\text{total}}}{\partial \alpha} \left( \frac{\partial F(x)}{\partial x} + 1 \right) = \frac{\partial E_{\text{total}}}{\partial \alpha} + \frac{\partial E_{\text{total}}}{\partial \alpha} \frac{\partial F(x)}{\partial x}$$

In other words: When we back-propagate the error to the layer before $x_0$, we consider both the error that is propagated regularly through the network — the chain $\frac{\partial E_{\text{total}}}{\partial \alpha} \frac{\partial F(x)}{\partial x}$ is exactly what we would get if we completely forget about the skip connection — but also *add* the error at the output layer $\frac{\partial E_{\text{total}}}{\partial \alpha}$.

In this example we skipped two weight layers, and the second weight layer was actually the output layer of the network. Needless to say that we can generalize this structure to skip an arbitrary number of layers. Also, the end-point of the skip connection does not need to be the output layer of the network. We can also of course add multiple skip connections. The important point to remember is that *residual networks* [HZRS16] use skip connections.

## 2.13 Overfitting and Underfitting

There is a joke among German university students that goes along the lines that a professor asks an engineering student to rote memorize a phone book.[4] The engineering students answers "Why?" Whereas when the professors asks a medical student the same questions the answer will be "Until when?".

Rote memorization is just that — you will learn the facts that you study but not more. Much focus is put by most schools of medicine into rote memorization such that later when doctors are faced with quick decision making they are able to immediately recall their important knowledge.

But rote memorization is simply not enough. We need to extract patterns and common themes, we need to generalize in order to be able to face new situations. For neural networks we have the two terms *underfitting* and *overfitting*.

Underfitting means we have been just lazy and did not learn enough. If we query our network it will spit out answers but these answers are not better than just guessing. We need to train the network enough in order to be able to learn — and this itself means we need enough training material and training time.

---

[4]Yup it's an old joke and will likely no longer work in near future.

Overfitting means we have learned the training data but did not generalize enough. Our networks performs very well on the training data but when faced with slightly different data the network performs poorly. There are multiple potential reasons for that and it is not always easy to spot the reasons. Typical solutions are to add more training data such that the network also sees such kind of slightly different data or if there is just no more training data to artificially create some by slightly distorting the existing training data to make the network more robust. But it could also be that our network is just too complex and over-engineered and reducing the complexity of it might force the network to generalize more.

We need not to delve into the reasons and how to overcome these remedies in all detail but just have to remember that these phenomena can occur and that sometimes decisions w.r.t. network architecture or training are rooted in avoiding both overfitting and underfitting.

## 2.14   Summary

We have seen that there is lot of parameters to choose from when designing a neural network. We have learning rates and batch sizes, activation functions, different layers and the question to put them where and in which order, and also the question on how deep the network should be.

It is difficult to choose the parameters in advance, and in fact subject to current research. Designing networks in practice therefore involves lots of practical experimentation and fine tuning.

We will discuss in detail the network structure of modern chess AI systems such as Google's AlphaZero in Chapter 4. Here, we have covered all required network elements to understand such network structures.

What's missing then?

Let's first create a hypothetical neural network that plays chess. First, we need

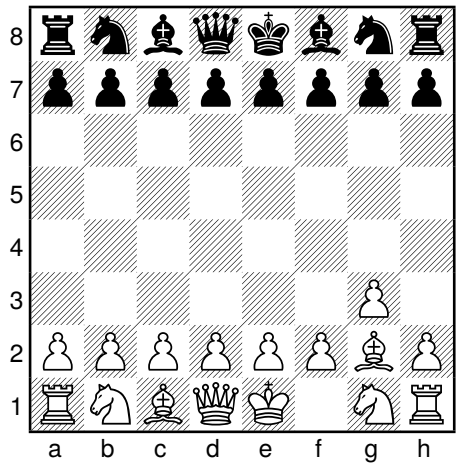

Figure 2.20: White bishop on g2.

to define the input of the network. That should be a chess position that encodes all information that define the state of the game. A naive way would be to simply take the binary representation of string with Forsyth-Edwards-Notation (FEN)[5]. Such a representation is far from being optimal: We would expect that positions with a similar structure would result in similar inputs to the network. But consider the positions depicted in Figure 2.20 and Figure 2.21 which have the FEN strings

    FEN: rnbqkbnr/pppppppp/8/8/8/6P1/PPPPPPBP/RNBQK1NR w KQkq - 0 1

and

    FEN: rnbqkbnr/pppppppp/8/8/8/5BP1/PPPPPP1P/RNBQK1NR b KQkq - 0 1

We have different FEN strings, which results in quite different bit strings. There are better encodings of course which we will encounter in Chapter 4.

---

[5]A FEN string encodes the position of all pieces, as well as information about and dates back to the 19th century.

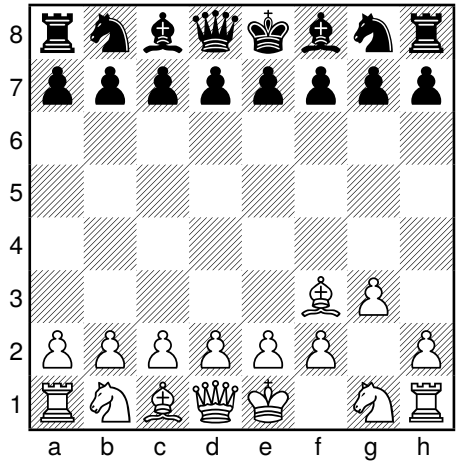

Figure 2.21: White bishop on f3.

We then build a fancy deep neural network and throw in all the tools we have learned so far. For the output we can chose to consider either regression or classification. Both more or less yield the same result:

1. In a regresssion model, we would simply output the expected likeliness that White wins (+1.0), Black wins (−1.0) or that the game is a draw (0.0). The output layer could for example use hyperbolic tangent as an activation function.

2. For a classification model, we would enumerate all finitely many possible moves on a chessboard. Suppose these are $n$ moves. We then create an output layer with $n$ nodes. We could add a softmax layer after that. For example if there are three possible moves, an output of $(0.3, 0.6, 0.1)$ would indicate that the first move has a change of 30 percent to be played, the second move has a change of 60 percent to be played, and the last move has a 10 percent change of being played.

Note that both models can be used to play chess. Whereas the first network just outputs a position evaluation, we can use this in the following way to play: Consider all moves, apply each move, and write down the output value of the regression model for that position. Next compare all output values, and then chose the move that leads to the next position with the highest evaluation.

What's missing is of course training data. Ideally we'd have a Grandmaster sitting next to us while training that given a chess position immediately answers with the best move. If such a GM is not available, the next best thing is probably to use a large chess database with strong Grandmaster games, go through all positions and then take the move played in a position as the expected output.

Unfortunately, this alone doesn't seem to work, even with all the advanced techniques discussed. Or at least noone so far was able to design and train such a network in a way that matches the best traditional chess programs. We need some more ideas on how to make a computer play chess which involve searching techniques. These are subject of the next chapter.

The first closest thing to a successful neural network that plays chess was *Giraffe*, an experimental implementation created by Matthew Lai as a master thesis project [Mat15].

Namely the first part of Lai's master thesis was just the above mentioned network with the regression model. Lai used a clever way of encoding chess positions together with a (not very deep) neural network with two hidden layers, each using ReLU's as the activation function. The **tanh** function was used to generate outputs between $-1$ (Black wins) and 1 (White wins). After quite some training, Giraffe assessed a set of test positions with an approximate ELO rating of 2400.

This doesn't seem a lot since traditional chess programs at that time were reaching ELO ratings of 3000. But this is quite an achievement since it was the first time that it was shown that a neural network can be trained such that it achieves IM or low GM chess expertise when *assessing* a chess position.

Note that it is really just that: After training we have a static network structure plus pre-computed weights that, given a position and one forward computation through the net immediately answers with a reasonable assessment of the position. It is even more incredible once we compare it to traditional chess programs, which use tree-search techniques.

The first use of neural networks for chess that actually resulted in a strong chess engine was by David et al. [DNW16]. There, they combined a strong neural network together with alpha-beta search. The neural network was used to evaluate chess position, and the alpha-beta search provided the necessary tactical strength. Their program, dubbed *DeepChess* achieved significant strength. While it was not among the strongest chess engines at that time, it was able to beat Robert Hyatt's well-known open source program *Crafty* [HN97].

**3**

# Searching

In the seminal work "Think Like a Grandmaster" [Kot71] Alexander Kotov
teaches an easy and straight-forward method to think like a grandmaster. His
method to find the best move in a position is based on these simple steps:

- consider all *interesting* moves in a given position

- then for each of these moves, calculate all the resulting variations. Once
  you calculated a line however, never come back to that line, just take your
  initial analysis for sure.

- calculate all lines for the opponent

- using this "analysis tree" of variations, finally assess the position and choose the most promising move

The problem with this great approach is that I do not know any amateur or grandmaster who thinks like that. But maybe the underlying reason is that I do not know any grandmaster in the first place. However, it sounds like a strategy that a computer could execute.

First note that a few things are slightly vague here. In particular how deep shall we calculate? Unless we are in an endgame where some lines actually end up in a terminal position, i.e. a mate, a stalemate or a draw, we will have to stop somewhere and assess the current position: Is it equal? Or does one side have an advantage?

We have seen in the last chapter how to create an extremely powerful and sophisticated evaluation function using a deep neural network with various layers. Such an evaluation function however was not available in the early days of computer chess. And as of now, there is no way to combine a deep neural network based evaluation function with minimax or alpha-beta search — but more on that later.

So let's start with something that probably every kid is taught when starting chess: Counting pawns. And slightly tweak that by considering piece placements, i.e. making sure that pieces that are placed in the center are considered to be of more value than pieces that are placed on the rim.

We'll use the numbers in Table 3.1 as a base. By giving a bishop ten points more than a knight, the evaluation function will favor the bishop pair. Also two rooks are rated higher than one queen. All this is debatable of course, but then again: This is just a very rough evaluation. We can slightly tweak this by adding resp. subtracting points depending on where a piece is placed. This is depicted in Table 3.2. For example, a knight placed on e4 would not be rated with 310 points, but rather with $310 + 20 = 330$ points. Again, this is debatable, especially when it comes to complex middlegames or endgames, but it suffices as a rough heuristic.

Table 3.1: Evaluation by Pawn Counting

| Pawn | Knight | Bishop | Rook | Queen |
|------|--------|--------|------|-------|
| 100  | 310    | 320    | 500  | 900   |

Table 3.2: Evaluation by Piece Placement

|   | A | B | C | D | E | F | G | H |
|---|---|---|---|---|---|---|---|---|
| 8 | -50 | -40 | -30 | -30 | -30 | -30 | -40 | -50 |
| 7 | -40 | -20 | 0 | 0 | 0 | 0 | -20 | -40 |
| 6 | -30 | 0 | 10 | 15 | 15 | 10 | 0 | -30 |
| 5 | -30 | 5 | 15 | 20 | 20 | 15 | 5 | -30 |
| 4 | -30 | 0 | 15 | 20 | 20 | 15 | 0 | -30 |
| 4 | -30 | 5 | 10 | 15 | 15 | 10 | 5 | -30 |
| 2 | -40 | -20 | 0 | 5 | 5 | 0 | -20 | -40 |
| 1 | -50 | -40 | -30 | -30 | -30 | -30 | -40 | -50 |

There are better evaluation functions of course. For example you could try to detect whether the game is at the beginning, in the middle game or whether there is an endgame position. Depending on that it might be better to either move the king outside of the center into safety or towards the center. In middle game positions you could evaluate king safety, i.e. has the king castled? Are pawns in front of the king? Did these pawns move? Are there pawns on the seventh resp. second rank?

Ideally of course we would use a deep neural network for evaluation, i.e. a network as introduced in the previous chapter.

To understand classical minimax and alpha-beta search however, from now on we will just a assume that we have an evaluation function that can be computed *fast*, like the trivial one above.

## Implementing the Evaluation Function

There is an excellent and easy to use chess library dubbed `python-chess` for `Python`. With it, it is straight-forward to implement the evaluation function that we described above. First, we define an array with the piece square values in Listing 3.1.

```
Listing 3.1: Piece Square Table
1 pieceSquareTable = [
2   [ -50,-40,-30,-30,-30,-30,-40,-50 ],
3   [ -40,-20,  0,  0,  0,  0,-20,-40 ],
4   [ -30,  0, 10, 15, 15, 10,  0,-30 ],
5   [ -30,  5, 15, 20, 20, 15,  5,-30 ],
6   [ -30,  0, 15, 20, 20, 15,  0,-30 ],
7   [ -30,  5, 10, 15, 15, 10,  5,-30 ],
8   [ -40,-20,  0,  5,  5,  0,-20,-40 ],
9   [ -50,-40,-30,-30,-30,-30,-40,-50 ] ]
```

The central data structure of `python-chess` is a `board` that encodes a chess position. Given a board, we simply iterate through all possible squares and depending on the piece that is placed on the square, we sum the value of the piece together with the value from the piece-square table to get an overall score. We do this both for White and for Black. Finally we return the difference of the overall score for White and Black. This is depicted in Listing 3.2. We can quickly test the evaluation function. Figure 3.1 is a position where it is White to move and he can checkmate.

The evaluation function will of course not consider the checkmate and just calculate the evaluation based on counting pieces and the placement of the pieces.

```
Listing 3.2: Computing the Evaluation Score
1 def eval(board):
2     scoreWhite = 0
3     scoreBlack = 0
4     for i in range(0,8):
5         for j in range(0,8):
```

```
6              squareIJ = chess.square(i,j)
7              pieceIJ = board.piece_at(squareIJ)
8              if str(pieceIJ) == "P":
9                  scoreWhite += (100 + pieceSquareTable[i][j])
10             if str(pieceIJ) == "N":
11                 scoreWhite += (310 + pieceSquareTable[i][j])
12             if str(pieceIJ) == "B":
13                 scoreWhite += (320 + pieceSquareTable[i][j])
14             if str(pieceIJ) == "R":
15                 scoreWhite += (500 + pieceSquareTable[i][j])
16             if str(pieceIJ) == "Q":
17                 scoreWhite += (900 + pieceSquareTable[i][j])
18             if str(pieceIJ) == "p":
19                 scoreBlack += (100 + pieceSquareTable[i][j])
20             if str(pieceIJ) == "n":
21                 scoreBlack += (310 + pieceSquareTable[i][j])
22             if str(pieceIJ) == "b":
23                 scoreBlack += (320 + pieceSquareTable[i][j])
24             if str(pieceIJ) == "r":
25                 scoreBlack += (500 + pieceSquareTable[i][j])
26             if str(pieceIJ) == "q":
27                 scoreBlack += (900 + pieceSquareTable[i][j])
28      return scoreWhite - scoreBlack
```

We then create a board with the corresponding FEN string and call the evaluation function (cf. Listing 3.3).

**Listing 3.3: Calling the Evaluation Function**

```
1 board = chess.Board("r1bqkb1r/pppp1ppp/2n2n2/4p2Q/2B1P3/8/PPPP1PPP/
      RNB1K1NR w KQkq - 4 4")
2 print("current evaluation")
3 print(eval(board))
```

White is considered to be slightly worse with a value of −55 since Black has placed more pieces in the center. As we can see, the evaluation function give a very rough assessment of the position but naturally fails to consider any kind of tactical threats. That's why we need search algorithms.

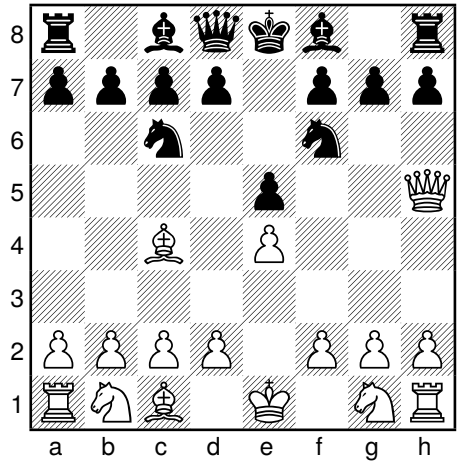

Figure 3.1: Mate in one.

## 3.1   Minimax

Let's reconsider Kotov's suggestion on how to calculate or how chess players should calculate in general, but define things more formally and without any disambiguity — so that a computer can understand and execute the approach algorithmically. But before let's take a look at a concrete example so that we can understand the general principle. For that we consider an endgame position with three opposing pawns, depicted in Figure 3.2. To make things easier we will only consider the upper right corner of the position and forget about the king position. We will also slightly change the winning conditions: White wins if he can place a pawn on the third rank. Black will win if he can block or capture all White's pawns. Note that there are no draws possible. We will also assume we have an evaluation function that can assess any position, and that will output a 10 if White wins, and a 0 if Black wins. There is no explanation how this evaluation function works; we will just assume it is there. All this is simply to get an example with fewer possible moves and variations such that it is still possible to manually calculate all options.

Before we start, some more definitions. When illustrating move sequences and resulting positions, we do so by using a *tree*. An example is depicted in Figure 3.3. On top, there is the *root* node. From the root node there are different *branches* we can follow which lead to other nodes, the *child* nodes of the root node. At the bottom there are nodes which have no successors. These are dubbed *leaf* nodes. It has some resemblance to a real tree if you consider this upside down.

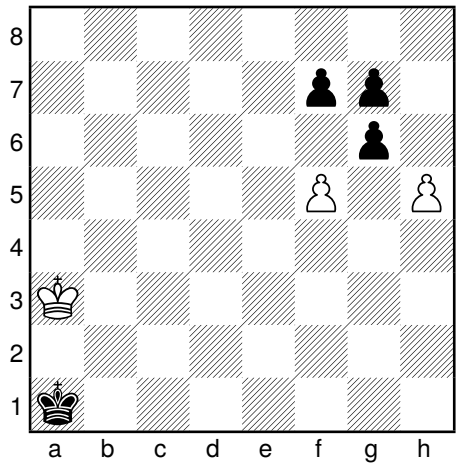

Figure 3.2: Endgames with opposing pawns

We will start with position (1) in Figure 3.3. It is White to move. There are four possible moves:

- White can move the left pawn forward.

- White can capture the center black pawn with his left pawn.

- White can capture the center black pawn with his right pawn.

- White can move the right pawn forward.

We now need to consider each of these. We start by executing the first option and move the left white pawn forward. Now it is Black's turn. She has three options: Capture the white pawn on the right resulting in position (4), move forward the middle pawn resulting in position (5), or capture the left pawn resulting in position (6). We could go on and consider all possible replies by White for each of these moves, but let's assume we stop here and call our evaluation function. Since in positions (4), (5) and (6) White will always have a free pawn and thus be able to reach the third rank, the evaluation will return an evaluation of 10 for each of these positions. As mentioned, we'll leave open how the evaluation function gets to this conclusion, we'll just assume it is there.

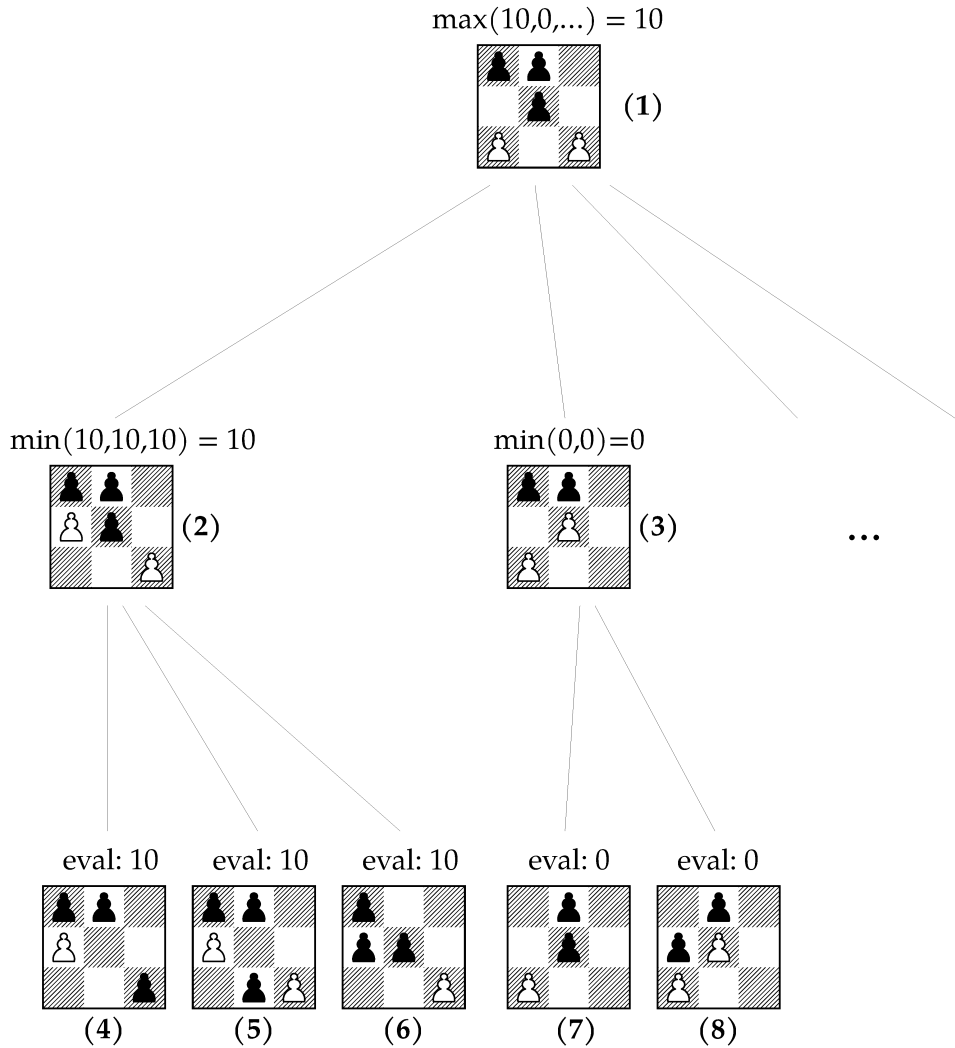

Figure 3.3: Minimax Search Tree.

We can do the same for other possible moves. For example if White in the initial position captures the center pawn with his right pawn and we consider all possible replies from Black, we end up in positions (7) and (8). These evaluate to 0, since White cannot break through and get a passer anymore.

Now how can we *use* this information of the evaluation of positions of leaf nodes? Suppose we want to calculate the evaluation of node (2). It is Black to move here. We will not consider hope chess, we will always assume that Black plays the *best* possible move (w.r.t. the evaluation of the child nodes). In other words, Black tries to *minimize* the evaluation outcome. He will choose the move that leads to the child node with the smallest evaluation. Here he can chose between three nodes, however all result in an evaluation of 10. The *minimum* of 10, 10 and 10 is however 10. Similar for evaluation of node (3), all moves lead to a position which is evaluated with 0. Here, the minimum is 0.

So far this is all not very interesting. So let's have a look on how these evaluations of node (2) and (3) help us to get a better evaluation of the root node. In the root node, it is White to play. White will of course chose the move that leads to the highest possible evaluation, i.e. he wants to *maximize* his outcome. For node (2) he now knows that if Black plays perfectly, the position is evaluated with 10. For node (3), the position was evaluated with 0. We haven't calculated it, but assume that for all other moves from the root node, we will also end up with evaluations of 0. White chooses the *maximum* of all possible options, i.e. he will play the left pawn forward leading to a node with an evaluation value of 10. As this is the maximum, we consider the root node also to be evaluated with a value of 10, i.e. winning for White.

We formulate this approach as an algorithm:

- First we create a search tree. Given a position, we consider all moves for the side whose turn it is, say White. For each move we consider then all replies by Black. Again, for each of these moves by Black, there are a number of moves by White and so on.

- At some point, we have to stop. Either because there are no more legal moves (i.e. the game is finished due to checkmate or a draw), or just because we ran out of time and space since the tree grows very quickly. A simple way to limit the size of the search tree is to always stop at a fixed distance from the root, i.e. at a fixed *depth* of the tree.

- For each *leaf node* of the generated tree, we have to evaluate the position by our evaluation function. This could be the pawn counting function that we introduced earlier. We augment this evaluation function for positions where White won by checkmate with evaluating the position as $\infty$, with drawn positions as $0$, and with positions where Black won by checkmate with $-\infty$.

- Now we propagate the results back to the root node step by step. Given a node where all child node have evaluations we

  - take the *maximum* value of all child nodes if it is White's turn

  - take the *minimal* value of all child nodes if it is Black's turn

This procedure is known as *minimax* [vN28]. Note that *minimax* is quite independent of chess. In particular it can be applied to all two-player games similar to chess, as long as there is an evaluation function that, given a position of the game, can give a numerical assessment of the position.

Using `python-chess`, minimax has a straight-forward implementation, shown in Listing 3.4.

**Listing 3.4: The minimax algorithm**

```
1 def minimax(board, depth, maximize):
2     if(board.is_checkmate()):
3         if(board.turn == chess.WHITE):
4             return -10000
```

```
5          else:
6              return 10000
7      if(board.is_stalemate() or board.is_insufficient_material()):
8          return 0
9      if(maximize):
10         best_value = -99999
11         for move in board.legal_moves:
12             board.push(move)
13             best_value = max(best_value,
14               minimax(board, depth-1, not maximize))
15             board.pop()
16         return best_value
17     if(minimize):
18         best_value = 99999
19         for move in board.legal_moves:
20             board.push(move)
21             best_value = min(best_value,
22               minimax(board, depth-1, not maximize))
23             board.pop()
24         return best_value
```

The function `bestmove` takes three parameters: The board that should be evaluated, the current depth (to limit the search depth) and a boolean parameter that indicates whether we should maximize or minimize the evaluation of the current board.

If we have a checkmate, instead of returning infinity we return a very large positive (resp. negative) integer depending on whether White checkmated Black or vice versa. If the game is drawn, we return 0. In all other cases we generate all legal moves of the current position. For each move we apply this move to reach the next position. Then it remains to distinguish if it is White's turn and we want to take the *maximum* result w.r.t. that board position, or if it is Black's turn and we want take the *minimum* for that position. After considering all legal moves, we have effectively computed an evaluation of the current board and return this evaluation that was stored in the variable `best_value`.

In fact, we have already implemented a chess engine — albeit a very simple one. Given a position, we can use *minimax* to get the next best move; cf. Listing 3.5.

**Listing 3.5: Compute Best Move in Position**

```
1 def getNextMove(depth, board, maximize):
2     legals = board.legal_moves
3     bestMove = None
4     bestValue = -99999
5     if(not maximize):
6         bestValue = 99999
7     for move in legals:
8         board.push(move)
9         value = minimax(board, depth - 1, (not maximize))
10        board.pop()
11        if maximize:
12            if value > bestValue:
13                bestValue = value
14                bestMove = move
15        else:
16            if value < bestValue:
17                bestValue = value
18                bestMove = move
19    return (bestMove, bestValue)
```

First, we generate all possible legal moves in the current position. If it is White's turn, we are looking for the move with the highest score, and initialize the evaluation score for the best move in the variable bestValue with a very low number, so that any move returned by *minimax* will be better than this default value. If it is Black's turn we seek to minimize the score, and thus initialize bestValue with a very high number. Next we iterate through all legal moves, and compare the score to the best one we've found so far. If we have found a better move, we remember it in the variable bestMove. After iterating through all moves, we finally return the best move found so far.

## 3.2   Alpha-Beta Search

With the *minimax* algorithm we have to visit a lot of nodes in the tree, i.e. search and evaluate a lot of positions. How many? Let's start a small experiment. Consider again the position in Figure 3.1.

We can use this position for two purposes. First, let's see if our *minimax* implementation actually works and finds the (obvious) checkmate Qxf7. Second, let's see how many nodes have to be visited to find the checkmate. Using our implementation of `getNextMove` this is straight-forward. First let's fix a depth *d* of halfmoves to search the tree. Then we have to create the `board` object with the position and call `getNextMove`. In Python this consists of just the two lines shown in Listing 3.6

```
1 board = chess.Board("r1bqkb1r/pppp1ppp/2n2n2/4p2Q/2B1P3/8/PPPP1PPP/
     RNB1K1NR w KQkq - 4 4")
2 print(getNextMove(4, board, True))
```

For Figure 3.1 and depth 3 this actually takes a few minutes on my machine. Again this is Python, and far away from an efficient chess engine implementation. The checkmating move Qxf7 is found eventually. Adding a counter gives us the visited nodes — in this example we had to visit 46828 nodes. Can we do better and reduce the number of nodes that we visit, yet still be sure to find the checkmate?

Let's have another look at the example that was introduced with *minimax*, depicted in Figure 3.4. We start at the root, then visit node (2), and further expand to nodes (6), (7) and (8). Our expansion stops here, we call our evaluation function, and propagate the result back to node (2). Here we take the minimum of the three child nodes, which results in an evaluation of 10 for node (2). We then go back to the root node, and start expanding the next possible move, resulting in a visit to node (3). There are two possible moves available at node (3). We start with capturing back with the pawn, visit node (9) where we reach our search depth limit and evaluate the position to get an evaluation value of 0.

At this moment let's pause a little bit before doing any further work, and consider the information that is currently available to us. In node (3) it is Black to play. The left child gave him a 0. If the other child is larger than 0, Black will choose the best (minimal) node below, and that is precisely the child with a 0. If the other child is even smaller than 0, Black will chose that one of course. In other

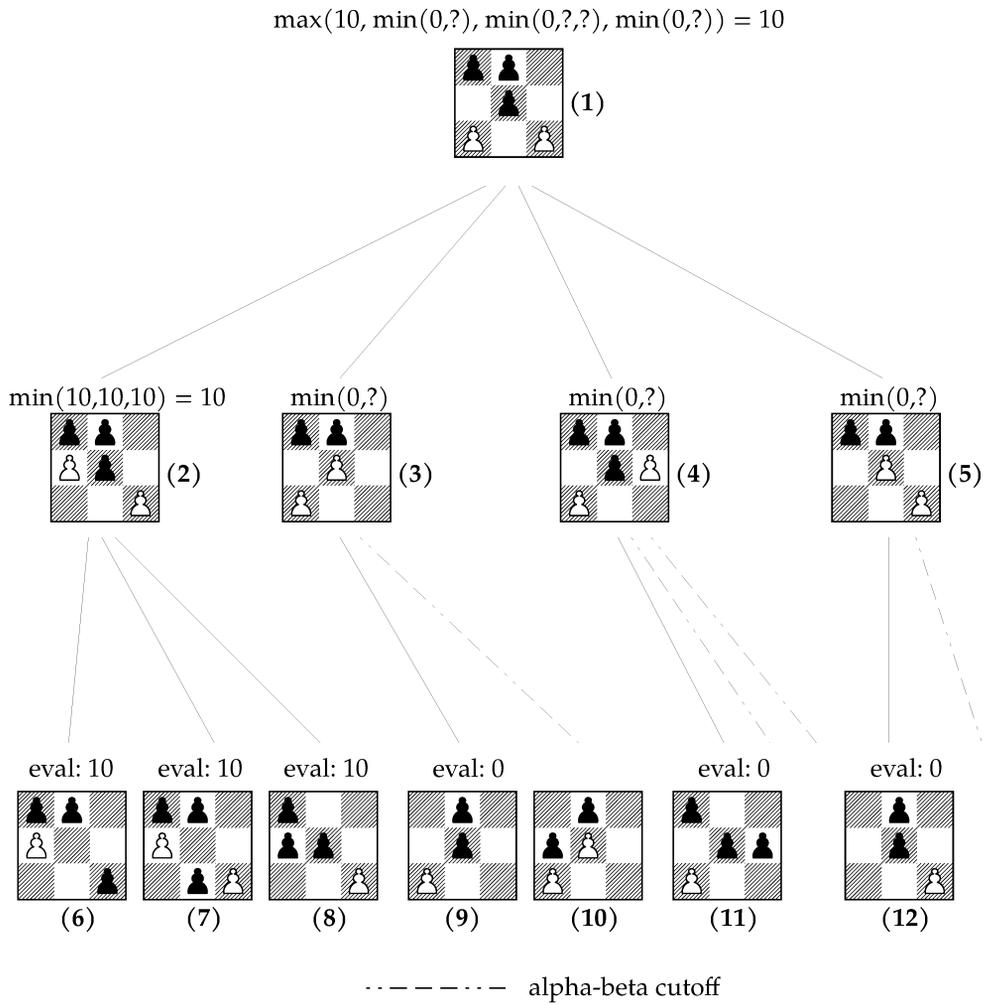

Figure 3.4: Alpha Beta Cutoff.

words, Black has *at least* a node with a value of 0 available, possibly even lower.

Now let's go up further in the tree, back to the root node (1). Here, it is White to play. White tries to *maximize* the outcome of the game. We know that Black can *at least* get a 0 if White chooses the child node (3). But from our previous evaluation we already know that the node (2) will get White a 10, so he will always prefer node (2) over node (3). In other words, the choice of White does not depend on the evaluation of node (10) any more. We can stop any further investigation below node (3), and instead directly hop over to node (4). After evaluating node (11), we also know that Black can get value 0 here. In the same manner we can stop any further investigation and hop over to node (5). The same spiel continues.

In the end, we saved visiting a lot of nodes. In fact, we only ever fully visited the leftmost branch of the tree, i.e. *all* the nodes below node (2) until we hit our search depth limit. For all other nodes, we skipped the search and evaluation of several parts of the tree. We did so by always recording the current best evaluation w.r.t. the current subtree for White (Black), and we call this the *alpha (beta)* value. If we can cut off a branch and skip visiting its nodes, we say that we did an *alpha (beta)* cutoff. Hence the name of the algorithm: *alpha-beta search*.

The algorithm is depicted in Listing 3.7. Here we reuse the evaluation function from the *minimax* implementation. The computation of the best move in a current position is also the same as before. We only replace the search algorithm itself.

**Listing 3.7: Alpha Beta Search**

```
def alphaBeta(board, depth, alpha, beta, maximize):
    if(board.is_checkmate()):
        if(board.turn == chess.WHITE):
            return -10000
        else:
            return 10000
    if depth == 0:
        return eval(board)
    legals = board.legal_moves
    if(maximize):
        bestVal = -99999
```

```
12        for move in legals:
13            board.push(move)
14            bestVal = max(bestVal, alphaBeta(board, depth-1, alpha,
                  beta, (not maximize)))
15            board.pop()
16            alpha = max(alpha, bestVal)
17            if alpha >= beta:
18                return bestVal
19        return bestVal
20    else:
21        bestVal = 99999
22        for move in legals:
23            board.push(move)
24            bestVal = min(bestVal, alphaBeta(board, depth - 1,
                  alpha, beta, (not maximize)))
25            board.pop()
26            beta = min(beta, bestVal)
27            if beta <= alpha:
28                return bestVal
29        return bestVal
```

Let's match the example with the implementation. We initialize the algorithm with a very small value for alpha, and a very large value for beta, indicating that we cannot cut any branch in the beginning without searching. Now let's focus on lines 12-19 and node (1). For each move we compute the evaluation of the subtree, and compare it with our current alpha value and take the maximum of both. Running alpha-beta search on node (2) gives us a best value of 10, which is our new alpha value. Now we start visiting node (3) with an alpha value of 10. After visiting node (9) and going back to node (3) we have a beta value of 0 — from our current information, this is the best that Black can do. At this point the beta value — value 0 — is *smaller* than the alpha value — value 10 — (line 17). Instead of finishing the for-loop that starts in line 12, we instead immediately return our evaluation, thereby cutting off node (10).

Running alpha-beta search with the same depth as minimax in the position shown in Figure 3.1 returns in a fraction of the time compared to minimax. The node counts is down from 46828 to 9050. Alpha-beta search allowed us to skip lots of nodes. For deeper searches the effect becomes even more important!

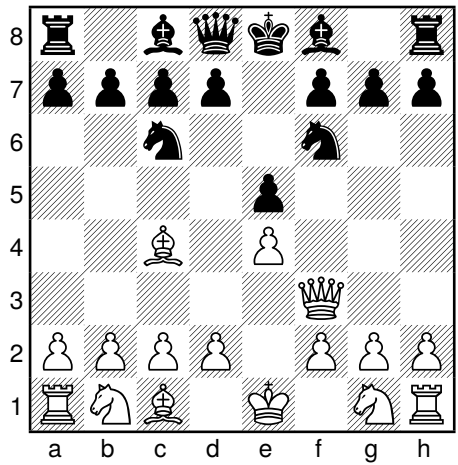

Figure 3.5: Quiescence Search

## 3.3  Advanced Techniques

Almost all (classical) chess engines are alpha-beta searchers, i.e. they combine
alpha-beta search with a fast-to-compute and handcrafted evaluation function.
To create a competitive chess engine some additional advanced techniques have
to be employed however.

One major issue is how to efficiently *implement* the chess logic. How do you
represent the board, how do you quickly generate all legal moves for a given
position? These very technical issues are not discussed here, as they are rather
independent of the algorithmic aspects of engine programming.

From a logical perspective, there are also additional techniques required in
order to create a useful chess engine. The first, and probably biggest issue is
related to the depth of a search and captures. Consider the position depicted
in Figure 3.5. Imagine that we use alpha-beta search with a fixed depth, and
the position depicted in the figure occurs in the search tree just so that there
is one half-move left before we reach our depth limit. One legal move is Qxf6.
After that we end up in a leaf-node, so the evaluation function will be used to evaluate the position. And if we just count pawns then great — White is up a knight! This evaluation is reported back into the upper layers of the tree, and suddenly this position is deemed very advantageous for White. Of course that's completely wrong because Black will recapture with the pawn on g7 or the queen on d8 and will then be up a queen against a knight.

The solution to this problem is to search for a fixed depth, but if a position is not deemed *quiet*, we search some more. This is dubbed *quiescence search*. The question is of course how to define when a position is quiet or not. A common simple heuristic for a given position is to consider all re-captures w.r.t. one field until no more re-captures are possible. So in Figure 3.5, after White's Qxf6 we would continue the search but not w.r.t. all possible moves by Black, but only by possible re-captures, i.e. consider the position after 1.Qxf6 Qxf6 and 1.Qxf6 gxf6. As there are no more re-captures possible, the position is quiet and the search stops. The evaluation would then be more realistic and the disastrous effect of the exchange becomes obvious to the engine.

The concept of quiescence search is as old as computer chess itself, and already mentioned in Shannon's legendary paper [Sha50] where he writes: *A very important point about the simple type of evaluation function given above (and general principles of chess) is that they can only be applied in relatively quiescent positions.*

Another major issue is the order in how moves are searched. If we have limited time, it makes a huge difference whether we first look at *interesting* moves, or just consider all possible moves in random order. Thinking about the example above, it might not make a lot of sense to look at all capture moves of the queen first, as it is unlikely that there is a huge material gain — in most cases, we would simply trade the queen for some lower valued piece and just loose material. There are static ways to do it – i.e. heuristically evaluate the position – or methods that employ information we gathered during the tree search itself. Consider for example the case where White makes a threat. There might be one specific move that counters this threat, resulting in a beta-cutoff in the alpha-beta algorithm, i.e. as soon as Black makes this move we can stop searching since we know that White's threat was not that good in the first place. In our alpha-beta cutoff example that is e.g. the Black's move leading from position (3) to (9). We consider this as a *killer move*. It was very good in this position, and therefore we will also consider it in similar positions first. This is dubbed *killer heuristics*.

An analysis of move orderings and the relation to alpha-beta was given in Eric Thé's master thesis [Eri92]. Killer heuristics were invented by Barbara Liskov in her PhD thesis [Hub68].

Another simple yet very effective technique to improve search is the use of *null move pruning*. It is quite advantageous in chess to be able to make two moves in a row. Suppose we consider a move by white, then let White immediately execute another move (skipping Black's turn), and then evaluate the position. If the position does not give White an advantage despite being able to move twice, then the first move was probably a bad move to begin with. Null move pruning was used quite early in chess programs, and the first time it was mentioned in the academic literature is most likely due to Adelson-Velsky et al. [AAD75].

Another simple trick to speed up search is transposition tables. A lot of move sequences in chess lead to the same position, and once we have analyzed one position, we should try to re-use the result instead of re-creating the full search tree below that position. To do so, we simply reserve a small amount of memory where we store positions that we have encountered as well as their evaluation. Ideally this is not just the evaluation score computed by the evaluation function but rather the result of a deeper search. Then when we encounter a new position, we compare the current position to all positions stored in memory and if we find a hit, we simply re-use the existing evaluation instead of re-examining it by search. Here it is important to be able to quickly check if two chess positions are equal, and we can so by *hashing*. This is why this technique is often dubbed *hash maps*. Of course we have only finite memory and therefore have to delete some (older) positions in order to store new ones once the hash map is full. When using computer chess engines, there is often a parameter to adjust the size of the hash map.

Last, a huge amount of applied research is spent in tuning the evaluation function. We can include as much chess knowledge that we want. Hsu, one of the creators of the groundbreaking Deep Blue chess program for example writes in his memoir [Hsu04] about the famous Deep Blue vs Kasparov match that they incorporated an evaluation feature "potentially open file", i.e. rooks would get higher scores if they were placed on files that could *potentially* open up later on after some capture sequence. He describes how that guided Deep Blue in one particular position into placing the rooks optimally. Of course computing a complex evaluation function uses time, which is then lost to explore the search tree. Thus we have to make some compromise between a sophisticated (but time-consuming) evaluation function where we can search only a few layers down in the tree versus taking a dead-simple but fast evaluation function with which we can search much deeper.

The ground breaking open-source chess program Fruit[1] for example used a rather simple evaluation function. However since it could thus search deeper, it beat a lot of competitors. This illustrated that a sophisticated evaluation function is not always an advantage if it slows down search too much.

Of course, ideally we would not handcraft an evaluation function, but rather automatically create an near-optimal evaluation function by using neural networks, as we have explored in the earlier chapters. But more on this challenge later on.

Only the most important optimizations are mentioned here. There is a plethora of techniques, refinements and optimizations. These kind of optimizations, together with implementation details, long differentiated good from very good chess programs. This is also what made chess programming so fascinating to programmers. A small neat idea or heuristic on how to improve the search together with some implementation optimization and your program gained another 50 or 100 Elo points. With neural networks on the horizon, these days seem to be over.

## 3.4   A Note on Deep Blue

For alpha-beta searchers we need both fast search speed and good positional evaluation. The case of DeepBlue greatly illustrates how important a good evaluation function is. Or in other words: An alpha-beta searcher with a weak, or rather *incorrect* evaluation function will result in a weak chess engine, no matter how powerful the search is.

---

[1]https://www.fruitchess.com

When it comes to alpha-beta searchers, Deep Blue was probably the culmination of all 90s technology combined. A true chess monster. But let's start at the beginning.

During the late 70s and early 80s, computers became a commodity item. Driven by the microcomputer revolution, it was now feasible to not only implement the chess logic but also achieve reasonable club player strength.

Common and affordable were 8-bit CPUs, like the ones typically found in home computers of that era, e.g. the Zilog Z80 or MOS 6502. But dedicated chess computer units were also common, and they used embedded chips you probably never heard of, like the Hitachi 6301Y.

These chips were cheap all-purpose commodity products and naturally very resource constrained. Or slow as a dog, to put it bluntly. Nevertheless enthusiastic specialist programmers implemented impressive programs on them.

There was another direction of chess programming however that was more centered around university research. Remember that chess was still a defining problem in artificial intelligence research at the time. These researchers had access to supercomputers. And prototypical hardware design tools. These enabled them to design special purpose circuitry for dedicated tasks such as chips optimized for digital signal processing.

It was this idea out of which Deep Blue was born. Namely to create dedicated chess chips: special circuitry for all the costly operations in chess; especially move generation and position evaluation. And not only that, Deep Blue consisted of a massively parallel CPU design that was able to execute alpha-beta search in parallel. This resulted in a chess computer that could search and evaluate positions several magnitudes faster than anything that was available on commodity hardware at the time; cf. Table 4.4 to get an intuition of its search speed.

However such a design came also with several challenges:

Creating a parallel version of alpha-beta search looks trivial on paper. Just execute different branches (e.g. different subtrees w.r.t. different moves in a given position) on different processors. However anyone who has ever conducted experiments in parallel programming will understand that this is an incredibly difficult task. You have to manage inter-process communication without having dead-locks, i.e. a situation where one processor waits for the result of a computation of the second processor and vice-versa — so they will wait forever. You have to balance the processor load evenly between the processes. And you also have to simply implement the algorithm correctly, which is notoriously difficult.

I was once attending a scientific conference where I met someone working at the research division of a very large and widely known software company. He told me that the programmers of the software products were joking that you should always create multi-threaded (i.e. executing in parallel) implementations, even if multi-threading is technically not required. The reason is that if there is a bug in the product, a customer will have a hard time to deterministically reproduce the bug, and you won't get bothered with customers and not reprimanded by management for your bugs.

The researcher also said that he wasn't really hundred percent sure though whether the programmers meant that as a joke or for real.

The same goes for hardware design. Remember the famous Pentium FDIV bug? The first generation of Pentium processors had a bug that resulted in incorrect floating point calculations. How come that such a large and skilled design team at Intel failed to spot the bug prior to release? It's because debugging a chip is notoriously difficult once it's put into silicon — especially with 90s technology. In fact, the problems with such large CPU designs led to an advance in model-checking and formal methods in order to ensure correctness of designs in the late 90s. But when the DeepBlue team developed their chips such technology was not readily available.

The source code and hardware of DeepBlue is not openly available. But from Hsu's own account of the events [Hsu04] we can infer that a significant amount of time was spent with bug-hunting. In fact Hsu describes that he spent weeks debugging problems with ensuring that the en-passent rule was recognized correctly. He also mentions that they had a "ghost queen" problem, i.e. under certain conditions a queen would suddenly appear on the board in the corner and hence screw up the evaluation of a position completely. He also mentions how DeepBlue just crashed several times during games and had to be rebooted. Anybody who has ever implemented a chess engine from scratch can probably relate.

Also remember that it is crucial for an alpha-beta searcher to have a good evaluation function available. We will see the effect of an evaluation function later when we talk about NNUE technology. Whereas DeepBlue's evaluation function could be fed with several parameters and was implemented in hardware, it was still required to actually select suitable parameters and weight them accordingly. If you do not implement chess knowledge in the evaluation function and just rely on, say, pawn counting, you will get an incredibly powerful pawn snatcher who will utterly fail in any position where some positional assessment is required. In fact, DeepBlue actually lost against Fritz with White at the 1995 World Computer Chess Championship in Hong Kong, despite having magnitudes more computing power available.

After DeepBlue lost its first match against Garry Kasparov in 1996, Grandmaster Joel Benjamin was hired by IBM to improve DeepBlue's chess knowledge, i.e. its evaluation function.

It is interesting to look at all this from two different perspectives. From the developers perspective you have an incredibly powerful system which is notoriously difficult to program for, has several bugs, and lacks a good evaluation function.

From an outsider's perspective without a background in computer science you have a chess computer whose developers claim how powerful it is, but which actually plays bad chess and loses to a chess program running on a standard home computer like Fritz. It is not difficult then to completely misjudge DeepBlues true capabilities.

There was a lot of controversy about the famous 37.Be4 move from the second game of the 1997 Kasparov versus DeepBlue rematch. After the game Kasparov stated:

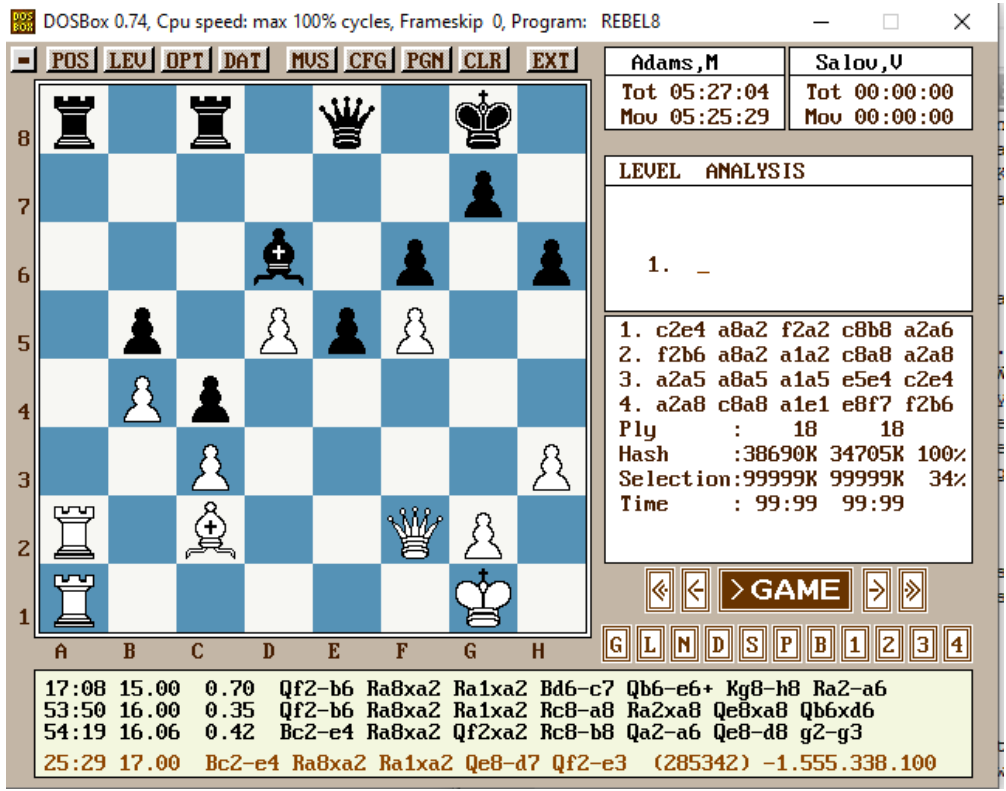

Figure 3.6: Rebel 8 in DosBox eventually finds the move Be4.

"We saw something that went beyond our wildest expectations of how well a computer would be able to foresee the long term positional consequences of its decisions. The machine refused to move to a position that had a decisive short-term advantage – showing a very human sense of danger [...] Even today, weeks later, no other chess-playing program in the world has been able to correctly evaluate the consequences of Deep Blue's position."[2]

We must put this in context: This was stated after the match ended and Kasparov lost. During the match he alleged that the DeepBlue team must have cheated, and that the move in question (37. Be4) must have been the result of human intervention, as most chess programs could not resist of trying to snatch a pawn with 37.Qb6 in that position instead of playing 37.Be4. In other words, this statement reiterates the narrative that cheating happened, as it alleges that since no (other) chess program was able to play that move, DeepBlue probably also wasn't able to find it without human intervention.

---

[2]Garry Kasparov: IBM owes mankind a rematch. Time, vol. 149, no. 21, 1997.

Let's have a look how Rebel 8.0 evaluates the position. Back then Rebel was known to be not only one of the strongest chess programs available but also one that had a comparatively good positional evaluation compared to other programs. Rebel 8.0 was released in 1996, well before the 1997 rematch with the controversial move in question, and is nowadays a free download on Ed Schröder's (Rebel's main programmer) web-page [3]. Computing resources were very limited at the time though. As an experiment I let run Rebel 8.0 for a few hours on my current machine in the DOS emulator DosBox. At first Rebel did not even consider 37.Be4 among its four best lines and strongly preferred 37.Qb6, trying to grab the b6 pawn. But after a few hours of calculation it suddenly prefers 37.Be4, as Figure 3.6 undoubtedly shows — the four best lines in descending order are shown in the middle box in the right[4].

So what can we conclude from that? It first shows that programs with good positional evaluation were very well able to see the move in question, contradicting Kasparov's statement. We can also see that for a strong chess program, thorough positional evaluation is crucial and that we cannot compensate for a weak evaluation function with pure processing power. And last, we can also see that it was very easy to misjudge the technical capabilities of DeepBlue without having knowledge about the intricacies of (hardware-based) chess engine implementations.

---

[3]`http://www.rebel13.nl`

[4]Note that the game information shows Adams, M. vs Salov, V. This is simply the result of my laziness of editing the game meta-data of a previously opened game.

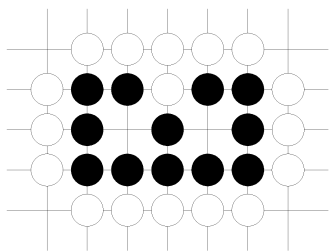

Figure 3.7: A group with two independent eyes, thus alive.

## 3.5   Monte-Carlo Tree Search

### 3.5.1   The game of Go

In the previous sections we have seen that alpha-beta search is a very powerful algorithm with which we can create powerful chess engines. So why should we look at alternative approaches for searching?

Let's have a look at the game of *Go*. Go looks intriguingly simple on a very shallow first sight, but is incredible deep. For those who are not familiar with Go, let's recapitulate very briefly the basic concepts and rules of Go. Instead of squares, 19 times 19 lines are printed on the board, and pieces are placed on the crossings, dubbed *points*. There is only one kind of piece, a *stone*. Players move alternatively (opposed to chess, Black has the first move) and put stones on the board. Once placed, stones are not moved around anymore — but they can be captured. The goal of the game is to occupy as much territory as possible. Territory is secured if there is a group of stones that encircle a territory. Enemy stones can be captured by encircling them with own stones. In order to maintain territory and make sure that the stones that encircle the territory are not captured by enemy stones, there need to be a certain number of unoccupied points. Stones that encircle territory are called *groups*. Consider the following examples:

- In Figure 3.8, Black has encircled territory. But this territory is attacked by White's stones, which itself encircle the black ones. There is only one free

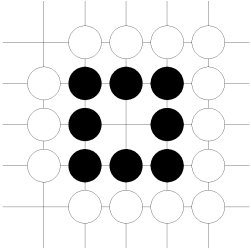

Figure 3.8: A group with just one eye, thus not alive.

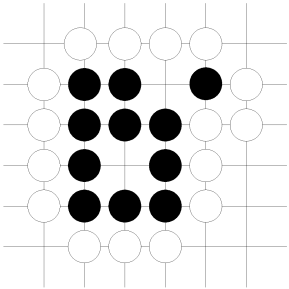

Figure 3.9: Another group with two eyes at first glance. One eye however is not a real eye, and the group is not alive.

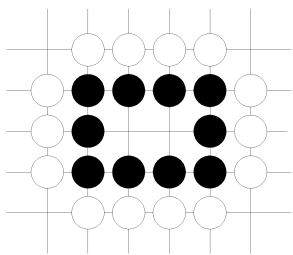

Figure 3.10: A group with two connected free points. These do not make up two eyes, hence the group is not alive.

point — an *eye* — within Black's territory. If it is White's turn, she will just place one stone on the last unoccupied point, thereby capturing all black stones at once. Hence, the black group is *not alive*, as the territory cannot be maintained.

- In Figure 3.7, Black has a group with two eyes. If White places a stone in one of the eyes, this would not result in a capture as Black has still one other eye. However the white stone placed in the eye would be itself surrounded by black stones. This would be a kind of suicide move, which is not allowed in Go. Therefore, the black group in the figure is *alive*.

- in Figure 3.9, one might at first sight think that the black group is alive. But consider White placing a stone on the upper right free point. This free point is not a real eye: If White places a stone there, it would result in a capture of the stone right next to the point, thereby not being a suicide move. The remaining black group has only one eye, and is not alive. Therefore, the whole group is not alive.

- Similarly, an *eye* is really just one free point. For example in Figure 3.10, the black group is not alive. You can try to figure out yourself why that is. Naturally, there are several edge cases and a lot of tactical theory on how to move to make sure groups are alive or how to kill moves.

This suffices to get a glimpse of the complexity involved in Go. There are tactics, such as the whole theory of groups and how to encircle stones as mentioned above. Compared to chess however there seems to be much more focus on positional understanding and pattern recognition, due to the sheer size of the board. Stones can be placed at seemingly unrelated locations far apart on the board, and later when more and more stones are placed then suddenly their relation matters. This brings us to a first major observation when we compare Go with chess: Opposed to chess, there is no simple evaluation function available when trying to assess an arbitrary position. Amateurs struggle with that, and it takes years of practice and talent to develop a positional understanding of the game.

In chess, we can count pieces. In Go that doesn't make much sense, because you could have way more stones placed on the board building a group, but if these stones are all captured at once later in the game because it turns out that you cannot keep the group alive, then these stones are worthless.

Building suitable evaluation functions is something that computer-go developers have struggled for ages. Before the revolution that happened with AlphaGo, complex evaluation functions incorporating lots of heuristic pattern detection were used. These were however costly to compute. Remember that for alpha-beta search to work, we need a fast and reasonable accurate evaluation function. If the evaluation function is so slow that in practice we can only search one move ahead, we basically do not search at all, but mostly rely just on the evaluation function to determine the next move.

Another complicating factor is the complexity of the game. As mentioned, Go is placed on a 19 times 19 board. Therefore the beginning player (Black) has $19 * 19 + 1 = 362$ moves to choose from — in Go it is possible to pass, hence the extra move. When one stone is placed, the opponent has $362 - 1 = 361$ possible moves and so on. Capturing stones and building groups of course limits the amount of possible moves, but it is still quite high.

Compare that with the amount of possible opening moves in chess. In the initial position, White can make two possible moves with each pawn, resulting in 16 possible moves. In addition there are four possible moves with the white knights, resulting in overall 20 possible opening moves for White. Black has then 20 possible replies. After moving some pawns the amount of possible moves increases of course, but we can still see that the amount of possible moves in a chess position is several magnitudes smaller than in Go.

According to Matsubara et al. [MIG96] we can estimate the branching factor (i.e. the average number of possible moves in a given position) for Go with 250 and for chess with 35. Note that in addition, in chess there are often nonsensical moves that are easy to spot — think of the previous section and quiescence search, where we can spot nonsensical captures. Whereas in Go it is not that easy. Also in Go, games just have way more moves until the game ends due to the board size. It is also interesting to compare the branching factors of chess and Go with Shogi, the Japanese variant of chess. It differs from western chess by having a slightly larger board ($9x9$) and that captured pieces can be put back into the game – similar to Crazyhouse or Bughouse chess. Intuitively this creates more complexity, and the branching factor is estimated to be around 80. Interestingly, alpha-beta search works (just barely) for Shogi, and Shogi programs have only recently reached grandmaster strength. In comparison, their strength has always been behind compared to computer chess, and despite very novel and revolutionary techniques, today they still are nowhere near as powerful as their chess counterparts. We will have a look at these revolutionary techniques, which have also been ported back to current chess programs, in the chapter about NNUE.

To summarize: Alpha-Beta does not work for Go, because

- of the high branching factor of the game and
- the fact that there is no fast and good evaluation function available.

Therefore it makes sense to take a look at search techniques that work even with high branching and do not rely on handcrafted evaluation functions.

### 3.5.2   MCTS/UCT

In order to tackle the goal of creating an algorithm that can handle games with a large branching factor and does not rely on an evaluation function, we'll take an inspiration from chess opening theory.

When professional chess player prepare openings, they rely on their own analysis as well as statistics. You probably have already seen opening databases. In such databases, information about played games is stored and accumulated. For example when opening the `Lichess.org` opening browser with the initial position, we get the following information for the moves 1.e4 and 1.a3. The information is accumulated for games of top players.

- For 1.e4, we know that this move was used in more than a million games. About 33 percent of all games were won by White, 42 percent were drawn, and 25 percent were won by Black.

- For 1.a3, only 574 games were played by top players. Here 29 percent were won by White, about 40 percent were drawn, and 32 percent were won by Black.

It seems natural to think that 1.e4 better than 1.a3 by looking at the statistics. We might be a little wary to be confident about that, since there were so few games played with 1.a3. It could be that 1.a3 is an excellent move and the small sample size is not representative. Another aspect to consider here is that the default opening book of `Lichess.org` considers only games by higher-rated players. Clearly, good players chose better moves on average, and therefore this should give a better indication on what is good and what is bad.

But what if we do not have such a database available? How about playing random games? Think of the following situation. You are in position where there are only two moves; one move wins a queen, and another move allows the opponent to move his queen away into safety. We could execute each move, and then for both of the resulting position play a few thousand *random* games. By that I mean to just *randomly* select valid moves in every position that follows, until the game ends in a win for White, a draw, or a win for Black. We could then compile the statistics for these random tryouts. We expect that for the move that wins the queen, that on *average* we should have a higher winning ratio.

And that's the whole idea of *monte-carlo tree search*. To determine the next move, just execute each valid move, play a number of random games, compile the statistics for these random tryouts, and compare the results to determine the next best move. Such an approach has several advantages:

- There is no need for an evaluation function. We just play random games until the game ends. This also means that monte-carlo tree search is completely independent of any domain-specific knowledge, and can thus be used for *any* kind of full-information deterministic two-player game.

- We can potentially handle games with a large branching factor. We just fix a number of random playouts and don't explore all subsequent replies by each player. Of course this has the potential to fail: Suppose you are White, in a tricky endgame position, and trying to find the next move. Suppose further that there are two moves. For the first move by White, almost all replies by Black result in a win for White, but there is just one *specific* reply where Black saves a draw. For the other move, White always wins with perfect play (there might be still possible draws, if White i.e. accidentally stalemates). Then relying on random playouts might miss this one specific reply for the first move, and report the two moves as statistically equal, or even falsely claim that the first move is better. On the other hand, Black's specific reply might be also missed by a human player, and even with these imperfect statistics this approach might still make a strong computer opponent.

- We do not rely on a huge database, but can still create reliable statistics — except for the issues mentioned above — as random playouts are usually *fast* to execute. We can therefore quickly play and generate *lots* of games for each position of the tree.

In order to describe monte-carlo tree search more formally, we can break down the algorithm into four main operations: *selection*, *expansion*, *simulation* and *backpropagation*.

- **Selection** In this step, we select a node in the search tree according to some rules. Initially we start with the current position as the root node of the search tree. After going through some iterations we will have created a larger search tree. In that case we move down in the game tree until we find and fix a leaf node. Of course we will need to find a strategy to select a move in every step. Such a strategy should select moves such that the most promising moves are selected and that we do not spend too much time in unpromising nodes. We will later see how such a strategy can be constructed regardless of any knowledge about the game by just using statistics. For now we will just assume that such a strategy exists.

- **Expansion** In this step we create a new node by applying moves at the node selected in the *selection* step. In the most simplest version of the algorithm, we just add precisely one move and add the resulting single node. This move could be selected at random, or according to some other strategy. We will stick with the simple version of just adding one node here.

- **Simulation** In this step we run a computation to get statistical information about the new node. The most straight-forward way is to play a certain number of *random* games, i.e. select random moves until the game finishes in a win, draw or loss.

- **Back-Propagation** In this step we take the statistical information that we computed by executing the random games and propagate it back up to the root of the game tree. We make sure to store and keep statistical information in each node on the winning statistics that we acquired for the new node.

We can visualize these operations as follows: We start at the root node, and first *select* moves subsequently until we reach a leaf node (Figure 3.11). Here we *expand* the leaf node by applying a (random) move, and create a new child node (Figure 3.12). From this child node we play random games (Figure 3.13) and finally propagate the results through all parent nodes back to the root node (Figure 3.14).

There is one important wording that we need to make clear here. Usually, the word *leaf* in a game tree denotes nodes where no move is possible. For chess, these are positions where the game is won, drawn or lost. Here we denote such nodes as *terminal* nodes. Leaf nodes on the other hand are nodes where not all possible child nodes have been expanded (and visited).

Consider Figure 3.15: On first sight, one might consider only nodes marked 1/1 and 3/5 as leaf nodes. After all, for the node marked 2/5, we have already

Figure 3.11: *Selection*: Starting from the root node at the top, a sequence of moves is selected such that we end up in a leaf node (bottom node marked in bold)

.

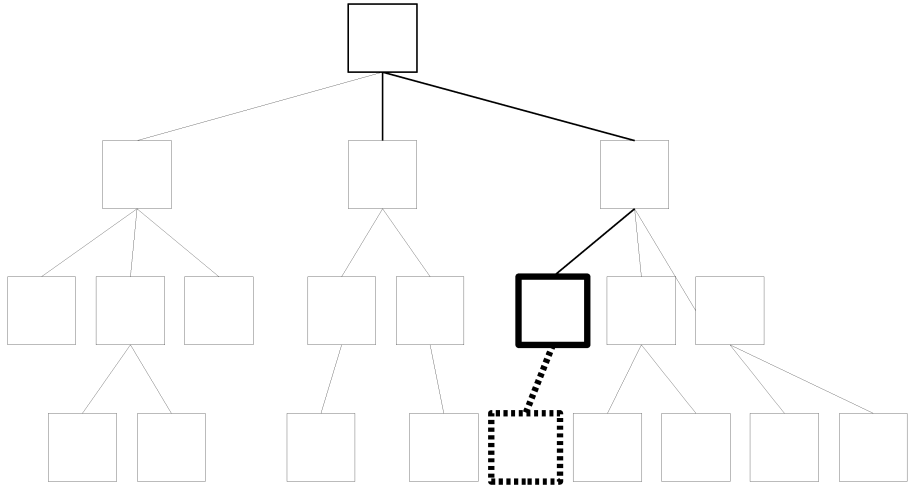

Figure 3.12: *Expansion*: At the leaf node, a move is selected at random, and a new node is generated by applying that move.

expanded a child. But there are other actions that are possible from the node marked 2/5 that have not been taken yet, so there are still un-expanded child nodes. Therefore, this is a leaf node.

Let's check a small example with real numbers to understand how the four operations work in practice. Initially we start with the current position as the root node. For each node, we record two parameters: a) the current evaluation of the node and b) the number of times it was visited. Have a look Figure 3.16. Here some iterations were already run, and there are values for the two numbers in each node. In the left part of the figure, we *select* the leftmost node and apply the *expansion* step. Then we run one *simulation* at that node. Suppose that this simulation consists of just one game played, and suppose further that we played a game (not chess) where the outcome of the game is either win (1) or loss (0). So we record 1/1 for this newly expanded node: It was one win, and we visited the newly expanded node exactly once.

In the right part of the figure, we employ *back-propagation* to update all parent nodes up to root of this newly expanded node. Here there is only one node. We update the visit counter by changing it from 5 to 6, and we also update the evaluation (add one win). We would then continue to select a node, expand a previously non-visited node, run a simulation, and update all parent nodes.

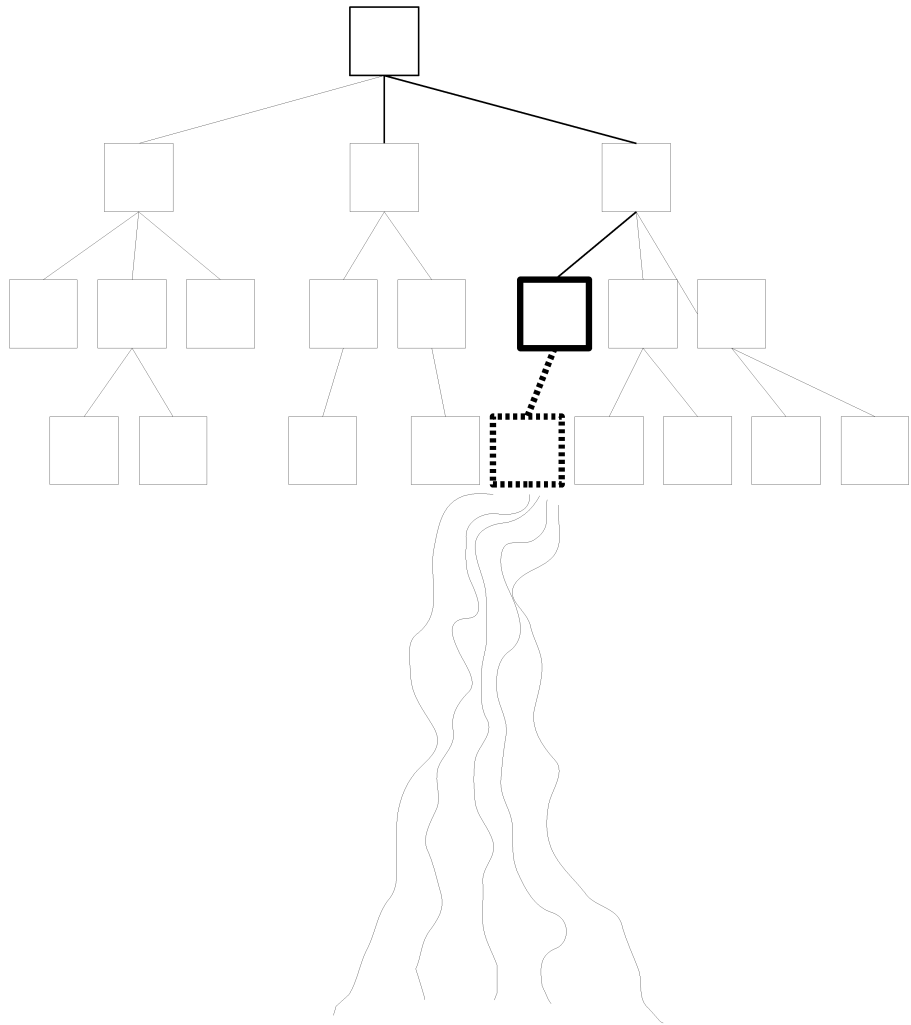

Figure 3.13: *Simulation*: A number of random games is is played. The win/-draw/loss ratio is recorded.

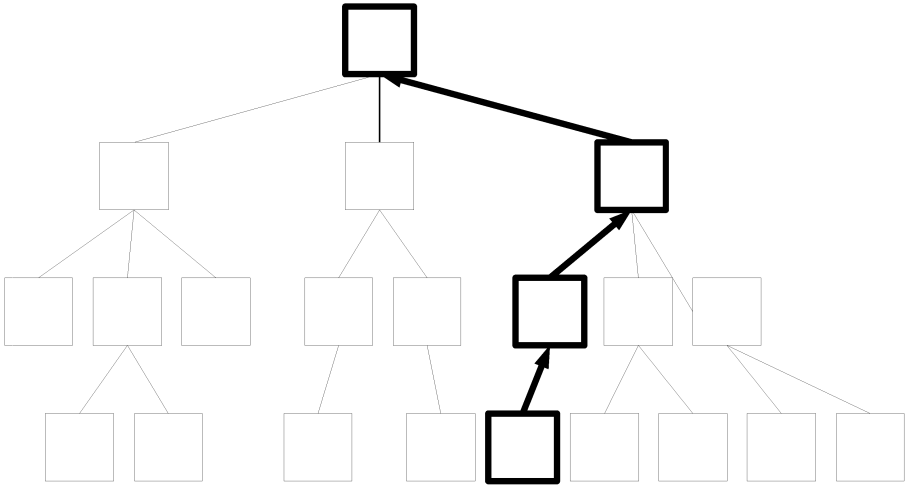

Figure 3.14: *Back-Propagation*: The previously computed win/draw/loss ratio is used to update all parent nodes up to root.

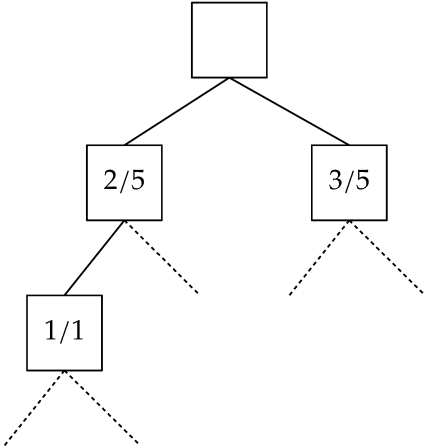

Figure 3.15: The node marked 2/5 is a *leaf node*

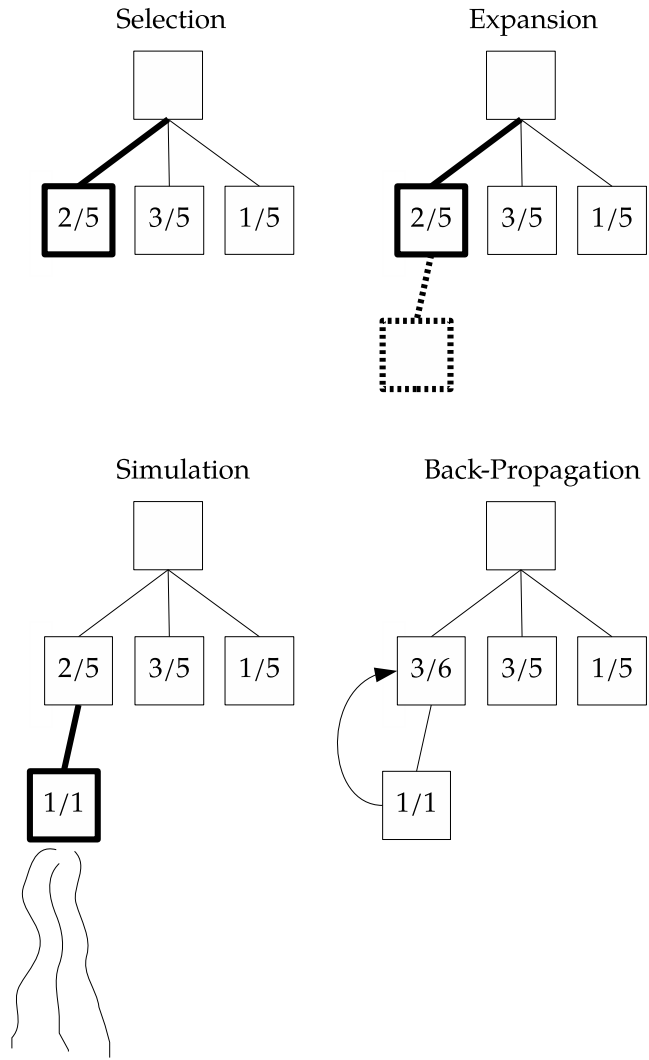

Figure 3.16: An example of MCTS selection, expansion, simulation and back-propagation.

We simplified things here a little bit. Not all games have a binary outcome. In chess for example either White wins, there is a draw, or White loses the game. Looking at arbitrary games, we will have some kind of *evaluation* that our simulation generates. Note that this should not be confused with an evaluation function as for e.g. alpha-beta search. There, the evaluation function evaluates non-final positions of the game. Here, we apply a simulation by random playout, and our evaluation result is based on the result of this playout. For example for chess, we could encode win/draw/loss as $(1, 0, -1)$ or by $(1.0, 0.5, 0.0)$. In back-propagation, parent nodes up to the root node are then updated according to the following formula. Given a new evaluation $E$, the average evaluation $M$ and the number of times a node was visited $V$, updates are defined by

$$M' = \frac{M * V + E}{V + 1}, \quad V' = V + 1$$

Consider for example Figure 3.17. As in the previous example, we have selection and expansion on the left side. The simulation results in an evaluation of 0.1. The parent node is then updated as

$$M' = \frac{0.4 * 5 + 0.1}{5 + 1} = 0.35 \text{ and } V' = 5 + 1 = 6$$

There are still two major issues left open that you are probably wondering about at this point. The first one is: How is the *selection* operation defined? Obviously selection has a huge impact on the outcome of MCTS. The second one is: Once we finish our MCT search, how do we select a move? For the first issue, there are two natural strategies:

- *Explore*: We should try to visit *all* nodes and try not to search too selectively. Otherwise we might overlook an excellent move because we never select and expand the corresponding node.

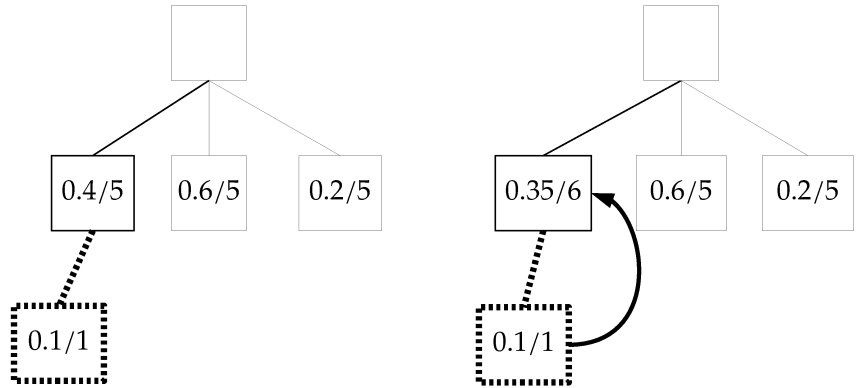

Figure 3.17:  Another example of MCTS back-propagation where the win-ratio is stored as a fraction.

- *Exploit*: If we identified a promising node, we should re-visit and analyze that node as often as possible, as this might be the best move in that position.  By re-visiting this node, we get a higher amount of confidence that this is a good move, indeed.

These two strategies are contradictive in nature.  The formula below is denoted UCT (Upper Confidence bounds applied to Trees) and is based on the *multi-armed bandit problem*, a heavily analyzed problem in probability theory.  The UCT value for a node $i$ with average evaluation $M_i$ which has been visited $V_i$ times is given by:

$$\text{UCT}(M_i, V_i) = M_i + c * \sqrt{\frac{\log(V_i)}{V_{\text{parent}}}}$$

Here $V_{\text{parent}}$ is the number of times the parent node of node $i$ has been visited, and $c$ is a small empirically chosen bias value.  Whenever we are at a branching point in the tree, we calculate the UCT value for all child nodes, and then select the child node with the highest UCT value.

We will skip the mathematical details on how to derive the UCT formula.  It suffices to say that it provides a balance between exploration and exploitation.

For more details see [ACF02] where a detailed analysis of the multi armed bandit problem is given.

There might be smarter ways to do the selection operation. If only we had some kind of oracle that could give us an assessment of which moves are more promising in a given position, so that we can immediately direct the node selection and expansion in the right direction and skip unpromising nodes. Like a fancy neural network, that can give us a quick assessment of the winning probability of each child node... Without revealing too much at this point, this is precisely the intuition behind most of the neural-network based engines that we will have a look at in the next chapter.

The second open issue was how to select a move once we finish our MCT search. There are various strategies. We could select the direct child of the root node with the highest average evaluation. On the other hand, UCT provides us with a good ratio between exploration and exploitation. Meaning that if a leaf node has been selected over and over again, the selection step already identified the best child by taking into account the UCT value which itself is based on $M$, the average evaluation of a node. The most common, and somewhat slightly unintuitive strategy is therefore to select the direct child of the root node that has been visited most, i.e. the one with highest value $V$.

We have now gone through the basic MCTS/UCT algorithm. Let's have some fun and implement this in Python to create a simple, MCTS-based chess program. Since it is in Python, it will be very slow and inefficient, but straightforward to implement. Moreover it practically illustrates the algorithm and underlying concept. The core data structure here is a tree node, cf. Listing 3.8.

**Listing 3.8: MCTS - Tree Node**

```
1  class TreeNode():
2
3      def __init__(self, board):
4          self.M = 0
5          self.V = 0
6          self.visitedMovesAndNodes = []
7          self.nonVisitedLegalMoves = []
8          self.board = board
```

```
9          self.parent = None
10         for m in self.board.legal_moves:
11             self.nonVisitedLegalMoves.append(m)
12
13     def isMCTSLeafNode(self):
14         return len(self.nonVisitedLegalMoves) != 0
15
16     def isTerminalNode(self):
17         return len(self.nonVisitedLegalMoves) == 0 and len(self.
               visitedMovesAndNodes) == 0
```

In the tree node we store the current evaluation in variable `M`, as well as the visit count `V`. We also maintain two lists: One that stores all moves that we already expanded at least once, and the corresponding child nodes. The other list stores all moves that we have not yet expanded, but could expand later. It is then trivial to check if a tree node is a leaf node (if there are possible, but non-expanded moves it's a leaf node) and whether it's a terminal node (there are no more possible moves, because e.g. it's a checkmate).

Listing 3.9 shows the four operations of MCTS, i.e. selection, expansion, simulation and back-propagation.

**Listing 3.9: MCTS - Operations**

```
1 def uctValue(node, parent):
2     val = node.M + 1.4142 * math.sqrt(math.log(parent.V) / node.V)
3     return val
4
5 def select(node):
6     if(node.isMCTSLeafNode() or node.isTerminalNode()):
7         return node
8     else:
9         maxUctChild = None
10        maxUctValue = -1000000.
11        for move, child in node.visitedMovesAndNodes:
12            uctValChild = uctValue(child, node)
13            if(uctValChild > maxUctValue):
14                maxUctChild = child
15                maxUctValue = uctValChild
16        if(maxUctChild == None):
```

```
17              raise ValueError("could not identify child with best
                    uct value")
18          else:
19              return select(maxUctChild)
20
21  def expand(node):
22      moveToExpand = node.nonVisitedLegalMoves.pop()
23      board = node.board.copy()
24      board.push(moveToExpand)
25      childNode = TreeNode(board)
26      childNode.parent = node
27      node.visitedMovesAndNodes.append((moveToExpand, childNode))
28      return childNode
29
30  def simulate(node):
31      board = node.board.copy()
32      while(board.outcome(claim_draw = True) == None):
33          ls = []
34          for m in board.legal_moves:
35              ls.append(m)
36          move = random.choice(ls)
37          board.push(move)
38      payout = 0.5
39      o = board.outcome(claim_draw = True)
40      if(o.winner == PLAYER):
41          payout = 1
42      if(o.winner == OPPONENT):
43          payout = 0.5
44      if(o.winner == None):
45          payout = 0
46      return payout
47
48  def backpropagate(node, payout):
49      node.M = ((node.M * node.V) + payout) / (node.V + 1)
50      node.V = node.V + 1
51      if(node.parent != None):
52          return backpropagate(node.parent, payout)
```

In the selection step, we first test if the node is a leaf node. If so, we return the node — this is the choice of our selection. Otherwise we iterate through all child nodes, compute the uct-value for each node, and select the one with the highest uct-value.

In the expansion step, we remove one move from the list of those moves that we have not considered yet. We create a new child node, and set all member values of the child node accordingly. Last, we add the move as well as the newly generated child node to the list of the node, where we maintain the visited children. We return the child node as the expanded node.

In the simulation step, we continue to randomly select legal moves of the current board position, and apply them to the board until the game ends with a win, draw or loss. Depending on the outcome we return a payoff of 1.0 (win), 0.5 (draw) or 0 (loss).

In the back-propagation step, we update the average evaluation as well as the node count according to the previously defined formula. Then we recursively apply this step to all parent nodes up to the root node.

We can test the implementation with the checkmate example shown in Figure 3.1 as illustrated in Listing 3.10.

**Listing 3.10: MCTS - Checkmate**

```
1 board = chess.Board("r1bqkb1r/pppp1ppp/2n2n2/4p2Q/2B1P3/8/PPPP1PPP/
      RNB1K1NR w KQkq - 4 4")
2 root = TreeNode(board)
3 for i in range(0,200):
4     node = select(root)
5     # if the selected node is a terminal, we cannot expand
6     # any child node. in this case, count this as a win/draw/loss
7     if(not node.isTerminalNode()):
8         node = expand(node)
9     payout = simulate(node)
10    backpropagate(node, payout)
11 root.visitedMovesAndNodes.sort(key=lambda x:x[1].V, reverse=True)
12 print([ (m.uci(), child.M, child.V) for m, child in root.
      visitedMovesAndNodes[0:10]])
```

First we create the board position, and the root of the tree. Then we apply a number of iterations of MCTS (here 200). Here we need to make sure that we do not accidentally try to expand nodes that are terminal, i.e. where the game ended. For example when looking at the child node after applying the move Qxf7, there are obviously no moves to expand anymore. A simulation then starts directly at this node. The simulation itself is then just determining that the game is in a final state, and will immediately compute the payoff. After applying MCTS we take a look at the root node to determine the best next move. We sort all moves according to the number of visits to that node. We note that the most-visited node was the one with Qxf7 which was visited 15 times.

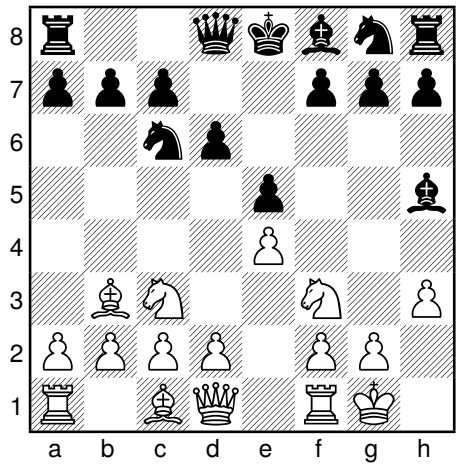

Figure 3.18: Legal's trap

How about more complex positions? Let's try Legal's trap, shown in Figure 3.18. Here 1.Nxe5 wins a pawn, since after 1...Bxd1? 2.Bxf7+ Ke7 3.Nd5++ wins by checkmate. The best reply by Black after the move 1.Nxe5 is 1...Nxe5, after White answers 2.Qxh5 and is a pawn up. Note that here the white bishop is placed on b3, whereas the position is typically presented with the bishop on c4. The reason is that the exchange gets slightly more deep then until the advantage materializes, i.e. we need to check the sequence 1.Nxe5 Nxe5 2.Qxh5 Nxc4 3.Qxc4 with a depth of five half-moves. Here we try to keep the search times low. Unfortunately, MCTS fails to identify the advantage with 2000 iterations. Is it that difficult? How about alpha-beta search. When we apply our simplistic alpha-beta searcher with `getNextMove(5, board, True))`, the correct move Nxe5 is found with a an evaluation of 280. How come that MCTS fails here?

The first issue is that after playing Nxe5, we notice that it is seemingly a very bad move. If we execute random games from that position, and Black captures the white queen, there is only one very *specific* combination that enables White to win the game. In order to detect this, and in order to make sure the *selection-step* moves in the right direction along that sequence of moves, we need *a lot* of iterations. Just a few random playouts will not pick up that sequence of moves, and probably end in favor of Black, especially if the queen is captured and White does not reply accordingly. This is a peculiarity of chess, where very specific tactics often occur, and far outweigh any positional aspects. This is different in other games such as Go for example.

The second issue is that we need a lot of playouts. However move generation is actually quite computationally intensive in chess. Chess moves seem intuitive for the advanced player, but the rules are actually quite complex:

- there are a lot of different pieces, and each has it's own rules

- pieces may be prevented from moving if that places the own king in check

- the knight can jump over other pieces whereas other pieces can not

- pawns have incredible complex rules. They can move two fields in the initial position, but only one field otherwise. Except for captures. And en-passent. And then when they reach the final row they become another piece

- castling is a special move in that two pieces are moved at once. Several complex conditions must be met prior to castling and after executing the castling move.

All that has to be computed, and fast move generation is actually very important when creating powerful chess engines. In addition to the issues mentioned above, Python is a very slow language — it trades ease of programming with execution speed. This is why the engines sketches shown in this book are so slow.

Now compare that with Go. There is only one kind of piece, a stone. A move consists of placing that stone on a free cross on the board. Done. Computing the possible legal moves is quite trivial.

All this affects heavily the amount of simulations that we can do in a certain amount of time. And this is why textbook MCTS as presented here is actually quite a bad choice for computer chess. There are of course exceptions to this rule. We will come back to that in the next chapter.

## 3.6 Summary

Let's quickly summarize: Alpha-Beta is a powerful search algorithm for game trees. Its drawback is that it requires domain-specific knowledge for the evaluation function. Also, the branching factor of the game must not be too large — in the extreme case, alpha-beta will search only one level deep, and essentially boil down to just directly applying the evaluation function once.

Monte-Carlo Tree Search is domain independent and can be applied to any deterministic two player game. If both alpha-beta and MCTS work, usually alpha-beta is more accurate and faster. But for games with a very high branching factor and/or no good evaluation function, MCTS can provide a powerful alternative to alpha-beta search.

Speaking of domain independence — is it a feature or a curse? In terms of chess we *could* actually use a lot of domain knowledge. After all, for a computer it *is* much simpler to assess a position in chess compared to say Go. Even pawn counting provides a meaningful evaluation in a lot of cases.

As for the timeline of MCTS: Brügmann [Brü93] first applied Monte-Carlo techniques, i.e. random playouts to the game of Go. Coulom [Cou06] then applied Monte-Carlo techniques to tree search and combined these two. Finally, Kocsis et al. [KS06] made the connection with the multi-armed bandit problem, and defined UCT.

# 4

# Modern AI Approaches - A Deep Dive

> I had seen some so-called research in AI that really deserved the bullshit label.
>
> Feng-hsiung Hsu

AlphaGo was the breakthrough. The major achievement that both researchers and the public will talk about in future generations. Go was not just another game that was mastered by computers, it was *the* defining open question in artificial intelligence. As mentioned before, the extremely high branching factor and the difficulty of defining objective criteria to evaluate a game position were the reasons that Go was deemed unsolvable at the time being and in the future. Therefore this major achievement should not be understated.

Still, one has to put this achievement in context. The theoretical groundwork had been laid out before. The foundation of neural networks and learning dates back to the 1980s and 1990s — there is a vast history and here we only mention [Fuk80, LBD+89, LBBH98] as examples. Network designs and structures were further explored and enhanced in the early 2000s. Still nobody deemed it possible to create a Go engine with these techniques. Not only are very deep and large networks often difficult to control — when a network fails to converge there is often no direct cause that one can identify, and it takes a lot of practical experimentation to create suitable network designs for the problem in question. But also the sheer amount of computational power was estimated to be too large. No university could afford to spent that amount of computing power to a board game with a high possibility of failure and wasted resources. The DeepMind team with the backing of Alphabet Inc. however, could.

Nevertheless, if we take a look at the approach that was taken — and we will, in the next section — one cannot deny that it looks a little bit crude and overly complicated. This is not untypical of a scientific breakthrough: You just try out a bunch of different things, re-combine them and mangle with different techniques until you find something that finally works. Still, it is very interesting to study the techniques that were used in this breakthrough, which is subject to Section 4.1.

However, the simplifications introduced by Alpha(Go)Zero are by no means less of an achievement. The whole learning pipeline was simplified and combined with the search part — making the overall algorithm much more efficient and easier to implement as well. The techniques of AlphaGoZero and AlphaZero are introduced in Section 4.2 and Section 4.3 — and if you are merely interested in state of the art neural-network engines for *chess*, it makes sense to directly start from there.

Unfortunately, the implementations of AlphaGo and Alpha(Go)Zero were never made public. Leela Zero (Go) and Leela Chess Zero were attempts to re-implement the algorithms presented by the DeepMind team in an open-source manner and make them available to the general public. Whereas DeepMind had a server farm at their disposal, the enormous computational power required for training was provided by volunteers all over the world who shared their computer's power to the benefit of the global chess community. As developers were re-implementing the algorithms presented in the scientific papers of the DeepMind team, the underlying hardware that the DeepMind team used (specifically designed processors dubbed tensor processing units or TPUs) was not available. Moreover research in neural network design progresses fast, and all this required some slight changes in the architecture of Leela Chess Zero compared to Alpha(Go)Zero. LeelaChess Zero is subject of Section 4.4

Fat Fritz (and especially later Fat Fritz 2) made quite some headlines when they were introduced to the public by Chessbase GmbH and sold commercially - after all, they are heavily based on open-source implementations with certain changes. We will take a look at Fat Fritz in Section 4.5 and also try to give a (subjective) evaluation whether we should put Fat Fritz more in the copycat category or in the legitimate improvement category.

Another real achievement and scientific breakthrough happened without much publicity: Efficiently Updateable Neural Networks, abbreviated NNUE (in reverse order). NNUE was originally developed for Shogi. In chess, the results by AlphaZero were quickly reproduced by a worldwide community of volunteers. The Shogi community is naturally much smaller, and therefore no powerful widely available AI based engine appeared; the results of AlphaZero could not be reproduced without the much needed computing power. Yu Nasu, a computer science researcher and computer Shogi enthusiast came up with the novel idea of designing a neural network for position evaluation that was optimized to run on off-the-shelf CPUs. Whereas the network itself is certainly not even remotely as powerful as a deep convolutional network like the ones used for AlphaZero, it is on the one hand still more powerful than a handcrafted evaluation function, and on the other hand fast enough to be combined with alpha-beta search. The beneficial impact of this network design was quickly identified by the Shogi community but somehow overlooked by the chess or AI community in general. Part of the reason maybe that the original paper about NNUE was available in Japanese only.

The computer Shogi scene took a lot of inspiration from the Stockfish project, e.g. the very efficient implementation of (alpha-beta) search with all optimizations and search enhancements. A Japanese AI programmer that goes under the pseudonym *yaneurao* then felt that the Shogi community should "give back" to the chess community, and created a port of the NNUE network structure back to Stockfish, replacing the handcrafted evaluation function of Stockfish by the neural network. This yielded a chess engine that not only beat LeelaChess Zero, but was also capable of running on commodity hardware — with no graphic card at all! This amazing step forward is subject of Section 4.6.

Last, but not least, Chessbase took the existing open source code from the Stockfish project with the NNUE enhancement and trained the neural network in a slightly different way. The result was then Fat Fritz 2, being subject of Section 4.7.

Most attempts at using neural networks for chess engines aim at making chess engines *stronger*. A completely different research direction is to use them to make engines play more *human-like*. An experimental engine that takes this direction is *Maia* which is subject of Section 4.8.

## 4.1    AlphaGo

At the heart of AlphaGo are five neural networks:

1.  A policy network that gets as input a Go position (plus some extra features) and outputs probabilities for moves.[1] This network was trained with millions of positions from games of top Go players, i.e. for training the input was the Go position, and the expected result was the move of the top player with probability one, and the other moves with a probability of zero. This network is dubbed the *SL policy network*. Here, SL stands for *supervised learning*.

2.  The SL policy was then improved by policy gradient reinforcement learning into the *RL policy network*. RL stands for *reinforcement learning*. The underlying idea is here to start with the SL policy network, play lots of games against itself and observe the result. With these observations as training data, the network is then improved and this is the first iteration of the RL network. Take the RL network, let it play against the SL policy network, obtain training data, and improve the RL network. Then continue this approach and iterate a lot of times, but each time take random previous instances of the RL network as opponents (as if you would play against lots of different opponents with different strengths and weaknesses). Finally, obtain the RL policy network.

---

[1]The term *policy* network stems from AI research. Here, a policy defines how an actor behaves in a given situation. In the context of board games, this simply means that the actor is a player, the situation she is in is defined by the board position, and the policy defines the next move.

3. A policy network dubbed *rollout policy* that takes as input tactical information and patterns of a Go position and the last move that led to that position and outputs probabilities for moves. This network is very simplistic, being a linear softmax classifier. Its advantage is that getting move probabilities for a Go position is several dimensions faster than using the large SL policy network — reported values are two microseconds to compute move probabilities for the rollout policy compared to three milliseconds(!) for the SL policy network. The network is trained similar to the SL policy network, i.e. by taking positions from games of top Go players together with the moves chosen by the players as training examples.

4. A policy network dubbed *tree policy*. This one is very similar to the rollout policy network, but uses more input features. It is thus slightly slower than the rollout policy, but still much faster than the SL or RL policy networks. Whereas the rollout policy network is used during the simulation step of MCTS, the tree policy is used during the expansion step. Training for this network is done in the same ways as for the rollout policy.

5. Last, train a *value network*. The value network receives as input the current position, and outputs a value, essentially predicting how likely it is to win or lose from that current position. More precisely, it outputs a value that predicts who is going to win if the game continues from the current position and both players play according to a given policy. To train this value network, first the (final) RL policy network was used to play a lot of games, thereby creating training data consisting of positions as input, and results of how the game ended when played according to the RL policy network.

After these five neural networks are created, we're done with all training. The Go engine then works as follows: Monte Carlo Tree Search is used for a given position to determine the next best move. The MCT search however differs heavily from plain UCT-based MCTS. Both the expansion and simulation steps are guided by the fast SL policy network and the value network[2].

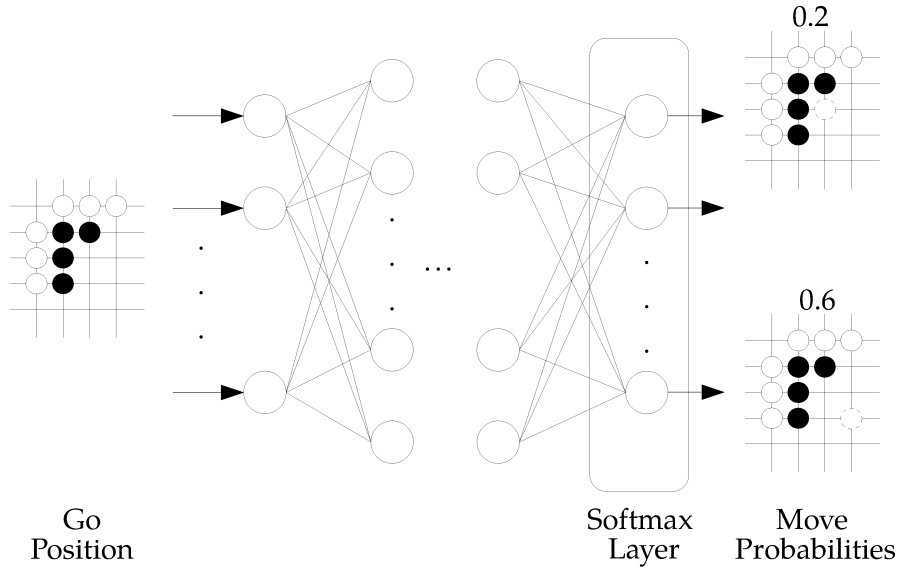

Figure 4.1: AlphaGo's SL Network Structure.

---

**Listing 4.1: RL Network Training**

```
1  RL Network := SL Network
2  Opponent    := SL Network
3  for 1...n training loops:
4      - play lots of Games of RL vs Opponent
5      - RL Network := ReinforcementLearning(RL Network, Games)
6      - Opponent   := random select RL Network from 0...i-1
```

---

A graphical overview of the architecture and training is given in Figure 4.1, Figure 4.2 and Listing 4.1. We will now take a deep dive into each of these components. First we'll examine the networks in detail. Then we will have a look at the modified Monte Carlo Tree Search.

---

[2]The RL policy network is not used at all at this stage. But note that the RL policy network was essential to actually create the value network in the first place

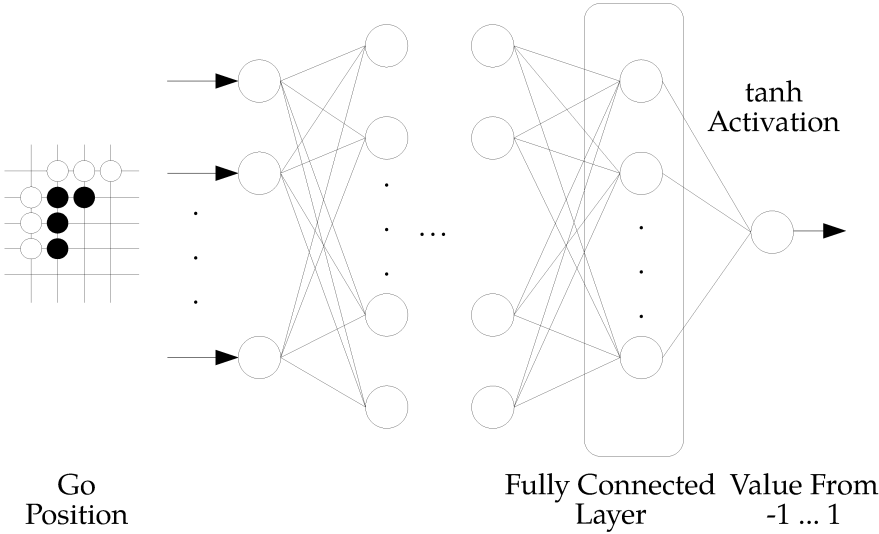

Figure 4.2: AlphaGo's Value Network Structure.

### 4.1.1   The SL Policy Network

**Network Input**    The input of the network consists not only of the current state of the board, but also some extra features that give additional information. All these input features are encoded as *binary values*. Of course we could think of other ways to encode input features, e.g. with some integer encoding of numbers as is typical in computing. However designing the input as simple as possible helps the network to train faster, as it directly sees the input and does not have to spent time learning the encoding.

A go board is 19x19 lines, which naturally results in 19x19 values. Such a 19x19 array is called a *plane*. Each value of the plane is binary, i.e. can be either 0 or 1. All in all there are 48 input planes. In other words, the input to the network are $19x19x48 = 17328$ binary values. You can imagine that this is a *large* network indeed.

But let's start with the encoding of the current board position. If we have only binary values available, we need three plains to represent the current position: One plane where a 1 indicates that there is a black stone placed at that position and where a 0 denotes that there is not a black stone placed there. Then one more plane to encode the white stones, and finally one plane to encode all empty fields. This is very similar to the encoding that the successor AlphaGoZero used; cf. Figure 4.6. Note that here, we always encode a board position from the perspective of the player whose move it is. That is, the first plane contains the position of the black stones if it is Black to move, and it contains the position of the white stones if it is White to move. The second plane contains the position of the stones of the respective other side. The third plane contains the positions of all empty points. The idea of this network is to create a neural network that outputs the best move for a given position for the side that is about to move. It really does not matter if the stones of that side have White or Black color.[3]

Then there are two categories of additional features used as an input. The first category is some simple additional information that makes it easier for the network to identify the border of the Go board. It has to do with convolutional filters of which the neural network makes heavy use of.

Recall how a convolutional filter works (cf. Chapter 2). In particular let's take another look at Figure 2.13. A convolutional filter works by putting an $n \times n$, for example a $3 \times 3$ kernel over an image (or here a 19x19 Go board representation). Then, depending on the filter size, a certain number of pixels (here fields of the Go board) are *convoluted* into one pixel. If we do this in a straight-forward manner over an image of say, size 4x4, we end up with a smaller board after applying the filter. Since we want to keep the size of the Go board over all layers of the network, *zero padding* is applied. For the SL network, the 19x19 board input is zero-padded into a 23x23 board. However doing this without any additional input makes it hard for the neural network to distinguish whether 0's at the edge are actually really zeros that are a result of the board position, or are a result of the zero padding. In other words, the edge of the board gets fuzzy or blurred. The solution to this is to add a constant plane of size 19x19 that has just ones, and another constant plane of size 19x19 of just zeros.

The other category of input features is encodings of simple properties of the board position. Technically they are not required. A network could simply *learn* these features itself by lots of training. However in order to reduce the size of the network and training time, it makes much sense to pre-calculate this simple information and add it as input features, so that some burden of learning is taken away from the network.

A lot of these features are trivial to recognize for those who are familiar with the game of Go. What is actually interesting is not necessarily the choice of these input features, but rather the *one-hot* encoding style, which is used for all of these additional features. One-hot encoding means you have one bit for every category. Therefore, we will not discuss all of these features, but only discuss the encoding of *liberties* as an example.

---

[3]Technically this is not true. As Black has an advantage by moving first, there is compensation in the form of *komi* points given to the white player. But this is not relevant if we want to find the best *move* in a given position, only if we want to predict the game *outcome*. Hence the color (and thus the komi) is encoded as an input feature in those networks that predict outcomes.

A liberty in Go is simply an empty adjacent point next to a placed stone. When there are connected stones of one color, their liberties become important. Remember that for a *group*, liberties are needed for the group to survive; cf. Section 3.5.1. Since we want to pass this information as input features to the neural network, we create eight 19x19 planes. In the first plane, each point with a placed stone that has one liberty is set to 1, and the other points are set to zero. In the second plane, each point with a placed stone that has two liberties is set to 1, and the other points are set to zero. We repeat this scheme up to the seventh plane. Finally in the eight plane all points with stones that have eight *or more* liberties are set to 1, and the other points are set to 0. This style of encoding is known as *one hot encoding*.

This makes it easy for the network to recognize if a group is alive or not and whether it can be captured or not, simply by determining if there are enough liberties.

The same encoding style is used for four more of such kind of tactical information about the current board position. Together these make up for 19x19x8x5 input features. Three more planes are used to encode simpler information and only take one plane each.

All in all we have for the network the following features: The points with stones of the player, the points with stones of the opponent and all empty points (three planes), a constant plane of ones and a plane of zeros (two planes), one-hot encodings of how many turns since a move was played, the number of liberties, how many stones of the opponent would be captured if a stone would be placed at a given point, how many of one's own stones would be captured if a stone would be placed at a given point, how many liberties there are if a stone would be placed at a given point (eight planes encoding this information for each point for each for these five features), whether a stone placed at a given point is a ladder capture (one plane), a ladder escape (one plane), and whether a move is a legal move at a given point in the first place (one plane).

**Network Output**   The network output consists of $19 \times 19 + 1 = 362$ move probabilities, i.e. for all possible moves, including the pass move, a probability is given, and the sum of all probabilities add up to 1. This is achieved by applying a softmax activation function at the final output of the network.

**Network Architecture**   The architecture of the SL policy network is shown in Figure 4.3. The first step consists of zero-padding the $19 \times 19$ planes into $23 \times 23$ to not shrink the input when adding a subsequent convolution layer. As discussed before, in order to make sure the network recognizes the size of the board as $19 \times 19$, two constant planes of ones and zeros are added as input features.

Next a convolution layer is added with 192 filters[4], stride 1, and kernel size of $5 \times 5$.

The next layers make the network a truly deep one. The following structure is repeated ten times: First the network is zero-padded to have $21 \times 21$ planes, next a convolution layer is applied with 192 filters, kernel size $3 \times 3$ and stride 1, and afterwards a rectifier linear unit (ReLU) is applied.

Finally a convolutional layer with only one filter, kernel size $1 \times 1$ and stride 1 is applied, before using a softmax activation function that generates the output of the network.

**Network Training**   There is a free online platform where players can compete against other Go players, the KGS Go Server[5]. The DeepMind Team acquired the database of all games, extracted games from higher-rated players, and sampled about 30 million positions together with the move played by the expert player in that position. These positions together with the move were then used to train the SL policy network by stochastic gradient *ascent* to maximize the log-probabilities of the moves; cf. Section 2.3.

---

[4]Actually, 192 filters were used in the AlphaGo version that played matches against the European Go Champion Fan Hui. They also tested different numbers of filters, but in their setup 192 seemed to give the best trade-off between strength and evaluation time.

[5]`http://www.gokgs.com`

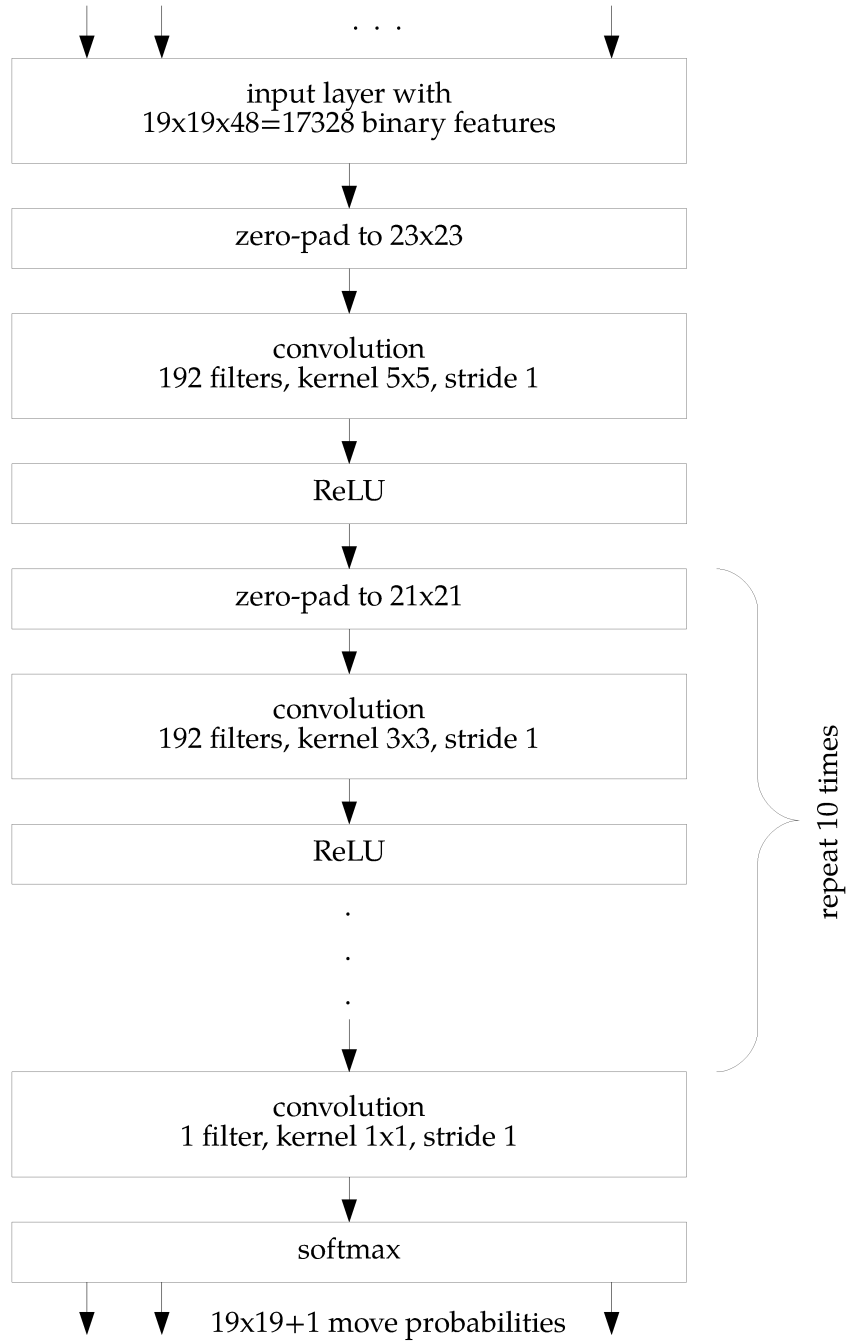

Figure 4.3: AlphaGo's SL and RL policy network architecture.

The trained network predicted 57 percent of expert moves from a test set of positions. This actually does not sound much, but was an increase of more than 10 percentage points compared to the state of the art.

Unfortunately, there are no precise numbers given on how much it took to train the SL policy network, and what hardware architecture was used (much more precise information is given for AlphaZero). However from [SSS+17] we can infer that *"AlphaGo Lee (the version that played against elite Go player Lee Sedol) was trained over several months"*, even though that leaves open how much time was spent on training the SL policy network, and how much time was spent on improving the SL policy network to create the RL policy network.

### 4.1.2 The RL policy network

The RL policy network is created by taking the SL policy network and improving it by applying *reinforcement learning*. In other words, network input, network output and network architecture are the same as with the SL policy network; cf. Figure 4.3. We thus have to take a look at how the reinforcement learning procedure works.

**Policy Gradient Reinforcement Learning** The original paper by the Deep-Mind team does not explicitly mention why the policy network was not further trained by more human experts. That is to create examples by sampling positions from games of very good players together with the move played by the player in that position. The simple truth might be that there were no more than 30 million positions available. The question is now how to get more training data. The idea is to generate training data by self play and apply *policy gradient reinforcement learning*.

Remember that when having expert games, the training of the policy network works roughly like this:

1. feed one (or a batch of) positions in the neural network

2. for each, get the output vector of move probabilities (via a softmax activation function)

3. compute the partial derivatives of the log-probabilities, i.e. compute

$$\frac{\partial \log(p(m|s))}{\partial \omega}$$

where $p(m|s)$ is the probability that the network outputs for move $m$ that was played by the expert human player in position $s$.

4. use those partial derivatives to perform gradient ascent (resp. descent with the negated logarithm) and update the weights of the network.

In that procedure we train the move policy by using the move labels. We just assume that the human expert knows what he is doing. The intuition between policy gradient reinforcement learning is to just randomly generate samples, observe the result of the game, and then use the result to gradually tune the moves by gradient ascent such that the policy networks plays moves which are more likely to win the game.

After all, if the game was won in the end, we did *something* correct when playing the game. Some moves must have been good moves. The rough procedure, namely *policy gradient reinforcement learning* is then

1. play a game of *random* moves. When the game is finished, we have a result of +1 (Black wins) or −1 (White wins). Note that in general, there are no draws in Go.

2. feed one (or all) positions into the neural network and get an output vector of move probabilities (via a softmax activation function)

3. compute the partial derivatives of the log-probabilities. However if the game was won, we want to *reward* the moves that were taken. Hence we multiply the probabilities by the outcome of the game, i.e. compute

$$\frac{\partial \log(p(m|s))}{\partial \omega} * (+1)$$

if the game was won, and compute

$$\frac{\partial \log(p(m|s))}{\partial \omega} * (-1)$$

if the game was lost. Here, $p(m|s)$ is the probability that the network outputs the move $m$ that was *actually taken* in the random playout.

4. use those partial derivatives to perform gradient ascent and update the weights of the network.

The logic is here: The moves that were taken in positions that eventually lead to winning the game must be all good, so we should tune the networks such that the probabilities of taking these moves in those positions are maximized. On the other hand, the moves that were taken in positions that eventually lead to losing the game must be all bad, so we should tune the network such that the probabilities of taking these moves in those positions should be penalized.

Actually, is that true? Thinking of chess, there are several games with 1.e4 in the initial position where White wins, and then there are also several games were White loses. Is 1.e4 a good move in the initial position or not? Are we not alternatively maximizing and minimizing the probability of such a move when we encounter games where White wins and White looses? In other words, does the final outcome really have a lot to do with the action we took a long time before the end of the game?

Well, turns out it actually does. You just need to play *a lot* of games. Proving that policy gradient reinforcement actually converges to an optimal policy (albeit a possible local optimum, as this is gradient descent/ascent as introduced in Chapter 2) is complicated and involves lots of math. If you are interested, a good starting point is [SB18].

Policy gradient learning can easily fail to converge in practice. This is probably why they first trained a network by supervised learning (i.e. learning by using human expert moves) to kickstart a powerful network, and then used policy gradient reinforcement learning only to *improve* the existing network. There were also some tricks used to prevent overfitting of the network, i.e. the games are not really played randomly, but rather between the policy network and another policy network that is randomly chosen among some earlier iterations of the network. This training loop is depicted in Listing 4.1. Here we start by initially using the SL policy network both for the RL policy network and the opponent. Then we apply a training loop. First we play a lot of games of the RL policy network against the opponent.  Using these games and their outcomes, we improve the RL policy network by gradient policy reinforcement learning. Then we randomly choose a next opponent among the previous incarnations of the RL policy network, and repeat the training loop.  This training by playing against *random* opponents as opposed to going with a fixed policy network is crucial here since otherwise the RL policy network would stick too much to the current policy.

Very interesting is by how much the existing SL policy network was improved by policy gradient reinforcement learning, resulting in the RL policy network. They report that the RL policy network won approximately in 80 percent of all games against the SL policy network.

Another interesting fact is how strong the RL policy network was compared to existing Go programs.  The Go program *Pachi* originated from Petr Baudis master thesis on Monte Carlo Tree Search [Pet11]. Pachi implemented state-of-the-art techniques for computer Go and used MCTS to find best moves.  Even though Pachi was not the strongest Go program at the time — some commercial vendors had stronger programs available — it was one of the strongest (if not the strongest) open source Go program and ranked high at computer Go competitions.  However by just taking the RL policy network without any kind of search, the RL policy network won 85 percent of all games against Pachi. Because it's so amazing, let's repeat that: *Without any kind of search, just by evaluating a position by the RL policy network and playing the move with the highest probability, the RL policy network won 85 percent of all games against Pachi.*

In the end however, we should not focus too much on technical details of policy gradient reinforcement learning.  It is a very general algorithm that can be applied to all kinds of games.  But in the context of Go, Shogi and chess, the DeepMind team found a much more elegant, simpler and faster approach, which we will take a look at in the chapter about AlphaZero.

### 4.1.3   The Rollout Policy Network

In standard MCTS, we use *random* playouts of games during the simulation step.  For Go, this is very simple and fast to compute.  Remember that move generation in Go essentially just involves finding a free point and placing a stone on it. There are a few checks for illegals moves required depending on the specific Go rules: Namely, there are different rules for Go depending on how these games developed in their respective country, but the differences are minimal. They center mostly around *Ko*, i.e. a situation where two single stone captures would repeat the same board position, and *suicide moves*, i.e. where you would capture your own stones which is not allowed. But all in all, it's very fast to do random rollouts of Go games compared to say chess, where the rules are much more complicated and legal moves are more costly to compute.

So random moves are fine, but it would be even better if we could do only *slightly* better than playing random moves, yet still very quickly decide on the next move. That is, to avoid obvious tactical blunders that even an amateur would spot. This is precisely the intention behind the rollout policy network. The rollout policy network is a *linear* softmax based network. Yes, linear means just one network layer; or formulated differently, just multiplying the inputs by a large matrix and applying softmax as the output function. But all in all this is still better than playing randomly, and still very fast. The DeepMind team reports that the rollout policy selects a move within just two microseconds. That's several dimensions faster than the SL or RL policy network, which use 3 *milli*seconds!

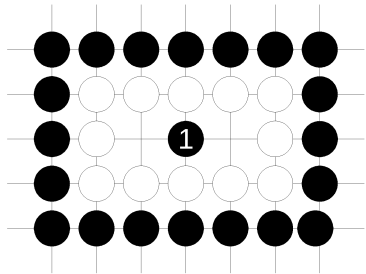

Figure 4.4: Nakade.

The rollout policy network centers around avoiding tactical mistakes and recognizing typical patterns during play. These patterns are somewhat handcrafted and are based on both the last move that led to the current position, and on patterns w.r.t. potential moves in the current position. They are all binary features, i.e. a 1 indicates that the pattern in question is present, and a 0 indicates that it is not present.

To give a better intuition on how this works, let's consider a small example, namely *nakade*. Consider the board in Figure 4.4. Can this group be alive? It is similar to the discussion of Go rules at the beginning of this Chapter. If Black can place a stone inside at the marked position (and this is what nakade actually means, i.e. putting a stone *inside*) than she has two eyes and the group is alive. If however it is White's move, we have the following variations. White will put a stone at the marked position. Then

1. if Black puts a stone left or right next to the white stone, White can put a stone on the last free point and capture everything.

2. if Black does not put a stone inside, then there are two points left open. White will put one stone next to the marked position, and only one point is unoccupied. Then

   (a) if White does nothing, Black will place a stone at the last unoccupied point and capture everything

   (b) if White places a stone at the last unoccupied point and captures Black's stones, two points are emptied. Black will put a stone on one of these unoccupied points and White will lose the group, too.

As you can imagine, there are a number of typical patterns that can occur on a Go board that fit into this tactical *nakade* category. Obviously they are highly important when figuring out a next move. What is done for the rollout policy network is to scan the $19 \times 19$ points, scan at each point whether a move there is a capturing move and whether this would match a *nakade* pattern. Then set a 1 if this is true, and a 0 otherwise.

The input of the rollout policy network consists of a 109747 bit input vector. Each bit indicates if one of the patterns was detected for a given point w.r.t. the last move that led to the position. Note that using such kind of handcrafted pattern detection to identify tactical motifs is not uncommon in computer Go, and had been heavily used in state-of-the-art Go programs. Compared to those however, the amount of handcrafted features that AlphaGo uses is rather low.

### 4.1.4 The Tree Policy Network

The tree policy network is also a *linear* softmax based network. Essentially, it is the same network as the rollout policy network, but uses approximately a 30 percent larger input vector. Simply the amount of patterns was increased. This increases computational cost when calculating the move probabilities for a given position, but it is still much faster than utilizing the SL policy network.

### 4.1.5 The Value Network

**Network Output**  Let's start with the output of the network for once before taking a look at the input. Suppose you have a move policy that, given a Go position, tells you what the next best move is. Now suppose further you are given an arbitrary Go position and have the policy play against itself from that position until the game ends. The output of the value network is just a single value between −1 and 1. It denotes the expected result of the game from a given position that is played according to a given policy. It is important not to confuse this and interpret it as if the value network answers the question "Who will win the game?", because instead it answers the question "Who will win the game if the game continues according to the given move policy?". The latter question is much more limiting, since the answer could be totally worthless if the move policy is garbage. In other words, we need a very good move policy in order to make the value network useful!

**Network Input**  The input to the value is similar as for the SL and RL policy network, namely a Go position as well as additional features. As mentioned before, for the SL and RL policy network the Go position is always encoded from the perspective of the player whose turn it is. Consider Figure 4.6 for example where it is Black to move. Therefore the position would result in the three planes encoding the position of the black stones, the White stones and the empty stones. If it is White's turn however, the encoding would simply switch the colors; the bit plane for the black stones becomes the one for the white stones and vice versa. This is very handy, since the network can train to predict the best move regardless of having to care about colors and turns. After all, we just need output probabilities for all possible next moves, and to get this output we don't need to care about the color.

For the value network the situation is different of course. We need to be precise if it is Black's or White's turn. This is not just about switching signs! Similar to chess, having the first move could make a huge difference in a given position; cf. the footnote about *komi* when discussing the SL policy network. Hence, the input of the value network is the same as for the SL/RL policy network *plus* an additional plane consisting of either only ones or only zeros that indicates whether it is Black or White to move.

**Network Input**    The structure is pretty much the same as for the RL and SL policy networks. The input layer is the same except that we need to now also input one plane that encodes the side that is about to move. The main difference is of course the output layer. Instead of outputting move probabilities, we want to predict the outcome of a game. Adding a fully connected layer with a tanh activation function is used for this. The tanh ensure that we always get an output in the range $-1$ (White wins) and 1 (Black wins). Ideally the network would always output either 1 or $-1$. However since the network is unable to exactly predict the outcome for every possible Go position it will output intermediate values.

**Network Training**    The network is trained by 30 million Go positions together with the outcome of the game when continuing from this position by self-play with the RL policy network. These Go positions itself stem from 30 million fully self-played games by the RL policy network from the initial position. One position plus the outcome was then selected from this game.

The average length of a Go game is within a range of about 200 moves. In other words one game of Go gives us about 200 positions. Then why should we not just self-play 150000 games, and sample 150000 * 200 = 30 million positions from it? It surely would be much faster to just play 150000 instead of 30 million games!

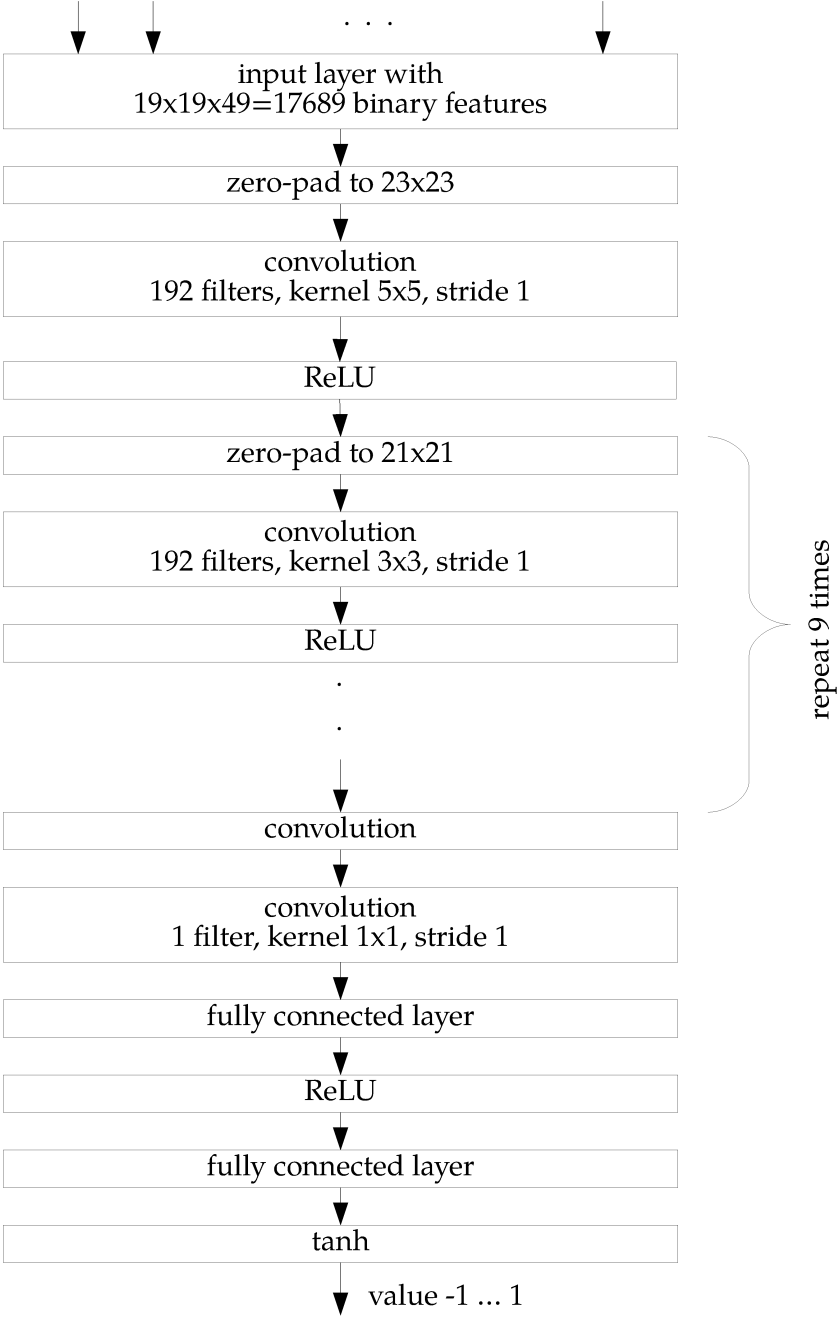

Figure 4.5: AlphaGo's Value network architecture.

The reason is overfitting. When the network was trained with full games, the network would adjust too much to the specifics of these games. In other words the network tended to *memorize* full games, instead of abstracting the underlying patters and actually *learn*[6] from the games.

As for the training method itself, stochastic gradient descent is used w.r.t. minimizing the mean-squared error between the predicted outcome of the network and the actual result that is the outcome of self-play by the RL policy network.

### 4.1.6   MCT Search

After all networks are trained, the actual Go engine is based upon MCT search. The MCT search follows the standard MCTS principle, i.e. consists of the steps *selection*, *expansion*, *simulation* and *backpropagation*. However where UCT-MCTS was solely based on random tryouts, the MCTS procedure here makes heavy use of the trained neural networks:

- During the *expansion* step, both the tree policy network as well as the SL policy network are used to identify promising moves and prioritize on these identified moves. The tree policy network is used to very quickly get probabilities for moves in the leaf-position. The SL policy network takes longer to process the leaf-position, but as soon as its output is available, the probabilities of the tree policy network are replaced by the probabilities of the SL policy network.

- During the *simulation* step, the current position is evaluated in two ways: First, by applying the value network and getting an expected result. Second, by games self-played by the rollout policy network. These two results are then combined using some weighting factor.

Let's have more detailed look into how this works. In UCT-MCTS we stored two values, the visit count as well as the number of winning games (or some win ratio). Moreover we could describe all steps as operations on nodes. Since we here have probabilities on moves, we associate the following three values with *edges* (i.e. with the moves that connect Go positions):

1. A $Q$ value; dubbed the action value. Here, we will simply call it the *move value*. This value denotes how good a move actually is. It is thus a similar measure compared to the number of winning games in standard UCT-MCTS, but its calculation takes all the network outputs into account.

2. A value $N$; dubbed the visit count. This is the same as in standard UCT-MCTS.

3. A value $P$; dubbed *prior probability*. This stores the probability of a move given by the tree policy network (to get a quick estimate) and as soon as an evaluation of the SL policy network is available, it permanently stores the output value of the SL policy network.

---

[6]I find the similarity to humans interesting. Especially beginners in chess often tend to memorize opening lines without really understanding the ideas and structures instead of tackling more difficult areas like endgames where probably much more can be learned about the game. At least personally I cannot say that I am free of guilt in that area...

The steps of MCTS then work as follows to update these values $Q$, $N$ and $P$:

- *Selection.* Starting from the root, a successor node is selected until a leaf-node, i.e. a node with non-visited moves is found. At each step during selection in a position, the next move is chosen by comparing the sum $Q + u$ of the $Q$ value of that move as well as some bonus $u$. This sum is compared w.r.t. all moves, and the one with the highest value is selected. The bonus $u$ is proportional to $\frac{P}{1+N}$[7]. If we start the search and we do not encounter a leaf-node, we will have visited the node very few times, say once, i.e. $N = 1$. The value $P$ is just a constant probability given by the SL policy network. For the sake of the example, suppose this is 0.5. The initial bonus would thus be $\frac{0.5}{2} = 0.25$. Now let's say that we have visited the node 99 times. The bonus then decays to $\frac{0.5}{100} = 0.005$ which is neglectable. In other words: Initially we have not run a lot of simulations, and therefore trust the SL policy network to pre-select good moves. Once we run more and more simulations, we trust more and more the result of the MCT search and minimize the effect of the SL policy network.

---

[7]The correct formula to compute $u$ for some move $m$ is actually $c \cdot P \cdot \frac{\sqrt{\sum_{m'} N_{m'}}}{1+N_m}$. Here $\sum_{m'} N_{m'}$ sums over the node count of all moves available in the current position, and $c$ is some empirically chosen constant factor that adjusts the effect of $u$

- *Expansion.* In this step, prior probabilities $P$ for each possible move are computed by the SL policy network. Since that takes a lot of time, the prior probabilities are first computed by the tree policy network, and once they are available from the SL policy network, immediately replaced.

- *Simulation.* The leaf node is then evaluated in two kinds of ways: First the value network outputs an expected outcome, say $v$. Then a game is played by self-play using the rollout policy network, which results in a result, say $z$. These are then summed using a *mixing parameter* $\lambda$ into a leaf evaluation value $V$, i.e. $V = (1 - \lambda)v + \lambda z$.

- *Backpropagation.* We now go upwards from the leaf node with new evaluation value $V$ in the tree up to root. For each edge, i.e. each move, we update the visit count with $N = N + 1$ by one. Accordingly, the $Q$ value of that move is $Q = \frac{1}{N+1}V$.

Now you have some intuition about the calculations that go on during the MCT search of AlphaGo. There are some challenges which are technical in nature: Searching iteratively like this is simply too slow. Since all operations, especially the ones involving the SL policy network and the value network are costly, the MCTS algorithm was heavily parallelized. Computations of the SL policy network, the value network, playouts by the rollout policy network, the computation of initial probabilities by the tree policy network and MCTS operations on the tree happen in parallel and are aligned with each other. Therefore, MCTS operations are slightly more complex in practice. The backpropagation step for example can then be simply extended to increase the visit count w.r.t. all simulations that passed through that move. The $Q$ value of a move is then updated by taking the mean w.r.t. all simulations that passed through that move. Unless we want to explicitly implement AlphaGo ourselves, this is of minor importance however.

### 4.1.7 Odds and Ends

In many ways the whole approach taken looks slightly odd. That is not meant in any way to criticize the tremendous achievement that AlphaGo constitutes. It looks however similar to a pattern that often occurs when great achievements were made. First one discovers one particular part that works extremely well. Based on that part one tries to create a complete solution by trying out various things, and prioritizes solutions based on engineering rather than questioning the whole solution in the first place. After making everything work, one can then afterwards re-study the solution, identify strengths and weaknesses and optimize and simplify everything.

We can only speculate in which order things were discovered. One explanation could be that training the SL policy network by millions of Go positions from the KGS Go server was the first step. It was not the first time this was tried, but the best previous approach reached an accuracy of 44.4 percent [CS15] whereas the SL policy network of the DeepMind team reached an accuracy of 57 percent. This alone constitutes a major improvement. We can then speculate that based on this achievement they tried to make MCTS work efficiently. The introduction of the rollout policy and tree policy networks suggests that the performance of MCTS was a major obstacle. This is likely the reason that they implemented a fully parallel MCTS. They report that the final version of AlphaGo used 40 search threads with 48 CPUs (for the search) and 8 GPUs (for the evaluation of the neural networks).

AlphaGo beat other programs by several magnitudes. The best Go program at the time was CrazyStone by French computer scientist Rémi Coulom. Whereas CrazyStone was approximately slightly below 2000 Elo, AlphaGo was just short of 3000 Elo points (it crossed the 3000 Elo margin with an even more parallelized version that used 1202 CPUs and 176 GPUs). One should mention however that Rémi was quick to adapt the neural network based ideas into CrazyStone and as of today it is again a very strong Go program.

AlphaGo also beat European champion Fan Hui in a small match over ten games by 8 vs 2. An enhanced version later beat the at the time best Go player Lee Sedol in a five game match by 4 to 1. Again, the tremendous achievement by the DeepMind team cannot be overstated.

Still we list some oddities which serve as a good reason to look into the even more tremendous achievement that was AlphaGoZero, described in the next Section.

- There are no more than five(!) different neural networks involved. The RL policy isn't even used directly, it just serves as an intermediary to train the value network. Moreover the SL policy network, the RL policy network and the value network share almost the same architecture (with the exception of the output layer), yet they are trained completely independent. Can this be streamlined?

- Oddly, the SL policy network is used to create the prior probabilities during MCTS expansion. One would expect that the stronger RL policy network is used for that. However the RL policy network was actually weaker than the SL policy network. The DeepMind team hypothesized that it is *presumably because humans select a diverse beam of promising moves, whereas RL optimizes for the single best move*. Nevertheless they report that the RL policy network *indeed* was stronger than the SL policy network when training the value network. So the reinforcement learning of the SL policy network to get the RL policy network seemed to be worthwhile.

- For simulation, both the value network *and* the fast rollout policy are used to get an evaluation for the position. Oddly, none of them alone reaches even closely the 3000 Elo level of everything combined; the value network alone reaches roughly 2200 Elo points, and the rollout policy network reaches approximately 2400 Elo points.

- After training all networks, a huge computational power is used to perform MCT search during which much insight is gained w.r.t. the game of Go itself. Yet this insight is not made use of for improving the networks, instead the MCT search is re-started (modulo some caching) in each new state of the game. In contrary, the rather universal gradient policy reinforcement learning is used to improve the SL policy network. Surely there is potential for utilizing the results of the MCT search somehow.

It is clear that the DeepMind team identified these issues. After all, all those issues were addressed and solved by the "zero"-style engines, which are subject of the next sections.

## 4.2 AlphaGo Zero

In case you started the chapter at the beginning and slightly struggled with the AlphaGo chapter with the five networks and their different training methods, policy gradient reinforcement learning and a complex MCT search, here is some word of comfort: It gets easier. Much easier actually.

Speaking in terms of metaphors: The breakthrough of AlphaGo can probably be described as building a pyramid by moving huge stones from one point to another by sheer brute force and muscle power. The breakthrough of AlphaGo Zero is then figuring out that the same can be accomplished much easier by inventing the wheel and putting the stones on a cart. The result is the same if not more efficient, but the method is way more elegant.

In AlphaGo Zero we have just one single network. The input of the network is much simplified, but more on that later. The output of the network combines the policy network(s) and value network of AlphaGo into one single structure. And instead of employing various different kinds of training methods for each network as done for AlphaGo, here the training is done by just using the results of MCT searches.

### 4.2.1 The Network

The network is a deep residual neural network, i.e. it is a deep convolutional neural network with skip connections. Compared to the networks used in AlphaGo, the use of skip connections is the biggest difference. Moreover it combines the policy and value network that were used in AlphaGo into one single network.

**Network Input**   The input of the network is the location of the Black and white stones of the current and the last seven positions as well as the side whose current turn it is. The last seven positions are required since there are rules w.r.t. avoiding repetitions of positions. The positions are always encoded from the perspective of the player whose turn it is. In Go, Black moves first. Having the first move is a slight advantage in Go. As mentioned before: To adjust for that there is a rule called *komi*. When a Go games finishes, one counts the occupied territory to determine the winner. Komi means that White gets a few extra points (agreed upon before the start) when counting the territory. Therefore it is important to know whether it is Black's or White's turn in a given position, because a position might look advantageous to Black but for example is just equal or even better for White when we consider komi. Aside from komi though, a Go position is mirror-invariant.

The input consists of 17 binary planes; each of size $19 \times 19$. The first plane encodes the current position of the stones of the current player. A bit is set to 1 if a stone of the player is placed on a point and in all other cases (a stone of the other player or empty) the bit is set to 0. The next seven planes encode the positions of the stones of the current player of the last seven positions that occurred previously in the game. The next eight planes encode the same for the other player. Finally there is one plane with all bits set to 1 if it is Black to play, and 0 if it is White to play. Figure 4.6 gives a brief example of this encoding. In particular no Go-specific features like tactical information are provided as the input. The input really is just the bare minimum to determine the current state of the game.

**Network Output**   The output of the network is constructed of two *heads*, the *policy head* and the *value head*. The policy head outputs probabilities of all possible $19 \times 19 + 1$ moves — the one is added here for the option of passing, which is a legal move in Go. The value head outputs a number in the range $-1...1$ where $-1$ denotes that the player with the current move loses from this position, whereas 1 denotes that he wins from this position. Compared to AlphaGo, this network architecture allows to combine the policy and the value network that were used in AlphaGo into a single network. The benefit w.r.t. the computational effort is obvious.

**Network Architecture**   The overall network architecture is depicted in Figure 4.7. Again, it is interesting to compare the architecture to the one used in AlphaGo. Besides the two heads for the output of the move probabilities and the value, the biggest change is the heavy use of *residual layers* (skip connec-tions) compared to just using convolutional layers alone. As we have discussed in Section 2.12, residual layers can help to train a deep network much more efficiently which is apparently the case here. The feature inputs are initially fed to a convolutional layer with 256 filters, a $3 \times 3$ kernel and stride 1, followed by batch normalization and rectifier nonlinearity. Next there is a chain of *residual blocks*. The DeepMind team experimented both with a smaller network of 19 residual blocks and a larger network of 39 residual blocks. Each residual block consists of two convolutional layers with 256 filters, a $3 \times 3$ kernel and stride 1 followed by batch normalization and rectifier nonlinearity. Each residual block also features a skip connection that connects the input of the residual block directly to the output.

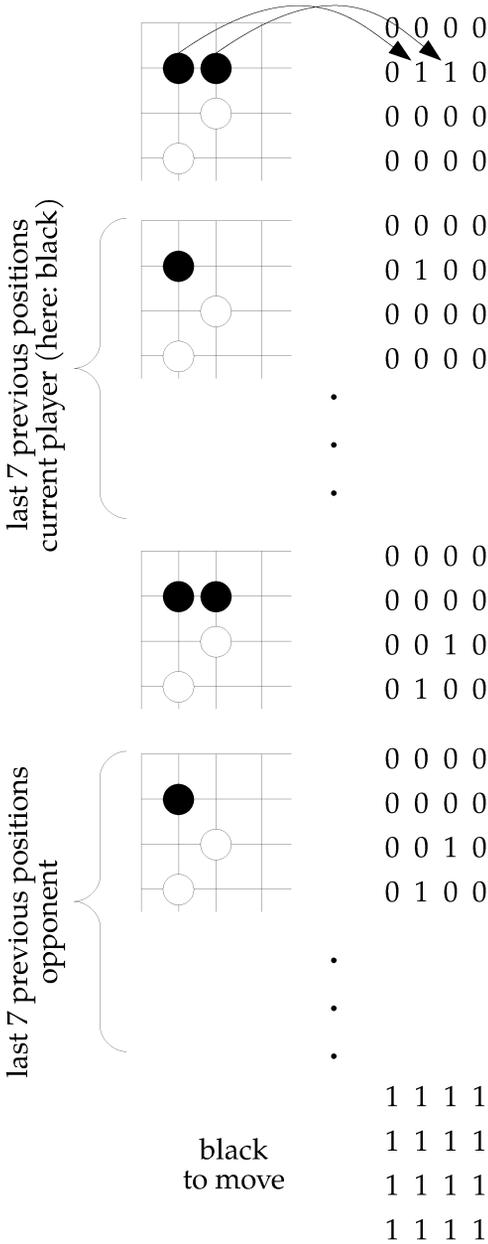

Figure 4.6: AlphaGo Zero: Input Encoding.

After these residual blocks, the output is fed into two heads. The *policy head* features one convolutional layer with two filters, a $1 \times 1$ kernel and stride 1 with batch normalization and rectifier nonlinearity followed by one fully connected layer with $19 \times 19 + 1$ outputs. The outputs correspond to the probabilities for each possible move on the Go board including the pass move.

The *value head* starts with one convolutional layer with one filter, a $1 \times 1$ kernel and stride 1 followed by — surprise, surprise — batch normalization and rectifier nonlinearity. This is then fed into a fully connected layer that has 256 outputs, followed by rectifier nonlinearity. Finally these outputs are connected to the inputs of one final fully connected layer with a single output and tanh activation which provides the value in the range $-1 \ldots 1$.

## 4.2.2   MCTS and Reinforcement Learning

Now suppose we already have finished training our network and are about to play a game of Go. We will now use MCTS to find the best next move in a given position. The MCTS approach is similar to the one used in AlphaGo but naturally much simpler since we have just one single network. Very interesting here to note is that *no random rollouts* are used in the simulation step, we just employ the network for the evaluation of leaf nodes. This is a major difference compared to classical MCTS. Let's discuss the four steps, namely *selection*, *expansion*, *simulation* and *backpropagation*. As before in AlphaGo we operate on a search tree where nodes correspond to Go positions; and edges that connect nodes correspond to moves. Each edge stores four values, the move value $Q$, some helper variable $W$ that stores the sum of move values to support the computation of $Q$, the visit count $N$, and a *prior probability* $P$. These values are initialized as $Q = W = N = 0$, and an evaluation of the current position with the network provides the initial probability value for $P$.

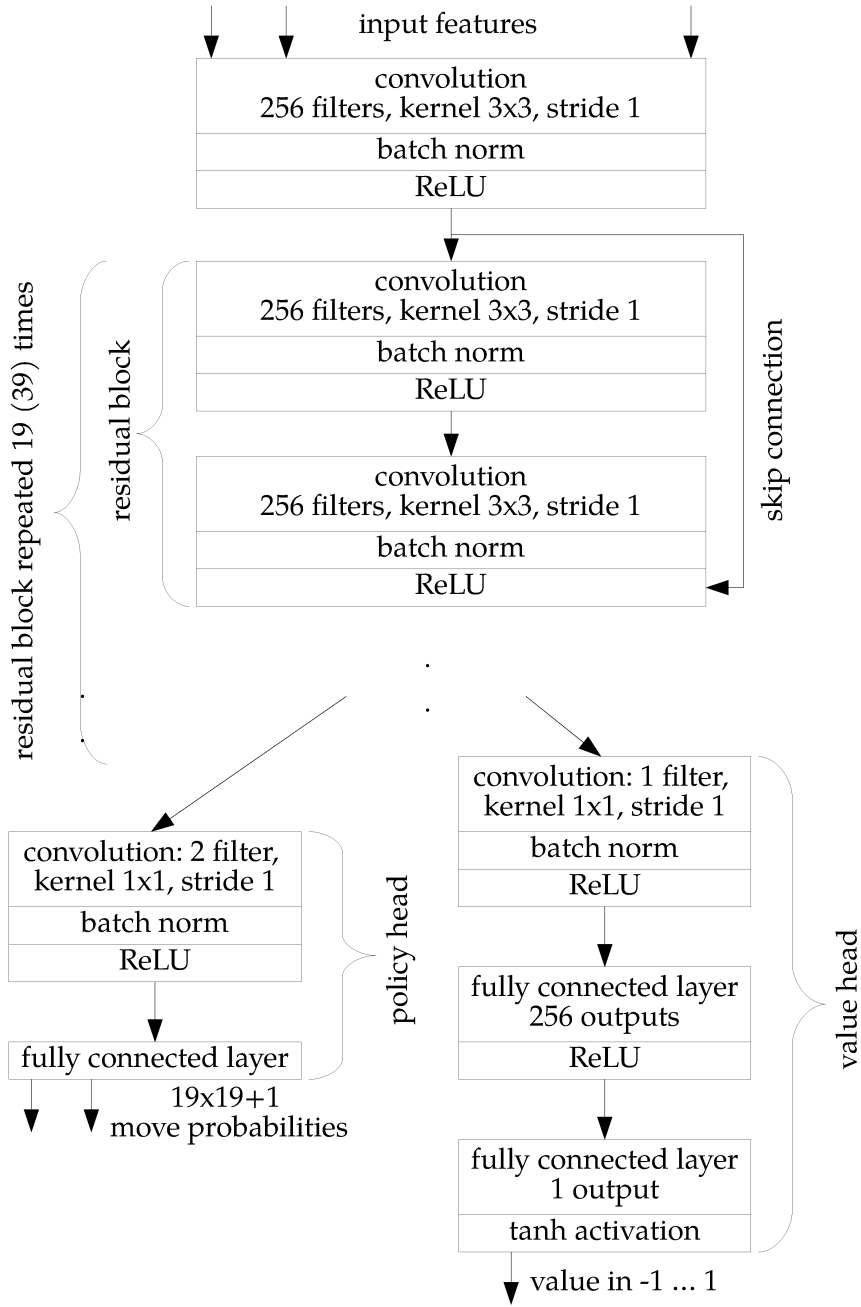

Figure 4.7: AlphaGo Zero: Network Architecture.

- **Selection**. We start at the root and at each step choose the next node by taking the one that maximizes $Q + u$. Given a move $m$, the value $u$ is computed as $c \cdot P \cdot \frac{\sqrt{\sum_{m'} N_{m'}}}{1 + N_m}$. Here $m$ is the move that leads to the next node, $\sum_{m'} N_{m'}$ sums up the node count of all possible moves (i.e. practically spoken that's the node count of the parent), and $c$ is some constant that adjusts the impact of the overall bonus value $u$. Similar as in AlphaGo, the value $u$ ensures that we first explore those moves with an initially high prior probability $P$, but later when we have searched more, we focus more on those moves with a high move value $Q$ — if we revisit the move $m$ over and over then $N_m$ grows and therefore $u$ gets smaller. We continue to select moves until we find a leaf node, i.e. a node with a position which has unvisited moves.

- **Expansion**. The position of the leaf node is evaluated by the neural network to get some evaluation value $v$. The network also provides probabilities $p_m$ for each possible move $m$ in the position of the leaf node. Then the leaf node is expanded by adding an edge for each possible move to a successor position. Each edge is initialized with values $Q = W = N = 0$ and the probability $P$ is set as $P = p_m$.

- **Simulation**. There is no explicit evaluation step anymore. The evaluation is combined with the expansion step: The computation of the evaluation value $v$ replaces the rollout simulations that we typically do in MCTS. We simply rely on the network for simulation.

- **Backpropagation**. Starting from the leaf node, the values $Q, W$ and $N$ are updated up to root. Each edge that is passed through is updated by $W = W + v$, $N = N + 1$, and $Q = W/N$.

Finally after the search (time) is over and a move has to be selected, we compute the following policy value for each possible move $m$ in the root position

$$\pi_m = \frac{N_m^{1/\tau}}{\sum_n N_n^{1/\tau}}$$

assuming $n$ possible moves at the root. The move $m$ with the highest value $\pi_m$ is then selected. This is almost the same as simply taking the move with the highest visit count, but the value $\tau$ slightly adjusts the level exploration.

As always, to get a better intuition of how this all works, let's consider a small toy example with actual numbers. Consider Figure 4.8. There are only two moves available in the root position. We have to decide whether we take the left or right move. For that we compute the value $U = Q + u$. For the left move we have

$$u = c \cdot P \cdot \frac{\sqrt{N_{\text{left}} + N_{\text{right}}}}{1 + N_{\text{left}}} = c \cdot 0.1 \cdot \frac{\sqrt{3 + 2}}{1 + 3} \approx c \cdot 0.56$$

For simplicity let's assume that our constant factor $c$ that adjusts the impact of $u$ to be simply set to 1. Then we have $U = Q + u = 0.2 + 0.56 = 0.76$. For the right move we obtain

$$U = Q + u = 0.4 + 1.0 \cdot 0.7 \cdot \frac{\sqrt{3 + 2}}{1 + 2} \approx 0.4 + 0.74 = 1.14$$

Since $0.56 < 1.14$ we select the right move. Now we continue and have to again compare the left and right move. For the left move we get

$$U = Q + u = 0.5 + 0.6 \cdot \frac{\sqrt{1 + 1}}{2} = 0.5 + 0.42 = 0.92$$

and for the right move we have

$$U = Q + u = 0.4 + 0.7 \cdot \frac{\sqrt{1 + 1}}{2} = 0.4 + 0.49 = 0.89$$

Since $0.92 > 0.89$, we take the left move. We are now in a leaf position and have to apply *expansion* and *simulation*.

The expansion step is shown in Figure 4.9. During expansion we generate all possible legal moves in that position and for each move create a successor node. Moreover we feed the position of the leaf node into our network and obtain move probabilities. For each legal move we store the corresponding probability as prior probability $P$, i.e. associate it with the edge. The remaining parameters $N$, $W$ and $Q$ are initialized to 0.

We already fed the position of the leaf node into the network and not only obtained move probabilities but also an evaluation value $v$ of that position. The *simulation* step is just that, i.e. obtaining the value $v$ for that position. During *backpropagation* we use that value $v$ to update all values going from the leaf node upwards in the tree up to the root note. This is illustrated in Figure 4.10.

The first edge on that path is updated by adding the value $v$ to the accumulated values $W$, i.e. $W := W + v = 0.5 + 0.7$. Moreover we update the visit count $N := N + 1 = 2$, and obtain the new move value $Q := \frac{W}{N} = \frac{1.2}{2} = 0.6$. We follow to update the next edge above with $W = 0.8 + 0.7 = 1.5$, $N = 2 + 1 = 3$ and $Q = \frac{1.5}{3} = 0.5$. This finishes one iteration of MCTS.

Note in particular how the comparatively good position evaluation of the leaf node $v = 0.7$ affects the move value of the upper layers, i.e. how the evaluation of the right move just below root is changed from $Q = 0.3$ to $Q = 0.5$.

Now we've seen how MCTS works when a powerful neural network is available to provide move probabilities and a position evaluation. But how do we train the network in the first place? Now here comes the real magic that makes AlphaGo Zero (and all derivatives) so elegant and beautiful: We just randomly instantiate the network, self-play games and then use the results of MCTS to train and improve the network. Then we continue.

Let's get down to the details. It works like this: First we self-play a game with MCTS. To compute the values during MCTS, we employ the current generation of the network. The self-played game will provide us with a sequence of positions and moves until the game ends with a win for Black or White. Let's call this result $r$. In each step of the MCT search we computed probabilities for all moves in the position. Note that these move probabilities are not just the result of one evaluation of the network (these are the prior probabilities $P$) but rather the policy values $\pi_m^i$ that we computed in each step $i$ for each possible move $m$. Let's refer to the vector of all policy values for all possible moves in step $i$ as $\pi^i$. In other words we have a sequence of triples

$$(s_1, \pi^1, z_1), (s_2, \pi^2, z_2), \ldots, (s_n, \pi^n, z_n)$$

until the games end. Here $z_i$ is the result of the game from the perspective of the player whose turn it is, i.e. we need to switch the sign of the result $r$ depending whether it was Black or White to move in that position.

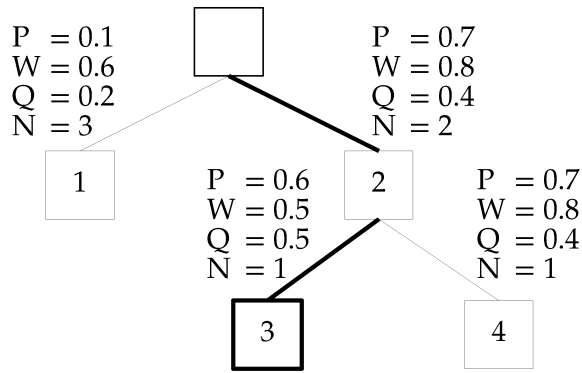

Figure 4.8: AlphaGo Zero: Selection.

But that's it! We have just generated training data for our network. We feed in a position $s_i$ into the network and expect as the output for the move probabilities $\pi^i$ and as the outcome of the game the value $z_i$. Again, note that these will be more precise than what the network currently outputs since they are the result of the MCT search.

Now we have an incredible powerful training pipeline! We can do this training for a number of games, and then replace the network that is used during the MCT search with the improved network and continue. We should mention here however that some tricks are used to generate even more training data. Note that a Go position is invariant to e.g. rotation and reflection (mirroring) — for example just rotating the board by 90 degrees does not change anything w.r.t.

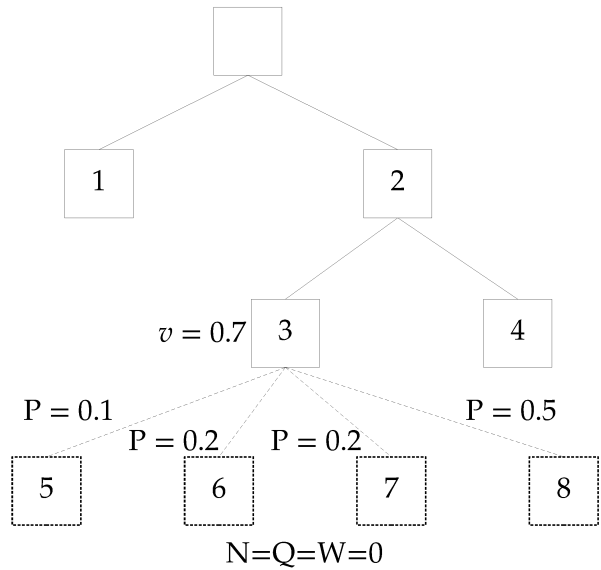

Figure 4.9: AlphaGo Zero: Expansion and Simulation.

the game state. For each Go position we therefore get all in all eight positions (rotations and reflections of the board) for free, thus significantly increasing the amount of training data. This also makes the network more robust as it will likely detect patterns in a position even if it only saw a rotated or mirrored position during training.

What is left to do is to define how to specifically train the network. We employ gradient descent for that. But how to compute the loss? Note that our network has two heads, one that outputs move probabilities and one that outputs a value that denotes whose is going to win the game when continuing from this position. For move probabilities we usually compute the loss as log probabilities or by cross entropy. For a value we have the mean squared error as a measure of loss. Cross entropy and mean squared error can simply be added together to compute an overall loss. Let $\pi$ denote the move probabilities that we expect as the outcome, $\mathbf{p}$ the probabilities that are output by the network, $z$ be the expected result of the game and $v$ the output value of the network, and $\Theta$ be the weights of the network. The overall loss of the network is given by

$$\text{loss} = (z - v)^2 - \boldsymbol{\pi}^T \log \boldsymbol{p} + c\|\Theta\|^2$$

Here $c$ is a regularization parameter. Regularization is just some technique to avoid overfitting; we'll ignore the details here.

Then there is of course the question how to organize the self play. We could immediately replace the old network by the improved network after just one self-played game. In AlphaGo Zero a slightly different approach was used. First a number of games are played with the current network. For these games the parameter $\tau$ is set to 1 for the first moves in order to make sure that it is not the case that the same positions are played over and over again and different positions are generated in the opening phase. Later $\tau$ is set to almost 0 to get the best moves according to the MCT search. After a number of games are played, the old and the new network are compared against each other by playing a number of games. Only if the new network convincingly beats the old network, the new network will replace the old network. This ensures that we make constant progress and are not stuck in a local minimum during training. The reported numbers are 25000 games to train one iteration of a network, 400 games between the old and the new network to compare the performance between the two, and a winning threshold of 55 percent that decides if the new network is really stronger than the old one.

If you take a look at the simulation step you might wonder why we need to create a network that outputs move probabilities. After all, during evaluation only the value output $v$ is really used to update the $Q$ values which eventually decide which move we take. But make no mistake: The prior probabilities that are generated during the expansion step are equally important in directing the MCT search! It is these probabilities that make sure that MCTS explores the most promising and strongest moves and does not waste time on legal but weak moves. It is precisely this approach of ruling out bad moves by just assessing the position *without* any tactical calculation of the search tree that makes AlphaGo Zero so human-like in the approach. Professional players are strong at tactics, but what's even more important is that they know *when* to calculate by identifying promising moves and not waste time on calculating inferior moves.

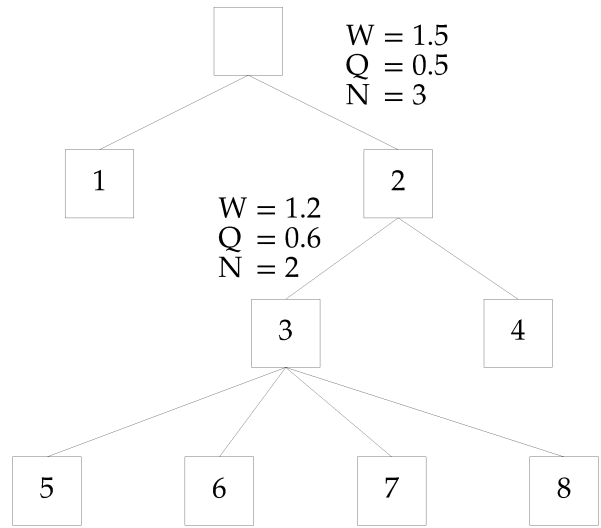

Figure 4.10: AlphaGo Zero: Backpropagation.

### 4.2.3   Odds and Ends

The computational effort compared to AlphaGo was significantly reduced by AlphaGo Zero. It seems that during play the network itself is much stronger than the one used by AlphaGo. Whereas AlphaGo used 48 tensor processing units[8], AlphaGo Zero beat AlphaGo despite using only four TPUs by 100:0. According to the data the DeepMind team provided, a great strength improvement seems to stem from switching from an architecture based on pure convolutional blocks to one with residual blocks.

An interesting comparison is with supervised learning, i.e. asking the question:

---

[8]proprietary dedicated chips for machine learning developed by Google that are even more efficient than modern GPUs

What is more efficient and leads to a stronger engine, using positions sampled from real games of strong professional players, or training by reinforcement learning using self-play as done for AlphaGo Zero. For that, the DeepMind team generated a network with a similar architecture based on residual blocks and trained it by position samples from expert games from the KGS Go server. Then they compared the performance of these two networks using the same MCTS approach for finding the best move in a given position.

We will return to the pros and cons of supervised vs reinforcement learning in the context of chess in Chapter 5, but for now let's recapitulate the findings by DeepMind:

There are three very interesting facts to note. The first is that initially the supervised network gained a high rating very quickly, beating the network trained by reinforcement learning. After about 20 hours of training time on their hardware however the network trained by reinforcement learning became strictly stronger than the supervised one. After about 40 hours of training, the supervised network reached a plateau at a rating level of about 3500 Elo. The network trained by reinforcement learning continued to improve and maxed out at a rating level of over 4000 Elo.[9]

Second the networks ability to predict outcomes of games was tested. To test their performance, the networks were fed positions from a dataset of games of professional Go players. As before, initially the network trained by supervised learning was better at predicting the outcome of games, but after about 15 hours of training, the network trained by reinforcement learning was strictly better at predicting the outcome of games on that dataset.

Now for the third, somewhat surprising fact. The networks were also tested on the question: Which one is better at predicting expert moves? Namely given a Go position from this database of expert games, the task was to predict which move was played by the expert in that position. And it turns out that the net-work trained by reinforcement learning was strictly worse than the network trained by supervised learning for any given time of training. How can this be if the network trained by reinforcement learning is stronger by 500 Elo compared to the one trained by supervised learning? Well it looks as if the network trained by reinforcement learning gained *superhuman* strength. It apparently discovered patterns during self-play that humans do not understand. It therefore predicts the *best* move according to its superhuman understanding and strategy in a given position and not the *weaker* move that was played by the expert player. This is a fascinating discovery and strong indication that computer Go has passed a significant milestone. From a computer scientist's view this is probably even more proof that AlphaGo Zero has superhuman strength than the demonstration matches against Lee Sedol. Although they were of course much more spectacular than some diagram in a paper.

---

[9]We should note that with the larger network with 40 blocks AlphaGo Zero even reached a peak rating of 5185 Elo. An interesting side node is that this large 40 block network reached an Elo rating of 3055 *without* any kind of search, i.e. just by selecting the move with the highest probability in a given position.

Except maybe for computer scientists.

## 4.3   AlphaZero

AlphaZero is essentially AlphaGo Zero adapted to chess and Shogi plus some slight simplifications for the training pipeline. The general approach, i.e. the network structure and training pipeline is mostly the same. Due to its impact in the chess world this section nevertheless tries to describe everything in full detail with the risk of some repetitions. The idea is that you can read this section mostly on its own with some occasional peeking into the section about AlphaGo Zero.

Let's start with a brief overview to get an intuition how AlphaZero works:

- There is one large and deep neural neural network. The *input* to the network are chess (resp. Shogi or Go) positions. The output is twofold: First probabilities for every possible chess (resp. Shogi or Go) move are output. These indicate how good a move is in that position. Second, a value is output for that position. This value indicates how likely it is that White wins, Black wins or that there is a draw.

- Once the network is trained the engine works by using MCT search to find

the best move in a given position. MCTS differs from textbook MCTS in that it is guided by the network: during *selection* the network is queried for move probabilities in a position. Child nodes are then selected w.r.t. the moves with the highest probabilities.[10] Second, for *simulation* there are no random playouts. Instead the network is queried with the position of the node, and the value that the network outputs is employed instead of using results of random playouts.

- The network itself is also trained using MCTS. A number of games are self-played by the engine. For each move an MCT search is conducted using the current state of the network. The MCT search results in a more accurate assessments of a position w.r.t. the evaluation value and best moves. Positions of these games together with the results of the MCT search for that position as well as the final outcome of the played game are then used as training data for the network.

As mentioned, the approach is very similar to AlphaGo Zero. There are a number of differences though. The rules for chess and Shogi are far more complex than for Go. Not only are there several different pieces but the rules on how they move are also more involved. It is therefore in particular interesting to take a look at the input encoding of a chess or Shogi position as well as the output (encoding) of the network. The other major difference is some simplification in the training pipeline.

**Network Input** There are numerous ways to encode a chess position into a sequence of bits. Probably every chess player is for example familiar with Forsyth-Edwards-Notation (FEN). One could simply take the bit string from the ASCII values of the letters of the FEN string. This however is extremely inefficient. Not only has the network to learn chess knowledge — i.e. who is going to win and which moves are best — but also learn to decode the position first. Defining an encoding such that the network does not have this burden is thus quite important. The following approach was used by the DeepMind team for AlphaZero, even though they mentioned that the overall approach worked also well for other *reasonable* encodings.

---

[10]Here, some exploration factor is employed such that the MCTS does not solely rely on the network for selection, especially in the initial phase.

The basic building block of the encoding is a *plane* of size $8 \times 8$, similar as done for Go. There are 119 planes used in AlphaZero to encode all necessary information about the game state for chess. Each plane consists of only bits, i.e. zeros and ones. Several planes are required to encode one position:

- Six planes are required to encode the position of the pieces of the white player, i.e. one plane for the position of the white pawns, the white rooks, the white knights, bishops, queens and the king. In a plane a bit is set to one if there is a piece of that type on the square, and set to zero otherwise. For example if White has pawns on d4 and e4, the plane for the white pawns is set to 1 at indices $(3, 3)$ and $(4, 3)$. Here we start counting from zero and the first index refers to the file and the second one to the rank.

- Another six planes are required to encode the position of the pieces of the Black player.

- Another two planes are used to record the number of repetitions of the positions. The first plane is set to all ones if a position has occurred once before and to zeros otherwise, and the second plane serves a similar purpose to encode if a position has occurred twice before.

- In order to cope with (i.e. detect possible forced draws) three-fold repetition the above mentioned planes for the eight previous positions are encoded as well. If the game just started, these history planes are simply all set to zeros. Of course creating a history of previous positions can also help to detect patterns during training.

This basically encodes the most important aspects of a game state in chess. There are however a few more features required. There is one additional plane that simply encodes the color of the player whose turn it is, i.e. a plane of just ones or zeros if it is White resp. Black to move. Another four planes are used to encode castling rights: One for white-kingside castles, one for white-qeenside castles, one one for black-kingside castles, and finally one for black-queenside castles. These planes are set to all ones if the right to castling exists, and to zeros otherwise.

Finally we need two counters. One counter encodes the total move count[11] and one counter is for progress. The latter counts the number of moves where no progress has been made, i.e. no capture has been made and no pawn has been moved (the 50 moves rule). The counters are given directly as numbers as a (real valued) input to the network.

An example of the input encoding for the six planes that encode the position of the pieces is given in Figure 4.11. The position is from Capablanca vs Corzo, Havana 1901 and one of many interesting endgames of Capablanca.

As mentioned, the same approach was used for Go and Shogi. The input encoding for Go is straight-forward (one plane for the black stones, one plane for the white stones, and one plane to encode the color of the player to move), and the input encoding for Shogi follows a similar approach as used for chess.

**Network Output** The output for the evaluation value is simple. Just output the value. The challenge is the output for the move probabilities. After all, depending on the position there are different numbers of possible legal moves. On the other hand we need a constant number of outputs for our network.

The general idea to cope with that is to consider every possible source square and then enumerate all possible moves of a "superpiece" that can move like a queen and a knight. For each of these possible moves we create one output. Naturally there will be outputs that have a positive probability but that correspond to illegal moves. One can cope with that by simply setting these move probabilities to zero and re-compute the remaining probabilities such that their sum equals one.

To illustrate the approach let's have a look at Table 4.1. We have three columns, one column for the source position, one column for the direction in which the piece moves, and one column for the number of squares that the piece moves. Each row corresponds to one output of the network.

---

[11]It is not clear why the total move count is required. This feature is e.g. omitted in Leela Chess Zero.

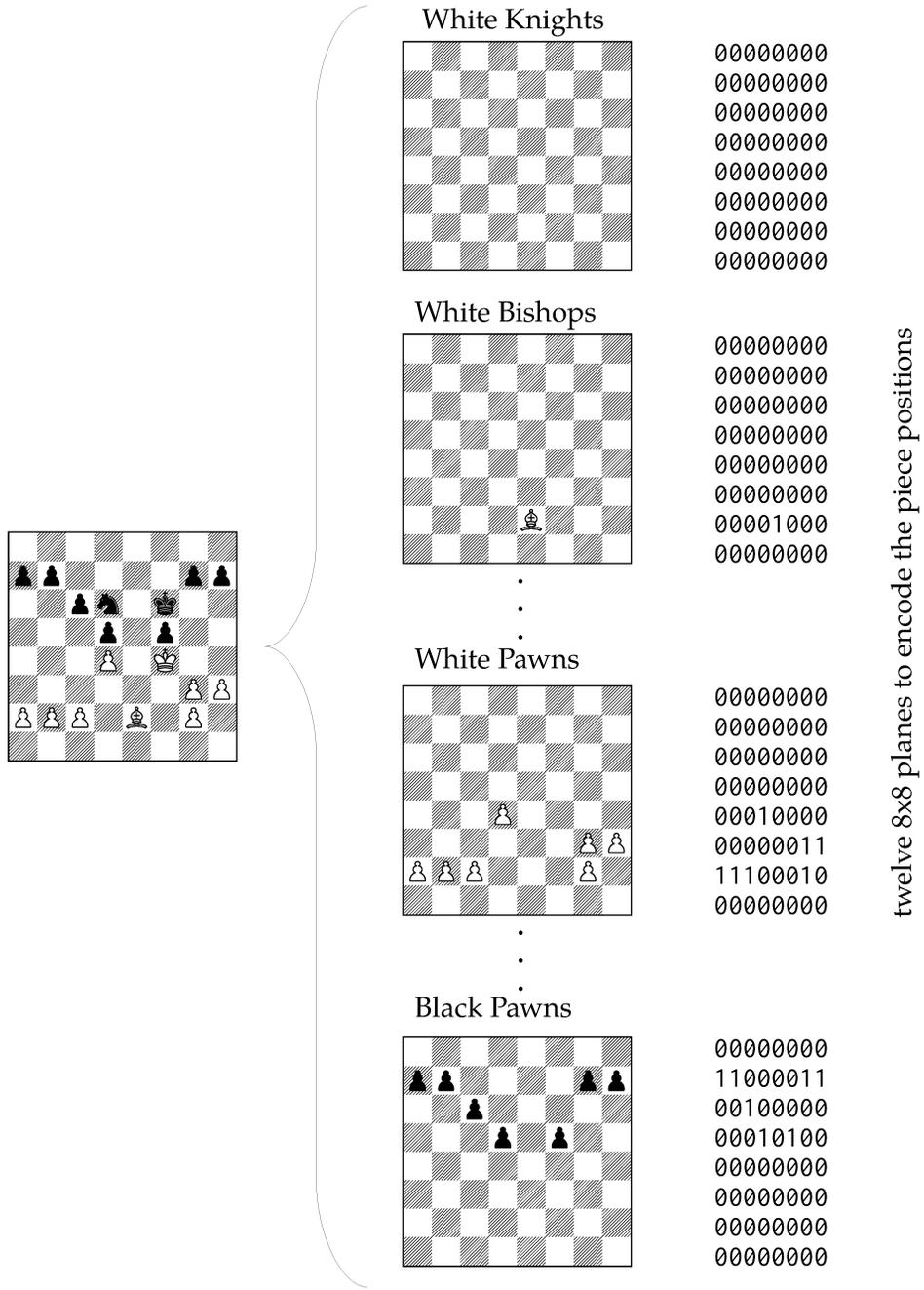

Figure 4.11: Alpha Zero: Input Encoding of Pieces.

Table 4.1: Output Move Encoding - "Queen-like" Moves

| Source Square | Direction | Number of Squares |
|---|---|---|
| a1 | Up | 7 |
| a1 | Up | 6 |
| ⋮ | ⋮ | ⋮ |
| a1 | Up | 1 |
| a1 | Up Right | 7 |
| ⋮ | ⋮ | ⋮ |
| a1 | Up Right | 1 |
| ⋮ | ⋮ | ⋮ |
| ⋮ | ⋮ | ⋮ |
| h8 | Down Left | 1 |

We start by considering a piece at position a1 that moves upwards seven squares. This corresponds to the move a1a8 for whatever piece was on a1. The next output of the network is the same except that the piece moves now only six squares. This corresponds to the move a1a7. We continue like that where we enumerate all possibilities for all potential source squares, directions and number of squares until we have encoded all potential (queen-like) moves.

How about knight moves? We follow the same approach to encode potential knight moves, i.e. for each combination of source square, and the eight potential knight moves (two up and right, two right and up, two right and down, two down and right, two down and left, two left and down, two left and up, two up and left)[12] we create one output. The approach is illustrated in Table 4.2.

Last we need to care about potential underpromotions. Promotions can occur when a pawn moves to the eighth (first) rank, or if a pawn captures a piece upper left or upper right at the eighth rank resp. first rank. This is is depicted in Table 4.3. Any combination of source square, move type (advance, capture to the left or capture to the right) as well as the promotion's piece type are considered. This is again inefficient since a pawn promotion from a1 can never happen — but it makes sense to simply keep the $8x8$ structure for the simplicity of implementation. Pawn moves that queen and are encoded in the outputs for queen-like moves are assumed to promote to a queen.

---

[12]A quite unknown fact is that there is an additional chess rule in Japan that says that you get an extra queen if you move your knight up, up, down, down, left, right, left, right, on a square on the b-file and then on a square on the a-file.

Table 4.2: Output Move Encoding - "Knight-like" Moves

| Source Square | Directions |
|:---:|:---:|
| a1 | two up, one right |
| a1 | two right, one up |
| ⋮ | ⋮ |
| a1 | two up, one left |
| a2 | two up, one right |
| ⋮ | ⋮ |
| a1 | two up, one left |
| ⋮ | ⋮ |
| ⋮ | ⋮ |
| h8 | two up, one right |

This all looks slightly cumbersome. After all, we create outputs for potential moves that can never happen. If there is a pawn on a7, then no matter what piece there is on a7, a move Qa1a8 is not possible, so a positive output of the network has no meaning. If there is a rook on a1, a move from a1 in the upper right direction can never happen. However the encoding ensures that each potential move corresponds to one specific output of the network. After all, it is completely irrelevant which piece is actually placed on a square. As mentioned, we can simply mask out illegal outputs to zero. Given training positions and corresponding expected move probabilities where illegal moves are masked, the network will learn by itself that it should not try to do illegal moves. And if the network outputs a probability greater than zero for an illegal move during play, we can simply set this probability to zero and and re-calculate the probabilities such that the probabilities of the legal moves add up to one.

Table 4.3: Output Move Encoding - Underpromotions

| Source Square | Move Type | Promotes To... |
| :---: | :---: | :---: |
| a1 | advance | knight |
| a1 | advance | bishop |
| a1 | advance | rook |
| ⋮ | ⋮ | ⋮ |
| a2 | advance | knight |
| a2 | advance | bishop |
| a2 | advance | rook |
| a2 | capture left | knight |
| a2 | capture left | bishop |
| a2 | capture left | rook |
| a2 | capture right | knight |
| a2 | capture right | bishop |
| a2 | capture right | rook |
| ⋮ | ⋮ | ⋮ |
| h8 | capture right | rook |

The last task is to calculate how many outputs we actually get if we follow this encoding approach. For the first table we have 64 potential source positions, eight possible (queen) directions and seven possible number of squares that a piece moves, i.e. $64 \times 8 \times 7 = 3584$ outputs. For the knight-like moves we have 64 potential source positions and for each source position eight possible knight moves, i.e. $64 \times 8 = 512$ outputs. Last for the underpromotions we have 64 potential source squares (even though technically only 16 are legally possible since a pawn must be on the second (Black) or seventh (White) rank to queen with the next move), three possible move directions (advance the pawn, capture to the left or capture to the right) and three possible underpromotion piece types (knight, bishop and rook). Thus we have $64 \times 3 \times 3 = 576$ outputs. All in all there are thus $3584 + 512 + 576 = 4672$ outputs. Again, this illustrates the sheer size and required computational effort to train such a network.

Why such a complicated output structure? An alternative output structure would be to simply enumerate all possible moves, i.e. a1b1, a1c1, a1d1 and so on. In addition to enumerating all combinations of source-square and to-square we need some additional outputs for promotions; for example we need both a7a8 for a normal piece to move from a7 to a8 as well as a7a8q for a pawn that queens. This would give less than 2000 outputs, and is also referred to as a "flat distribution" in the AlphaZero paper [SHS+17, SHS+18].

Compared to that, the approach used for AlphaZero is very source-square centric, i.e. the network might be able to train faster by roughly guessing *which* piece to move in a position (i.e. which source-square to select), and only then refine with more training batches *where* to actually place that piece. And indeed, according to the AlphaZero creators they tried both encoding styles and both worked, but the slightly more complicated one with the 4672 outputs trained faster.

Note that AlphaZero used a similar encoding for Shogi, whereas the outputs for Go were encoded as a flat output just enumerating all legal moves. But remember that moves in Go are of trivial nature, it just means placing a stone on a point — there is no source and target square, and there are no different pieces like in chess or Shogi but just stones.

### 4.3.1 Network Architecture

Not much to see here. Except for the network input and output — i.e. the input vector that encodes a chess or Shogi position and the encoding of all possible moves — the network architecture follows precisely the one used for AlphaGo Zero based on resnets. For their experiments, the DeepMind team used the "small' version with 19 residual blocks. For an overview of the architecture cf. Figure 4.7.

## 4.3.2 MCTS and Reinforcement Learning

The approach used for MCTS combined with reinforcement learning is essentially the same as the approach for AlphaGo Zero. The major difference is a slightly different training pipeline. All the details are already described in Section 4.2, one just needs to think of chess positions and chess moves instead of Go positions and Go moves. Therefore we only briefly recapitulate the approach here.

Assume we have a trained network that, given a chess position, outputs move probabilities that denote how good a move is and also outputs an evaluation value in the range $-1, \ldots, 1$ that denotes whether the game from the perspective of the current player with perfect play of both sides is going to be a loss, draw or a win. Then given a position, MCTS is used to find the best move in that position. Here MCTS differs from textbook MCTS-UCT:

- During *selection* we don't use the standard UCT value computation. Instead in each step, the network is queried to get move probabilities for that position. The child node with the best move probability is selected. There is also some small correction factor added such that MCTS also explores moves that are currently deemed good but not the best moves. This is in order to facilitate exploration and not to miss an excellent move that is currently not evaluated as the best move by the network.

- During *simulation* we don't play random games at all to compute an evaluation value. Instead the network is queried to get the evaluation value. This value is used in the subsequent backpropagation.

Again, this is absolutely the same procedure as the one for AlphaGo Zero.

The training pipeline is also almost the same as the one used in AlphaGo Zero. Initially we start with random values for the weights of the neural network. Then we generate training data by using MCTS with the current network to self-play games. In each step of the game, MCTS is used to find the best moves and get move probabilities (they used 800 simulations in each MCTS step). Note that these move probabilities differ from the move probabilities of the network, since the *search* will generate more knowledge about the position than what the network currently has. When the game finishes we also have a result.

To illustrate this let's assume a self-played game generated the moves:

1.e4 e5 2.f4 exf4 3.Bc4 Qh4+ 4.Kf1 b5 5.Bxb5 Nf6 6.Nf3 Qh6 7.d3 Nh5 8.Nh4 Qg5 9.Nf5 c6 10.g4 Nf6 11.Rg1 cxb5 12.h4 Qg6 13.h5 Qg5 14.Qf3 Ng8 15.Bxf4 Qf6 16.Nc3 Bc5 17.Nd5 Qxb2 18.Bd6 Bxg1 19. e5 Qxa1+ 20. Ke2 Na6 21.Nxg7+ Kd8 22.Qf6+ Nxf6 23.Be7# 1-0

Yes, this is the famous game Anderssen vs Kieseritzky, the "Immortal Game" and yes, it is highly unlikely that a self-played game with the network will generate precisely this sequence of moves. Let's assume this for the sake of argument. For each step, we have a triple of (current position, move probabilities generated by MCT search, game result). For example we could have:

1. (initial position, $[e4 : 0.4, d4 : 0.4, c4 : 0.1, \ldots], 1 - 0$ )

2. (position after 1.e4, $[e5 : 0.3, c5 : 0.3, e6 : 0.2, \ldots], 1 - 0$ )

3. $(\ldots, \ldots, \ldots], 1 - 0$ )

4. (position after 22...Nxf6, $[Be7 : 0.99, Nxf6 : 0.01, \ldots], 1 - 0)$

But these are exactly the training data for our network! The input of the network is the position of the game, and the expected result is the vector of move probabilities and the result of the game as the expected evaluation value. As for the loss function, network and training: Again I refer to Chapter 4.2.2 since it's all the same. What changes really is only that we have chess positions and chess moves instead of Go positions and Go moves, and that there is the possibility of a draw as an outcome of the game. Same for Shogi by the way.

As already hinted in the introduction of this chapter: The major difference compared to AlphaGo Zero is that the training there was done with several specific iterations. First a network was fixed as the *current network*. Then games were self-played using MCTS and the current network was used to support the MCT search and generate training data as mentioned above. The current network was then trained with this data to generate a new network. This newly generated network was then compared with the current network. Only if the newly generated network could beat the current network in a number of games with a winning ratio of 55 percent or higher, the current network was replaced with this newly generated network, and the cycle was repeated. This was to ensure that we are not stuck in a local minimum and make steady progress during learning.

It turns out however that this is apparently not necessary. Alpha Zero was trained by continously updating the network and immediately using the updated network without any kind of evaluation, e.g. by matches of the old and newly generated network. Just to clarify: Learning is still done with batches of positions of course.

### 4.3.3 Odds and Ends

AlphaZero builds upon the success of AlphaGo and AlphaGo Zero. It not only simplifies a lot of the complexity of AlphaGo but is also an evolutionary step as it contains the very much simplified training pipeline of AlphaGo Zero. This should not be misunderstood in the sense as that *implementing* this training pipeline is absolutely easy — while it still involves some programming effort, it especially also still requires huge computational resources. Nevertheless we will implement a (computationally) simplified version of AlphaZero in Chapter 5.

Nevertheless from an algorithmic level, everything is very much simplified. For example they completely eliminated not only the need for supervised learning (i.e. the need of a huge training data set, which is often not available) but also for policy gradient reinforcement learning.

Especially if you managed to survive all the math details about networks, and maybe even the complicated chapter about AlphaGo, you might very likely be able to appreciate this simplicity and elegance.

One aspect that should be stressed here is that AlphaZero is very much oblivious to the specific rules of a game. In AlphaGo Zero some specific properties of the game (the symmetry of positions to generate more training data) were utilized.

This is not the case with AlphaZero; they even tried out Go *without* exploiting the symmetry of positions. Except for the specific rules of a game that must be utilized for move generation during MCT search and the design of the network input and output, AlphaZero is more like a framework to solve any kind of deterministic two player game. It really does not matter whether it is Go, Shogi, chess, Xiangqi, Reversi, Othello, Amazons or any other kind of such board game.

What is also remarkable is the huge difference in the approach compared to alpha-beta searchers: We have a very narrow search tree where only a fraction of all possible legal moves are actually explored. Moreover compared to alpha-beta searchers an almost insane amount of time is spent on evaluation rather than on searching. Whereas alpha-beta searchers search millions of positions employing a very fast evaluation function, Alpha Zero searches only a fraction of those positions, since querying the network takes a lot of time and computational resources.

This very much resembles how humans play chess. A huge amount of time is spent on evaluating positions. This capability of being able to evaluate a position without calculating much tactics, i.e. just having a "feeling" for the position and grasping its positional nuances is probably the most difficult task to learn for a chess player. There are very scholarly books about that even for club players like "Reassess Your Chess" [Sil10] or "Chess Strategy for Club Players" [Gro09]. Such books help, but still lots of coaches are convinced that the only way to get that chess understanding is to play lots of serious tournament games. And that's precisely what AlphaZero does to get good, mimicking the human approach to chess.

## 4.4   Leela Chess Zero (Lc0)

It is December 2017 and DeepMind released the AlphaZero preprint on the internet, including all those sample games against Stockfish. The whole chess and computer science crowd is excited - but DeepMind does not release AlphaZero to the public. Everyone wants to actually *use* these new technologies. The magic happened. We know *that* it works. But nobody can use it. What to do?

There were immediate efforts to reproduce the results for Go after AlphaGo Zero: Gian-Carlo Pascutto started Leela Zero, a distributed effort to get a strong Go engine released under an open-source license. This code was then ported and adapted to chess by several volunteers. The result was LeelaChess Zero, or short Lc0. This made an AlphaZero-like engine accessible to everyone. The port did not follow AlphaZero in every detail, taking freedom in changing things if they did not fit or if better approaches were found. Let's have a look under the hood and see how Lc0 works.

### 4.4.1 The Network

**Network Input**   The input encoding follows the approach taken for AlphaZero. The main difference is that the move count is no longer encoded — it is technically not required since it's just some superfluous extra-information. We should also mention that Leela Chess Zero is an ongoing project, and naturally improvements and code changes happen. The input format was subject to such changes as well, for example to cope with chess variants such as Chess960 or Armageddon, or simply to experiment with encodings. The encoding described here is the *classic* encoding, referred to in source code as INPUT_CLASSICAL_112_PLANE. For those who want to look up things in code, the relevant source files are lc0/src/neural/encoder.cc and lc0/src/neural/encoder_test.cc.

The input consists of 112 planes of size $8 \times 8$. Information w.r.t. the placement of pieces is encoded from the perspective of the player whose current turn it is. Assume that we take that player's perspective. The first plane encodes the position of our own pawns. The second plane encodes the position of our knights, then our bishops, rooks, queens and finally the king. Starting from plane 6 we encode the position of the enemy's pawns, then knights, bishops, rooks, queens and the enemy's king. Plane 12 is set to all ones if one or more repetitions occurred.

These 12 planes are repeated to encode not only the current position, but also the seven previous ones. Planes 104 to 107 are set to 1 if White can castle queenside, White can castle kingside, Black can castle queenside and Black can castle kingside (in that order). Plane 108 is set to all ones if it is Black's turn and to 0 otherwise. Plane 109 encodes the number of moves where no capture has been made and no pawn has been moved, i.e. the 50 moves rule. Plane 110 used to be a move counter, but is simply set to always 0 in current generations of Lc0. Last, plane 111 is set to all ones. This is, as previously mentioned, to help the network detect the edge of the board when using convolutional filters.

**Network Output**   Originally Lc0 used a flat distribution over all possible moves[13], i.e. start with a1b1, a1c1 and continue up to h8g8 plus the moves promoting a pawn. This results in 1858 outputs. Later, a bug in the network structure was found. Before adding the fully connected output layer, the policy head of the network convolved the last residual output into an output of "only" 128 neurons.[14] These 128 neurons were then fully connected to the 1858 network outputs. Clearly there is an information bottleneck — such bugs can easily be overlooked — as the number of 128 intermediate neurons is just very low. Consequently, the network output was changed and the structure was changed to one similar to AlphaZero[15].

### 4.4.2   Network Architecture

Lc0 is an ongoing project and therefore there is no "fixed and final" architecture. Instead the architecture is subject to changes and optimizations. The architecture described here is the current one (as of 2021), but we also point out differences to the architecture previously used.

Essentially, the network architecture follows closely to the one used for AlphaZero, but replaces the ResNet building blocks by Squeeze and Excitation blocks. Also, as was the case for AlphaZero, the network is characterized by how many of these blocks are used, and also how many filters are used in the convolutional layers. Typical combinations of the number of blocks and filters are 10 blocks with 128 filters, 20 blocks with 256 filters and 24 blocks with 320 filters.

---

[13]cf. the array in "decode_training.py" where training data is decoded for debugging purposes
[14]`https://github.com/glinscott/leela-chess/issues/47`
[15]`https://lczero.org/blog/2019/07/end-of-era`

The architecture is shown in Figure 4.12. The input features are passed to one initial convolutional layer. Then, the main building blocks follow. One block consists of two convolutional layers, followed by a Squeeze and Excitation block and ReLU activation. In case you are wondering why there is no batch normalization: This is employed directly in the convolutional layers. Normalization is here *folded* into each convolutional layer — a small optimization technique.

Finally there are three heads. The first head is the policy head that outputs move probabilities, consisting of two convolutional layers (no activation function). The output used is the $73 \times 8 \times 8 = 4672$ output move encoding that was also used in AlphaZero.

The value head consists of a fully connected layer and softmax activation. It has three outputs instead of the single output of AlphaZero. The three outputs explicitly model the probability of *winning*, *drawing* and *loosing*.

Recently a third head has been introduced, dubbed the *moves left head*. This head consists of a convolutional layer, followed twice by a fully connected one with ReLU activation. The single output predicts the remaining number of moves until the game finishes. It is not absolutely clear how and if this is going to be a useful head, but

- it prevents that random moves are output that do not worsen the position and do not lead to a forced draw (i.e. do not violate the 50 moves rule), but also do not lead to the end of the game. In other words this might prevent random moves that do not make any progress

- in tournament controls it might be useful to use this information to allocate more or less calculation time at a given point in time during the game. For example we could identify the opening phase, middle game phase and end game phase by the number of moves left and allocate more time during the middle game.

As mentioned before this is the current state-of-the-art architecture used by Lc0. The previously used network had a few small differences:

- The *value head* was a fully connected layer that output one value and used tanh for activation. This single value denotes the winning probability — as done for AlphaZero. In other words, there is only one value indicating the winning probability, instead of the triple explicitly modelling the winning, drawing and loss probability as done in the current network architecture.

- The third head, modelling the number of moves left in the game, does not exist.

- The *policy head* output used a flat head, i.e. associated one move (source square plus destination square) with one output, resulting in 1858 outputs for all possible moves.

### 4.4.3    MCTS and Reinforcement Learning

Conceptually the whole training approach is closely modeled after the one used for AlphaZero. The main problem that Lc0 faced and faces is that the DeepMind team had access to a sheer unbelievable amount of computing power to train the network.

In order to replicate the results by DeepMind, the authors of Lc0 did a clever trick by giving the computer chess community a chance to contribute to the effort in a united manner. Interestingly, the resource-intensive part in training the network is not training itself, i.e. feeding in training data and updating its weights by backpropagation. Of course this is no easy task either, but can be handled by a reasonable powerful computer system.

The difficult part is actually to generate training games. After all, during each step of generating a training game we do not only have to apply MCT search excessively, but also query the network in its current state over and over again. Therefore it is precisely this "training by self-play"-part that has been done as a community effort.

Figure 4.13 gives a brief overview. The heart is the network that is hosted on a central sever instance. The network itself — or rather its weights — is then distributed to volunteers. They use the current instance of the network to generate millions of training games using self-play of the current network combined with MCT search; the same approach that was used in the case of AlphaZero.

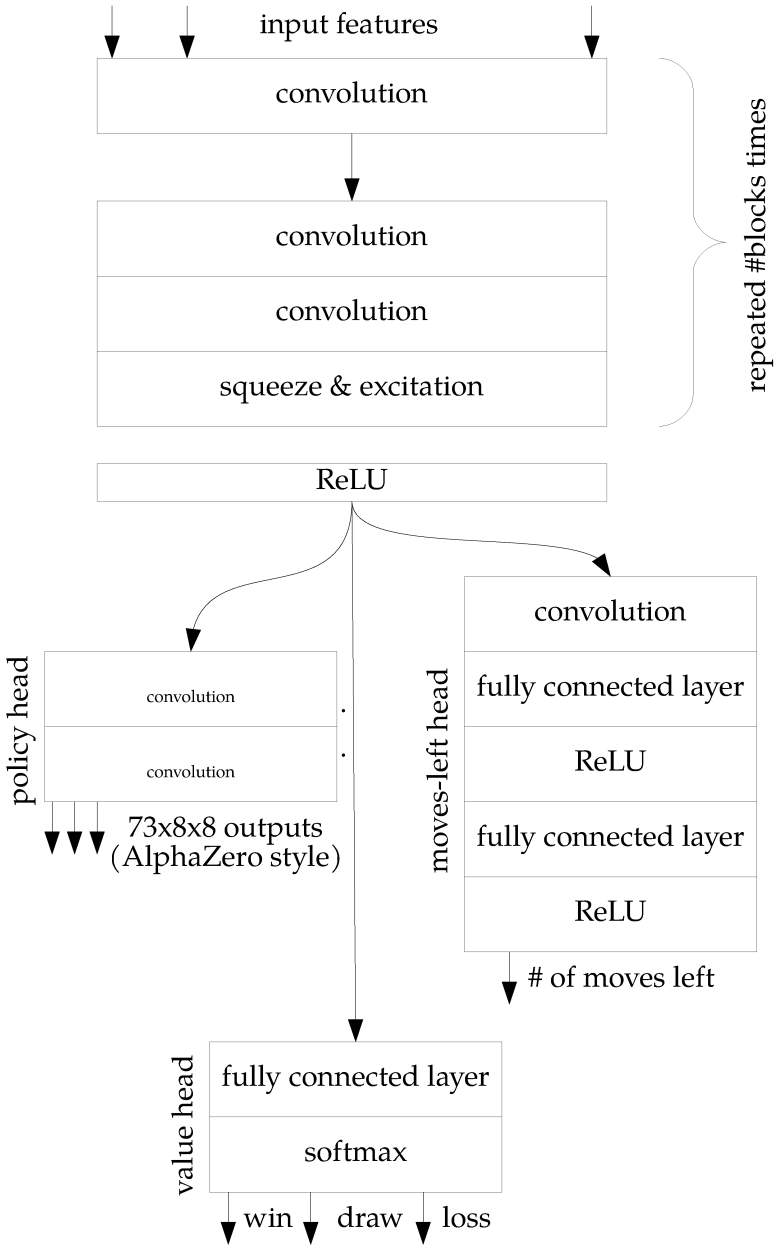

Figure 4.12: Lc0: Network Architecture.

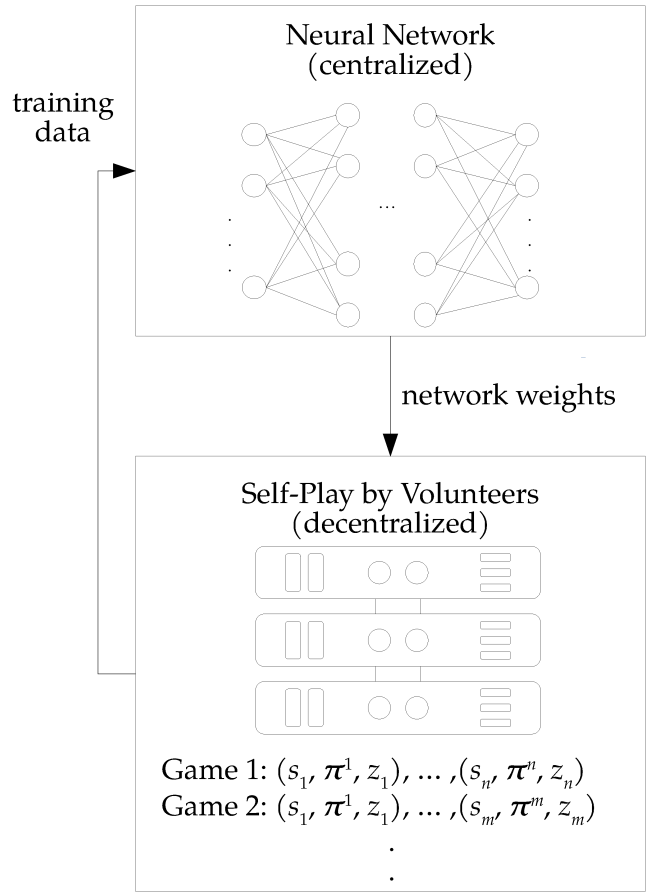

Figure 4.13: Lc0: Training Pipeline.

Once enough self-played games have been generated, the states of these games (i.e. the triple of current position, vector of move probabilities as well as the result) are sent back to the central server. This training data is then used to train the current network and improve its weights. The updated network is then distributed to the volunteers and the whole cycle continues.

As a practical remark w.r.t. using Lc0: Users of chess engines are accustomed to get an evaluation of the current position in pawns. For example a value of 1.00 means White's position is better to the equivalent of being a pawn up.

The current version of Lc0 outputs win, draw and loss probabilities — as neural network based chess engines get more and more common, users will get more and more accustomed to interpret this different kind of output.

In the previous version of the network architecture, the value head would output only a value in the range $[-1, 1]$. Given the output of the value head $Q$, this was then converted to a pawn-based evaluation by the empirically chosen formula

$$\text{Pawns} = 111.714640912 * \tan(1.5620688421 * Q)$$

This means an evaluation of Lc0 of 0 (probably a draw) results in a 0.00 pawn-based evaluation. An evaluation of Lc0 of 0.5 (White is clearly better) results in 1.11 pawns, and 0.7 (White will likely win this) results in a pawn based evaluation 2.15 (more than two pawns up).

### 4.4.4 Odds and Ends

The impact of Lc0 should not be underestimated. AlphaZero made a huge impact in the chess (and Go and Shogi) community, but the results could not be independently reproduced. It's not that the DeepMind team hid anything, but the required computing power was simply too difficult to get for an outside party. Creating the neural network structure, the MCTS implementation and the distributed training framework is a monumental task. As of writing this, the Github repository where the code of Lc0 is hosted counts 1323 commits of code changes. If we assume two hours of work that go into one commit, we are at roughly 2600 hours of work. If we further assume an 8 hours work day, that's 325 days of work! Or in other terms: In the US there is an average of 1700 hours of work per year per employee, meaning that if we assume a single developer, that's 1.5 years(!) of work. All done on a volunteer basis for free![16].

Lc0 not only brought the power of neural-network based engines to the masses, it also made it available to elite Grandmasters as well. It is apparently used heavily by Grandmasters to find opening novelties and thus changes indirectly the chess that we play today. All thanks to the Lc0 developers and their volunteer work!

## 4.5 Fat Fritz

Fat Fritz is a commercial, neural network based chess engine sold by ChessBase GmbH. Before we delve into its inner workings, let's quickly recapitulate on supervised learning.

In the previous chapters you learned about the quite elegant reinforcement learning algorithm discovered by DeepMind. Self-play generated games which itself consist of triples

$$(s_1, \boldsymbol{\pi}^1, z_1), (s_2, \boldsymbol{\pi}^2, z_2), \ldots, (s_n, \boldsymbol{\pi}^n, z_n)$$

Here $s_i$ is the current chess position of the game, $\boldsymbol{\pi}^i$ is a vector of probabilities for each move in that position. These are the result of an MCT search of the current position based on the current neural network. Last, $z_i$ is the result of the game.

Instead of using this reinforcement learning approach one can of course just use existing chess games, for example games played by strong players, and train the network with these games. In that case we of course do not have a vector of move probabilities in each position, but just the move that was played by the player. However we can simply treat this as a vector of move probabilities where we set the probability of the move that was played at 1.0 (i.e. 100 percent probability) and the other moves to 0.0 (i.e. probability 0). In other words if we have an implementation of the neural network with the associated self-training pipeline, it is trivial to apply supervised learning as well.

---

[16]As an interesting side note, the free tool `git-estimate` available at `https://github.com/lui gitni/git-estimate` that calculates work hours out of the logs of the version control system git confirms my very rough estimate by guessing an overall amount of 2178.32 hours of work

With Fat Fritz, Chessbase took the existing open-source implementation of Leela Chess including its training pipeline but with an untrained network, and applied both supervised and unsupervised learning. According to their marketing announcements[17]

> The philosophy behind Fat Fritz has been to make it the strongest and most versatile neural network by including material from all sources with no such 'zero' restrictions, such as millions of the best games in history played by humans, games by the best engines including Stockfish, Rybka, Houdini, and more, endgame tablebases, openings, and so on. If it was deemed a possible source of improvement, 'zero' or not, it was used. Even millions of exclusive self-play games were created[...]

From this description it is unclear how many and which games were used for supervised learning, and how much training was spent on reinforcement learning. As Chessbase is also a vendor of chess databases, we can assume that their main product *Mega Database* was used, a collection of approximately 8 million chess games dating from the year 1475 until now. They also sell a correspondence chess database with approx 1.6 million games. As nowadays correspondence players utilize chess engines, these games are usually of very high quality. Last, there are several endgame tablebases sold by Chessbase as well as opening books. It is unclear however if any of the opening books overlap with the *Mega Database*. To my best knowledge there is no information on which kind or how many engine matches were played.

There are several questions that arise now:

1. Does this approach result in a strong chess engine?

2. What are the advantages and disadvantages of employing supervised learning in that context? Does it make sense from a scientific standpoint?

---

3. Which of the three — supervised, unsupervised or a mix of both — will result in the *strongest* engine?

4. What is the added value compared to Leela Chess Zero — after all, Fat Fritz retails for 60 to 80 Euro whereas Leela Chess Zero is a free community effort.

The first question can be answered easily in that Fat Fritz is a strong chess engine that was competitive. The other questions are more complicated though.

### 4.5.1   A Critical Evaluation

So what are the advantages and disadvantages of employing supervised learning in our context?

First we must understand that when we train a network, we want it to learn from the training data that we feed in. Ideally the network extracts and learns patterns and structures and thus evaluates chess positions similar as the evaluation of the training data, and also computes similar move probabilities compared to the ones in the training data. If we have bad training data, say chess games from amateurs with lots of blunders, then our network is trained to blunder as well. Thus a big disadvantage of supervised learning is simply *availability* and *quality* of training data.

Even if we have enough training data and our network achieves good performance, there is the open question whether our network is saturated or if we are just stuck in a local maximum and could even improve the strength of our network by more training data.

On the other hand *if* such training data is available, it is often easy to assess the quality of the training data. It is also often easier to train a network with supervised methods than to train one with unsupervised methods. Even though it has to be noted that in this particular case of training chess/Go/Shogi engines, the training pipeline developed by DeepMind is particularly simple.

Another big advantage by supervised learning with high quality training data is that it is usually much *faster*. Think of AlphaZero: The weights of the network are initially random. This more or less random network is then used in combination with MCTS to generate self-play games. These games will naturally of very poor quality initially and it will take quite some time until the network as well as the quality of the generated games improve.

If we instead directly train the network with high-quality games of grandmasters, the initial progress will be much faster. Which brings us to the next question: Which of the three — supervised, unsupervised or a mix of both — will result in the *strongest* engine?

ChessBase is as a commercial operation selling chess products and engines. We cannot expect scientific evaluations or peer-reviewed papers from them like in the case of DeepMind. In other words aside from marketing, no investigation of supervised learning vs. reinforcement learning in the case of computer chess has been made. Therefore it is difficult to get an objective view on whether the approach used to train Fat Fritz is scientifically valid or not.

Our next best information is the scientific data provided by DeepMind. However they only thoroughly compared supervised vs. reinforcement learning for the case of Go, not for chess. We can assume with large certainty that the result would hold for chess as well, but as of now, there is no data to prove it.

We have talked about the data provided by DeepMind in their paper on AlphaGoZero [SSS+17] in Section 4.2, but let's briefly summarize what they investigated and re-think of it in the context of Fat Fritz. They used the network architecture of AlphaGo Zero and then trained that network once with supervised learning using the KGS dataset and once with reinforcement learning according to their self-learning pipeline. The KGS dataset they used contained approximately 30 million positions [SHM+16].

The first comparison is training time vs. Elo rating. Even after only a few hours of training, the supervised network achieved an ELO rating of nearly 3000, whereas the network trained with reinforcement learning required almost 15 hours[18] of training to achieve that level. After 20 hours of training however the network trained with reinforcment learning strictly surpassed the supervised one and eventually reached an ELO of over 4000 after 70 hours of training when the experiment was stopped. The supervised network leveled out after about 40 hours of training.

---

[18]15 hours on their computing cluster which of course translates to months or even years of training time on a standard home computer

The conclusion we can draw from that is that supervised learning will achieve good results faster, but with enough training time a network trained with reinforcement learning will surpass it. This of course holds especially when we run out of training data. But as mentioned in Section 4.2, other data from DeepMind suggest that this effect is not just because of a lack of training data but rather that reinforcement learning results in a superior engine.

Namely the next thing they compared was training time vs. prediction accuracy w.r.t. a dataset of moves played by strong human players (the GoKifu dataset). That is, given a position from that dataset and the move that the human expert player selected, the network is queried to predict that move. Here at any point in time, even after 70 hours of training, the network using supervised training was strictly better at this task with a prediction accuracy of about 53 percent reached after 70 hours of training. The network trained with reinforcement learning only reached about 49 percent (this is actually a large gap; getting even a few more percent is a challenging task). This looks strange, isn't the network trained with reinforcement learning the stronger player? The answer will become clear very soon.

Last they compared training time vs. the mean squared error of predicting game outcomes on the GoKifu dataset, i.e. given a position, who is going to win the game? Here initially the network trained by supervised learning has a lower error, but after approximately 15 hours of training the error of the network trained by reinforcement learning is strictly smaller then the network trained by supervised learning. After 70 hours of training, the difference is an MSE of about 0.18 (reinforcement learning) vs. 0.21 (supervised learning).

Let's summarize: After a sufficient amount of training

1. the network trained with reinforcement learning yields the stronger engine by more than 500 Elo points

2. the network trained with reinforcement learning is worse at predicting human expert moves than the network trained by supervised learning on games of human players

3. yet the network trained with reinforcement learning is better at predicting game outcomes. In other words it has a better positional understanding and evaluation of a position.

This leads to the conclusion that the network trained by reinforcement learning gained superhuman strength. Of course it is bad at predicting moves of human expert players because it chooses not to play such inferior moves! Or as the Deepmind team put it: AlphaGo Zero achieved *superhuman performance*.

A scientific evaluation for supervised vs. reinforcement learning for chess would solve the question whether these findings would hold for chess as well. Aside from the above mentioned research questions regarding supervised vs. reinforcement learning, an additional question would be how a network behaves if it is first trained (kickstarted) by supervised learning and then improved by reinforcement learning and vice-versa.

We must also not forget two differences to Go that might impact the results. According to the data provided by Deepmind [SHM+16], the KGS dataset contained games that resulted in 30 million positions. MegaBase contains about 8 million chess games. If we assume a game length of 20 moves (to account for transpositions in the opening and endgame), we have already 160 million positions available for training — an order of a magnitude more positions! Second, chess is a very drawish game and existing classical chess engines already reached a very strong level of play. Of course AlphaZero beat existing classical alpha-beta searchers convincingly in chess, but the ELO gap was nowhere near the situation in computer Go or computer Shogi. With smaller margins in improvements, it might be more difficult to see clear trends and separate them from artifacts, like e.g. being stuck in a local minimum in training.

Assuming however that the results for Go transfer to chess we can try to make sense w.r.t. Fat Fritz' performance. Clearly training with supervised learning with games of human expert players will result in getting a strong engine with much less training time compared to the reinforcement learning approach. Training without reinforcement learning might miss "superhuman" moves since such *superhuman* moves are apparently missing in training data composed of games by *human* players. In such case there is no reason to believe that training by supervised learning does give any advantage when the final goal is to create the *strongest* possible engine.

## 4.5.2   Free and Open Source and the GNU GPL

Leela Chess Zero is developed by volunteers as free software under a particular license, the GNU General Public License. Software is usually distributed in binary form which only a computer can really understand. During development software is written in source-code and then translated to binary form by a compiler. In order to modify, improve or change software we need the source code.

While I am not a lawyer, the GNU GPL works as follows: Software is distributed with its source code. Anyone can can change, modify or improve the software — but if you do so you have to make your changes public under the same license. The intention is that others can benefit from your changes as well.

As you have gained an understanding on how neural networks and MCTS works you can imagine that Leela Chess Zero consists of several components:

- the (fast and efficient) implementation of the neural network and its associated data-format of the weights for distribution

- the input and output encodings of chess positions

- the training pipeline

ChessBase took all these components without any significant changes. Instead they changed a few parameters as well as the name of the engine and the authors notice[19]. They then invested a lot of computing resources to train the network in a different way, apparently mostly by supervised learning, packaged the engine together with a graphical user interface and sold the whole package for around 60 Euro.

---

[19]`https://github.com/LeelaChessZero/lc0/compare/release/0.23...DanielUranga:fritz`

It is perfectly valid to sell free software that is licensed under the GNU GPL. Of course this makes only sense if there is some perceived added value. If such added value does not exist, one could simply download the (original) version of the software for free.

Linux distributions work like that. Vendors collect various free software components like the Linux kernel, a graphical user interface, drivers and more, package them together with a custom install tool, add a manual and provide customer support in particular for the case where Linux is run as a server, and ship and sell that package as a Linux distribution. Users are willing to pay for that since it is very cumbersome to download, install and combine these software packages all by yourself. Whereas there are also a lot of free Linux distributions, a lot of users — especially the ones that run mission critical systems — choose to buy the commercial version with commercial support. The makers of Linux distributions on the other hand also use some part of their profits to invest in the further development of key components of Linux such as the kernel. This is perceived as a symbiosis and thus software developers are motivated to spend their free time and resources to improve Linux and its codebase.

In the case of Fat Fritz there was no giving back in terms of code to Lc0. Moreover Chessbase was not always very clear about the fact that FatFritz is just Leela Chess Zero with the network trained in different way. For example on my German retail package of Fritz 17, Leela Chess Zero is nowhere mentioned. In fact when buying it I was not aware that Fat Fritz was based on Lc0; I honestly thought Chessbase independently developed their own in-house neural network based engine from scratch.

The perceived benefit of Fat Fritz is at least questionable since as we discussed, it can be asked if applying supervised learning to train the network does yield any benefit compared to just training the network by reinforcement learning — as done by the Lc0 community. Their network(s) can be downloaded and used for free.

Understandably Fat Fritz was not really warmly welcomed by the chess enthusiasts and programmers community. And the situation somewhat escalated when Fat Fritz 2 was introduced — but that is subject to one of the other chapters.

We should however mention here that computer chess has unfortunately a history of alleged GPL violations. Chess engines used to be black art, and there was a small community of programmers who lived well from developing chess engines; first for dedicated chess computers and then later for home and MS-DOS computers. This all started to change when French programmer Fabien Letouzey developed his chess engine Fruit and made it freely available under the GNU GPL. Not only was it a strong chess engine, coming in second at the World Computer Chess Championship (WCCC) 2005, it was also clean and well-written code. A lot of programmers took inspiration from Fruit. Most adhered to the GNU GPL releasing their sources as well, but some did not. Allegedly we must say, as no court case has ever been filed by Fabien.

Then the commercial chess engine Rybka by Vasik Rajlich started to appear and quickly gained traction winning the Computer Chess Championships in 2007, 2008, 2009 and 2010. When allegations were made that Rybka's code base was sourced in Fruit, things got nasty. After an investigation, the International Computer Games Association — the organization behind the WCCC — concluded in June 2011 that Rybka was plagiarized, stripped Rybka from all its titles and banned Vasik Rajlich for life from competing in WCCC events.

As Rajlich refused to provide his source code to the investigation committee, the evidence for a GPL violation was circumstantial, and based on the analysis of the binary. The whole issue remains a heated debate up until today.

Which is not to say that the GNU GPL is not enforceable. Open source enthusiast Harald Welte started the project `gpl-violations.org` when he noticed that many large companies borrowed Linux source code without adhering to the GPL. Several court cases were filed and to my best knowledge they never lost a case. Their most spectacular case was probably against Fortinet which used the Linux kernel for their product FortiOS without providing its sources. The result was a settlement where Fortinet would provide all sources in question[20]. Nevertheless suing in court is an expensive and risky endeavour and `gpl-vi olations.org` could only do that with the support of volunteers as well as donations. It is thus understandable that no court case has been filed in the Rybka controversy by Fabien Letouzey, even though that would probably end the dispute, one way or the other.

---

[20]`http://www.cnet.com/news/fortinet-settles-gpl-violation-suit`

The very successful chess engine of the Fritz chess software package from ChessBase was developed by Frans Morsch. When he retired after Fritz 13 after a long career in chess engine programming, ChessBase was left without an engine author. Despite the open dispute around Rybka, they then hired Vasik Rajlich to write the engine for Fritz 15 and Fritz 16. As he likely didn't start from scratch, we can assume that both were based on Rybka.

It seems ChessBase really isn't afraid to fish in muddy waters.

Again and again.

## 4.6 Efficiently Updateable Neural Networks (NNUE)

By achieving superhuman strength in Go — one of major long-lasting open challenges in artificial intelligence — AlphaZero rightfully made a deep impact not only in the Go, Shogi and chess world, but also even in mainstream media.

But there has been another revolution going on which mostly went unnoticed aside from a very interested crowd in the computer Shogi and computer chess world. That revolution is efficiently updateable neural networks (NNUE). But let's start at the beginning.

Shogi is mainly a Japanese thing. Whereas Go is a fundamental different game compared to chess and widespread in Asian countries and even gained traction in the west, Shogi can be considered as a Japanese variant of chess. Similar to western chess it likely originated in India and derived from the game Chaturanga, but whereas Chaturanga developed in the middle east and the western hemisphere into modern chess, it developed into Shogi in Japan and the chinese game Xiangqi. The Shogi scene is thus much smaller, and the computer Shogi scene is even more smaller.

As mentioned before, one major difference between Shogi and chess is that captured pieces can re-enter the game, similar to Crazyhouse or Bughouse chess.

The branching factor, i.e. the average number of possible moves in a position is thus much larger than in chess (but smaller than in Go). Similar to chess, the best traditional computer Shogi programs before AlphaZero were alpha-beta searchers with custom evaluation functions. But whereas for Go and chess the results of AlphaZero were quickly reproduced by an international community due to Leela Zero and Leela Chess Zero, that never happened with Shogi.

Then in 2018, computer Shogi enthusiast Yu Nasu posed the question of why not combine alpha-beta search with a neural network for evaluation. For chess, this had been already done with DeepChess [DNW16], which reached grandmaster level but was never close to the state of the art of traditional chess programs. How come? In fact why did AlphaZero use MCT search during training and play, instead of alpha-beta search?

To understand the general problem with such an approach we need to think about the time that a neural network requires to compute output values from the input. That required time is very small or very large, depending on the context of the underlying use-case. Suppose that one pass through a neural network requires 100 milliseconds of computation time and consider the face comparison task that we described in Chapter 2, i.e. a border official queries a neural network with a photograph taken of a person upon entry together with the photo stored on that person's electronic passport. Then the network computes a score value that indicates whether these two photographs show the same person. If that takes 100 milliseconds that is a perfectly valid time.

Now imagine that we want to use such kind of network for alpha-beta search. This means we can only process ten chess positions per second — that is far too slow! To just give an intuition, Table 4.4 shows typical numbers of processed chess positions per second.[21] Of course a neural network as capable as the very deep network used for AlphaZero would be nice to have for alpha-beta search. But even a much simpler network would do — as long as it is more powerful than a handcrafted evaluation function, it will increase engine power. Moreover we have to note that while AlphaZero's results against Stockfish and state of the art Shogi programs were impressive, the gap between them was nowhere near as large as compared to the best available Go programs.

---

[21]Of course, the number are highly dependent on which computing power is available on the system. The numbers for Stockfish 8 and AlphaZero are the ones reported by the Deepmind team.

Table 4.4: Chess Engines Speed

|             | Positions per Second | Search Algorithm |
|-------------|---------------------|------------------|
| DeepBlue    | 200,000,000         | Alpha-Beta       |
| Stockfish 8 | 70,000,000          | Alpha-Beta       |
| AlphaZero   | 80,000              | MCTS             |

In 2018 Yu Nasu proposed to use a very specific kind of network which he dubbed *efficiently updatable neural networks (NNUE)*. And this is where he used really all tricks of the trade. He considered all layers of abstractions: The network architecture, the nature of Shogi gameplay, as well as the hardware architectures of modern CPUs. At the end it all fits neatly together and his NNUE network achieved a speed that made it usable for alpha-beta search in computer Shogi. As a computer scientist I have to say that I was full of excitement and enjoyment when finally understanding this very neat architecture.

Motohiro Isozaki teamed up with Yu Nasu to integrate an NNUE network into his state-of-the art open source computer Shogi program YaneuraOu. The whole NNUE revolution went largely unnoticed in the computer chess community. First, the network was specific to Shogi and it was not clear whether the idea would work for western chess as well. Second the original paper is in Japanese only[22] and the language barrier is simply very high.

As mentioned the open-source computer Shogi scene is smaller and limited in resources compared to western chess. When developing YaneuraOu, Motohiro Isozaki used GPL licensed code from Stockfish and adapted it into YaneuraOu which is also open source software licensed under the GPL; in particular he took inspiration of the well-engineered search algorithm of Stockfish.

Nevertheless he felt that merely abiding to the license terms was not enough. In a true open source spirit he felt obliged to give back, and ported an experimental version of the NNUE network back to Stockfish. This made some huge impact and eventually resulted in Stockfish 12 which uses NNUE for position evaluation instead of the old-fashioned handcrafted evaluation function. NNUE brought an increase of around 80 Elo[23] points to Stockfish and as of now, Stockfish is the number one engine beating Leela Chess Zero by a significant margin.

---

[22]I provide a rough translation at `http://github.com/asdfjkl/nnue`

No book about computers is complete with really bad car analogies. So let's get it over with: When AlphaZero is the Benz Patent-Motorwagen, then NNUE is the Ford Model T of computer chess. We should not underestimate AlphaZero's achievement, but due to the huge computing power required, the underlying chess engine was not available for the general public. Leela Chess Zero tried to remedy that, but even Leela Chess Zero requires significant hardware to have it run at an acceptable speed — the best and latest graphic card is a must. With NNUE we can run the best of both worlds — classic alpha-beta search and a neural network for position evaluation — on a standard personal computer without requiring an expensive graphic card.

Now let's take a deep dive into NNUEs. As always, we start with the input and output of the network before taking a look at the architecture.

### 4.6.1   Network Input and Output

The output is the most simple one. We have one single output node and that value denotes the evaluation of the position in centipawns.

The input on the other hand is slightly... quirky. It is dubbed HalfKP which stands for Half-King-Piece relationship and is a binary encoding of a chess position. For that we first take the role of the player whose current turn it is. Then we enlist all possible triples

(own king position, own piece type, position of that piece)

as well as

(own king position, enemy piece type, position of that piece).

---

[23]for current chess engines that's *a lot*

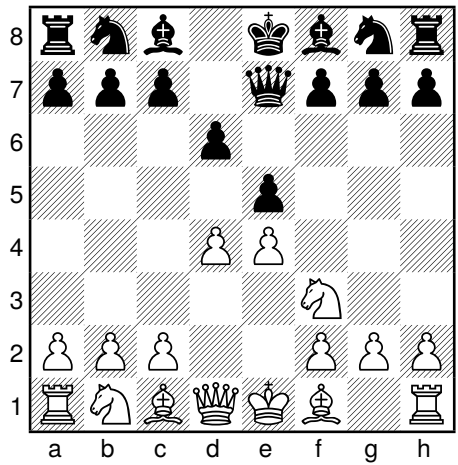

Figure 4.14: Typical position in the Philidor opening. White to play.

This is the first half of the encoded position. Then we take the role of the other player and repeat it. This is the second half of the encoded position.

It is best illustrated with a small example. Consider Figure 4.14. Black just played Qe7 and it is White to play. Thus we first take the role of the white player. We fix a king position and a piece type other than the king — note that we distinguish between our own and enemy pieces — and start enumerating them. Assume we always start at a1 and work towards h8:

- own king on a1, own pawn on a1: $0$[24]

- own king on a1, own pawn on a2: 0

- own king on a1, own pawn on a3: 0

- own king on a1, own pawn on a4: 0

- . . .

---

[24]This is of course impossible and will always be zero. Omitting it is also possible, but including it makes programming the encoding slightly easier.

This is pretty boring but once we are at "own king on e1" things start to get interesting:

- own king on e1, own pawn on a1: 0

- own king on e1, own pawn on a2: 1

- own king on e1, own pawn on a3: 0

- own king on e1, own pawn on a4: 0

- . . .

- own king on e1, own rook on a1: 1

- own king on e1, own rook on a2: 0

- . . .

- own king on e1, enemy pawn on a1: 0

- . . .

- own king on e1, enemy pawn on a7: 1

- . . .

- own king on e1, enemy rook on h8: 1

You probably get the idea. Next we repeat that for the enemy player (here this is Black). We get ones for "enemy king on e8, own pawn on a2", "enemy king on e8, enemy rook on a1" and so on as well as "enemy king on e8, enemy pawn on a7" and "enemy king on e8, enemy rook on a8".

The let's just quickly calculate on how many bits we require for this encoding. Let's start with the first half, i.e. fixing the position of our own king. There are 64 possible squares for the king. Next there are ten possible pieces that we can consider except for our own king and the enemy king, namely our own five piece types (rook, knight, bishop, queen and pawn) and the same again for the enemy. There are potentially 63 positions for these pieces each as a piece cannot be on the same square as the king, but as mentioned before when implementing this in software it is probably easier to just consider all 64 possible squares. This makes $64 \times 64 \times 10 = 40960$[25] input bits, and thus $2 \times 40960 = 81920$ input bits for both halves. The whole idea of this rather odd encoding will become clear in the next section. But without going into details let's note a few properties of this encoding.

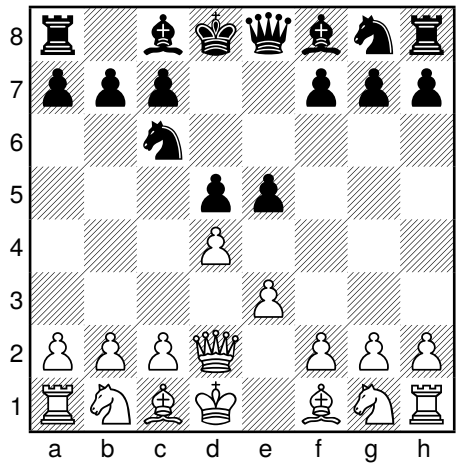

Figure 4.15: Typical position in the Philidor opening. White to play.

1. Mirrored positions result in same input values since we consider a position from the player whose move it is, agnostic of the actual color. Consider for example Figure 4.15 and assume that it is Black to move. Then the encoding is exactly the same as for the position depicted in Figure 4.14.

2. With the exception of the end game, kings are typically moved much less in a chess game than other pieces. Thus when making a move, only a few bits of the input change. Consider Figure 4.16 which occurs from the position of Figure 4.14 after the move Nc3. In principle only the following bits change:

   - own king on e1, own knight on b1 from 1 to 0

---

[25]In Stockfish it's actually a few more bits, but that is an artifact of the straight-forward port from Shogi where more input bits are required.

- own king on e1, own knight on c3 from 0 to 1

- enemy king on e8, own knight on b1 from 1 to 0

- enemy king on e8, own knight on c3 from 1 to 0

Of course we also have to take into account that the perspective of the player whose current turn it is has changed. Therefore we have to also flip "own" with "enemy" but this is a mere technicality:

- enemy king on e1, enemy knight on b1 from 1 to 0

- enemy king on e1, enemy knight on c3 from 0 to 1

- own king on e8, enemy knight on b1 from 1 to 0

- own king on e8, enemy knight on c3 from 1 to 0

3. The input encoding is highly inefficient in that there are way more input bits than necessary. A more straight-forward way to encode inputs is to just use bitboards as done in e.g. AlphaZero. There we take 8x8 bits for each piece type and set a bit at position (x,y) to one if that piece is present on (x,y) and to 0 otherwise. Now there are six possible pieces for the white player (king, queen, bishop, knight, rook, pawn) as well as six possible pieces for the black player. All in all we merely need $64 \times 6 \times 6 = 768$ bits opposed to 40960 bits in our case. As inputs will be associated with weights when building our network, the network will be heavily over-parameterized, i.e. there are more parameters in the network than (theoretically) required, especially when considering the amount of possible training data. It is known that overparameterized networks tend to learn better and generalize well. From a stochastic viewpoint it is an interesting problem to understand why this works, and this is subject of current research [DZPS19, ALL19]. Usually however this overparametrization is part of the inner network structure, i.e. by creating a deeper network by adding more hidden layers or increasing the size of the hidden layers. Here it is explicitly modeled by the input layer. The reason for this rather odd choice is entirely due to optimizing the network for ultra-fast computation, while still maintaining some sort of overparametrization in the hope that the network will learn well and generalize from the training data. How this input encoding allows ultra-fast computations is subject to the next section.

4. The encoding focuses heavily on the relation of pieces to their king. While it is true that a king rarely moves (at least during the opening and middle game), this choice also strikes as slightly odd. This choice of encoding seems to be entirely based in the game of Shogi. As mentioned before, Shogi is more similar to Bughouse or Crazyhouse chess in that captured pieces can be put pack in the game as own pieces. Thus there is an increased chance of being checkmated. Moreover Shogi has no castling move. Instead, players manually build a fortress around their king by putting pieces around it such as gold or silver generals and pawns. Such kind of fortress is known as a castle (kakoi) in Shogi. There are different kinds of castles and a whole theory around the different castle types. Apparently the positional relation of own and enemy pieces to one's own king are quite important. As mentioned, NNUE was originally developed for Shogi and later ported to Stockfish, and it was empirically discovered that such kind of encoding works for chess as well — but there is currently no satisfying theoretical explanation for that.

## 4.6.2   Network Architecture

The whole network is shown in Figure 4.17. It is a comparatively small network with only three hidden layers. The input bits are divided into two halves: The first half comprises all bits where we consider the location our (or rather the player whose turn it is in the current position) own king plus the pieces in relation to our own king. The second half comprises all bits where we consider the location of the enemy and the pieces in relation to that position.

The next layer consists of two halves of 256 nodes each. Each of the halves of the input layer is fully connected to the corresponding half of 256 nodes (but there is no connection between non-corresponding halves). This layer is then fully connected to a next layer of 32 nodes — finally mixing everything together — which is itself fully connected to another layer of 32 nodes. This layer is then

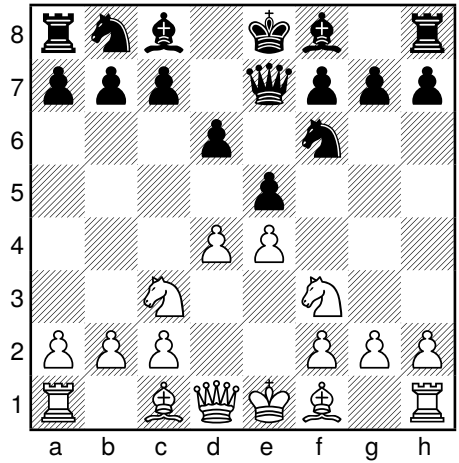

Figure 4.16: Mirrored Philidor opening position. Black to play.

fully connected to a final single node which outputs the evaluation value.

Throughout the whole network, (clipped) ReLU is used as the activation functions. Values smaller or equal to zero are mapped to 0, values in between 0 and 1 are mapped to themselves, and values larger or equal than 1 are clipped to 1.

There is one non-standard oddity: The weights that are used to connect the first half of the input layer (i.e. the half of one's own king) to the corresponding half of the first layer are the *same* weights that are used to connect the second half of the input layer to the second half of the first layer. They are shared in the sense that *mirrored* piece-square relations share the same weight.

Consider for example the weight that corresponds to the arrow that connects the bit "own king on e1, own pawn on d2" to the first node of the first layer. This is the same weight that is used for the bit "enemy king on e8, enemy pawn on d7". At first sight this really does not make any sense, especially if we consider that it should make a huge difference if it is Black or White to move and thus the weights should be different. While this is correct and sharing weights in that way is certainly restricting in terms of expressiveness[26], this special arrangement makes it possible to efficiently compute the two 256 values of the first layer for a given position as we will see below. First however we will take a look on how to efficiently compute the values of the second, third and final layer.

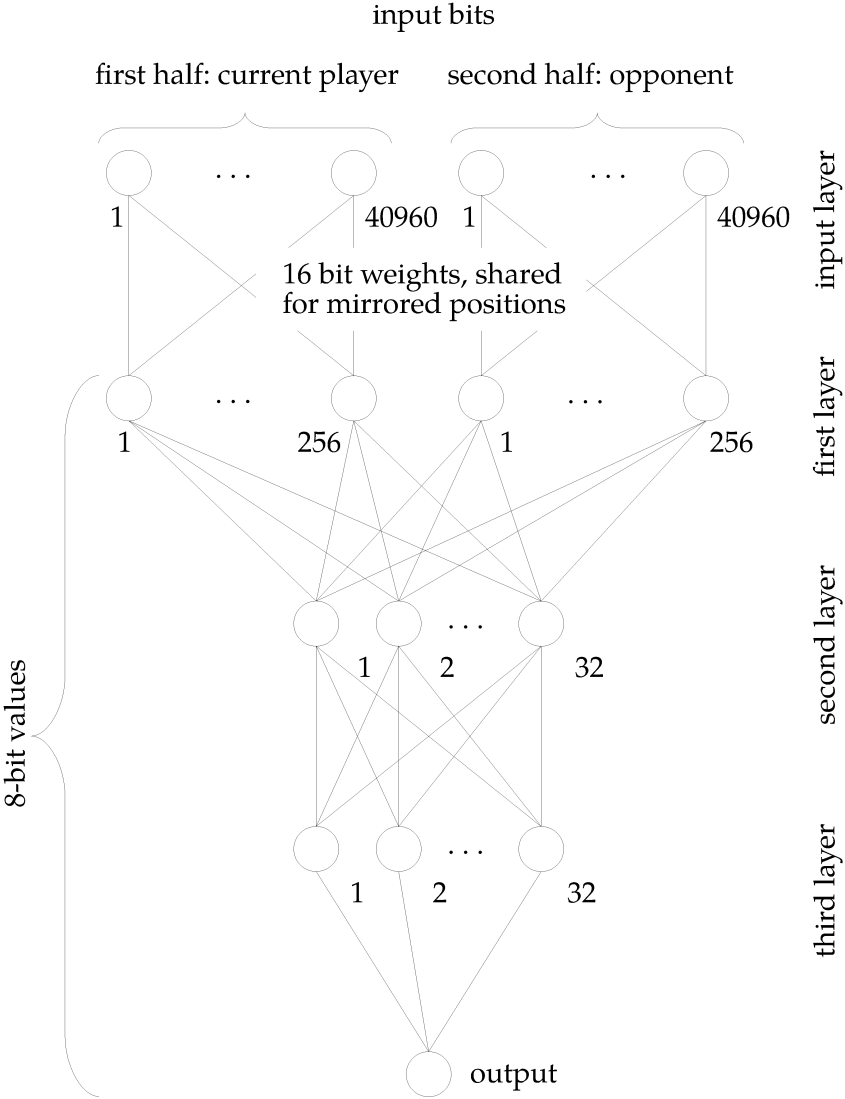

Figure 4.17: NNUE network architecture.

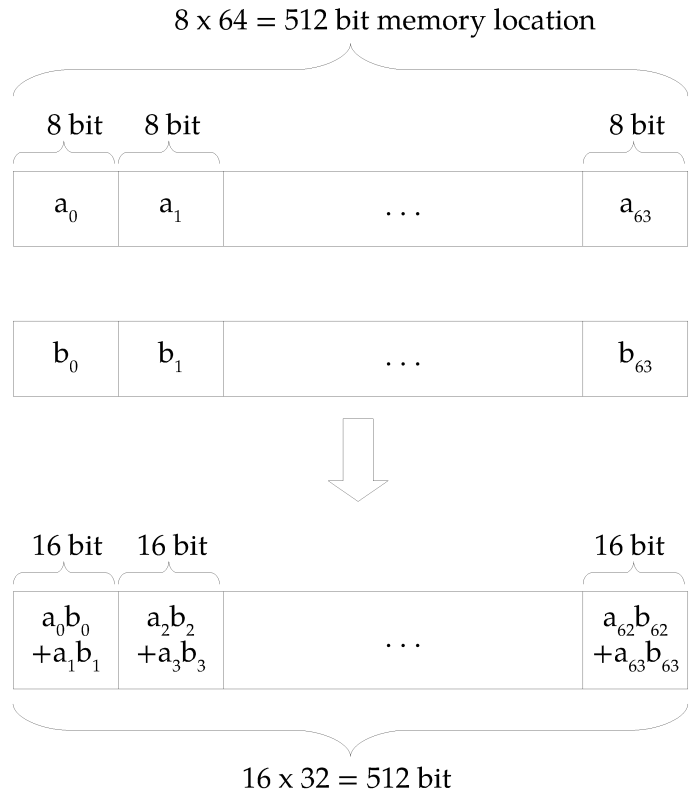

Figure 4.18: (V)PMADDUBSW.

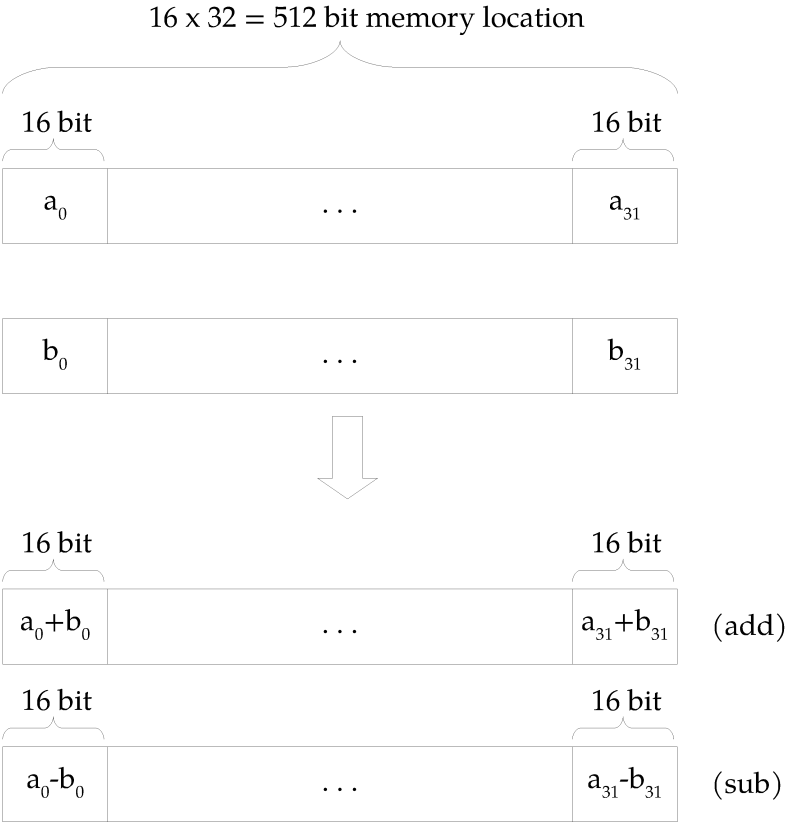

Figure 4.19: (V)PMADDW and (V)PMSUBW.

The general underlying idea is that for these layers, we consider a specific way of representing the weights by 8 bit values so that we can use acceleration features of modern CPUs, namely single-instruction-multiple-data (SIMD) extensions.

We already talked briefly about vectorization in Chapter 2 but let's dive further into the topic now. I should mention that in the further discussions I will completely omit the bias weights of the network for the sake of clarity. Adding biases just means we add constant values to the weight computation and all ideas work in the same way, but notation will become more intricate.

We will now take a look on how to compute the values for the second layer. The second layer receives 512 inputs from the first layer which are then used to compute the values for the 32 nodes of the second layer. Let write $w_j^i$ for the weight that connects input $j$ from the first layer to node $i$ of the second layer, and let's denote with $x_i$ the $i$th input. We thus have two matrices

$$\mathbf{w} = \begin{bmatrix} w_0^0 & w_1^0 & \cdots & w_{511}^0 \\ w_0^1 & w_1^1 & \cdots & w_{511}^1 \\ & & \vdots & \\ w_0^{31} & w_1^{31} & \cdots & w_{511}^{31} \end{bmatrix}, \text{ and } \mathbf{x} = \begin{bmatrix} x_0 \\ x_1 \\ \vdots \\ x_{511} \end{bmatrix} \tag{4.1}$$

Multiplying these matrices $\mathbf{wx}$ yields precisely the values of the second layer

---

[26]We can somehow take care of this misjudgement when interpreting the final output value by e.g. adding a small bonus to the evaluation for the side whose turn it is.

that we need:

$$\mathbf{wx} = \begin{bmatrix} w_0^0 x_0 + w_1^0 x_1 + \ldots w_{511}^0 x_{511} \\ w_0^1 x_0 + w_1^1 x_1 + \ldots w_{511}^1 x_{511} \\ \vdots \\ w_0^{31} x_0 + w_1^{31} x_1 + \ldots + w_{511}^{31} x_{511} \end{bmatrix} \tag{4.2}$$

CPUs are general all-purpose processing units. They can quickly add, subtract and multiply integer numbers. The size of these integers is determined by the word size of a CPU. A 64 bit CPU for example has registers where one register can hold 64 bit.

A CPU inherently however does not necessarily have good support for floating point numbers. Early computer systems like the 8 bit home computers of the 80s had no floating point support at all. Floating point operations had to be simulated in software which is very slow. Even early Intel chips had no such support — the first Intel chip with integrated floating point support was the 486 DX processor line. For 486 SX processors and earlier you had to buy a separate 80x87 co-processor. The reason for this is that space on the chip is very expensive and chip makers have think carefully on what users want and are willing to pay for. Even today with an integrated floating point processor, integer operations are simply more efficient to compute.

Since multimedia application, such as video decoding, became more and more popular, Intel decided to add various extensions for mathematical operations. They began with so called multimedia extensions (MMX) and later added single-instruction multiple-data (SIMD) extensions. As an example we will take a look at the SIMD operation (V)PMADDUBSW which works on two inputs. Often, such an input is a single integer value, but SIMD instructions are special. The bit size of each of these inputs depends on the version of (V)PMADDUBSW; various extensions have been added by Intel over time. For the ease of presentation we assume that both inputs have a size of 256 bits.

Consider Figure 4.18. The 256 bits of each input are split into 8 bit subwords

$$a_0, a_1, \ldots, a_{31} \text{ and } b_0, b_1, \ldots, b_{31}.$$

Each of these 8 bit subwords $a_i$ and $b_i$ are interpreted as an integer value in the range of $0 \ldots 2^8 - 1 = 255$. (V)PMADDUBSW then computes

$$a_0 * b_0 + a_1 * b_1, a_2 * b_2 + a_3 * b_3, \ldots a_{30} * b_{30} + a_{31} * b_{31}.$$

And this computation is very fast, as the circuit for this computation is present as silicon on the CPU.

The first step in optimizing the network for fast computation with a CPU was thus to forget about floating point numbers. Each intermediate value as well as the weights are therefore expressed as 8 bit integer values. Of course we loose precision as we can only distinguish $2^8 = 256$ different weight values. On the other hand we use the clipped ReLU as the activation function, so we will clip anyway at values below of 0 and above of 1, so we need precision mostly to express the range between $0 \ldots 1$.

We can then compute for each $i$ in the range $0 \ldots 31$ the value $w_0^i x_0 + w_1^i x_1 + \ldots + w_{511}^i x_{511}$ by multiple calls to (V)PMADDUBSW. Start with

$$w_j^0, w_j^1, \ldots w_j^{31} \text{ and } x_0, \ldots x_{31}$$

as inputs and use (V)PMADDUBSW to get

$$w_j^0 x_0 + w_j^1 x_1, w_j^2 x_2 + w_j^3 x_3, \ldots w_j^{30} x_{30} + w_j^{31} x_{31}.$$

Then repeat this with multiple calls for indices 32 to 512, and sum up all the results.

This is only the brief idea. The ReLU activation function can be computed with similar optimizations using SIMD extensions. As there are different SIMD commands depending on each processor generation, further optimizations are possible.

Nevertheless, these computations are not even the main bottleneck. Instead this is the input layer. After all, for the first layer we have 512 nodes and for the second and third layer we have 32 nodes. But for the input layer we have more than 80000 values to compute!

One important point of the input layer is to note that a lot of input bits will be zero. All bits with own king on the correct square will be set to one, but all the other bits where our king is on a different square will be zero and these are of course many more. The same holds for the enemy king. These inputs are then multiplied by weights and summed up to create the values of the first layer, but multiplying something with zero yields zero, and if we add up a lot of zeros we still have zero.

Now what happens if we have computed the values for the first layer from the input layer and then apply a move? Let's assume it was White to move, i.e. White was "our side" and thus comprises the first half of the input bits, whereas Black corresponds to the second half of the input bits. After the next move we switch sides and the first half has the input bits of Black (who is now to move and thus "our side"), and White will become the enemy side and thus fill the input bits of the second half. In both cases the bits w.r.t. White's move change. However note that weights are shared for a flipped position and therefore not much changes w.r.t. the computation of the values of previous positions. Both for the first and second half almost all input bits are the same as we have an (almost) flipped version of each half where only the bits corresponding to the last move changed. The idea is now to just add or subtract the differences that are the result of the move from the previous values to get the new values instead of recalculating *all* values. This will become clearer once we write everything down in matrix notation. Let $w_i^j$ be the weight of input $j$ that connects to node $i$ of the first layer, and let $x_i$ denote the input bit $i$. For the sake of the example we consider only one half of inputs — of course this re-computation has to be done for both sides. We have

$$\mathbf{w} = \begin{bmatrix} w_0^0 & w_1^0 & \cdots & w_{40959}^0 \\ w_0^1 & w_1^1 & \cdots & w_{40959}^1 \\ & & \vdots & \\ w_0^{255} & w_1^{255} & \cdots & w_{40959}^{255} \end{bmatrix}, \text{ and } \mathbf{x} = \begin{bmatrix} x_0 \\ x_1 \\ \vdots \\ x_{40959} \end{bmatrix} \tag{4.3}$$

and we want to compute

$$\mathbf{wx} = \begin{bmatrix} w_0^0 x_0 + w_1^0 x_1 + \ldots w_{40959}^0 x_{40959} \\ w_0^1 x_0 + w_1^1 x_1 + \ldots w_{40959}^1 x_{40959} \\ \vdots \\ w_0^{255} x_0 + w_1^{255} x_1 + \ldots + w_{40959}^{255} x_{40959} \end{bmatrix} \quad (4.4)$$

However a lot of the $x_i$ are zero and thus instead of multiplying the two matrices (as would be done on a general neural network framework with GPU support) it makes much more sense to compute the desired values iteratively.

1. We first initialize a result variable $r = \begin{bmatrix} 0 \\ \vdots \\ 0 \end{bmatrix}$

2. Then for each $i$ in the range $0 \cdots 40959$

   - if $x_i$ is zero then immediately skip and continue with $i := i + 1$

   - if $x_i$ is not zero, then set $r := r + x_i \begin{bmatrix} w_i^0 \\ \vdots \\ w_i^{255} \end{bmatrix}$

Note that $r = \mathbf{wx}$. Since there are only few cases where $x_i$ is different from 0, this will be a short loop. Moreover since $x_i$ is an input bit it can only be either 0 or 1, and thus

$$x_i \begin{bmatrix} w_i^0 \\ \vdots \\ w_i^{255} \end{bmatrix} \text{ simplifies to just } \begin{bmatrix} w_i^0 \\ \vdots \\ w_i^{255} \end{bmatrix}$$

Now we are almost there. Suppose that we are in a position where we computed the values of the first layer with the method above and obtained a vector $r$ that has one column and 255 rows. Assume we now execute a move that is not a king move. Then there is only one input bit that changes from 1 to 0 for the location where the piece is picked up — say $x_{42}$ — and one input bit that changes from 0 to 1 where the piece is placed, say $x_{55}$. To get the new values $r'$ we now have to take the old vector $r$, subtract the weights that are now zero due to $x_{42}$ and add the weights for $x_{55}$. In other words

$$r' = r - \begin{bmatrix} w_{42}^0 \\ \vdots \\ w_{42}^{255} \end{bmatrix} + \begin{bmatrix} w_{55}^0 \\ \vdots \\ w_{55}^{255} \end{bmatrix}$$

Note how *much* simpler this operation is than recomputing everything. Let's add a chess annotation symbol to the last sentence to make sure the emphasis conveys: Note how *much* simpler this operation is than recomputing everything!!

Now we also understand where the term *efficiently updatable neural network* stems from. We can efficiently update the network weights by computing incremental differences w.r.t. a move that changes the position.

We can even make this faster by using SIMD instructions VPADDW and VPSUBW that bulk add or subtract values and are depicted in Figure 4.19. They come for different bit sizes; for simplicity assume they operate on 128 bit values. The bits of each 128 bit input are split into 16 bit subwords $a_i$

$$a_0, a_1, \ldots, a_{32} \text{ and } b_0, b_1, \ldots, b_{32}.$$

Each of these 16 bit subword $a_i$ and $b_i$ is interpreted as an integer value in the range of $0 \ldots 2^{16} - 1 = 65535$. VPADDW then computes[27]

$$a_0 + b_0, a_1 + b_1, \ldots, a_{16} + b_{16}$$

and VPSUBW does the same thing for subtraction. If we represent the weights of the input layer as 16 bit values we can use multiple calls to VPADDW and VPSUBW to compute the difference as defined above. Since each vector holds 255 rows, we will require roughly $256/32 = 8$ calls to VPADDW and VPSUBW each.

That's pretty fast.

---

[27]Some issues can arise when the result does not fit into the desired 16 target bits but we won't delve into these technicalities here.

### 4.6.3    Training the Network

Training the network is pretty straight forward. We need pairs of positions and evaluation scores. We can bootstrap a network by taking positions from existing games or games of self play, and let an existing chess engine evaluate a position deeply to output a score. We then use these position/score pairs to train our neural network.

We can even apply some kind of reinforcement learning in that we let our bootstrapped engine with the trained neural network re-analyze deeply each position or new positions and output new evaluation scores. We then use these new pairs of position/evaluation score to further train the network and repeat.

Interestingly at the time of writing this, Stockfish uses Leela Chess Zero's training data, i.e. positions from self-played games of Leela and her evaluation scores.

Note that this is a much less complex network compared to say Leela Chess Zero. Training requires nowhere near the effort. Training a network is probably not something you would start on a Friday afternoon and expect it to be finished before 5pm, but creating one is certainly doable in a few weeks. More time and a lot of experimentation is of course required if you want to create a network that is strictly *better* than the existing one used by Stockfish.

It is nice to see how the open-source spirit of sharing ideas and code among Leela Chess Zero, Stockfish and YaneuraOu brought great progress and in the end benefited everyone.

### 4.6.4    NNUE in Practice

In order to evaluate the performance of AlphaZero, the DeepMind team conducted several test matches against the — back then — state of the art chess engine Stockfish 8. That version of Stockfish was a heavily optimized alpha-beta searcher with a carefully hand-crafted and semi-automatically tested evaluation function. AlphaZero beat Stockfish 8 convincingly. In particular AlphaZero showed superior positional understanding and identified long-term advantages which Stockfish was unable to spot. A detailed analysis of the games is provided in [SR19].

Let's have a look at one of the games, namely AlphaZero vs Stockfish Games: Game 8. An interesting analysis is given by Grandmaster Daniel King on his Youtube channel[28]. The game goes

1. d4 Nf6 2. c4 e6 3. Nf3 b6 4. g3 Bb7 5. Bg2 Bb4+ 6. Bd2 Be7 7. Nc3 c6 8. e4 d5 9. e5 Ne4 10. O-O Ba6 11. b3 Nxc3 12. Bxc3 dxc4 13. b4 b5 14. Nd2 O-O 15. Ne4 Bb7 16. Qg4 Nd7 17. Nc5 Nxc5 18. dxc5 a5 19. a3 axb4 20. axb4 Rxa1 21. Rxa1 Qd3 22. Rc1 Ra8 23. h4 Qd8 24. Be4 Qc8 25. Kg2 Qc7 26. Qh5 g6 27. Qg4 Bf8 28. h5 Rd8 29. Qh4 Qe7 30. Qf6 Qe8 31. Rh1 Rd7 32. hxg6 fxg6 33. Qh4 Qe7 34. Qg4 Rd8 35. Bb2 Qf7 36. Bc1 c3 37. Be3 Be7 38. Qe2 Bf8 39. Qc2 Bg7 40. Qxc3 Qd7 41. Rc1 Qc7 42. Bg5 Rf8 43. f4 h6 44. Bf6 Bxf6 45. exf6 Qf7 46. Ra1 Qxf6 47. Qxf6 Rxf6 48. Ra7 Rf7 49. Bxg6 Rd7 50. Kf2 Kf8 51. g4 Bc8 52. Ra8 Rc7 53. Ke3 h5 54. gxh5 Kg7 55. Ra2 Re7 56. Be4 e5 57. Bxc6 exf4+ 58. Kxf4 Rf7+ 59. Ke5 Rf5+ 60. Kd6 Rxh5 61. Rg2+ Kf6 62. Kc7 Bf5 63. Kb6 Rh4 64. Ka5 Bg4 65. Bxb5 Ke7 66. Rg3 Bc8 67. Re3+ Kf7 68. Be2 1 - 0.

Daniel King identified two key positions where Stockfish 8 failed to assess the position correctly and that were essential for the outcome of the game. And there is another position that is interesting in the sense that the current Stockfish identifies the move by Stockfish 8 in that position as a crucial mistake.

The first position is shown in Figure 4.20. AlphaZero is a pawn down but still finds this position very playable and thinks there is more than enough compensation. Let's compare the following engines: It is interesting to have Fritz 6 take a look at it for historical reasons. Stockfish 8 was the version AlphaZero played against. Stockfish 11 was the last Stockfish release prior to incorporating NNUE whereas Stockfish 12 was the first version to include it. Finally, Stockfish 14 was the latest Stockfish when writing these lines. Each engines was run for a few minutes with six threads on an Intel Core i3 10100F with 512 MB hash size; with the exception of Fritz 6 which of course only runs single threaded and with a mere 1 MB hashtable. The binaries are `stockfish_8_x64_popcnt.exe`, `stockfish_20011801_x64_modern.exe`, `stockfish_20090216_x64_avx2.exe` as well as `stockfish_14_x64_avx2.exe` for

---

[28]https://www.youtube.com/watch?v=Ud8F-cNsa-k

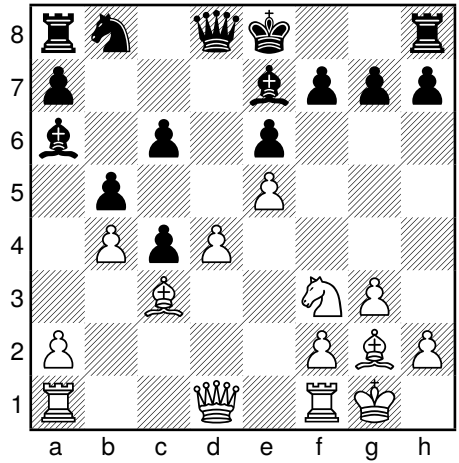

Figure 4.20: White to move. AlphaZero is a pawn down.

Stockfish 14. These evaluations are of course not tournament conditions but are just intended to give a more general hint how NNUE affects positional evaluations.

- Fritz 6 is happy to have snatched a pawn and thinks Black has an advantage of around -0.66.

- Stockfish 8 searching with around 7,500,000 nodes per second thinks Black has a slight advantage of approximately -0.12.

- Stockfish 11 agrees while also searching with around 7,500,000 nodes per second and evaluates this at -0.13.

- Stockfish 12 searches with approximately 4,600,000 nodes per second — the neural network evaluation takes time! Version 12 thinks this is an equal position (0.00) despite Black being a pawn up.

- Stockfish 14 (5,000,000 nodes per second) agrees with Stockfish 12 and considers this position absolutely equal.

The second key position according to Daniel King is shown in Figure 4.21. Here

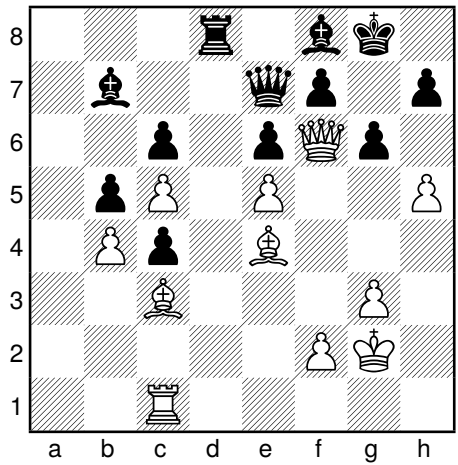

Figure 4.21: White to move. AlphaZero is able to assess this endgame.

the ability of AlphaZero to assess the endgame is crucial. Let's again check how our zoo of engines evaluates that position from Black's point of view.

- Fritz 6: I am doing great here and about to win. After all, I am a pawn up (-0.66).

- Stockfish 8: Ok, maybe White has some compensation for the pawn, but still I am a pawn up. It's equal then (0.00).

- Stockfish 11: I think something went wrong here (+0.25).

- Stockfish 12: You guys screwed up here. Like seriously. We are about to loose (+1.04).

- Stockfish 14: Yup, that definitely won't work out in the end (+1.12).

As can be seen, the impact of the NNUE evaluation is quite remarkable. When we compare Stockfish version 11 and 12, they differ in evaluation by 0.79 pawns — that's not only quite a difference but it also changes the perspective on the outcome of the game completely!

It is also interesting to let Stockfish 14 analyze Stockfish 8's moves and try to

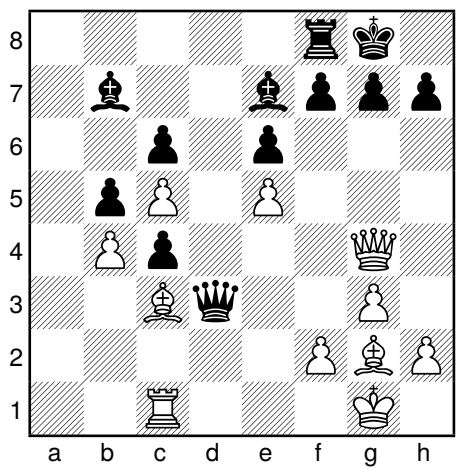

Figure 4.22: Black to move.

figure out where the biggest mistake happened. Turns out this is apparently
the position in Figure 4.22. Here Stockfish 14 thinks that White has already an
edge (+0.47) and suggests to play 22...Rd8 instead of 22...Ra8, the latter being
assessed as a blunder (+1.00). Stockfish 8 on the other hand thinks Black has a
slight edge here (-0.35) and does not see any complications and of course prefers
22...Ra8, and Stockfish 11 comes to the same conclusion. Stockfish 12 on the
other hand seems to also suffer from not being able to look far enough ahead.
It prefers 22...Ra8 and evaluates this at 0.00 requiring a few more moves to
understand what's going on. Only after 23...Qd8 it realizes White's advantage
(+0.91). Stockfish 8 on the other hand still thinks everything is perfectly fine
after 23...Qd8 (-0.38).

And Fritz 6 thinks at this point that it is only a matter of time until Black wins
(-0.87). After all, Black is a pawn up! What could possibly go wrong?!

As we can see, Stockfish made a huge leap forward in assessing long-term
positional advantages. And this is mostly due to NNUE. Moreover we must
remember that Stockfish is running on rather low-end office computer, whereas
AlphaZero requires an array of specially developed dedicated TPUs or at least modern GPUs.

### 4.6.5   Summary

Handcrafting evaluation functions for chess has always been a tedious task with lots of trial and error. It is very difficult to formulate our human understanding of chess into an objectively correct mathematical evaluation function. In the podcast "Let's talk about chess", Eric van Reem asked the author of the chess engine Shredder Stefan Meyer-Kahlen whether he ever got important feedback by chess grandmasters on what went wrong in games played by his engine. Stefan said that one time when he inquired on what specifically was wrong about a move he got the reply "Nah you just don't play like that".

It is very difficult to translate "Nah you just don't play like that" into a mathematical formula that computes evaluation scores for a position.

Similarly in his book about the Deep Blue matches with Kasparov [Hsu04], Feng-hsiung Hsu writes about the Deep Blue team's struggle to create a good evaluation function despite the raw processing power of the machine. He excitedly tells that they handcrafted an evaluation feature of their chess chips [Hsu99] for "potentially open files", i.e. files that are currently not open but *could* open in the future so that it is wise to put rooks there for the case that they open later on. He also explains how this might have helped in one game against Garry Kasparov.

The idea to use neural networks to automatically construct evaluation functions is not new. However before NNUE the only alpha-beta searcher that used a neural network for evaluation was DeepChess [DNW16], and that achieved "only" Grandmaster strength. While this was quite an achievement on its own, it was far from the super-human strength of current chess engines at the time.

Despite all the smart design choice w.r.t. the network architecture and implementation tricks using SIMD extension: After Motohiro Isozaki's code patches were officially integrated into Stockfish, the search depth almost halved due to the time spent for computing evaluation scores. Yet the strength of the engine improved by 80 Elo points due to it's better understanding of chess positions.

Nevertheless the time penalty of course is indeed a weaknesses. The latest development versions of Stockfish use a hybrid approach where NNUE is only applied for quiet positions (where a deep positional understanding is required) and for open positions (where pawn snatching might be enough to win) the previous handcrafted but quicker evaluation function is used. Stockfish gained another 20 Elo points by that.

## 4.7   Fat Fritz 2

Fat Fritz 2 is based on Stockfish and it's NNUE implementation and created quite some controversy. In an official blog post by the Lichess operators it was called "a ripoff" [29]. Stockfish authors noted that they *feel that customers buying Fat Fritz 2 get very little added value for money*[30]. Tord Romstad, author of the chess engine Glaurung and one of the first developers of Stockfish said on Twitter: *I'm so disappointed in ChessBase for selling FF2. It might be legal, but it's morally questionable and shockingly unoriginal.*[31]

We will not join this discussion but simply describe the changes that Chessbase did, including their main artificial intelligence expert Albert Silver. They took the source code of Stockfish and[32]

- changed the name of the file that stores the network weights to "Fat-Fritz2_v1.bin"

- changed one constant value that is used to scale the output of the neural network into a centipawn value that is better understood by Stockfish's search function from 679 to 1210

---

[29]`https://lichess.org/blog/YCvy7xMAACIA8007/fat-fritz-2-is-a-rip-off`, accessed July 9th, 2021.

[30]`https://stockfishchess.org/blog/2021/statement-on-fat-fritz-2/`, accessed July 9th, 2021.

[31]`https://twitter.com/tordr/status/1359428424255823875`, accessed July 9th, 2021.

[32]`https://github.com/official-stockfish/Stockfish/compare/550fed3343089357dc89ec f78ce8eb4b35bcab88...DanielUranga:faef72afbf10273ca8688a4ba1c7863426c93c6e`, accessed July 9th, 2021.

- changed the authors to "Stockfish Devs and Albert Silver (neural network)" and the engine id string from "Stockfish" to "Fat Fritz 2".

- The original input weights were not invariant to flipping a position but to rotation, an artifact from Shogi. This was changed to flipping and involved altering the value 63 to 56 at one position in the source code

- the layer sizes of the network were changed as this: The size of the first layer was increased from $2 \times 256$ to $2 \times 512$, and the sizes of the second and third layers were decreased from 32 to 16

Of course they also trained their own network. As of writing this, the resulting engine is strictly weaker than the current version of Stockfish in the TCEC benchmark[33].

The retail version comes with the following note in the fine print: "The Fat Fritz 2 Chess Engine is based on the software Stockfish. The Fat Fritz 2 Chess Engine and the software Stockfish are licensed under the GNU General Public License Version 3. You will receive further information during installation."

I will leave it up to you to decide whether you find Fat Fritz 2 worth shelling out 80 Euro for, and whether you agree or disagree with the Lichess developers that Fat Fritz 2 is "a ripoff". The previous section that described all the genius that went into creating, designing and implementing NNUE provides the basis to understand and value the differences that ChessBase introduced.[34]

## 4.8 Maia

How to become really good at playing bad chess?

In 2016 I was still trying to get better at chess. I sort of have given up. Not the hope of becoming better, but the actual training due to time reasons. But even back than I was very time-constrained. One very effective method of improving is apparently to play long over-the-board games and later analyze these games and one's mistakes in great detail. Unfortunately, this is also very time-consuming which is why I tried to play serious games against the computer. After all you can save approximately half the time compared to playing against a human opponent as even with only one second for every move, the computer is more than strong enough.

---

[33]https://tcec-chess.com/

[34]As of writing these lines, a lawsuit has been filed by the Stockfish developers against Chessbase w.r.t. an alleged GPL violation.

The major drawback I stumbled upon however is that weakening the engine down to my patzer level often resulted in very unnatural play. After a few moves, the computer would out of the blue sack a bishop for a pawn without any compensation. After that sacrifice however it would play like a grandmaster until the end of the game.

The reason is how alpha-beta searchers are usually made weaker: The result of a search will be a list of moves and their evaluations scores. Usually we sort this list by the evaluation values and take the best move. To get a weaker move we can have the user specify a handicap value in centipawns, or let the user select a desired ELO value and approximate a corresponding centipawn value to that ELO number. We can also slightly randomize the value within a given range so that the engine plays with more variety. Then just subtract this handicap value from the evaluation value of the best move and find a move with an evaluation that is close to that result. Thus this move is roughly a handicap worse than the best move.

The problem is now to make sure that this selected move makes any kind of sense. We can check for very obvious blunders by comparing the evaluation value of the best move with the evaluation value of the second best move. If the second move is say −5.0 and the best move is 1.0, then it is obviously a blunder and we should play the best move instead.

But sometimes the situation is not so clear. Consider the position shown in Figure 4.23. It is White to move. On my system the chess engine Fruit 2.1 considers Bb3 the (obvious) best move with an evaluation score of 0.0. The second best move is d4 with an evaluation of roughly −1.5 and Fruit gives several lines all around −1.5 for a continuation, for example the line 7.b4 bxa4 8.d5 Na5. Now imagine we are playing with a handicap of say 2.00, i.e. the engine is supposed to play with a handicap of roughly two pawns in each position. Is d4 a blunder? By our definition above probably not, the difference is "only" 1.5 centipawns and we are playing with an even higher handicap.

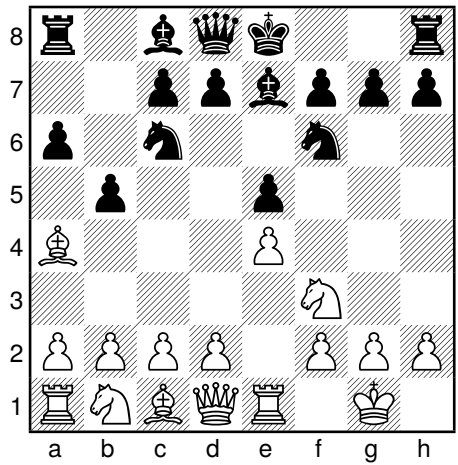

Figure 4.23: White to move. There is no reasonable alternative to *Bb3* that any chessplayer would play.

I think there is no discussion required to understand that any human chess player, even a complete patzer like myself, would move his bishop. In other words if a human player plays Black in this position as a training match against a computer, any move except from Bb3 would feel very unnatural.

We can try to remedy this situation by e.g. checking whether material is sacrificed and then always consider this a blunder, but even for such scenarios we can find examples where this leads to unnatural play. I mean even if we only consider the very best move an engine gives, these moves are at times very unnatural as watching any live commentary will prove, and moves are dismissed as "yeah but this is an engine move...". If we add some kind of artificial weakening, the situation certainly does not improve.

Then in January 2017 I stumbled upon the DeepChess paper [DNW16]. Needless to say I was impressed. I was quick to send out the following email to Eli David, one of its authors:

From: Dominik Klein
To: Eli (Omid) David

Dear Mr. David,

with high interest I read your paper on DeepChess. [...]

A frequent complaint from like myself is that computers are not
very good at playing _bad_. That is when trying to lower the
chess playing ability of computer programs, the most simple
approach is to limit search depth, or - what I've seen often
and what seems to be implemented in Fritz - is that less good
moves are considered (i.e. via uci- multi pv), then some
random centipawn value is chosen and within that threshold
a lesser good move is chosen.

These are very crude approaches and result in very un-human
play. I.e. for Fritz, the engine often sacs a piece early
on, and then plays like a grandmaster with very good technical skills.

Have you ever considered adjusting the training goal in
DeepChess to reflect human play? For example you took the
CCRL games as an input to train your neural networks.
Instead one could take a set of games played by amateurs
(i.e. FICS with players rated, say, ELO 1500) as input
in the hope that learning then will generate a position
evaluation that is incorrect but amateur-like (i.e. overlooking
typical patterns, like pins, longer combinations, positional
misjudgements etc.)

king regards

- Dominik Klein

to which he replied:

```
From: Eli (Omid) David
To: Dominik Klein
Completely agree with your observation. DeepChess may work
when trained on datasets of players with different ELO points.
I haven't had a chance to test it, but it is interesting to check...
```

Ok, this computer science researchers' lingo is a language of its own. Let me quickly translate this into plain English for you.

```
Hi Eli,
I have this super cool idea but neither the time nor the will
to put any effort into executing it. Could you do it for me?
- Dominik
```

```
From: Eli (Omid) David
To: Dominik Klein
Nope, you are on your own.
- Omid
```

Or in other words: One should never underestimate the hours of work that go into a research paper with "just" ten or twenty pages. Sketching out the approach, the programming, the nifty details — all that takes an incredible amount of time. And on top of that are the computational resources required to train such a deep network.

I was therefore very glad that someone independently came up with the idea and actually did put the effort into executing it. The result is Maia, an experimental research engine [MSKA20].

The idea is quickly explained. Young et al. created a network with a similar network architecture to the network of AlphaZero. But then they trained different networks, all by supervised learning. First they collected games from the free internet chess platform Lichess. They sorted these games according to the player's ratings, e.g. from 1200 to 1299, from 1300 to 1399 and so on. They considered only those games where both players were in the same rating range. Next they trained the network on these games (approximately 12 million games for each range) and tested move-matching accuracy on a number of games that were not used for training. Namely they queried their networks with a position from a real game and asked which move the network would select. Then they checked whether this move was actually played in the game by the human player.

Move matching accuracy varies with rating and among different approaches, but especially for lower rating levels the difference to existing engines is quite outstanding. For example for a rating level of 1100 their network achieves a matching accuracy of more than 50 perecent, whereas Stockfish is only in the 35 percent range. For higher levels the differences get lower — best engine moves coincide with the best moves humans play — but their network still outperforms Leela Chess Zero and Stockfish by 5 percent move accuracy.

So can we immediately turn these networks into chess engines? Sort of. There is still the problem of blunders of course. Especially on lower levels. There is especially the question on whether we should use these networks to search a position, at least a limited number of plys. Here they decided to not conduct any tree search, the move probabilities that a network outputs are solely responsible to select the next move.

These ranges of networks are dubbed Maia. And what's even more important is that they cooperated with the Lichess project and you can play against these networks online. So we can test how well Maia works in practice. As I am a bad chess player you should take my analysis with a grain of salt, but then again: If there is one thing that I *can* understand, then it is to tell if my opponent is a patzer like myself. Here is a quick blitz game that I played as White against the bot Maia1. It was initially trained to play around 1100, but its current Lichess blitz rating is 1487.

The game started with 1.e4 e5 2.Nf3 Qe7 3.d4 d6 4.Bc4

Now Maia answered with 4...Be6. While Philidor defense is a common opening among amateurs, the move Be6 feels slightly awkward, albeit not entirely impossible. First, most amateurs will remember the guideline "knights before bishops" w.r.t. development in the opening. Second, while 5.d5 is probably not even best in this position, it is a tempo move that is easily spotted even by an amateur.

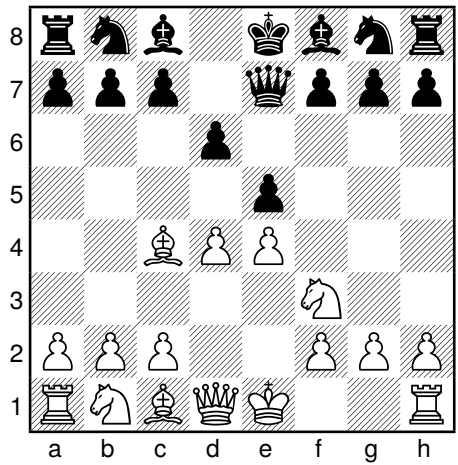

Figure 4.24: Author vs. Maia, position after 4.Bc4

5.d5 Bd7 6.Bg5 f6 7.Bh4

You can verify my claim that I am not a very skilled player here. The black-squared bishop will now have a very lone time for the rest of the game

7...g5 8.Bg3 h5 9.h4 g4 10.Nfd2 Bh6 11.Nc3 Bxd2+ 12.Qxd2 c6

According to Nimzowitsch you should attack a pawn chain at the base, i.e. at it's weakest point. But which amateur reads Nimzowitsch' lavish rants? Thus I think this is a very realistic mistake an amateur would make.

13.dxc6 Bxc6 14.Nd5 Bxd5 15.Bxd5 Nc6 16.O-O-O Nd4 17.c3 Nb5 18.Bc4 a6 19.Bxb5+ axb5 20.Qxd6 Qxd6 21.Rxd6 Rxa2

I am not sure whether this feels realistic. Even amateurs get the notion of open files and will identify that giving White the d-file is probably not a good idea. Rd8 might be even worse, but I think it is the somewhat more natural move for a weak player, instead of snatching the pawn on a2.

22.Kc2 Ne7 23.Rhd1 O-O 24.Rd7 Nc6 25.Rxb7 b4

While b4 might be objectively the best move here to get some kind of initiative,

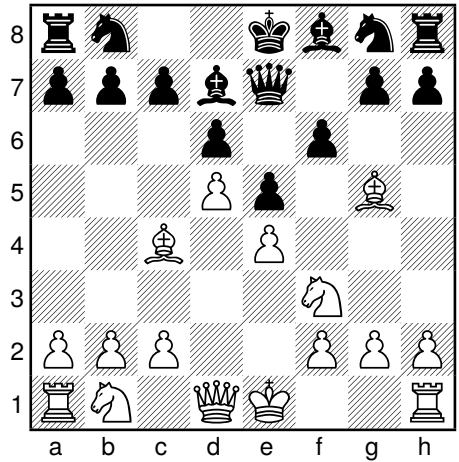

Figure 4.25: Author vs. Maia, position after 4.Bh4

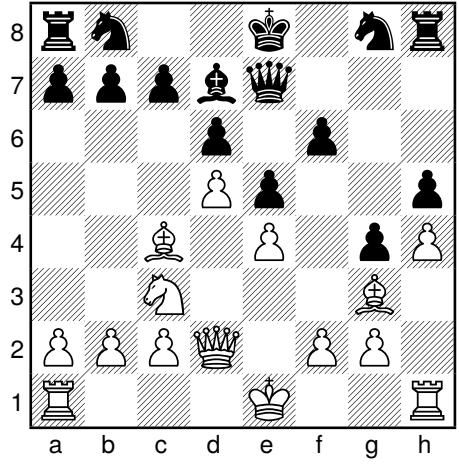

Figure 4.26: Author vs. Maia, position after 12...c6

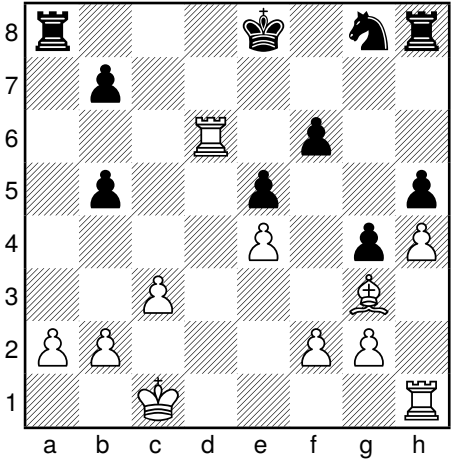

Figure 4.27: Author vs. Maia, position after 21.Rxd6

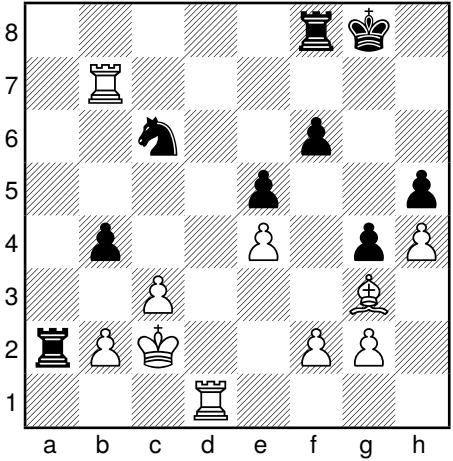

Figure 4.28: Author vs. Maia, position after 25.b4

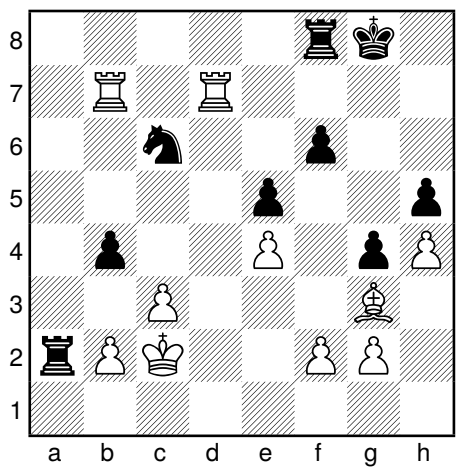

Figure 4.29:  Author vs.  Maia, position after 26.Rdd7

letting White invade the seventh rank with two rooks is also something I think most weak player would not allow or at least try to fight against.

26.Rdd7 bxc3

Now bxc3 feels very strange.  There is a very basic mate threat that most human amateurs will spot.  It seems this is where the decision to let the network spit out moves without any kind of calculation backfires.

27.Kxc3 Na5 28.Rg7+ Kh8 29.Rh7+ Kg8 30.Rbg7#

As you can see, Maia is not perfect yet.  But on the other hand the game also did not feel completely unrealistic and much closer to playing a human being than most of the other artificially weakened engines out there.  I am also quite sure that at some point the played games on Lichess will be used to improve the approach.

I encourage you to try it out on your own.  Not only is it free, but you can play against and thus train with the Maia bots at any time anywhere as long as you have an internet connection.

## 4.9 Conclusion

We have seen how neural networks outperform handcrafted evaluation functions. They can be trained by reinforcement learning through self-play or by supervised learning with existing games, but it seems that reinforcement learning will achieve better results in the end, as there is no inherent limit in the amount of training data. On the other hand supervised learning with high quality training material naturally provides faster training; and since the computing power required to train deep networks for two-player games such as Go, Shogi and chess is quite significant, this advantage should not be underestimated.

For computer Go there seems to be currently no alternative than to create a very deep network and train it. This will create a network with incredible positional understanding of Go — remember that even without any kind of search, the networks trained by Deepmind were very strong. Combined with Monte Carlo Tree Search we get a computer Go engine with superhuman strength. For other more complex strategy games — think computer games like StarCraft — with an incredible high branching factor this approach works as well [VBC+19]. This universality makes it also very appealing since we do not have to invest any kind of thought into encoding domain logic. Just throw the algorithms at your problem and the reinforcement learning procedure will figure everything out.

For Shogi and chess the situation is not so clear. Whereas in Go we had no strong — as in Grandmaster strength — computer programs, the lower branching factor of Shogi and chess allowed to implement alpha-beta searchers. And these already outperformed humans. In chess the breakthrough was the DeepBlue match in 1997, and for Shogi there were some interesting computer matches between 2007 and 2014 ultimately drifting the edge towards the computers. Therefore it is only logical that combining alpha-beta search with neural networks is likely the most promising approach, as shown by the NNUE network structure introduced by Yu Nasu.

AlphaZero has been described as a *game changer* [SR19]. This is certainly true for computer Go, but not necessarily for chess and Shogi. Sure, AlphaZero beat Stockfish convincingly, no doubt. But the ELO difference was comparatively low[35].  This is not meant to diminish the great achievement of AlphaZero, but it just hints that alpha-beta search is just a very powerful tool if the branching factor of a game is not too high.

What's the future like, then?  As industry demand is growing, a lot of resources are spent into making neural networks train and evaluate faster.  More and more chip vendors donate silicon space to integrate neural network accelarators into their chips.  Intel introduced a number of dedicated instructions dubbed Deep Learning Boost.  Apple added a "Apple Neural Engine" to their M1 chip.  And more and more, fast GPUs are integrated into CPUs as well.

Moreover there are also interesting developments on the software side.  Apple created their ML Compute framework, and Microsoft created DirectML, an abstraction layer that enables potentially any graphic card to speed up deep learning via Window's DirectX interface.  They even created an experimental port of tensorflow with DirectML support.

All this is unfortunately just not enough to create a deep neural network whose evaluation is fast enough to support a chess alpha-beta searcher.  Therefore encoding and architecture quirks will remain, even though the current NNUE architecture looks very much crafted for Shogi and it is likely that architectures more suited to chess will be discovered.

Over a larger time span though it seems likely that alpha-beta searchers with strong neural networks for position evaluation will become feasible and prevail over competing approaches.  With more and more computing power available we will likely also see NNUE quirks vanish and more "standard" networks to appear.  Nevertheless the computational aspect will remain very challenging due to the amount of network queries that alpha-beta search induces.

The question of creating computer opponents that do not just have human or super-human strength but rather make enjoyable human-like opponents is rarely investigated.  And if so the (applied) research with regard to entertainment computing tends to go more in the direction of creating game models that hook up users and create patterns of addiction, like all those freemium strategy

---

[35]Figure 1 in the AlphaZero paper [SHS+18] hints at less than 100 Elo points.

games or loot boxes; essentially masking gambling as computer games.

There is not much research on what makes games actually enjoyable for humans. Iida et al. [SPI14] investigated why and how games progress and are refined and relate that to the emotional impact they have on us. While it does not directly focus on creating a convincing human-like engine for chess or Shogi, it is nevertheless an interesting read. The development started by the team around Maia is therefore a refreshing step into a very interesting direction. This especially holds if we want to create chess engines for analysis that are not just very good in terms of raw ELO performance but rather play or suggest moves that can be understood (and later on played if used for game preparation) by mere humans. It is nevertheless this different kind of chess understanding that is the source of all the hype around AlphaZero.

# 5

# HexapawnZero

If you want a thing done well, do
it yourself!

---

Haruko Obokata

In order to get an intuition and a better understanding how AlphaZero works
in practice, it is worthwhile to implement it by yourself. However as we dis-
cussed, training a competitive network for chess takes enormous computational
resources, and the assumption is that you do not have a small data center in
your basement. We will therefore have look at a much simpler game. Often Tic
Tac Toe is used for such experiments, but while it is simple, it has unfortunately
not even a remote resemblance of chess.

The most simple game that somewhat looks like chess is probably *Hexapawn*.
It was invented by popular science writer Martin Gardner in 1962 [Gar62] to
showcase simple game mechanics and game trees. The rules are very simple:
Hexapawn is played on a 3x3 board. Each side has three pawns on the first
and third row respectively as seen in Figure 5.1. Pawns move as usual with the
exception of not moving two steps at the beginning — and there is obviously
no en-passent. A player wins if he can queen a pawn. Moreover there is no stalemate or draws: If it is a player's turn and he has no move, then he loses the game.

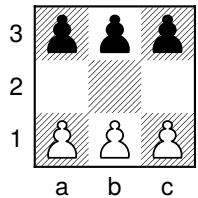

Figure 5.1: Hexapawn. Initial position.

Hexapawn is solved game, i.e. we know for each position who is going to win if he plays optimally. In fact, it is not difficult to verify that Black will always win the game if he plays with an optimal strategy: White can start with a2, b2 or c2, but the cases a2 and c2 are symmetrical. So let's start with a2. Black answers with bxa2, and then

1. If White plays bxa2, Black wins with c2, as White has no move.

2. If White plays b2 or c2, Black qeens with a1.

Now if White starts the game with b2 then Black answers axb2.

1. If White plays axb2, Black moves c2 and wins as White has no move.

2. If White plays a2 or c2, Black queens with b1.

3. If White plays cxb2, Black answers with c2. White's only legal move is a2 and Black queens with c1.

In other words this game is extremely trivial to solve and perfectly suited for our experiments.

In this section we will design a neural network akin to AlphaZero. It will accept states of the board as input and output move probabilities that denote how good a moves is resp. which move should be played, as well as a value that outputs the likely outcome of the game from the current state.

We will train this network with two different approaches and compare them. First, we will use supervised learning, i.e. we will generate training data. This training data consists of positions, the optimal move in a given position and the (calculated) outcome of the game in that position — note that such an outcome is known.

Second we will apply a "Zero"-like approach. That is we will implement the training loop of AlphaZero using Monte Carlo Tree Search in combination with self-play and reinforcement learning.

We will verify the effectiveness of our training by comparing the power of the trained supervised and "Zero"-network against an opponent who plays just random moves. A well-trained network that plays Black should beat such an opponent in the vast majority of all cases.

Let's start with the network architecture, i.e. the network itself, as well as network input and output.

## 5.1   The Network

**Network Architecture**   The network architecture is quite simplistic: We have an input bit vector of size 21. This input encodes a Hexapawn positions as well as the current turn, cf. below. The network itself has five hidden layers, namely connected layers each of size 128 with rectified linear activation.

Note that a network of this size is probably a complete overkill for such a simple game, but it is still computationally very cheap. Note also that we completely forget about network constructs such as convolutional layers or the like. These are required if we want to learn patterns and train our network to abstract strategic information and create a positional understanding. However the game tree of Hexapawn is so trivial that there really is no positional understanding required. It suffices if our network simply *remembers* all board states and the corresponding correct move and game outcome. This maybe puts the joy out of creating *HexapawnZero*, but it also makes training the network very simple.

The output of the network has two *heads*. The *policy head* is a fully connected layer with softmax activation and 28 outputs. These encode all possible moves of both the white and black player. Each output provides the move probability for the corresponding move and all outputs sum up to one.

The *value head* is just one output with tanh activation. It provides a value between -1 and 1 that denotes who is going to win in that position if both players play optimally.

For the policy head we will use categorical crossentropy to measure the difference between a training example and the output of the network. For the value head we will employ the mean squared error[1]. The overall loss of our network is then simply defined as the sum of the crossentropy and the mean squared error. We will train the network using good old gradient descent. It is straight forward to put this into `Python` code, create the network and let `keras` initialize it with random initial values. We will then save the current (untrained) state of the network for our experiments.

**Listing 5.1: Hexapawn Network**

```
1  inp = Input((21,))
2
3  l1 = Dense(128, activation='relu')(inp)
4  l2 = Dense(128, activation='relu')(l1)
5  l3 = Dense(128, activation='relu')(l2)
6  l4 = Dense(128, activation='relu')(l3)
7  l5 = Dense(128, activation='relu')(l4)
8
9  policyOut = Dense(28, name='policyHead', activation='softmax')(l5)
10 valueOut = Dense(1, activation='tanh', name='valueHead')(l5)
11
12 bce = tf.keras.losses.CategoricalCrossentropy(from_logits=False)
13 model = Model(inp, [policyOut,valueOut])
14 model.compile(optimizer = 'SGD',
15               loss={'valueHead' : 'mean_squared_error',
16               'policyHead' : bce})
17
18 model.save('random_model.keras')
```

---

[1]cf. Chapter 2

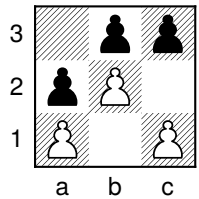

Figure 5.2: Hexapawn. White to move.

**Network Input** We will encode the position of the white and black pawns as bitvectors. For that we will iterate twice through the board; first for the white pawns and then for the black pawns. We go through the position starting with a3 and then moving to c3, next a2 to c2 and finally a1 to c1. Finally after encoding the pawn positions we will add three ones if it is White to move, and three zeros otherwise. An example of the encoding is shown in Figure 5.3, here this results in the 21 bit vector

$$000 \quad 010 \quad 101 \quad 011 \quad 100 \quad 000 \quad 111.$$

**Network Output** We simply enumerate all possible moves (of both Black and White) and associate each move with a unique number. Each number corresponds to one output of the policy head of the network.

We start with white moves forward and map them from 0 to 5:

| | | |
|---|---|---|
| $a1 - a2 : 0$ | $b1 - b2 : 1$ | $c1 - c2 : 2$ |
| $a2 - a3 : 3$ | $b2 - b3 : 4$ | $c2 - c3 : 5$ |

The same approach is used for the black moves forward:

$$a3 - a2 : 6 \qquad\qquad b3 - b2 : 7 \qquad\qquad c3 - c2 : 8$$
$$a2 - a1 : 9 \qquad\qquad b2 - b1 : 10 \qquad\qquad c2 - c1 : 11$$

Next we consider all possible captures by White:

$$a1 - b2 : 12 \qquad b1 - a2 : 13 \qquad b1 - c2 : 14 \qquad c1 - b2 : 15$$
$$a2 - b3 : 16 \qquad b2 - a3 : 17 \qquad b2 - c3 : 18 \qquad c2 - b3 : 19$$

and all possible pawn captures by Black:

$$a3 - b2 : 20 \qquad b3 - a2 : 21 \qquad b3 - c2 : 22 \qquad c3 - b2 : 23$$
$$a2 - b1 : 24 \qquad b2 - a1 : 25 \qquad b2 - c1 : 26 \qquad c2 - b1 : 27$$

Of course not all moves are possible in each position and during training or search the network may output illegal moves with a probability larger than 0. In such a case we have to mask out these probabilities.

## 5.2   Game Logic

Encoding the game logic requires some boiler plate code but is not too difficult. At the center is the object Board which encodes all information about a current position. The current state is stored in the list board. Index 0 here corresponds to the square a3, index 2 is c3 and so on. Accordingly we have a function that sets the starting position by setting the indices of board with the corresponding values.

**Listing 5.2: Board Initialization**

```
1 class Board():
2
```

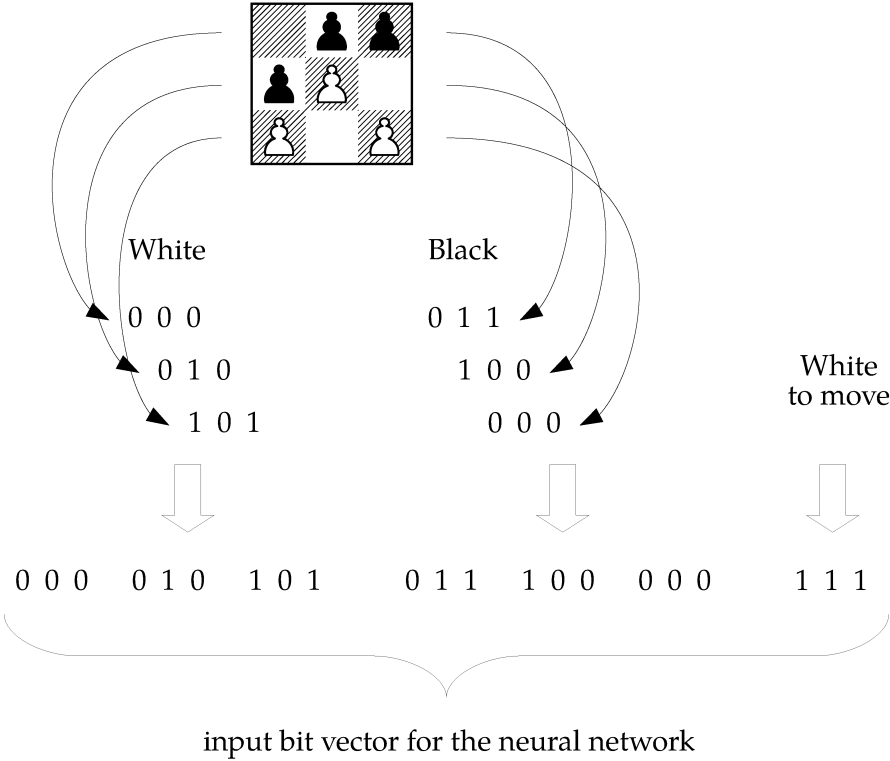

Figure 5.3: Hexapawn: Neural Network Input Encoding.

```
3    EMPTY = 0
4    WHITE = 1
5    BLACK = 2
6
7    def __init__(self):
8        self.turn = self.WHITE
9        self.outputIndex = {}
10       self.board = [self.EMPTY, self.EMPTY, self.EMPTY,
11                     self.EMPTY, self.EMPTY, self.EMPTY,
12                     self.EMPTY, self.EMPTY, self.EMPTY ]
13
14   def setStartingPosition(self):
15       self.board = [self.BLACK, self.BLACK, self.BLACK,
16                     self.EMPTY, self.EMPTY, self.EMPTY,
17                     self.WHITE, self.WHITE, self.WHITE ]
```

A move is encoded as a tuple where the first element is the index of the source-square and the second element is the index of the target square. To get the output index for a given move we create the python dictionary `outputIndex`. This dictionary stores information such that the string representation of a move yields the number of the output of the neural network. For that we use the convenience function `getNetworkOutputIndex`.

**Listing 5.3: Output Indices**

```
1  # white forward moves
2  self.outputIndex["(6, 3)"] = 0
3  self.outputIndex["(7, 4)"] = 1
4  self.outputIndex["(8, 5)"] = 2
5  self.outputIndex["(3, 0)"] = 3
6  self.outputIndex["(4, 1)"] = 4
7  self.outputIndex["(5, 2)"] = 5
8  ...
9  def getNetworkOutputIndex(self, move):
10     return self.outputIndex[str(move)]
```

Applying a move simply means setting the value of the target square to that of the source square, and then setting the source square to empty. Moreover we have to change the turn from White to Black (or vice versa). Also, we reset

`self.legal_moves` — this is the cache that contains all legal moves of the current position. As applying a move changes the position, we will need to re-compute all legal moves the next time they are required.

**Listing 5.4: Applying a Move**

```
1 def applyMove(self, move):
2     fromSquare = move[0]
3     toSquare = move[1]
4     self.board[toSquare] = self.board[fromSquare]
5     self.board[fromSquare] = self.EMPTY
6     if(self.turn == self.WHITE):
7         self.turn = self.BLACK
8     else:
9         self.turn = self.WHITE
10    self.legal_moves = None
```

To generate moves we iterate over all squares of the board and check whether the square is occupied with a pawn that has the color of the current player. The target square of a forward move of a pawn is +3 resp. −3. If that square is empty, a forward pawn move is a legal move.

Capture moves are slightly more intricate. For those we create two arrays, namely one array that contains target squares of white pawn captures and one for black pawn captures. For example `WHITE_PAWN_CAPTURES[4]` returns `[0,2]`, since from source square 4 (square b2) there are two potential capture squares, namely 0 (square a3) and 2 (square c3). Given a source square, we use these arrays to get all potential capture squares. If a capture square contains a pawn of the enemy color, we have a legal pawn capture move.

All moves are accumulated in `self.legal_moves` and returned. Here the array `self.legal_moves` acts as some kind of cache, i.e. once we computed all legal moves, we just return this list instead of re-computing all legal moves over and over again.

**Listing 5.5: Move Generation**

```
1 def __init__(self):
2     ...
```

```
 3      self.WHITE_PAWN_CAPTURES = [
 4          [],
 5          [],
 6          [],
 7          [1],
 8          [0,2],
 9          [1],
10          [4],
11          [3,5],
12          [4]
13      ]
14      self.BLACK_PAWN_CAPTURES = [
15          [4],
16          [3,5],
17          [4],
18          [7],
19          [6,8],
20          [7],
21          [],
22          [],
23          []
24      ]
25
26  def generateMoves(self):
27      if(self.legal_moves == None):
28          moves = []
29          for i in range(0, 9):
30              if(self.board[i] == self.turn):
31                  if(self.turn == self.WHITE):
32                      # check if we can move one square up
33                      toSquare = i - 3
34                      if(toSquare >= 0):
35                          if(self.board[toSquare] == self.EMPTY):
36                              moves.append((i, toSquare))
37                      potCaptureSquares = self.WHITE_PAWN_CAPTURES[i]
38                      for toSquare in potCaptureSquares:
39                          if(self.board[toSquare] == self.BLACK):
40                              moves.append((i, toSquare))
41                  if (self.turn == self.BLACK):
42                      toSquare = i + 3
43                      if(toSquare < 9):
44                          if (self.board[toSquare] == self.EMPTY):
```

```
45                              moves.append((i, toSquare))
46                    potCaptureSquares = self.BLACK_PAWN_CAPTURES[i]
47                    for toSquare in potCaptureSquares:
48                        if (self.board[toSquare] == self.WHITE):
49                            moves.append((i, toSquare))
50      self.legal_moves = moves
51   return self.legal_moves
```

The function isTerminal is used to check if a board is in a terminal position, i.e. if we have a winner. The function returns (False, None) if the position is not terminal, and it returns (True, Winner) where Winner is either Black or White if the position is terminal. A position is terminal if either Black or White have queened a pawn, or if it is White's turn (Black's turn) and he has no legal move.

**Listing 5.6: Detecting Terminal Position**

```
1  def isTerminal(self):
2     winner = None
3     if(self.board[6] == Board.BLACK or
4             self.board[7] == self.BLACK or
5             self.board[8] == self.BLACK):
6         winner = self.BLACK
7     if (self.board[0] == self.WHITE or
8             self.board[1] == self.WHITE or
9             self.board[2] == self.WHITE):
10        winner = self.WHITE
11    if(winner != None):
12        return (True, winner)
13    else:
14        if(len(self.generateMoves()) == 0):
15            if(self.turn == Board.WHITE):
16                return (True, Board.BLACK)
17            else:
18                return (True, Board.WHITE)
19        else:
20            return (False, None)
```

Encoding a Hexapawn position to a bitvector as input for the neural network is also straight forward. We create an empty list and then iterate over the whole board. Whenever we encounter a white pawn, we add a one to the list, and a zero otherwise. We do the same for the black pawns. Finally we add three ones[2] if it is White to move, and three zeros otherwise.

**Listing 5.7: Creating Network Input**

```
1 def toNetworkInput(self):
2     posVec = []
3     for i in range(0,9):
4         if(self.board[i] == Board.WHITE):
5             posVec.append(1)
6         else:
7             posVec.append(0)
8     for i in range(0,9):
9         if(self.board[i] == Board.BLACK):
10             posVec.append(1)
11         else:
12             posVec.append(0)
13     for i in range(0,3):
14         if(self.turn == Board.WHITE):
15             posVec.append(1)
16         else:
17             posVec.append(0)
18     return posVec
```

## 5.3 Supervised Learning

One main challenge in supervised learning is how to get high quality training data. Unfortunately there is no professional online Hexapawn league from where we could snatch game data. As the game is trivial, we could simply write down all possible game positions together with the best move and game outcome down ourselves; but despite the comparatively small game, that's still a tedious amount of work — there are 188 possible distinct game states as we will discover soon.

---

[2]Probably one bit would suffice for the network to recognize turns - but we really do not have to consider input size or performance issues for a game with such low game complexity.

We have already seen in Chapter 3 how *minimax* can be used to solve games, and that's what we are going to employ. We will start from the initial position and then try out every possible possible combination of moves until a game ends. In each position we will run minimax to get the best move and the game outcome. Since the game tree is of such shallow depth, minimax has no trouble searching until a terminal state is encountered.

**Generating Training Data**  First we define a function `getBestMove()` that takes as input a Hexapawn gamestate. The function then iterates through every possible legal move in that position, applies the move and runs minimax on the resulting position. It compares all outcomes to find the best possible move in that position. The game outcome as returned by minimax is also recorded. Note that when there are two or more possible moves that are best (i.e. winning) then only one of them is returned. But to play optimally that is all that is required.

We also create a recursive[3] function `visitNodes()` that takes as input a Hexapawn gamestate. This function first checks if the encountered position is terminal. If so we return, as a terminal position is useless as training data (there is no best move to play anymore). Otherwise the function generates all legals moves, applies each move, runs `getBestMove()` on the position and creates the training data. The training data consist of triples:

1. The game position itself, encoded as a suitable bitvector as input for the neural network

2. A move probability vector of size 28 corresponding to the output of the policy head of the neural network. Here we set all probabilities to 0 except for the best move for which we set the probability to one.

3. A value of either 1 or -1 (win or lose from the perspective of the White player) that corresponds to the desired output of the value head of the network.

These three features of the training data are stored separately in the lists `positions`, `moveProbs` and `outcomes`. We also put the currently encountered position into a Python dictionary in order to later count the amount of distinct positions that we encountered. Afterwards the function applies each move and calls itself recursively to generate training data for the subsequent positions.

---

[3]i.e. a function that calls itself

Finally we create a board with the initial position and run `visitNodes()` on that position. We save the generated training data as `numpy` arrays. As output by the program (the length of the dictionary), there are 188 distinct positions. We now have one training example for every possible game state. That should be enough to train our network!

**Listing 5.8: Creating Training Data**

```
1  def getBestMoveRes(board):
2      bestMove = None
3      bestVal = 1000000000
4      if(board.turn == board.WHITE):
5          bestVal = -1000000000
6      for m in board.generateMoves():
7          tmp = copy.deepcopy(board)
8          tmp.applyMove(m)
9          mVal = minimax(tmp, 30, tmp.turn == board.WHITE)
10         if(board.turn == board.WHITE and mVal > bestVal):
11             bestVal = mVal
12             bestMove = m
13         if(board.turn == board.BLACK and mVal < bestVal):
14             bestVal = mVal
15             bestMove = m
16     return bestMove, bestVal
17
18 positions = []
19 moveProbs = []
20 outcomes = []
21
22 gameStates = {}
23
24 def visitNodes(board):
25     gameStates[str(board)] = 1
26     term, _ = board.isTerminal()
27     if(term):
28         return
29     else:
```

```
30          bestMove , bestVal = getBestMoveRes(board)
31          positions . append(board . toNetworkInput())
32          moveProb = [ 0 for x in range(0,28) ]
33          idx = board . getNetworkOutputIndex(bestMove)
34          moveProb[idx] = 1
35          moveProbs . append(moveProb)
36          if(bestVal > 0):
37              outcomes . append(1)
38          if(bestVal == 0):
39              outcomes . append(0)
40          if(bestVal < 0):
41              outcomes . append(-1)
42          for m in board . generateMoves():
43              next = copy . deepcopy(board)
44              next . applyMove(m)
45              visitNodes(next)
46
47  board = Board()
48  board . setStartingPosition()
49  visitNodes(board)
50  print("# of reachable distinct positions:")
51  print(len(gameStates))
52  np . save("positions", np . array(positions))
53  np . save("moveprobs", np . array(moveProbs))
54  np . save("outcomes", np . array(outcomes))
```

**Training and Evaluation**   Training itself is almost trivial. We have already created the network architecture and saved a copy of the network with randomly initialized weights. All that remains to do now is to load that initial network, the training data and train it; here we use 512 epochs with a batch size of 16. This is probably a slight overkill, but due to the comparatively low amount of training data the network is still trained in seconds, even on a CPU. To get some more insight we can print out the shape of the training data. As we can see, there were 118 (non-terminal) game states that we encountered during training data generation.

Listing 5.9: Supervised Learning

```
1  model = keras . models . load_model("../common/random_model.keras")
```

```
2
3 inputData = np.load("../minimax/positions.npy")
4 policyOutcomes = np.load("../minimax/moveprobs.npy")
5 valueOutcomes = np.load("../minimax/outcomes.npy")
6
7 print(policyOutcomes.shape)
8
9 model.fit(inputData,[policyOutcomes, valueOutcomes], epochs=512,
      batch_size=16)
10 model.save('supervised_model.keras')
```

It remains now to see how successful we were with our training. A simple way to evaluate the network's performance is to have it play against a player that randomly selects moves. Here we create two functions. The first is `rand_vs_net`. It takes as input a game state and then applies moves until the game ends in a terminal state. If it is White to move, we generate all legal moves use `random.randint` to randomly select one of them. If it is Black to move, we query the network for move probabilities. The network might want to play illegal moves, so we mask out probabilities for illegal moves by setting them to 0. Technically we actually do it the opposite way: We create a vector of zeros, and then assign the network's probabilities to that vector but only for legal moves. Next we iterate through all legal moves and by using the corresponding move index for that move we find the maximum probability as output by the network. This is then the move we apply. If we reach a final position, we return the winner.

We also crate a second function `rand_vs_rand` where both White and Black play against each other by randomly selecting moves.

Next we play a hundred games twice with each function. Here we create a board with the initial position and let the random player and the network compete against each other as well as the two random players. Each time we record the number of wins and losses. Note that the game was solved and Black will always win if he plays optimally.

It is also worthwhile to let the random player play against the untrained network with randomly initialized weights. While it is unlikely, we could also have been just lucky, i.e. the randomly selected initialized weights were already such that
the network chooses an optimal move in every positions or a lot of positions —
in other words it plays better than randomly selecting moves.

Listing 5.10: Network Evaluation

```
1  model = keras.models.load_model("supervised_model.keras")
2  # model = keras.models.load_model("../common/random_model.keras")
3
4  def fst(a):
5      return a[0]
6
7  def rand_vs_net(board):
8      record = []
9      while(not fst(board.isTerminal())):
10         if(board.turn == Board.WHITE):
11             moves = board.generateMoves()
12             m = moves[random.randint(0, len(moves)-1)]
13             board.applyMove(m)
14             record.append(m)
15             continue
16         else:
17             q = model.predict(np.array([board.toNetworkInput()]))
18             masked_output = [ 0 for x in range(0,28)]
19             for m in board.generateMoves():
20                 m_idx = board.getNetworkOutputIndex(m)
21                 masked_output[m_idx] = q[0][0][m_idx]
22             best_idx = np.argmax(masked_output)
23             sel_move = None
24             for m in board.generateMoves():
25                 m_idx = board.getNetworkOutputIndex(m)
26                 if(best_idx == m_idx):
27                     sel_move = m
28             board.applyMove(sel_move)
29             record.append(sel_move)
30             continue
31     terminal, winner = board.isTerminal()
32     return winner
33
34  def rand_vs_rand(board):
35      while(not fst(board.isTerminal())):
36          moves = board.generateMoves()
```

```
37          m = moves[random.randint(0, len(moves)-1)]
38          board.applyMove(m)
39          continue
40      terminal, winner = board.isTerminal()
41      return winner
42
43  whiteWins = 0
44  blackWins = 0
45  draws = 0
46  for i in range(0,100):
47      board = Board()
48      board.setStartingPosition()
49      winner = rand_vs_net(board)
50      if(winner == Board.WHITE):
51          whiteWins += 1
52      if(winner == Board.BLACK):
53          blackWins += 1
54  all = whiteWins + blackWins + draws
55  print("Rand vs Supervised Network: "+str(whiteWins/all) +
56          "/"+str(blackWins/all))
57
58
59  whiteWins = 0
60  blackWins = 0
61  draws = 0
62  for i in range(0,100):
63      board = Board()
64      board.setStartingPosition()
65      winner = rand_vs_rand(board)
66      if(winner == Board.WHITE):
67          whiteWins += 1
68      if(winner == Board.BLACK):
69          blackWins += 1
70  all = whiteWins + blackWins + draws
71  print("Rand vs Rand Network: "+str(whiteWins/all) +
72          "/"+str(blackWins/all))
```

If we run the script we will get an output similar to Table 5.1[4]:

---

[4]Running this on your computer, your numbers might vary slightly due to the randomly initialization.

Table 5.1: Supervised Network Evaluation

| Matchup: Random Player ... | White Wins | Black Wins |
|---|---|---|
| vs Trained Network | 0% | 100% |
| vs Untrained Network | 34% | 66% |
| vs Random Player | 62% | 38% |

As we can see, the randomly initialized network indeed plays slightly better than just randomly selecting moves. Our trained network however wins all games. It seems the training worked out pretty well.

## 5.4 "Zero-like" Reinforcement Learning

We now pursue a "Zero"-like approach to train our network, that is reinforcement learning in combination with Monte Carlo Trees Search. The general outline is:

- initialize the network with random weights

- self-play a number of games. At each move, MCTS is used to get move probabilities over all legal moves in that position. The MCTS itself is guided by the network which is queried during MCTS for move probabilities and game outcomes. Apply the best move and continue until we end up in a final state of the game. We can also try out other moves than the best one in order to explore and get a broader knowledge about the game.

- We gathered a number of positions together with move probabilities and the final outcome of the game. We use these to train the network and repeat where we use the improved network in the next round of self-play games for MCTS.

The main task is now to implement MCTS such that it is guided by the network, and to implement the training loop. To evaluate the trained network we will use the same approach as for supervised learning, i.e. we will let the network compete against a random player. But first let's start with MCTS.

**Monte Carlo Tree Search**    Our MCTS approach follows closely the algorithm
that is sketched out by AlphaZero and AlphaGoZero. We distinguish between
*nodes* and *edges*.

Listing 5.11: MCTS Nodes and Edges

```python
class Edge():
    def __init__(self, move, parentNode):
        self.parentNode = parentNode
        self.move = move
        self.N = 0
        self.W = 0
        self.Q = 0
        self.P = 0

class Node():
    def __init__(self, board, parentEdge):
        self.board = board
        self.parentEdge = parentEdge
        self.childEdgeNode = []

    def expand(self, network):
        moves = self.board.generateMoves()
        for m in moves:
            child_board = copy.deepcopy(self.board)
            child_board.applyMove(m)
            child_edge = Edge(m, self)
            childNode = Node(child_board, child_edge)
            self.childEdgeNode.append((child_edge,childNode))
        q = network.predict(np.array([self.board.toNetworkInput()])
            )
        prob_sum = 0.
        for (edge,_) in self.childEdgeNode:
            m_idx = self.board.getNetworkOutputIndex(edge.move)
            edge.P = q[0][0][m_idx]
            prob_sum += edge.P
        for edge,_ in self.childEdgeNode:
            edge.P /= prob_sum
        v = q[1][0][0]
        return v

    def isLeaf(self):
```

```
36          return self.childEdgeNode == []
```

A node keeps the current state of the board. There are outgoing *edges* from nodes, and one edge corresponds to a particular move. During expansion of a node we first generate all legal moves and for each move an edge that contains the information. Each edge then leads to a new node that keeps the board position that results from applying the move to the current board position.

During expansion we also query the neural network with the current board position. The policy head of the network provides move probabilities for all the possible moves, and the value head provides an evaluation of the position. We store each move probability in the corresponding edge. We return the evaluation of the value head to the caller of the expansion function, so that MCTS can use this information for backpropagation. MCTS also needs to know whether a node is expanded or not. This is trivial to check by testing if there are any childs at all.

Beside the move and a link to the parent node, an edge stores the four important values N, W, Q, and P. As described in Chapter 4, N counts how many times the node was visited, W is the total reward, Q is the total reward divided by the number of times we visited that node, and P is the initial probability that was provided by the network during expansion.

Note that we store these values as absolute values and not from the perspective of the player whose turn it is. Thus if the position is favourable for Black, the total reward will be negative.

Having classes built that represent edges and nodes, we are ready to implement Monte Carlo Tree Search. During creation of our MCT searcher we need to set several parameters: The network that will be used during the search, the root node where we will start our search as well as parameters $\tau$ that is used to derive the final move probabilities and $c_{\text{puct}}$ that is used during selection for computation of the UCT value (cf. Chapter 4).

Remember that MCTS consists of four steps: Selection, Expansion, Evaluation and Backpropagation. Let's start with selection. Here we move from the root node down to a leaf node. If we are at a leaf node we are finished and return that node. Otherwise we will consider all possible move in that position, i.e. all edges.

**Listing 5.12: MCTS Class**

```
1 class MCTS():
2     def __init__(self, network):
3         self.network = network
4         self.rootNode = None
5         self.tau = 1.0
6         self.c_puct = 1.0
```

We compute the UCT value for each and then select the edge that maximizes the sum of the total reward of that edge and the UCT value, i.e. edge.Q + uctValue(edge). Remember that the UCT value was computed as

$$c_{\text{puct}} \cdot P \cdot \frac{\sqrt{\sum_{m'} N_{m'}}}{1 + N_m}$$

Note that $\sum_{m'} N_{m'}$ is just the parent visit count.

No matter if it is Black or White to play, we seek the best move from the perspective of the current player. Since we store rewards independent on whose turn it is, we need to invert the reward value Q, as we seek the child that maximizes the sum, but the best possible reward Q for Black is negative. Last, if there are several childs with the same value of the total reward plus UCT value, then we randomly select one child in order to have more exploration. Finally we move to the selected child and repeat.

**Listing 5.13: MCTS Selection**

```
1 def uctValue(self, edge, parentN):
2     return self.c_puct * edge.P * (math.sqrt(parentN) / (1+edge.N))
3
4 def select(self, node):
5     if(node.isLeaf()):
6         return node
7     else:
```

```
8            maxUctChild = None
9            maxUctValue = -100000000.
10           for edge, child_node in node.childEdgeNode:
11               uctVal = self.uctValue(edge, edge.parentNode.parentEdge
                     .N)
12               val = edge.Q
13               if(edge.parentNode.board.turn == Board.BLACK):
14                   val = -edge.Q
15               uctValChild = val + uctVal
16               if(uctValChild > maxUctValue):
17                   maxUctChild = child_node
18                   maxUctValue = uctValChild
19           allBestChilds = []
20           for edge, child_node in node.childEdgeNode:
21               uctVal = self.uctValue(edge, edge.parentNode.parentEdge
                     .N)
22               val = edge.Q
23               if(edge.parentNode.board.turn == Board.BLACK):
24                   val = -edge.Q
25               uctValChild = val + uctVal
26               if(uctValChild == maxUctValue):
27                   allBestChilds.append(child_node)
28           if(maxUctChild == None):
29               raise ValueError("could not identify child with best
                     uct value")
30           else:
31               if(len(allBestChilds) > 1):
32                   idx = random.randint(0, len(allBestChilds)-1)
33                   return self.select(allBestChilds[idx])
34               else:
35                   return self.select(maxUctChild)
```

We can combine expansion and evaluation in one function. If a node is terminal we do not to query the network for an evaluation value and can instead take the game result and propagate it as the reward back in the tree up to the root node. If the node is not terminal we expand the node, thereby yielding the evaluation value of the network for that node and propagate it back in the tree.

Backpropagation is straight forward. For each edge we increase the visit count, add the reward to the total reward $W$ and update the mean reward by dividing the total reward through the visit count. We then check if the edge has a parent node (and subsequent parent edge) and repeat backpropagation. If we instead reached the root we stop.

**Listing 5.14: MCTS Expansion**

```
1  def expandAndEvaluate(self, node):
2      terminal, winner = node.board.isTerminal()
3      if(terminal == True):
4          v = 0.0
5          if(winner == Board.WHITE):
6              v = 1.0
7          if(winner == Board.BLACK):
8              v = -1.0
9          self.backpropagate(v, node.parentEdge)
10         return
11     v = node.expand(self.network)
12     self.backpropagate(v, node.parentEdge)
```

**Listing 5.15: MCTS Backpropagation**

```
1  def backpropagate(self, v, edge):
2      edge.N += 1
3      edge.W = edge.W + v
4      edge.Q = edge.W / edge.N
5      if(edge.parentNode != None):
6          if(edge.parentNode.parentEdge != None):
7              self.backpropagate(v, edge.parentNode.parentEdge)
```

It remains to implement the actual MCT search procedure. Here we are given a node which is the root node where we start the search. Before starting the actual search, we expand this root node. Then we iterate a fixed number of times (here a 100 times) and execute *selection*, *expansion*, *evaluation* and *backpropagation* – the latter is called by the evaluation function directly. The policy value that is returned by an MCT search was defined as

$$\pi_m = \frac{N_m^{1/\tau}}{\sum_n N_n^{1/\tau}}$$

i.e. essentially turning the visit counts into probabilities by dividing each count through the sum of all visit counts for the root edge. As described in Chapter 4, the parameter $\tau$ can be used in order to facility more exploration. For Hexapawn we just set it to a fixed value of 1.0 during initialization of our MCTS object.

**Listing 5.16: MCT Search**

```
1 def search(self, rootNode):
2     self.rootNode = rootNode
3     _ = self.rootNode.expand(self.network)
4     for i in range(0,100):
5         selected_node = self.select(rootNode)
6         self.expandAndEvaluate(selected_node)
7     N_sum = 0
8     moveProbs = []
9     for edge, _ in rootNode.childEdgeNode:
10         N_sum += edge.N
11     for (edge, node) in rootNode.childEdgeNode:
12         prob = (edge.N ** (1 / self.tau)) / ((N_sum ** (1/self.tau
               )))
13         moveProbs.append((edge.move, prob, edge.N, edge.Q))
14     return moveProbs
```

**Training and Evaluation** To implement the self-play pipeline we define a class `ReinfLearn`. During initialization we provide the network to that class. The class has only one function `playGame()` which is called to self-play a game and collect the positions, move probabilities and outcomes of this game.

First we set up lists that will hold the positions (already encoded as network input bit vectors), the move probabilities and the outcome values. Next we set up a Hexapawn board in the initial position. We then start to play the game.

First we append the current position to the `positionsData` list. Then we set up the MCT search object by creating a root node from the current position and execute the MCT search. The result are move probabilities for each legal move. Next we create a move probability vector where we initialize each value with 0. Then we set the move probability for the legal moves to those values returned by the MCT search. Thus move probabilities for illegal moves in that vector are masked to 0.

We now need to select a move to play. Here we slightly deviate from the AlphaZero algorithm. The reason is that Hexapawn is a very small game that lasts only a few moves. Hence, the MCT search is very successful, i.e. even if the network is initialized with random values and not trained at all, the MCT search will almost always find the very best move in each position. If we now select the best move we will play the same game over and over again without exploring all possible game states.

The problem also exists in AlphaZero, and to have more exploration they used two parameters: First, they adjusted the parameter $\tau$ for MCT search. Second they added random noise at the root node during MCT search. Here we will resort to a simpler solution that works just as well. Note that `outputVec` contains probabilities that add up to one. We can interpret this as a *multinomial probability distribution* and randomly select a move according to that probability distribution. Without going into mathematical details, the way this works is best illustrated by an example. Suppose we have the vector of move probabilities

$$0.1, 0.2, 0.7$$

We now want to randomly select one of the three possible indices (and thereby the associated moves), but we do not want to do that randomly with the same probability. Instead we desire to select the index 0 only in about 10 percent of all cases, index 1 in about 20 percent of all cases and index 2 in about 70 percent of all cases. In other words we want to randomly select moves, but still favor those moves which are more promising according to the MCT search. This is precisely what calling `np.random.multinomial` and subsequent selection of a move does. Next we append the move probabilities for the current position to our list, apply the move and continue.

If we reached a final position we need to evaluate who won the game. We add this information $n$ times to our list of outcomes where $n$ is the number of positions that we encountered during the game. This means we now have collected $n$ training examples which consist of $n$ positions (in the list positionsData), $n$ sets of move probabilities (in the list moveProbsData) and $n$ game outcomes (the same value for each position of the game of course, in the list valuesData).

> **Listing 5.17: "AlphaZero"-like Training (1)**

```
1  def fst(x):
2      return x[0]
3
4  class ReinfLearn():
5
6      def __init__(self, model):
7          self.model = model
8
9      def playGame(self):
10         positionsData = []
11         moveProbsData = []
12         valuesData = []
13
14         g = Board()
15         g.setStartingPosition()
16
17         while((not fst(g.isTerminal()))):
18             positionsData.append(g.toNetworkInput())
19
20             rootEdge = mcts.Edge(None, None)
21             rootEdge.N = 1
22             rootNode = mcts.Node(g, rootEdge)
23             mctsSearcher = mcts.MCTS(self.model)
24
25             moveProbs = mctsSearcher.search(rootNode)
26             outputVec = [ 0.0 for x in range(0, 28)]
27             for (move, prob, _, _) in moveProbs:
28                 move_idx = g.getNetworkOutputIndex(move)
29                 outputVec[move_idx] = prob
30
31             rand_idx = np.random.multinomial(1, outputVec)
32             idx = np.where(rand_idx==1)[0][0]
33             nextMove = None
34
35             for move, _, _, _ in moveProbs:
36                 move_idx = g.getNetworkOutputIndex(move)
37                 if(move_idx == idx):
```

```
38                      nextMove = move
39                moveProbsData.append(outputVec)
40                g.applyMove(nextMove)
41            else:
42                _, winner = g.isTerminal()
43                for i in range(0, len(moveProbsData)):
44                    if(winner == Board.BLACK):
45                        valuesData.append(-1.0)
46                    if(winner == Board.WHITE):
47                        valuesData.append(1.0)
48        return (positionsData, moveProbsData, valuesData)
```

We are now ready to train our network. We load the randomly initialized model, setup the MCT search object and the reinforcement learner, and start training. Here we used 21 training loops, and save the current network at iteration 0, 10 and 20. In each loop we play ten games and obtain training data from these ten games. We then use this training data to train and improve the network and repeat.

```
1  model = keras.models.load_model("../common/random_model.keras")
2  mctsSearcher = mcts.MCTS(model)
3  learner = ReinfLearn(model)
4  for i in (range(0,11)):
5      print("Training Iteration: "+str(i))
6      allPos = []
7      allMovProbs = []
8      allValues = []
9      for j in tqdm(range(0,10)):
10         pos, movProbs, values = learner.playGame()
11         allPos += pos
12         allMovProbs += movProbs
13         allValues += values
14     npPos = np.array(allPos)
15     npProbs = np.array(allMovProbs)
16     npVals = np.array(allValues)
17     model.fit(npPos,[npProbs, npVals], epochs=256, batch_size=16)
18     if(i%10 == 0):
19         model.save('model_it'+str(i)+'.keras')
```

The evaluation of our trained network is the same as before. We simply reload (one of the) networks trained by reinforcement learning instead of the one trained by supervised learning.

| Matchup: Random Player ... | White Wins | Black Wins |
|---|---|---|
| ... vs Trained Network (iteration 0) | 0% | 100% |
| ... vs Trained Network (iteration 10) | 0% | 100% |
| ... vs Untrained Network | 34% | 66% |
| ... vs Random Player | 62% | 38% |

Table 5.2: Supervised Network Evaluation

On my system I get the results depicted in Table 5.2. Again, due to the random initialization of the network your numbers may vary, but we can see that even one iteration of training suffices to ensure that the trained network wins all games with Black.

## 5.5 Summary

We have seen how to implement an AlphaZero-like learning approach in Python. It is apparent that both supervised and reinforcement learning work here for Hexapawn. Clearly, supervised learning is much simpler to implement, but requires training data to be available. Whether such training data is readily available or not heavily depends on the problem in question.

Hexapawn is way too trivial to draw any conclusions as to what approach is better. Both, supervised and reinforcement learning create strategies that work a hundred percent, something that we cannot do for complex games such as chess. Especially, for our network here it is enough to simply remember all crucial positions, as there are very few game states. No abstraction or pattern recognition was required and therefore the network architecture did not require any advanced building blocks such as convolutional layers.

Note how small — in terms of lines of code — the implementation was, especially, but not only, for supervised learning. Note also how universal the approach is. To change the implementation to other problem domains it suffices to exchange the game logic. Everything else, i.e. the network architecture as well as MCT search, remain largely the same.

# 6

# Conclusion and Outlook

> I hate to write. I love to have
> written.
>
> —————————————————————
>
> Douglas Adams

I hope you had as much fun reading this book as I had writing it.

In fact these are stereotypical closing words and thinking about it, at times,
writing this was not so much fun at all. Instead it was a lot of work. If you
carefully read this book it must have been a lot of work, too. So congratulations
for that — I hope you feel that it was worth the effort and still enjoyed the whole
experience.

We have covered neural networks and their building blocks. These are used
not only in chess programming but of course in other domains as well. A lot of
"revolutionary" recent artificial intelligence based solutions are based on such
neural networks. Therefore taking a look at those concepts should enable you
to get an easier understanding of those solutions as well.

I hope that the introduction into current state-of-the art engines like Leela Chess
Zero and Stockfish NNUE also gave you a better understanding on what chess engines are capable of and what the current limits are, especially if we want to use them on commodity hardware.

Predictions are incredibly difficult, especially when they concern the future. But if I want to make one predictions it's that alpha-beta searchers will prevail for chess. Moreover if neural network circuitry and accelerators will become more common on desktop and mobile computers, it is likely that other, more powerful ways of combining alpha-beta search with fast neural network based position evaluation will start to appear. And hopefully it will enable chess programmers to use these techniques not only to build the strongest engine, but also to build learning tools and realistic sparring partners in order to help humans become better chess players.